\documentclass[12pt]{report}   
\usepackage[hidelinks]{hyperref}
\usepackage{graphicx}  
\usepackage[letterpaper, left=1in, right=1in, top=1in, bottom=1in]{geometry}
\usepackage{setspace}  
\usepackage{times}  
\usepackage[explicit]{titlesec}  
\usepackage[titles]{tocloft}  
\usepackage[backend=biber, sorting=nty, bibstyle=ieee]{biblatex}  
\usepackage{appendix}  
\usepackage{rotating}  
\usepackage{textcomp} 
\usepackage{indentfirst} 
\usepackage{booktabs,array,arydshln} 
\usepackage{amsmath} 
\usepackage[T1]{fontenc} 
\usepackage[utf8]{inputenc} 
\usepackage{newtxmath} 

\usepackage{soul}
\usepackage[abs]{overpic}
\usepackage{microtype}        
\usepackage{ccicons}          
\usepackage{listings}
\usepackage{courier}
\usepackage{color}
\usepackage{subcaption}
\usepackage{caption}
\usepackage{algorithm}
\usepackage{algorithmic}
\usepackage{tabularx}
\usepackage{siunitx}
\usepackage{enumitem}
\usepackage[table]{xcolor}
\usepackage{multirow}
\usepackage{makecell}
\usepackage{mwe}
\usepackage{adjustbox}
\usepackage{breakcites}
\usepackage{url}
\usepackage{amsfonts}

\bibliography{references}

\AtEveryBibitem{\clearfield{issn}}
\AtEveryBibitem{\clearlist{issn}}

\AtEveryBibitem{\clearfield{language}}
\AtEveryBibitem{\clearlist{language}}

\AtEveryBibitem{\clearfield{doi}}
\AtEveryBibitem{\clearlist{doi}}

\AtEveryBibitem{%
  \ifentrytype{online}
    {}
    {\clearfield{urlyear}\clearfield{urlmonth}\clearfield{urlday}}}

\DeclareUnicodeCharacter{0301}{\'{e}}

\begin{document}
\doublespacing  

\pagenumbering{gobble}

\begin{titlepage}
\begin{center}

\begin{singlespacing}
\vspace*{6\baselineskip}
Learning Mobile Manipulation\\
\vspace{3\baselineskip}
David Watkins\\
\vspace{18\baselineskip}
Submitted in partial fulfillment of the\\
requirements for the degree of\\
Doctor of Philosophy\\
under the Executive Committee\\
of the Graduate School of Arts and Sciences\\
\vspace{3\baselineskip}
COLUMBIA UNIVERSITY\\
\vspace{3\baselineskip}
\the\year
\vfill

\end{singlespacing}

\end{center}
\end{titlepage}

\pagenumbering{gobble}

\begin{titlepage}
\begin{singlespacing}
\begin{center}

\vspace*{35\baselineskip}

\textcopyright  \,  \the\year\\
\vspace{\baselineskip}	
David Watkins\\
\vspace{\baselineskip}	
All Rights Reserved
\end{center}
\vfill

\end{singlespacing}
\end{titlepage}

\pagenumbering{gobble}

\begin{titlepage}
\begin{center}

\vspace*{5\baselineskip}
\textbf{\large Abstract}

Learning Mobile Manipulation

David Watkins
\end{center}
\begin{flushleft}
\hspace{10mm}
	Providing mobile robots with the ability to manipulate objects has, despite decades of research, remained a challenging problem. The problem is approachable in constrained environments where there is ample prior knowledge of the environment layout and manipulatable objects. The challenge is in building systems that scale beyond specific situational instances and gracefully operate in novel conditions. In the past, researchers used heuristic and simple rule-based strategies to accomplish tasks such as scene segmentation or reasoning about occlusion. These heuristic strategies work in constrained environments where a roboticist can make simplifying assumptions about everything from the geometries of the objects to be interacted with, level of clutter, camera position, lighting, and a myriad of other relevant variables. The work in this thesis will demonstrate how to build a system for robotic mobile manipulation that is robust to changes in these variables. This robustness will be enabled by recent simultaneous advances in the fields of big data, deep learning, and simulation. The ability of simulators to create realistic sensory data enables the generation of massive corpora of labeled training data for various grasping and navigation-based tasks. It is now possible to build systems that work in the real world trained using deep learning entirely on synthetic data. The ability to train and test on synthetic data allows for quick iterative development of new perception, planning and grasp execution algorithms that work in many environments.
	
    To build a robust system, this thesis introduces a novel multiple-view shape reconstruction architecture that leverages unregistered views of the object. To navigate to objects without localizing the agent, this thesis introduces a novel panoramic target goal architecture that takes previous views of the agent to inform a policy to navigate through an environment. Additionally, a novel next-best-view methodology is introduced to allow the agent to move around the object and refine its initial understanding of the object. The results show that this deep learned sim-to-real approach performs best when compared to heuristic-based methods in terms of reconstruction quality and success-weighted-by-path-length (SPL). This approach is also adaptable to the environment and robot chosen due to its modular design.

\end{flushleft}
\vspace*{\fill}
\end{titlepage}

\currentpdfbookmark{Table of Contents}{TOC}
\pagenumbering{roman}
\setcounter{page}{1} 
\renewcommand{\cftchapdotsep}{\cftdotsep}  
\renewcommand{\cftchapfont}{\normalfont}  
\renewcommand{\cftchappagefont}{}  
\renewcommand{\cftchappresnum}{Chapter }
\renewcommand{\cftchapaftersnum}{:}
\renewcommand{\cftchapnumwidth}{5em}
\renewcommand{\cftchapafterpnum}{\vskip\baselineskip} 
\renewcommand{\cftsecafterpnum}{\vskip\baselineskip}  
\renewcommand{\cftsubsecafterpnum}{\vskip\baselineskip} 
\renewcommand{\cftsubsubsecafterpnum}{\vskip\baselineskip} 

\titleformat{\chapter}[display]
{\normalfont\bfseries\filcenter}{\chaptertitlename\ \thechapter}{0pt}{\large{#1}}

\renewcommand\contentsname{Table of Contents}
\renewcommand{\chapterautorefname}{Chapter}
\renewcommand{\appendixautorefname}{Appendix}

\begin{singlespace}
\tableofcontents
\end{singlespace}

\clearpage

\phantomsection
\addcontentsline{toc}{chapter}{Acknowledgments}

\clearpage
\begin{center}

\vspace*{5\baselineskip}
\textbf{\large Acknowledgements}
\end{center}

\hspace{10mm}
I would like to acknowledge all members of the Columbia Robotics Lab (CRLab) for making my stay there an enjoyable and rewarding experience. In particular, I would like to thank Adam Richardson, Bohan Wu, Caroline Weinberg, Chaiwen Chou, Feng Xu, Gavi Rawson, Ian Huang, Jack Shi, Jenny Li, Jiaheng Hu, Jonathan Sanabria, Lucas Schuermann, Madhavan Seshadri, Neil Chen, Shriya Balaji Palsamudram, Vaibhav Vavilala, Wenhao Li, Xuelong Mu, and Zizhao Wang for their support and collaborative effort during my Ph.D. I would like to acknowledge Amy Xu, Ashley Kling, Dr. Boyan Penkov, Chris Mulligan, Dave Chisholm, Emily Chen, Howon Byun, Katy Gero, Kevin Kwan, Khaled Atef, Levi Oliver, Lynne Weber, and Oriana Fuentes for their collaboration during my tenure at Columbia University. All the work they have done has broadened my understanding of the field of computer science.

My research colleagues have provided me with a wealth of support and knowledge throughout my studies. Dr. Iretiayo Akinola and Dr. Jacob Varley both collaborated and taught me about the world of robotics and research while making our work a joy. Dr. Carmine Elvezio and Henrique Maia provided me with plenty of jokes, guidance, and support during the difficult years of the pandemic while also providing great feedback about my research. To my collaborators Jingxi Xu, John Hui, and Dr. Vinicius Goecks, whose research expertise and excellent attitude towards computer science helped guide me through a variety of projects. I would like to acknowledge Dr. Caroline Yu, Dr. Cassie Meeker, Chad Dechant, Dr. Richard Townsend, and Dr. Travis Riddle, for collaborating with me on various research projects.

To my teachers and mentors, Ann Bozdogan, Anne Fleming, Bill Andersen, Dr. Buck Weaver, Cammy Morteo, Dr. Carole Srinivasan, Sister Catherine Clifford, Colm McGarry, Cynthia Daigle Xenakis, Dana Gurwitch, Edward Barry, Edward Kern, George Peterson, Gerald Herlihy, Hipolito Rivera, James Hyland, Dr. Janet Kayfetz, Jayant Srinivasan, Joyce Cavanaugh, Judy Landis, Kathleen Germaine, Kelly Naughton, Michael Alger, Dr. Pierluigi Miraglia, Shreyas Shah, Stephen Flynn, Walter Johnson, and Prof. Wayne Snyder, whose guidance throughout my life has molded me into the person I am today. I am grateful to each educator who has guided me to this point in my life.

The faculty at Columbia University have provided me with years of experience that helped me throughout my research experience. Prof. Jae Woo Lee, Prof. John Kender, and Prof. Paul Blaer, have all helped me through their anecdotes and mentoring. Thank you to Prof. Daniel Rubenstein, Prof. John Kymissis, Prof. Martha Kim, and Prof. Valerie Purdie-Greenaway, for their collaboration during my Ph.D. The members of my committee, Prof. Matei Ciocarlie, Dr. Michael Reed, and Prof. Shuran Song, deserve special thanks for the guidance and support provided to me throughout this process. I also extend thanks to my fellowship advisor, Dr. Nicholas Waytowich, who collaborated and advised me on novel research methods, and exposed me to a wealth of information.

My advisor, Prof. Peter Allen, without whom none of this would have been possible. Peter's knowledge and guidance gave me the tools to become the researcher I am now. He has given me the opportunity to improve the world through robotics. I am eternally grateful.

To my siblings, Jonathan Watkins, Emelie Watkins, and Matthew Watkins, whose humor and support over the course of my life have been invaluable resources to me. Our many moments of joy and fun together have helped to keep me engaged and youthful during my life and especially during my Ph.D.

To my grandparents, $\dagger$Rafael Valls, Lillian Emanuelli, Henry Watkins, and Jacqueline Watkins, whose infinite wisdom and support throughout my life has fueled my curiosity for science and discovery. I am extremely fortunate to have been able to share this part of my life with them.

Rarely does one get to extend as much gratitude as I have for both my parents, David Watkins, and Lillian Watkins. Without them, I would not have been able to start or finish my Ph.D. program. The wealth of support, validation, expertise, and solutions they have provided me over the 27 years of my life has been invaluable. I could not and would not ask for better parents than them.

\clearpage


\phantomsection
\addcontentsline{toc}{chapter}{Funding}

\clearpage
\begin{center}

\vspace*{5\baselineskip}
\textbf{\large Funding Sources}
\end{center}

\begin{flushleft}
\hspace{10mm}
Graduate study was supported by a fellowship from the U.S. Army Research Laboratory through the Oak Ridge Associated Universities and by NSF Grant CMMI 1734557.

This research was sponsored by the Army Research Laboratory and was accomplished under Cooperative Agreement Number W911NF-18-2-0244 and W911NF-20-2-0114. The views and conclusions contained in this document are those of the authors and should not be interpreted as representing the official policies, either expressed or implied, of the Army Research Laboratory or the U.S. Government. The U.S. Government is authorized to reproduce and distribute reprints for Government purposes notwithstanding any copyright notation herein.

\end{flushleft}
\clearpage


\phantomsection
\addcontentsline{toc}{chapter}{Dedication}

\begin{center}

\vspace*{5\baselineskip}
\textbf{\large Dedication}
\end{center}

\begin{flushleft}
\hspace{10mm}
I dedicate this thesis to my parents, David Vincent Watkins and Lillian Valls Watkins.

\end{flushleft}




\clearpage
\pagenumbering{arabic}
\setcounter{page}{1} 

\phantomsection
\addcontentsline{toc}{chapter}{Preface}

\begin{center}
\vspace*{5\baselineskip}
\textbf{\large Preface}
\end{center}

\begin{flushleft}
\hspace{10mm}
Before you lie the dissertation "Learning Mobile Manipulation", the basis of which is a solution to the problem of robotic mobile manipulation. This document has been written to fulfill the graduation requirements of the Computer Science graduate program at Columbia University in the city of New York. I was engaged in researching and authoring this dissertation from January 2020 to May 2022. 

My initial background before starting my Ph.D. was specifically in software design principles and programming languages. When I started working in my advisor's, Peter Allen's, lab, I was able to leverage my understanding of parallel programming to optimize algorithmic approaches to robotics. Over the course of my Ph.D., I have leveraged a software-oriented approach to robotics that has allowed me to create robust solutions that work in both real and simulated contexts. I have gained skills in robotic navigation, manipulation, and deep learning through the many different projects I have worked on.

This project is a culmination of work done over the course of my Ph.D. in robotic manipulation, shape understanding, robotic navigation, and next-best-view planning. The research questions were formulated together with my advisor, Peter Allen. This research was challenging, but through this investigation I was able to produce a unique solution to mobile manipulation. 

My motivation for this work is to provide a unique perspective on mobile manipulation. The field of robotics is currently undergoing a transition between algorithmic approaches from the past 60 years of robotics research and new deep learned approaches. These approaches are both valid, but which approach, or a hybrid of the two, will perform best remains to be seen. In my thesis I have put effort into building a hybrid approach to solve the problem of mobile manipulation. 

The audience of this dissertation is aspiring roboticists, veterans of the field of robotics, or people looking to implement mobile manipulation systems. . I constructed this thesis to document my research conducted during my Ph.D. first, but second to illustrate the challenges a new researcher may face when trying to build complicated end-to-end systems that require deep learning and control methods. 

I would like to acknowledge the editors of this thesis, Dr. Janet Kayfetz and Prof. Peter Allen. The chapters of this thesis are based on previously published work. Each of these chapters have shared authors who have contributed knowledge and writing to help me accomplish this research. The co-contributors are as follows:
\begin{enumerate}
	\item \textbf{Chapter 3} Jingxi Xu, Nicholas Waytowich, and Peter Allen
	\item \textbf{Chapter 4} Jacob Varley and Peter Allen
	\item \textbf{Chapter 5} Jacob Varley and Peter Allen
	\item \textbf{Chapter 6} Peter Allen, Henrique Maia, Madhavan Seshadri, Jonathan Sanabria, Nicholas Waytowich, and Jacob Varley
	\item \textbf{Chapter 7} Jacob Varley, Krzysztof Choromanski, and Peter Allen
	\item \textbf{Appendix B} Vinicius G. Goecks, Nicholas Waytowich, and Bharat Prakash
\end{enumerate}
\noindent Thank you to all my co-contributors for your help and expertise over the course of my graduate studies.

I hope you enjoy your reading.

David Watkins

\end{flushleft}


\titleformat{\chapter}[display]
{\normalfont\bfseries\filcenter}{}{0pt}{\large\chaptertitlename\ \large\thechapter : \large\bfseries\filcenter{#1}}  
\titlespacing*{\chapter}
  {0pt}{0pt}{30pt}	
  
\titleformat{\section}{\normalfont\bfseries}{\thesection}{1em}{#1}

\titleformat{\subsection}{\normalfont}{\thesubsection}{0em}{\hspace{1em}#1}


\chapter{Introduction}
\label{ch:intro}

\section{Mobile Manipulation}
The field of mobile manipulation derives its direction from the set of skills used by humans as they interact with their environment. As toddlers learn to walk by 11-12 months of age, they explore and reason about their environment, and they also learn to manipulate objects within that space, for example stacking, sorting, and counting. These same skills encompass the multiple facets of robotic mobile manipulation research. A robot that can successfully manipulate its environment must move throughout an environment or space to reason about its environment and to manipulate objects within that space. It is these skills and abilities translated into a complexity of multiple systems operating in concert that roboticists have worked to address for decades.

While researchers are currently able to build robotic mobile manipulation systems to manipulate objects within an environment relying on specific constraints on the problem, building a general-purpose mobile manipulation system remains a challenging engineering problem. The engineering difficulty increases exponentially as the number of intertwined systems increases. More specifically, mobile manipulation requires navigation, manipulation, and robotic control systems where each requires its own set of pipelines, with overall complexity additionally impacted as the environment becomes cluttered, messy, or as dynamic elements like people in the space move around. Robots, unlike people, face three critical issues when interacting in a space: 1) they are not localized in the environment, 2) they do not know the objects they will manipulate beforehand; and 3) they do not know how to leverage tactile information or multiple views of an object. This thesis addresses these issues by proposing a solution to localization-free robotic mobile manipulation to manipulate unseen objects leveraging multiple object views.

\begin{figure}[H]
    \centering {
        \includegraphics[width=\linewidth]{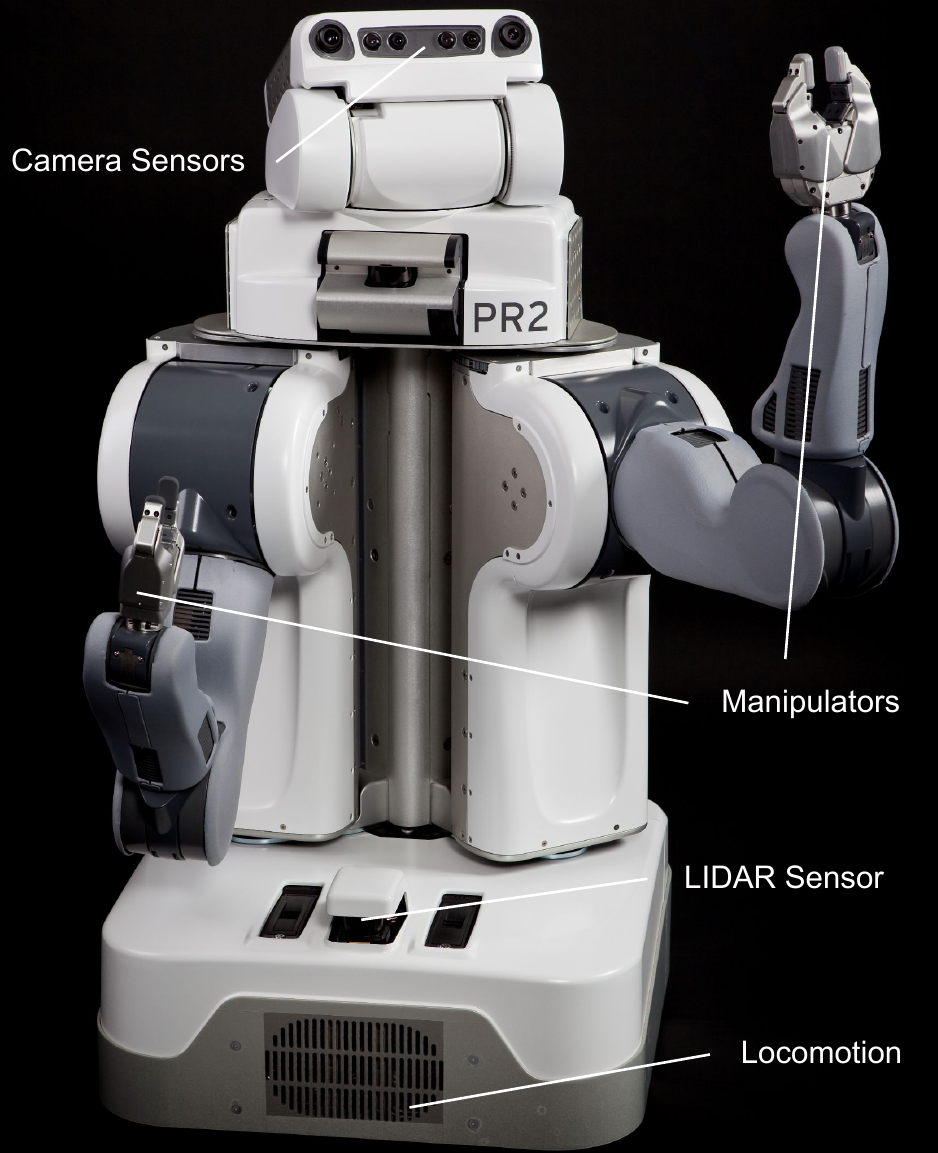}
    }
    \caption{The PR2 is emblematic of mobile manipulation. It has a set of arms for interacting with its environment. It uses a set of wheels for locomotion. For sensory information, it has a LIDAR sensor, a set of cameras, as well as optional force torque and tactile sensing. It is important to highlight these traits as they describe what the target robot for this research is. }
    \label{fig:pr2}
\end{figure}

A robotic mobile manipulation pipeline can be expressed as a state machine where the agent is trying to reach a goal state via a sequence of transitions. There are many goal configurations that a robotic agent may want to achieve, such as lifting an object or opening a door. Important attributes of a mobile manipulator include that it can move around its environment, capable of acquiring sensory information about its surroundings, and has a manipulator that can interact with objects in a workspace. There have been many examples of mobile manipulators, but one is particularly emblematic of the title "Mobile Manipulator": the Willow Garage PR2. The PR2 has two arms each with a parallel jaw gripper, a LIDAR sensor, a series of cameras, and wheels for locomotion. Each of these traits are highlighted in \autoref{fig:pr2}. Highlighting these specific set of traits is important as it establishes what kind of robot would be relevant for this thesis. It also helps illustrate how each state in the proposed mobile manipulation state machine maps onto a real-world example:

\begin{figure}[ht!]
    \centering {
        \includegraphics[width=\linewidth]{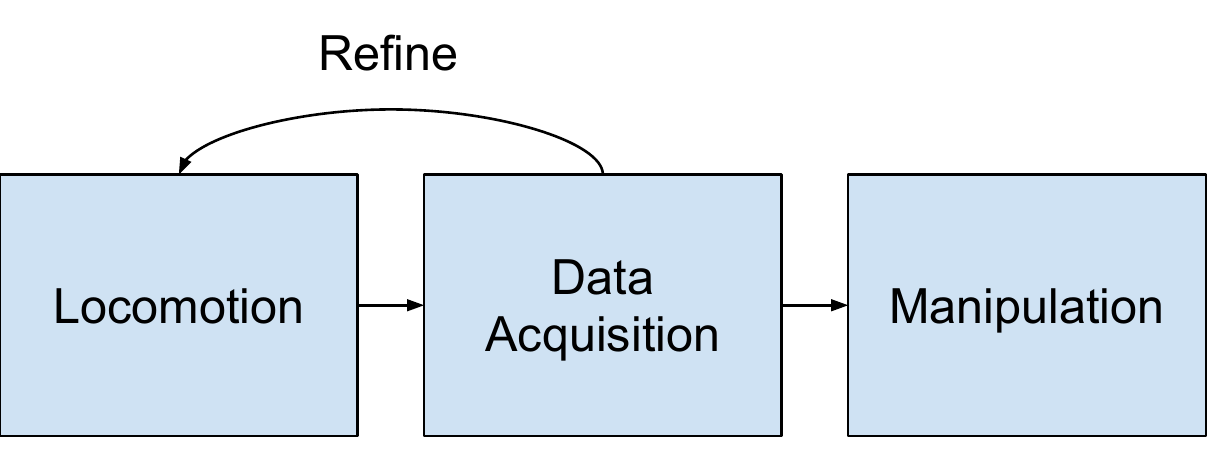}
    }
    \caption{The state transition diagram of a mobile manipulator. The agent starts from some starting configuration eventually accomplishing a target task. Simplifying a mobile manipulation pipeline in this paradigm helps to illustrate what is required to innovate research in this field. }
    \label{fig:mobilemanipulationstatediagram}
\end{figure}

\begin{enumerate}
    \item \textbf{Locomotion} The robotic agent will move through its environment to a target position or configuration of its joints to allow itself to begin executing a manipulation task. This task can be done by a variety of robotic agents including humanoid, drone, quadrupedal, and aquatic. Once at the target location, the agent will then begin a data acquisition stage. 
    \item \textbf{Data Acquisition} The agent needs to collect information about its target object using cameras, tactile sensors, temperature sensors, and any modality that provides information about the target. This stage informs the agent as to whether it should continue searching for additional information from another position relative to the object to increase the success of a manipulation step. If the agent decides that it has received enough information, it will then move to a manipulation planning step. 
    \item \textbf{Manipulation} The agent will utilize the information it collected to plan a series of moves allowing it to execute a manipulation task on the environment. This manipulation task could be picking up a piece of fruit in a kitchen, opening a cabinet, or loading a dishwasher. Manipulation planning and execution is a combination of robotic control systems and algorithmic approaches to utilize information about the agent's environment to better accomplish a task. 
\end{enumerate}

\noindent
The state transition diagram is shown in \autoref{fig:mobilemanipulationstatediagram}. 

Mobile manipulation encompasses such a wide variety of robotic fields and disciplines that it can be challenging to address in a single work. But each time researchers address robotic mobile manipulation, the work brings roboticists closer to a general-purpose mobile manipulation platform: one that will be able to aid humans in their homes, workplaces, and lives to accomplish more than they could do on their own. This dissertation is no different. 

\section{Problem Statement}
\noindent
The impetus for mobile manipulation research can be stated as follows:
\textit{There is no clear data-driven approach for localization free robotic mobile manipulation. Such manipulation systems are not yet generalized to arbitrary locations or unseen objects in a real-world environment considering sensor noise, task location, workspace configuration, and other environmental variables. Current systems constrain the problem to conform to global positions in the environment or restricting the problem to a known object database. The lack of a natural basis for coordinate systems in the real-world is preventing these agents from handling more complex scenarios outside of controlled lab and factory settings where researchers can choose and identify experimental constraints. Constraining the set of objects also prevents generalization to new domains of problems where knowing object geometry beforehand is not possible. }

\section{Approach}
\begin{figure}[ht!]
    \centering {
        \includegraphics[width=\textwidth]{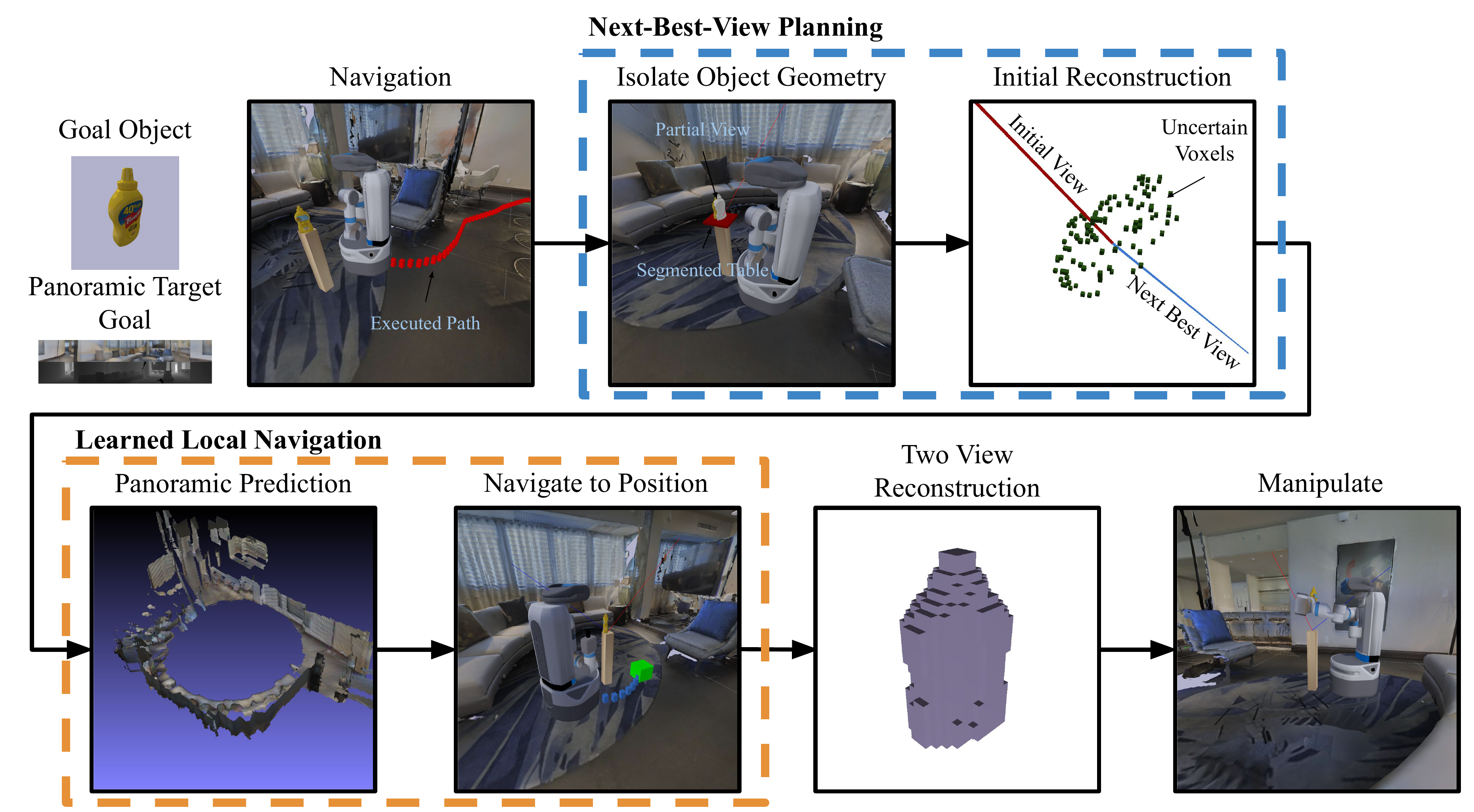}
    }
    \caption{\textbf{Two-View Robotic Mobile Manipulation Pipeline} Given an image of the goal object and a panoramic target goal, the agent navigates to the region and aligns itself with the object. It then uses a single-view completion of the object to predict a next-best-view and utilizes a predicted panoramic image to navigate to that next-best-view. Upon capturing the next-best-view it refines its prediction of the object to calculate and perform a grasp on the object.  }
    \label{fig:mobilemanipulationpipeline}
\end{figure}

This dissertation presents a novel approach to mobile manipulation. The scope of mobile manipulation is restricted to manipulating an object indoors, in an environment the agent knows beforehand, with an object that the robotic agent can grasp. The agent is not aware of the object beforehand, nor is it allowed to localize itself at runtime. The agent can navigate to an object, understand the shape of that object, and manipulate the object without requiring it to know its position in the environment at runtime via transitioning from heuristic methods to data driven approaches. This system is described in five stages: (1) learned visual panoramic-target navigation, (2) single-view shape understanding, (3) next-best-view planning, (4) multiple-view shape understanding, and (5) manipulation. The agent utilizes deep-learned methods to determine a near optimal path to the object location without requiring a global position to the goal location, but instead a novel panoramic target goal image. Because the agent is a robotic mobile manipulator, the agent's initial understanding of the object can be updated using multiple views to refine the predicted shape of the object. Additionally, the system can utilize tactile and visual information for multi-modal refinement of the shape geometry. An overview of this pipeline is shown in \autoref{fig:mobilemanipulationpipeline}. 

A key idea to explore is the application of data driven techniques to enable robotic mobile manipulation. The various navigation, grasping, and shape understanding sub-tasks are approached using heuristics, or static rule-based approaches. These heuristic methods work if the underlying assumptions hold. For example, shape completion by symmetry works very well for symmetric objects, or navigation relying on odometry and lidar for localization assumes no sensor error in ideal circumstances. Unfortunately, these heuristic methods fail when utilized in situations where the underlying assumptions do not hold. In the presented framework, it employs data driven techniques making the systems more robust in comparison to heuristic approaches. This data driven approach enables this system to adapt to new situations when presented new training data. Training data can then be created at scale using simulation enabling the system to be exposed to a wide variety of conditions, making this approach flexible and useful to new robotics technologies.

\section{Procedure}
\label{sec:introduction_contributions}
This thesis presents a data driven approach for localization-free mobile manipulation of novel objects. Because the mobile manipulation pipeline follows a data-driven approach, it is generalizable to a variety of different environments. The procedure is as follows:
\begin{enumerate}
    \item The agent will receive an 8-image RGBD panoramic target goal where the target object is located. Using a learned navigation system which uses only visual information to navigate to the target location, the agent successfully navigates to the position and aligns itself with the object. This navigation pipeline allows the agent to move through an environment to arbitrary goals without localizing itself at runtime. The details of this system are addressed in \autoref{ch:learning_visual_navigation}. 
    \item Upon reaching the target location, the agent aligns itself with the object. It can utilize visual-tactile information to get a better understanding of the object. The agent can also utilize tactile sensors on its end-effector to capture information about the occluded side of the object. With the current-view and tactile information, the robotic agent utilizes a visual-tactile fusion convolutional-neural-network (CNN) to predict the geometry of the object more accurately than a single-view alone. This stage is addressed in \autoref{ch:visual_tactile_manipulation}.
    \item The agent may instead capture two views of the object. To utilize two views of the object without localizing the agent, a CNN model is needed that can leverage these two in-frame 2.5D images. The design and implementation of this network is described in \autoref{ch:two_view_shape_understanding}.
    \item The agent requires an intelligent methodology for predicting a next-best-view and navigating to it. The agent can utilize a single-view shape completion CNN to estimate the geometry of the object. This initial shape understanding is then used to predict a next-best-view of the object by considering only the parts of the predicted shape that the network was uncertain about. Using this prediction, the agent then navigates to this next-best-view by predicting the panoramic view at the new target location and reusing the visual navigation system to navigate there. Upon reaching the target location, the agent captures a second view of the object and uses the two-view shape completion architecture that uses two unregistered views to create a more accurate shape estimation. Once the shape is finalized the agent can manipulate the object. The mobile manipulation system is described in \autoref{ch:mobile_manipulation}.
    \item There is a lot of information potentially gathered by the agent while moving around the object in the form of multiple shape views. To leverage these views, a CNN model is needed to utilize an arbitrary number of in-frame 2.5D images. The use of performers, a form of attention layers, is used to refine the initial shape completion of an object with many views. The design and implementation of this architecture is described in \autoref{ch:performers}. 
\end{enumerate}

\noindent
The contributions of the work described in this thesis are:
\begin{itemize}
    \item An end-to-end system for mobile manipulation of household graspable objects utilizing novel learning algorithms
    \item An algorithm that takes an initial shape completion estimate of the manipulation target using voxel grid occupancy thresholding to plan the next-best-view
    \item An algorithm that uses a predicted panoramic goal and reuses the long-range learned image navigation system to navigate to the next-best-view locally
    \item A learned two-view shape completion method that creates a more accurate reconstruction for robotic manipulation using an initial view and a next-best-view
    \item A learned visual-tactile shape completion method that given the initial and tactile views creates a more accurate reconstruction for robotic manipulation
    \item A learned multiple-view shape completion method that can take an arbitrary number of views to refine its understanding of the object geometry
    \item A series of metrics and benchmarks that can be used to evaluate mobile manipulation systems in future work
    \item Ablation studies demonstrating the utility of system subcomponents as well as overall system performance
    \item An open-source dataset of trajectories, object point clouds, and object placements in real-world scanned environments to reproduce results
\end{itemize}

\chapter{Related Work}
\label{ch:related}

Mobile manipulation is composed of multiple subcomponents all working together to manipulate a robot's environment. This process has been studied and developed for decades, with one of the first informative work from Joshi et al. describing the process of modeling a robot with a mobile base while accounting for noise in the environment~\cite{mobilerobotjoshi1986}. This thesis describes how multiple subcomponents can come together to address mobile manipulation holistically. These subcomponents cover different subfields of robotics, including shape understanding, navigation, next-best-view planning, and manipulation. 

\section{Robotic Navigation}
\paragraph{Reinforcement learning methods for navigation} 
Previous work in visual navigation~\cite{THOR} provides a target-driven reinforcement learning framework for robotic visual navigation. Our method shares the same objective of navigating to the goal position using the goal image, the current image, and a sequence of history images. ~\cite{splmetric, habitat} trains a reinforcement learning (RL) agent to navigate in realistic cluttered environments using a PointGoal (e.g., a specific location of the goal target). They assume an idealized GPS which constantly provides the relative goal position of the target and use this information to train their agents. Both ~\cite{habitat} and ~\cite{THOR} claim that their learned policy generalizes across targets and environments. ~\cite{THOR} only evaluates their method on new targets that are several steps away from the targets that the agent is trained on, and the scene-specific layer must be retrained for the policy to work in a new environment. ~\cite{habitat} relies on an idealized GPS and the specific location of the goal and it generalizes to new environments by learning to imitate a bug-algorithm to follow the boundaries of its environment. Imagine a scenario where a person is placed into a building they have not seen before with nothing but an image of the place they need to get to. It would be unfair for us to expect this person to navigate to the target location in any efficient manner. Therefore, a robotic agent would be unable generalize to new untrained environments using vision alone. ~\cite{COMPLEX_ENVS, mirowski2018learning} evaluate reinforcement learning strategies and associated deep learning architectures. Both works presented a novel approach to navigating through structured environments and set a baseline for navigation research. The experiments in~\cite{COMPLEX_ENVS} use a synthetic 3D maze environment with a single goal which does not exhibit the complexity of real-world settings. ~\cite{mirowski2018learning} trains an agent to navigate a long-range path to the target goal using real-world Google map street views. However, they provide the agent with the coordinate of the goal rather than images. The street view also has extra information brought by signs and building characteristics. ~\cite{faust2018prm, francis2019long, chiang2019learning} present hierarchical robot navigation methods using reinforcement learning to learn local and short-range obstacle avoidance tasks. They propose sampling-based path planning algorithms as global map planners. These methods use 1D lidar sensor data and a dynamic goal position as input. ~\cite{gupta2017cognitive, khan2017memory} use value iteration networks~\cite{tamar2016value} to learn navigation strategies in simplistic synthetic simulated environments. ~\cite{mousavian2019visual} evaluates different representations for target-driven visual navigation using a semantic target and an off-the-shelf segmenter. ~\cite{bruce2017one} presents a method to navigate to a fixed goal in a known environment.

\paragraph{Supervised learning methods for navigation}
While most work in using learning methods for robotic navigation relies on deep RL because of its self-supervised convenience, supervised methods in navigation are less explored. ~\cite{richter2017safe, lind2018deep} apply CNNs to help robotic navigation but they reply on odometry and need the goal location specified. They train models to only execute simple tasks such as collision detection and position evaluation.

\paragraph{Datasets and simulators for navigation environments}
The broader area of active and embodied perception has received increased interest focusing on robotics navigation, task planning, and manipulation. New datasets for navigation environments have been created which feature fully scanned 3D homes and buildings such as the Stanford2D3DS~\cite{stanford2d3ds} and Matterport3D~\cite{Matterport3D} datasets. Additionally, a large synthetic dataset of homes is offered by both the SUNCG~\cite{suncg} and Gibson environments~\cite{GIBSON}. These new environments enable researcher to train an agent in simulation using real-world data and obtain training data much faster than would be possible in the real-world alone. Since the advent of these datasets, the MINOS~\cite{minos}, Gibson~\cite{GIBSON}, Habitat~\cite{habitat}, and AI2THOR~\cite{kolve2017ai2} simulators all offer simulation for real-world navigation. These simulators allow agents to be trained using ground truth positioning, fast rendering, and training RL agents at scale. The Gibson simulator uses PyBullet~\cite{PYBULLET} to simulate collisions with the environment as well as dynamic environment tasks. Additionally, it offers a framework, \textit{Goggles}, which takes RGB images from the simulator and uses a learned transfer model to render the image photo-realistically. 

\paragraph{End-to-end machine learning}
The term ``end-to-end machine learning'' is used for algorithms that learn purely from data with minimal bias or constraints added by human designers, besides the ones that are already inherently built-in to the learning algorithm. For example, deep reinforcement learning algorithms learning directly from raw pixels~\cite{Mnih2013,Mnih2016} and algorithms that automatically decompose tasks in hierarchies with different time scales~\cite{NIPS1992_d14220ee,vezhnevets2017feudal}. This often includes massive parallelization and distributed computation~\cite{espeholt2018impala,rudin2021learning} to fulfill the data requirement of these algorithms. Other works train robotic agents to navigate through their environment using RGBD information to determine optimal discrete steps to navigate to a visual goal~\cite{watkinsvalls2020}. 

\paragraph{Human-in-the-loop machine learning}
Learning from human feedback can take different forms depending on how human interaction is used in the learning-loop~\cite{Waytowich2018,goecks2020human}. A learning agent can be trained based on human demonstrations of a task~\cite{pomerleau1989alvinn,Argall2009,Bojarski2016}. Agents can learn from suboptimal demonstrations~\cite{brown2019extrapolating}, end goals~\cite{Rahmatizadeh2016}, or directly from successful examples instead of a reward function~\cite{eysenbach2021replacing}. Human operators can augment the human demonstrations with online interventions~\cite{Akgun2012,Akgun2012a,Saunders2017,goecks2018efficiently} or offline labeling~\cite{Ross2011,Ross2013} while still maintaining successful at the proposed task. Agents can learn the reward or cost function used by the demonstrator~\cite{Ng2000,finn2016guided} through sparse interactions in the form of evaluative feedback~\cite{knox2009interactively,MacGlashan2017,Warnell2018} or human preferences given a pair of trajectories~\cite{Christiano2017}. Additionally, agents can learn from natural language-defined goals~\cite{zhou2008inverse}. Finally, agents can learn from combining human data with reinforcement learning~\cite{rajeswaran2017learning,goecks2019integrating,reddy2019sqil}.

\section{Robotic Visual Shape Understanding}
\paragraph{Algorithmic shape understanding}
Several recent uses of tactile information to improve estimates of object geometry have focused on the use of Gaussian Process Implicit Surfaces (GPIS)~\cite{williams2007gaussian}. Several examples along this line of work include~\cite{caccamo2016active,yi2016active,bjorkman2013enhancing,dragiev2011gaussian,jamali2016active,sommer2014bimanual,mahler2015gp}. This approach can quickly incorporate additional tactile information and improve the estimate of the object's geometry local to the tactile contact or observed sensor readings. There have additionally been several works that incorporate tactile information to better fit planes of symmetry and superquadrics to observed point clouds~\cite{ilonen2014three,ilonen2013fusing,bierbaum2008robust}. These approaches work well when interacting with objects that conform to the heuristic of having clear detectable planes of symmetry or are easily modeled as superquadrics. 

\paragraph{Visual fusion reconstruction}
There has been much successful research in utilizing continuous streams of visual information like Kinect Fusion~\cite{newcombe2011kinectfusion} or SLAM~\cite{thrun2008simultaneous} to improve models of 3D objects for manipulation, an example being~\cite{krainin2010manipulator,krainin2011autonomous}. In these works, the authors develop an approach to building 3D models of unknown objects based on a depth camera observing the robot's hand while moving an object. The approach integrates both shape and appearance information into an articulated ICP approach to track the robot's manipulator and the object while improving the 3D model of the object. Similarly, another work~\cite{hermann2016eye} attaches a depth sensor to a robotic hand and plans grasps directly in the sensed voxel grid. These approaches improve their models of the object using only a single sensory modality but from multiple points in time. 

\paragraph{Visual deep-learning-based reconstruction}
Modern work in shape reconstruction has centered around pose estimation of known objects~\cite{qi2017pointnet,qi2017pointnet++}. These works utilize multi-layer-perceptrons (MLP) to encode points into a dense representation to then perform a classification task. This can be useful for scene segmentation, part identification, or pose estimation. Other work has divided the reconstruction and pose estimation into two different steps to allow for a more refined mesh prediction of a set of objects on a table~\cite{irshad2022centersnap}. Additional work with point cloud reconstruction has taken each layer of points to further refine the prediction of the object~\cite{lin2018learning}. Other shape completion systems utilize voxel representations to perform shape understanding by voxelizing the input point cloud into a fixed size voxel representation~\cite{Yang18,3DNVS}.

\paragraph{Visual-tactile shape understanding}
The idea of incorporating sensory information from vision, tactile and force sensors is not new~\cite{miller1999integration}. Despite the intuitiveness of using multi-modal data, there is still no consensus on which framework best integrates multi-modal sensory information in a way that is useful for robotic manipulation tasks. While prior work has been done to complete geometry using depth alone~\cite{dai2017complete,Dharmasiri2018}, none of these works consider tactile information. More recent works have looked at informing decision making using partial observations from tactile information~\cite{xu2022tandem}.

Varley et al. created a shape completion method using single depth images~\cite{varley2017shapecompletion_iros}. The work provides an architecture to enable robotic grasp planning via shape completion, which was accomplished using a 3D CNN. The network was trained on an open-source dataset of over 440,000 3D exemplars captured from varying viewpoints. At runtime, a 2.5D point cloud captured from a single point of view was fed into the CNN, which fills in the occluded regions of the scene, allowing grasps to be planned and executed on the completed object. The runtime of shape completion is rapid because most of the computational costs of shape completion are borne during offline training. This prior work explored how the quality of completions vary based on several factors. These include whether the object being completed existed in the training data, how many object models were used to train the network, and the ability of the network to generalize to novel objects, allowing the system to complete previously unseen objects at runtime. The completions are still limited by the training datasets and occluded views that give no clue to the unseen portions of the object. From a human perspective, this problem is often alleviated by using the sense of touch.

\paragraph{Implicit shape reconstruction}
New work in rendering unseen images using implicit representations have been a potential source for new research. Work done by Sitzmann et al. analyzed the effectiveness of generating meshes, images, and even sound from input data via implicit representations and novel use of sine as an activation function~\cite{siren}. Other work looked at generating novel views of an environment using only a single view~\cite{eslami2018neural}. Other researchers have utilized text-based mesh generation using these new implicit representations~\cite{texttomesh2022khalid}. 

\paragraph{Object datasets}
Many different object datasets exist for performing reconstruction. An early attempt at reconciling the disconnect between simulation and real is the YCB household object dataset~\cite{calli2015ycb}. This dataset consisted of over seventy purchasable real-world objects which could be used by researchers to perform experiments but also came with high resolution scanned meshes to perform simulation experiments. Additional datasets, such as Thingi10K, aim to provide a 3D printable set of meshes that can be used for a variety of mesh generation tasks~\cite{thingi10k}. The Grasp database dataset incorporated data from a variety of different graspable meshes that have well defined geometry~\cite{bohg2014data}. The ShapeNet dataset consists of labeled high resolution meshes of different objects that are useful for creating environments or visual object analysis~\cite{shapenet}. 

\section{Mobile Manipulation}
\paragraph{Autonomous mobile manipulation}
Fully autonomous mobile manipulation has long been an important goal in robotics, with particular focus on such wide-ranging applications as manufacturing, warehousing, construction, and household assistance~\cite{kalashnikov2018qt, mahler2019learning, jacobus2015automated}. Mobile manipulation encompasses a sequence of robot navigation, object detection, view planning, grasp planning, and grasp execution which makes it a challenging task. This is particularly evident when the task environment is dangerous to be explored by a human. Increasing efforts in this problem aim to map, traverse, and grasp in unknown and partially observable environments~\cite{schwarz2017nimbro,orsag2017dexterous,wang2020multi}. Active perception to set up a goal based on some current belief to achieve an action is a good model for how to potentially solve this problem~\cite{bajcsy2018revisiting}. 

Navigating accurately and efficiently in an environment is a crucial first step to achieving autonomous mobile manipulation. Traditional position and mapping focused algorithms include a Simultaneous Localization and Mapping (SLAM)~\cite{dissanayake2001solution} technique to plan a collision free path. Such techniques, while effective in mapping and localizing in an unknown environment, are sensitive to odometry errors and noise. Increasingly, reinforcement learning-based techniques have also been used to solve navigation in complex environments~\cite{THOR,francis2020long,mirowski2016learning}. However, reinforcement learning uses sparse rewards and requires extremely large amount of training episodes to achieve a good navigation model.

Mobile manipulation has been explored in a variety of contexts, including household mobile manipulation. One of the first examples of mobile manipulation is HERB~\cite{HERB} allowing a robotic agent to select grasping targets and navigate through an environment. However, it required checkerboard localization and precise sensors to plan tasks. Nevertheless, this work helped form the basis for robotic mobile manipulators, including the separation of manipulation and navigation tasks. Some works have come out more recently advancing on this initial vision that utilize global localization and point cloud reconstruction in household environments~\cite{gofetch, DOMel2017, Wu2020}. 

Indoor map exploration provides a mobile robot with many opportunities to navigate around the environment but becomes increasingly difficult when deprived of sensory information. Several authors have proposed reinforcement learning techniques for this task~\cite{THOR,anderson2018evaluation,bansal2020combining}. We share a common goal as Zhu et al.~\cite{THOR} where the task is to navigate using images. Some assume the presence of an idealized global localization system where they train the bot to reach the goal location~\cite{anderson2018evaluation, lind2018deep, richter2017safe, relomogen}. Another aims to solve the problem of traversing unknown environments by combining model-based control with learning-based perception where the task is to produce a set of waypoints leading up to the goal~\cite{bansal2020combining}. 

\paragraph{Grasp planning using visual information}
On approaching the goal, the mobile robot captures images nearby containing the object of interest. However, this image only represents the raw sensory data, which is incomplete due to the field of view and the approach angle. Grasp planning with such minimal and incomplete information is a challenging task prone to failures. Accurate shapes of objects improve grasp planning and execution success rate. Several works have proposed a deep learning approach to produce predictive mesh representations of partial views for unseen objects through either 3D convolutions or graph convolutions for robotic grasping~\cite{varley2017shape, dai2017shape, litany2018deformable, watkins2019multi}. Several geometric solutions to object 3D modeling have been proposed as well~\cite{williams2007gaussian, krainin2010manipulator, krainin2011autonomous, hermann2016eye}. 

\paragraph{Next-best-view planning}
A mobile manipulation system also needs to leverage the ability to acquire additional information by moving a vision system. To find candidate next-best-views, an algorithm to evaluate the quality of the current geometric understanding of an object is needed while considering the cost of acquiring additional views. Foundational work from Connolly did this by identifying positions of the sensor that will maximize data collected~\cite{connolly1985determination}. Much work came after Connolly et al. utilizing heuristic approaches to determine obstructions that would block future collected data~\cite{pito1995solution, Callieri2004RoboScanAA, chen2005vision, gomeznbvplanning}. A recent survey of next-best-view algorithms showed that Chen et al. had the best next-best-view performance~\cite{karaszewski2016assessment}. Another work from McGreavy et al.~\cite{mcgreavy2017next} looked at next-best-view planning using a cylindrical model to analyze the visibility of an initial object candidate and find an optimal view. Their model does not use a learned shape completion system and relies on a priori knowledge of objects. Some newer work in shape segmentation is also applicable for next-best-view work as they allow for better segmentation for unseen object geometry~\cite{xie2021unseen}. Other work in occlusion-based grasping has shown success in mapping voxels using registered views~\cite{kahn2015active}. One method to capture a next-best-view is to use an eye-in-hand camera as shown by Potthast et al.~\cite{Potthast2011NextBV}, however there are kinematic restrictions with respect to the workspace of the robot that prevent these views from being captured. Other works have presented object reconstruction under uncertainty that utilize an algorithmic approach to estimate the object's geometry while utilizing the odometry of the mobile robot~\cite{gomez2014}. 

Several mobile manipulation benchmarks exist to evaluate the performance of a system~\cite{Bostelman2016, Sereinig2020, Stuckler2012}. These benchmarks evaluate the performance of a mobile manipulator in a variety of contexts but fail to take advantage of modern simulators~\cite{GIBSON} that utilize real-world scanned data of household objects~\cite{calli2015ycb} and environments~\cite{Matterport3D, stanford2d3ds}.

\chapter{Learning Visual Navigation}
\label{ch:learning_visual_navigation}

To build a mobile manipulation system, a robot must be able to perform locomotion within its own environment. There are many ways to enable mobility in a robot, whether its driving itself via wheels, moving quadrupedal legs, flying via drone rotors, or even rolling its whole body. The goal of navigation can similarly be general. A robot may navigate through its environment to collect additional information about potential obstacles. A robot may be moving through an environment to map its surroundings. To refine the scope of this work, a solution to localization-free robotic navigation using visual information only is provided. Using visual information alone means the agent does not have to worry about odometry at runtime or perform localization computations using noisy RGBD data. The content in this chapter will explore how to build a system that can effectively navigate its own environment via training only in simulation. 

The basis for navigating through an environment is inspired by a human's ability to reason about their environment to plan navigation tasks and is fundamental to intelligent behavior. Therefore, it has been a focus of research in robotics for many years. Traditionally, robotic navigation is solved using model-based methods with an explicit focus on position inference and mapping, such as Simultaneous Localization and Mapping (SLAM)~\cite{dissanayake2001solution}. These models use path planning algorithms, such as Probabilistic Roadmaps (PRM)~\cite{kavraki1994probabilistic} and Rapidly Exploring Random Trees (RRT)~\cite{lavalle2000rapidly, kuffner2000rrt} to plan a collision-free path. These methods ignore the rich information from visual input and are overly sensitive to robot odometry and noise in sensor data. For example, a robot navigating through a room may lose track of its position due to the navigation software not properly modeling friction. 

The exploration of localization free robotic navigation will show that a history buffer of previously seen images, a panoramic target goal, and use of a custom autoencoder are all beneficial to the success rate of an agent navigating to a goal position. A robot's perception of its nearby environment can include many errors that are common to human beings. A history buffer helps prevent the robot from getting stuck in a loop of repeating actions between time steps. A panoramic target goal allows the agent to better reason about the goal location without having to worry about whether the goal image contains too little information for it. A custom autoencoder ensures that the embedding of the image contains relevant information and not features learned from an unrelated dataset. Each of these individually contribute and impact to the overall performance of the agent. Additionally, this framework can be mapped onto problems in other domains. This chapter will explore an experimental result using a similar design to build an agent to play Minecraft.

\section{Introduction}

\begin{figure}[H]
    \centering {
        \includegraphics[width=0.9\linewidth]{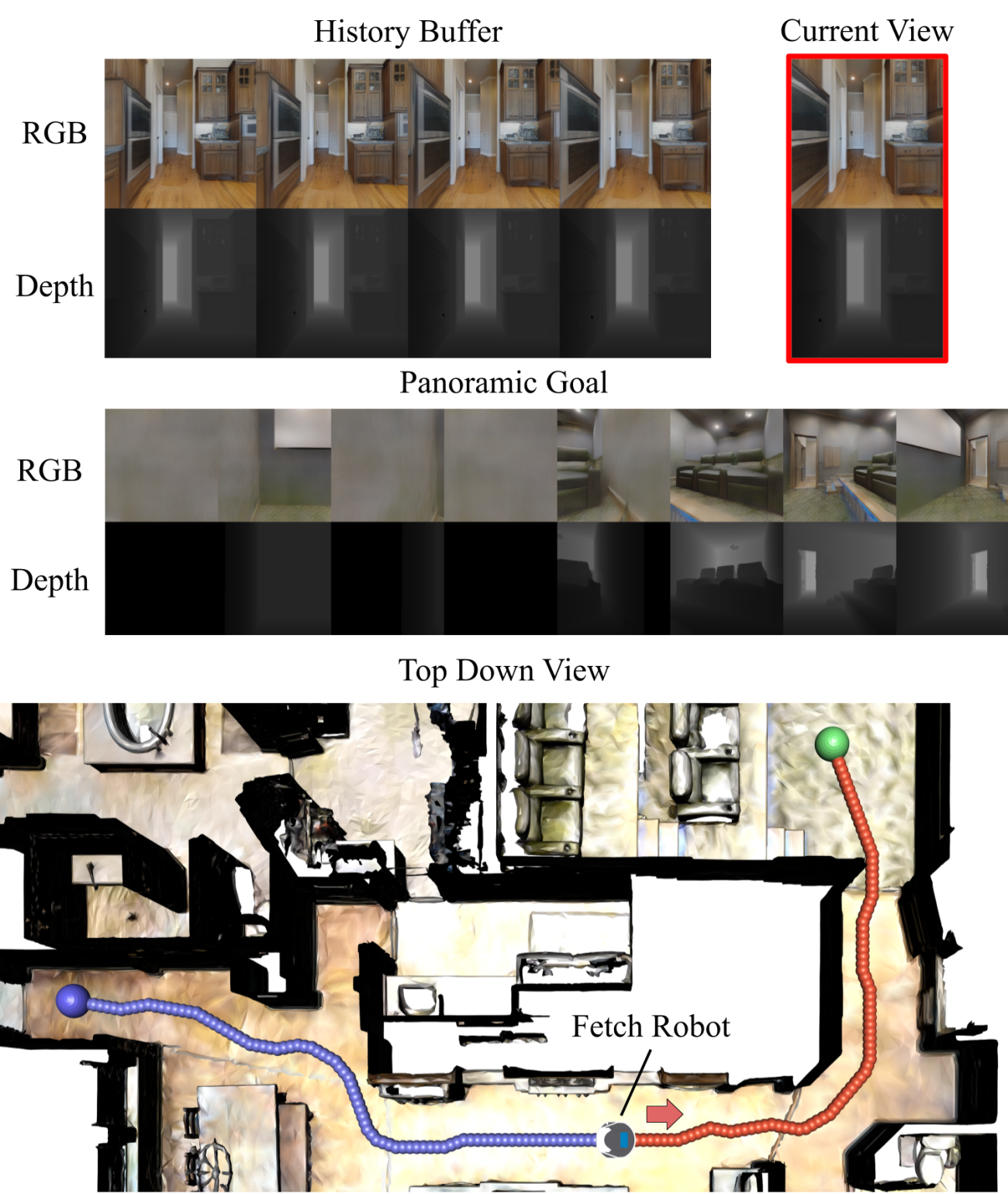}
    }
    \caption{A successful trajectory executed in \texttt{house17} from the Matterport3D dataset. The history buffer and current view are the state of the pipeline. The panoramic goal is 8 RGBD images each taken at a \ang{45} turn. The top-down view is the agent moving through the trajectory with the blue sphere as the start position and the green sphere as the goal position. Smaller blue spheres are positions that the agent has been to, and the orange spheres are the remaining positions. The images are taken at the current position of the robotic agent. }\label{fig:navigation_title_image}
\end{figure}

This work in navigation focuses on exploring supervised methods (in particular, imitation learning) to bring better performance to robotic visual navigation, while taking advantage of the current progress in robotic simulators and datasets to efficiently collect training data. 
A navigation pipeline is presented where the agent learns to navigate to unseen targets using the current RGBD view and a novel 8-image panoramic goal without using GPS, compass, map, or relative position of goals at runtime. Panoramic images are chosen as they allows for easier acquisition of training data for a policy model and they allow the agent to generalize better to unseen targets. A framework to efficiently generate expert trajectories in the Gibson~\cite{GIBSON} simulator using a 3D scan of the environment of interest is also shown. Additionally, a methodology is provided for discretizing a continuous trajectory into a series of \{\textit{forward}, \textit{right}, \textit{left}\} commands. Shown in \autoref{fig:navigation_title_image} is an example successful trajectory of this system. 

Why use imitation learning instead of a commonly used deep learning method: reinforcement learning (RL)? Model-free RL agents have performed well on many robotic tasks~\cite{kober2013reinforcement, mulling2011biomimetic, mnih2015human, andrychowicz2017hindsight}, leading researchers to rely on RL for robotic navigation tasks~\cite{THOR,francis2019long, COMPLEX_ENVS, faust2018prm}. Recent work in robotic visual navigation uses reinforcement learning which trains an agent to navigate to a goal using only the current and goal RGB images~\cite{THOR}. While reinforcement learning has the convenience of being weakly supervised, it suffers from sparse rewards in navigation, requires a substantial number of training episodes to converge, and struggles to generalize to unseen targets. The problem is further exacerbated when the navigation environment becomes large and complex (across multiple rooms and scenes with various obstacles), leading to difficult long-range path solutions. Two issues for RL approaches yet to be addressed are (1) lack of generalization capability to unseen target goals, and (2) data inefficiency, i.e., the model requires an enormous number of episodes of trial and error to converge. These problems become more pronounced in robotic visual navigation when training an agent in a complex and large environment across multiple rooms and distinctive styles of scenes and where long-range solutions are required. 

\begin{figure}[H]
    \centering {
        \includegraphics[width=0.9\linewidth]{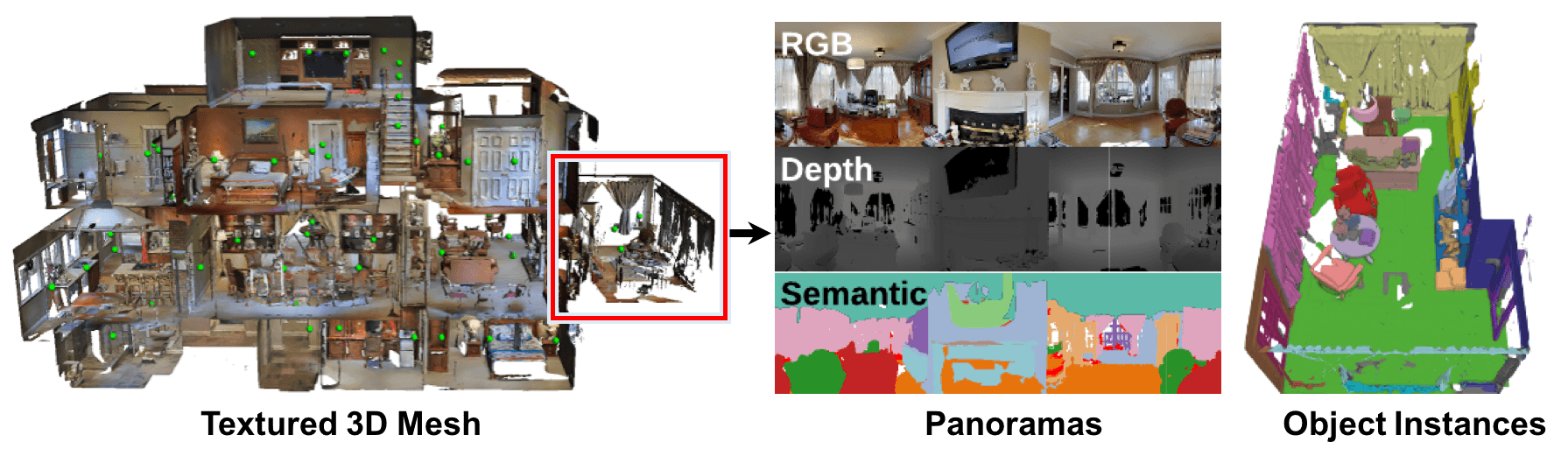}
    }
    \caption{An example environment in the Matterport3D~\cite{Matterport3D} dataset showing the mesh of the environment on the left, some of the panoramas used to generate the mesh, and the resulting semantic mesh of the environment with object categories. The dataset contains 80 individually scanned homes each semantically annotated with relevant labels. }\label{fig:matterport3d_teaser}
\end{figure}

When using an imitation learning methodology, one needs to acquire copious amounts of labeled training data to train the agent. Issues such as distributional drift, domain randomization, and gradient decay become more relevant when switching to this methodology. Thankfully, new advancements in annotated 3D maps of real-world data, such as Stanford2D3DS~\cite{stanford2d3ds} and Matterport3D~\cite{Matterport3D}, enable the collection of substantial amounts of indoor trajectory data. Training an agent in more of an environment is a way to increase robustness and decrease distributional drift. These kinds of data acquisition methods can be time-consuming or impossible using a real-world robot. It does require having a real-world scan of the environment, but companies such as Matterport~\cite{Matterport3D} or software such as RTABMap~\cite{labbe2019rtab} make indoor mapping much easier and affordable than they were previously. One such example indoor environment from the Matterport3D~\cite{Matterport3D} dataset is shown in \autoref{fig:matterport3d_teaser}.

Once the necessary environmental scans are captured, software to collect images and simulate a robot is needed. Simulation of robotic sensors and execution of their actions have existed in some form since the beginning of robotics research on a computer~\cite{beginningofrobotics}. To capture visual images along a trajectory, a modern simulator is needed that is capable of rendering images in color (RGB) and depth. RGB and depth are captured in simulation via rendering 1) meshes of these environments and 2) the robot via OpenGL. Simulators capable of collecting this data have arisen in the past few years in the form of MINOS~\cite{minos}, Gibson~\cite{GIBSON}, Habitat~\cite{habitat}, and THOR~\cite{THOR}. These systems enable simultaneous use of real and simulated environments for training, without the need for visual domain adaptation. While each has their own pros and cons, Gibson~\cite{GIBSON} is used as it features real-world and photo-realistic data generated from fully scanned 3D homes and buildings that allow for easier collection of demonstration data for supervised learning. At the time of this research, Gibson was the only simulator which allowed for us to simulate robotic grasping in addition to collecting visual information. Since the publication, Habitat~\cite{habitat} simulator from Facebook has added robotic grasping to their features as well as faster rendering times. Future work may want to utilize the features of this newer simulator to collect more data, however this system performs well despite not having collected data as quickly as other alternatives could have. 

\section{Method}

\subsection{Formulation}

The goal of this work is to enable the robot to autonomously navigate to a target position, described by a set of panoramic images taken at the goal, without providing any odometry, GPS or relative location of the target but only RGBD input from the robot's point of view. The agent is not required to achieve a particular orientation relative to the goal image at runtime. The problem is referred to as \textit{target-driven visual navigation} in the literature~\cite{THOR}, where the task objective (i.e., navigation destination) is specified as input to the model. Traditional learning-based visual navigation methods have focused on learning goal-specific models that tackle individual tasks in isolation, where the goal information is hard-coded in the neural network representations, leading to poor generalization to unseen\,/\,unexplored targets. Target-driven approaches learn to navigate to new targets without re-training, using a single navigation pipeline. 

The navigation pipeline, denoted as $\Pi$, takes as input the observation of the current state $s_i$ at time step $i$, the target information $g$, and outputs an action $a_i \in \{forward, right, left, done\}$.

\begin{equation}
    a_i = \Pi \left( s_i, g \right)
\end{equation}

The \textit{left}\,/\,\textit{right} action indicates turning the agent in place left\,/\,right 10 degrees and the \textit{forward} action moves the agent 0.1$m$ ahead. The unknown transition model $\Gamma$ of the environment updates the state, denoted by $s_{i+1} = \Gamma(s_i, a_i)$, when an action $a_i$ is executed. The objective is that given any goal $g$ in the map, a maximum number of steps $T$, and a success threshold $\zeta$, the navigation pipeline $\Pi$ can generate a sequence of actions $\{a_i\}, i\in[t]$, which satisfies (1) $t < T$, (2) $a_t = done$, (3) the final location of the robot is within $\zeta$ meters of the target location, and (4) the length of the path should be as short as possible. The navigation pipeline is fully automated as it learns to stop at the goal and does not require human intervention.

The state $s_i$ is the current RGBD visual observation and a history buffer of 4 concatenated past RGBD images, both of which are from the agent's viewpoint. The goal information $g$ is a set of 8 panoramic RGBD images. An example of the state and the goal information is shown in \autoref{fig:navigation_title_image}.

\subsection{Navigation Pipeline}

The navigation pipeline $\Pi$ consists of three separately trained models using neural networks, the autoencoder model $A$, the policy model $E$ and the goal checking model $G$. 

\begin{figure}[ht!]
    \centering {
        \includegraphics[width=\linewidth]{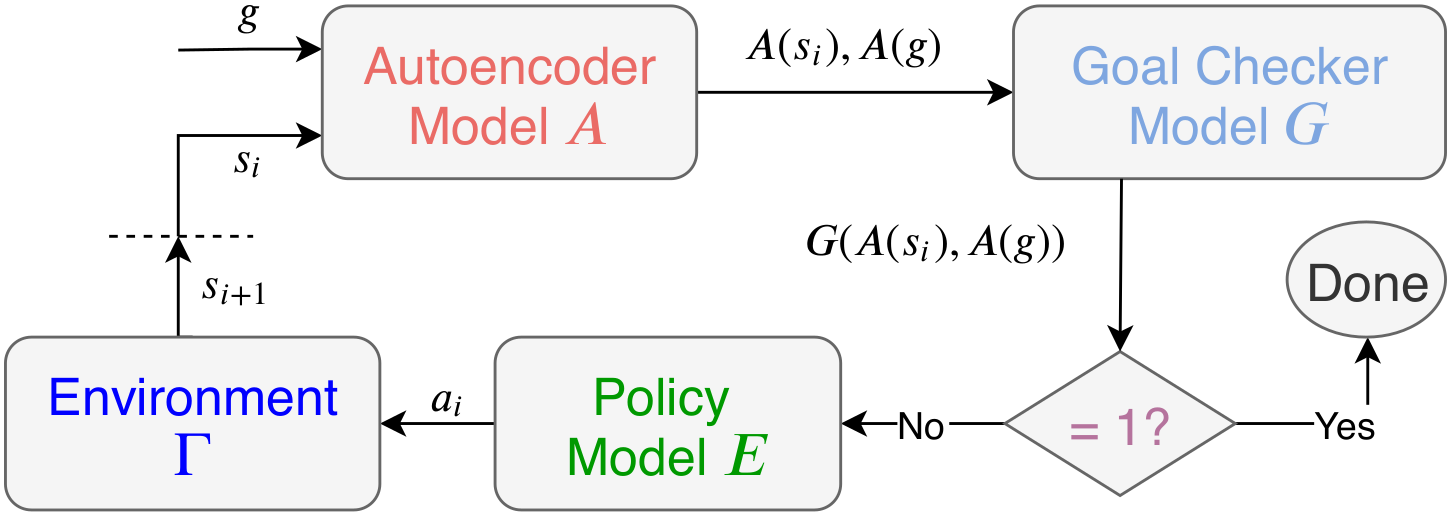}
    }
    \caption{Overview of the navigation pipeline. The flow starts when a new navigation task is received. The agent encodes each new image captured from the environment using a trained Autoencoder $A$. The encoded images are passed into the Goal Checker $G$ to determine if the agent is $done$. If the $G$ is not $done$, the agent then passes the current image, the previous set of images, and the panoramic target goal into a policy model to determine the next action to take. This action then updates the environment, and the process repeats until $done$.}
    \label{fig:overview}
\end{figure}

The autoencoder model generates latent representations (i.e., embeddings) for both the state $s_i$ and the goal $g$, denoted $A(s_i)$ and $A(g)$. The policy model takes two inputs, the embeddings of the current state and the embeddings of the target, and produces a probability distribution over three actions, $\displaystyle a_i \in \{\textrm{\textit{forward}, \textit{right}, \textit{left}}\} \sim E\left(A(s_i), A(g) \right))$. It then picks the action with highest probability from this distribution. The policy model is responsible for leading the agent towards the goal with as little exploration as possible. The goal checking model is a binary function which takes the same input as the policy model, and decides if the agent has reached the target or not, denoted by $G(A(s_i), A(g)) \in \{1, 0\}$, where 1 corresponds to \textit{done} and 0 corresponds to \textit{not done}. An overview of the navigation pipeline is shown in \autoref{fig:overview}.

\subsubsection{Autoencoder Model}
Because the input into the neural network models is RGBD images, the training is more efficient if embeddings are used instead of the raw input. Instead of extracting features from an intermediate layer of a pre-trained classifier such as ResNet-50~\cite{THOR, he2016deep}, an autoencoder is trained from images captured from the same environment.

\begin{figure}[ht!]
    \centering {
        \includegraphics[width=\linewidth]{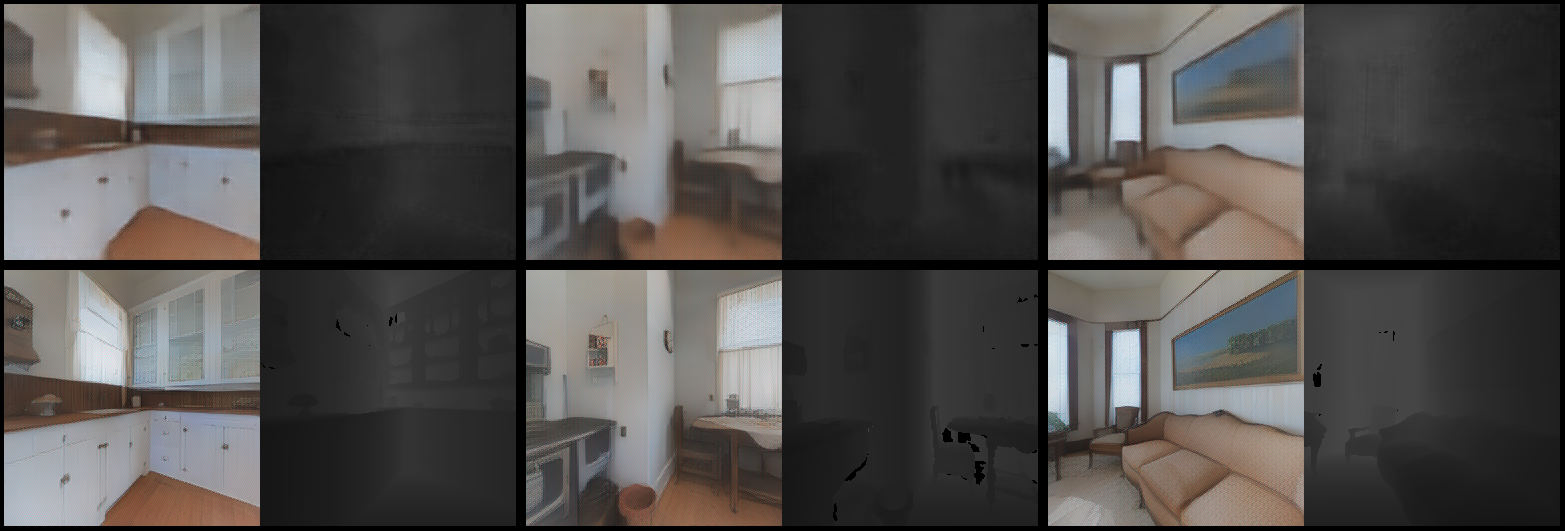}
    }
    \caption{An example of reconstructed images from the autoencoder model trained in the \texttt{house2} environment. The top row is three predicted output images (RGB image appended by depth image); the bottom row is the original images.}\label{fig:autoencoder_imgs}
\end{figure}

\begin{figure}
\centering
\begin{subfigure}[b]{\textwidth}
   \includegraphics[width=1\linewidth]{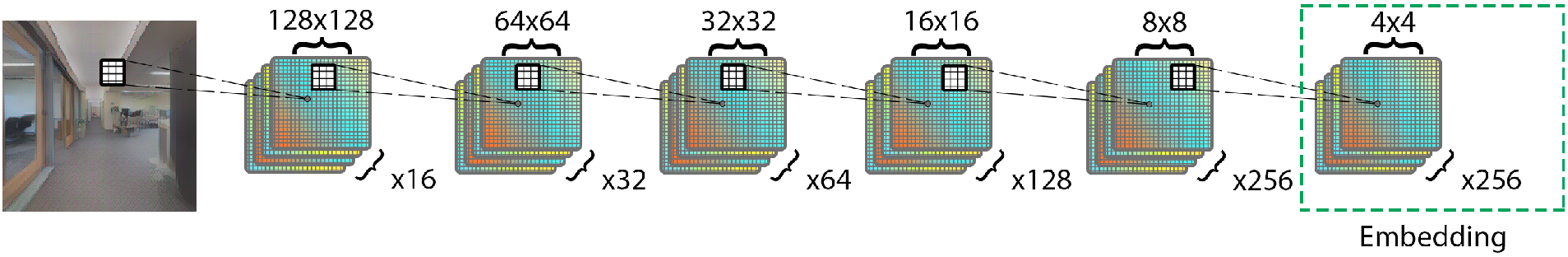}
   \caption{}
   \label{net:autoencoder} 
\end{subfigure}


\begin{subfigure}[b]{\textwidth}
   \includegraphics[width=1\linewidth]{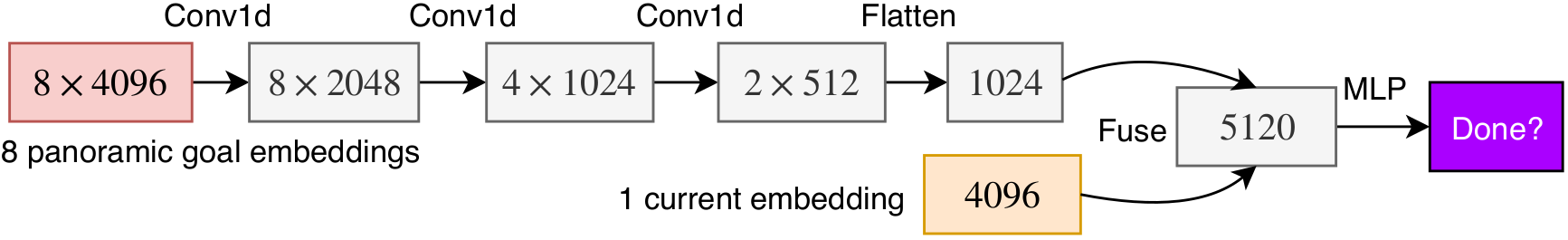}
   \caption{}
   \label{net:goal} 
\end{subfigure}

\caption{(a) \textbf{Encoder architecture of the autoencoder} Progression through each layer consists of a convolution with a stride of 2 followed by batch normalization and ReLU activation. (b) \textbf{Goal checking model architecture} \texttt{Conv1D} is the 1D convolution operation.}
\end{figure}

Like RedNet~\cite{jiang2018rednet}, the autoencoder network is based on a 6-layer CNN with batch normalization on every layer during training. The reconstruction half of the network is made up of an additional 6 transposed convolutional layers with batch normalization applied before each transposed convolution. Rectified linear unit (ReLU) is used as the activation function. The Adam optimizer~\cite{kingma2014adam} is used to minimize the mean squared error between the reconstructed and the original images. The autoencoder can compress a $256\times256\times4$ RGBD image into the $4096D$ latent space ($\times 64$ space savings). It is then used to encode each image of the state and each image of the panoramic goal. A detailed topology of the encoder section is pictured in \autoref{net:autoencoder}. An example of the autoencoder performance is shown in \autoref{fig:autoencoder_imgs}.

\subsubsection{Policy Model}
The policy model takes as input the embeddings of stacked observations and the panoramic goal images to generate the next action $a_i \in \{\textrm{\textit{forward}, \textit{right}, \textit{left}}\}$.

\begin{figure}[ht!]
    \centering {
        \includegraphics[width=\linewidth]{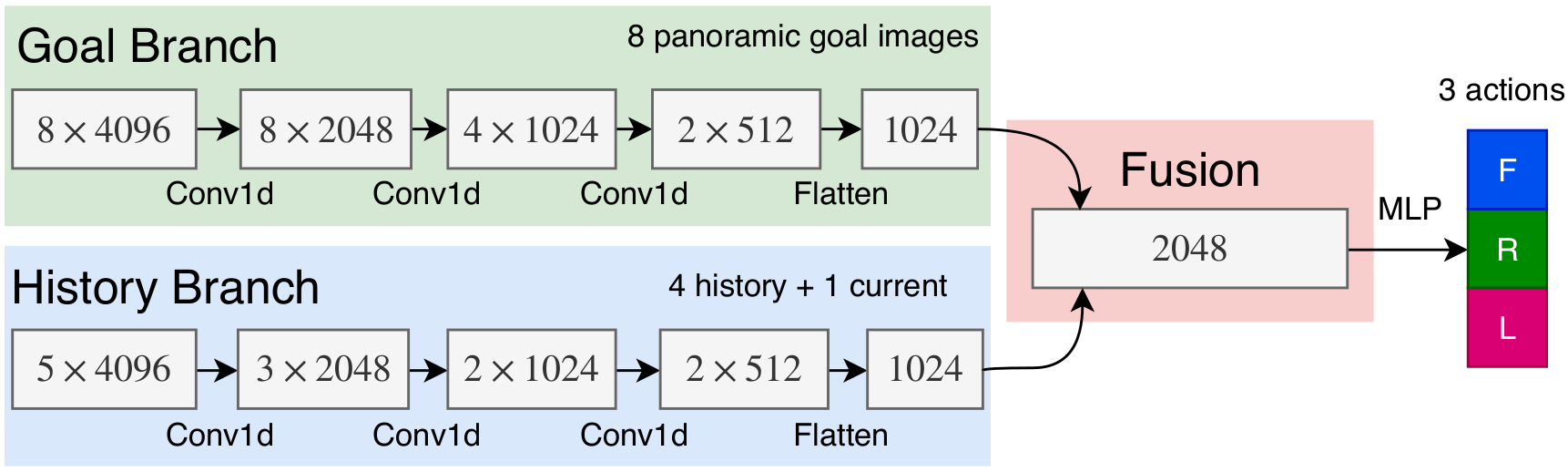}
    }
    \caption{The layout of the policy model network which takes the embeddings of a sequence of past observations, a set of panoramic goal images, and the current observation of the robot. Numbers inside each rectangle correspond to the input dimensions of each layer. \texttt{Conv1D} represent 1D convolution and \texttt{Flatten} represents the flatten operator on a matrix. The SoftMax is taken of the final logits from MLP to generate a distribution over 3 actions.}\label{fig:policy_model}
\end{figure}

The policy model is a fully-connected multilayer perceptron (MLP) as shown in \autoref{fig:policy_model}. Also evaluated is the performance of a variety of other deep learning architectures including convolution along the temporal dimension and long short-term memory (LSTM)~\cite{hochreiter1997long}, with different numbers of past images in the state and a different number of panoramic goal images, but the MLP architecture using 4 previous images and 1 current image outperforms other methods. Its larger number of parameters increases its ability to model complex functions. 

The embeddings of the state and the panoramic goal are first concatenated to form a $13 \times 4096$ matrix and then progress through 3 fully-connected layers followed by batch normalization (during training) and ReLU activation after each layer, to generate a $16D$ vector. The $16D$ vector passes through the last fully-connected layer to generate three logits. A SoftMax activation then outputs a distribution over three actions \{\textit{forward}, \textit{right}, \textit{left}\}. This SoftMax activation means the policy model always chooses the most probable choice of action. The Adam optimizer is used on the cross-entropy loss for back propagation. At testing, the action with highest probability is chosen deterministically.

As an ablation study, the performance of several deep learning architectures is evaluated on a subset of the \texttt{area1} environment. For this subarea, the model which performs best is picked from all experiments described in \autoref{sec:experiments}. The models to be evaluated include 1) fully-connected network with a 5-image history buffer, 2) fully-connected network with 5-image history buffer keeping every third image, 3) LSTM with a 15-image history buffer, 4) LSTM with a 25-image history buffer, 5) dual branch temporal convolutional network with a 15-image history buffer, 6) dual branch temporal convolutional network with a 25-image history buffer, 7) a fully connected single-image single-target model, and 8) a fully connected single-image panoramic-target model. A fully connected model with more images on the history buffer is not assessed because the VRAM on a 1080Ti does not provide sufficient space to train a model of this size. The LSTM and convolutional architectures are evaluated in lieu of a larger fully connected model as they are more scalable. The proposed model, using a 5-image history buffer, outperforms all the above architectures on the testing subset of \texttt{area1}. 

\autoref{fig:policy_model} illustrates the network architecture of the dual branch temporal convolutional network. There are two separate branches that deal with the state space (current observations stacked with history observations) and goal embeddings, respectively. The model applies convolution along the temporal dimension of the sequence of images. This is like a 3D convolution~\cite{tran2015learning} on image sequences but uses 1D convolution because of the latent space input embeddings. Compared to a fully connected multi-layer perceptron (MLP), it has the advantages of 1) being much more lightweight due to parameter-sharing in the convolution process, 2) scalable to longer history buffer, and 3) is better at capturing spatiotemporal features from the data. LSTM is also known to perform well on sequential data and has been widely used in the computer vision community for action recognition and video preprocessing, but it consumes a large amount of time and resources to train and does not perform well as shown in \autoref{sec:experiments}.

The $1024D$ output vectors from both branches are then fused to a $2048D$ vector. It passes through 2 fully connected layers and the SoftMax activation function to generate a distribution over three actions. The policy model chooses the one with the highest probability for the agent to take.

During the training phase, images (and in the case of the history policy, image histories) and commands were randomly selected from a training set of 700 trajectories. The cross-entropy loss was computed between the output of the policy network and the ground-truth commands. 

\subsubsection{Goal Checking Model}
The goal checking model takes in the embeddings of the current observation concatenated with the panoramic goal images and predicts whether the agent is at the target position, as shown in \autoref{net:goal}. 

This model is created in response to an optimization on the original architecture which had the policy model output a \textit{done} action when the robotic agent arrives at a goal position. The policy training data is too sparse for the agent to effectively learn identifying a goal location because it only has one positive example of \textit{done} at the end of each trajectory. All the other steps are negative examples for \textit{not done}. There is a significant imbalance in the number of positive and negative examples. In addition, at runtime the robot is likely to arrive at the target position from a different viewpoint than those in the panoramic goal images, but during training the policy model receives a view that is one of the panoramic goal images. An additional binary classifier is implemented to identify whether the agent has arrived at the goal location. When this model predicts a \textit{done} action the navigation pipeline terminates.

The goal checking model is a dual-branch network with 1D convolution over the panoramic goal branch. The $1024D$ vector from the goal-branch is then concatenated with the current embedding branch to form a $5120D$ fused vector, which then passes through an MLP with a hidden layer of 512 units to output the probability of the goal being reached. Weights are then updated using an Adam optimizer on the cross-entropy loss. While using the learned goal checker at runtime, to reduce noise, the agent does a \ang{360} rotation when its belief of reaching the goal is over 0.99. It calls the learned goal checker after each \ang{10} turn. The agent outputs \textit{done} only if the average probability is over 0.9.

\section{Experiments}\label{sec:experiments}
\begin{figure*}[t]
    \centering {
        \includegraphics[width=0.95\textwidth]{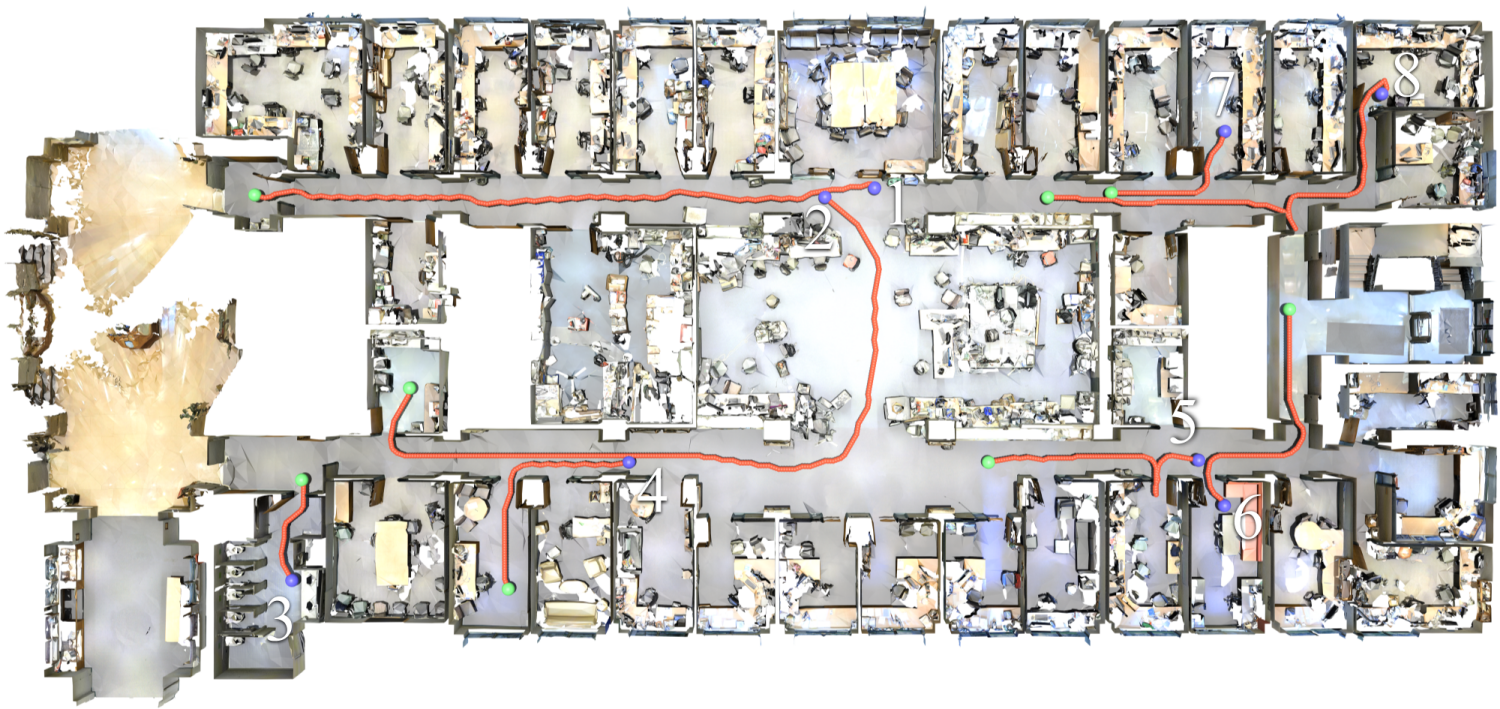}
    }
    \caption{Eight randomly selected non-overlapping successful trajectories in \texttt{area1}. Blue dots are start positions and green dots are goal positions. Trajectory $5$ and $8$ show recovery behavior which leads to a successful trajectory.} \label{fig:area1_demo}
\end{figure*}

The navigation pipeline is evaluated in 2 environments selected from the Stanford2D3DS dataset and 3 environments selected from the Matterport3D dataset. Metadata (including number of rooms and area) are shown in \autoref{table:results}. The Gibson simulator is used with a Fetch~\cite{fetch} robot and focus on how the navigation pipeline generalizes to unseen targets under the same trained environment. All experiments are conducted using an NVIDIA 1080Ti GPU. Examples of planned paths, recovery behavior, and experimental environments can be found in the attached video.

\subsection{Network Training Setup}
For each submodule (autoencoder, policy, and goal checker) in the navigation pipeline, the corresponding dataset needs to be generated for training and testing using the colored meshes loaded in the Gibson simulator. The model with least loss (autoencoder) or highest accuracy (policy, goal checker) on the test set is selected. Unless otherwise specified, a learning rate of $0.001$ is used.

\subsubsection{2D Environment Map}
\begin{figure}[t]
    \centering {
        \includegraphics[width=\textwidth]{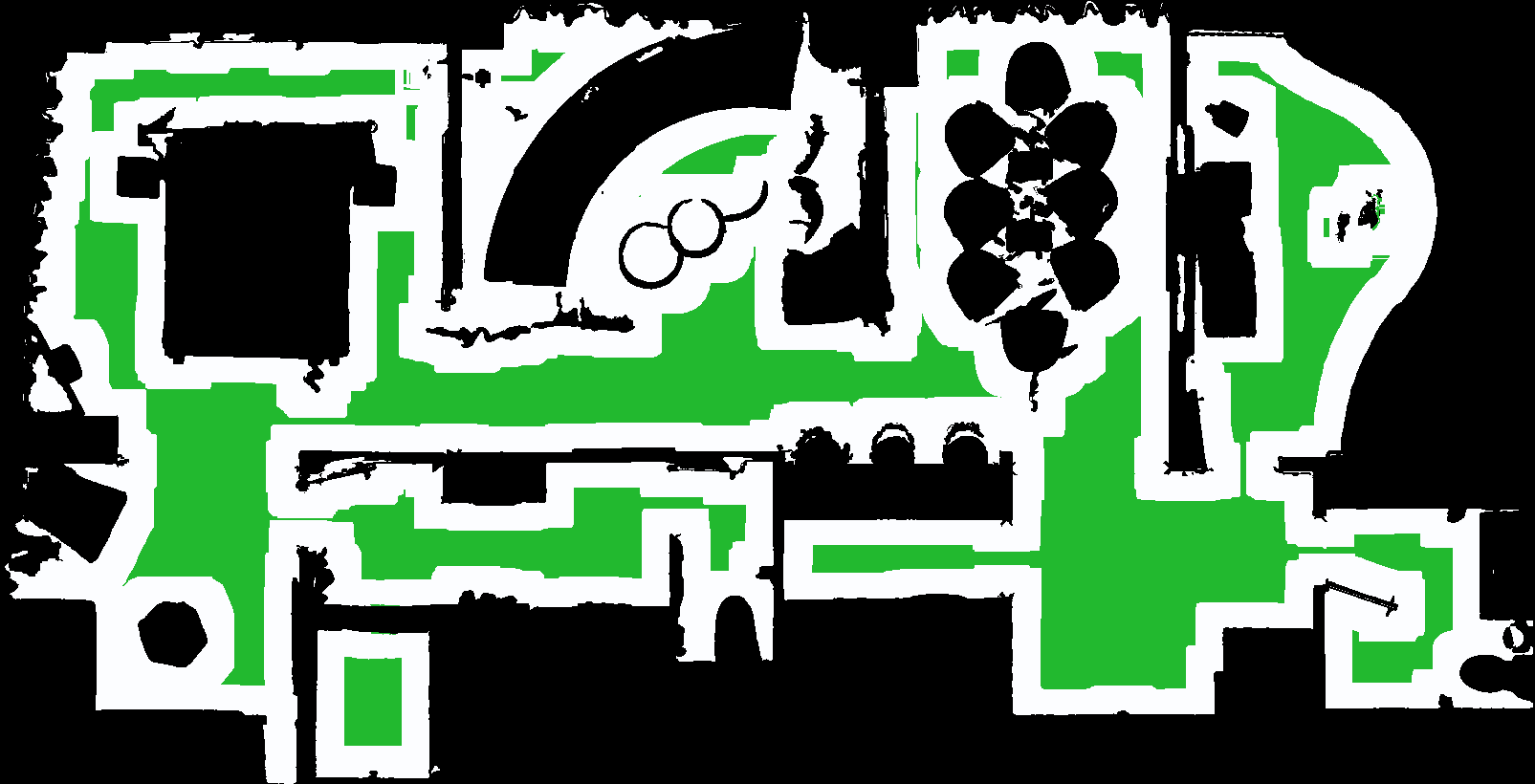}
    }
    \caption{2D map of the \texttt{house1} environment in the Matterport3D environment. Each pixel is 0.01m x 0.01m. The collision-free locations for the robot are shown in green, obstacles are shown in black and in-collision positions are shown in white. Some regions are non-contiguous and if trajectories were too short within their subregion, they were not used for navigation planning. }\label{fig:house1_map}
\end{figure}

Each environment is represented as a 3D mesh in both the Matterport3D and Stanford 2D3DS datasets. To plan expert trajectories in the simulator, a 2D map of the environment which contains obstacles the robot would encounter during execution of a trajectory is needed. The environment is discretized into a grid of $n x m$ cells where $n$ is the number of cells in the y-axis and $m$ is the number of cells in the x-axis. Each cell is a fixed dimension that can be changed depending on how granular the positions for the robot need to be for generating data. A cell resolution of $1cm\times 1cm$ was used during testing. At each position, the robot is modeled as a bounding box to check for a collision with the mesh of the environment. For a Fetch robot the collision bounding box is $0.6m \times 0.6m \times 1.6m$ where $0.6m$ is the diameter of the robot and $1.6m$ is the height of the robot. After checking a cell for a collision with the environment, an occupancy map of valid positions for the robot to navigate to is obtained. Due to the uneven nature of tthe target embeddinghe floor of the environment, collisions with faces of the environment mesh close to the ground level are ignored, which was $0.05m$ for each environment. Five maps of the scenes house1, house2, house17 are generated from Matterport3D and area1 and area2 from Stanford 2D3DS datasets. An example of the occupancy map for house1 is shown in \autoref{fig:house1_map}. With these positions calculated, the expert trajectories could then be produced. 

\subsubsection{Autoencoder}
For each environment's autoencoder dataset, 120$K$ collision-free locations are randomly sampled from the generated map and capture RGBD images at those locations in the simulator. A 0.9\,/\,0.1 train\,/\,test split is used. The autoencoder model is trained for 200 epochs which on average takes 12 GPU hours for each environment. 

\subsubsection{Policy} Using Dijkstra’s algorithm with costmap optimization, collision-free expert trajectories can be generated. To map the trajectory into to a sequence of \{${forward, right, left}$\} commands, each waypoint position is rendered. A novel discretization strategy used to turn waypoints into discretized commands. The agent looks ahead 25 steps to determine whether to turn left or right. The robot turns left or right whether the look-ahead position is turned at an offset greater than \ang{20} and outputs turns in increments of \ang{10}. Using only the next position would result in the robot constantly recalculating its direction and the robot would turn at every step to face a new direction. Forward commands are given if the robot is farther from the next position than $0.1m$, and if so, moves the robot forward in increments of $0.1m$ until under the threshold. The algorithm for this process is shown in Algorithm~\ref{alg:nav_gen_pseudocode}.

\begin{figure}[ht!]
    \begin{algorithm}[H]
    \caption{Trajectory Discretization}\label{alg:nav_gen_pseudocode}
    \begin{algorithmic}
        \STATE trajectories $T$ = \{ \[pose_{1}\], \dots , \[pose_{N}\] \} ;
        \STATE start\_pose = T[0];\\
        \STATE goal\_pose = T[0];\\
        \STATE angle\_threshold = $20^{\circ}$;\\
        \STATE commands = [];\\
        \STATE distance\_step = $0.1m$; \\

        \FOR{$t_i, t_{i+1}, t_{i+25}$ in $T$}
            \STATE heading\_direction = $t_i$ - $t_{i+25}$; \\
            \STATE current\_robot\_angle = start\_pose.yaw; \\
            \IF{current\_robot\_angle - heading\_direction $>$ angle\_threshold}
                \STATE command = $LEFT$ if angle\_difference $>$ 0 else $RIGHT$; \\
                \STATE commands.append($command$ for every 10 degree difference); \\
                \STATE final\_angle = heading\_direction; \\
            \ENDIF
            
            \STATE unit\_pose\_increment = [cos(final\_angle), sign(final\_angle), 0] * distance\_step; \\
            \STATE distance\_to\_goal = position\_distance(current\_position, next\_position); \\
            
            \IF{distance\_to\_goal $>$ distance\_tolerance}
            
                \STATE commands.append($FORWARD$ for every $distance\_step$ in $distance\_to\_goal$; \\
            \ENDIF
            
            \RETURN commands, poses;\\   
        \ENDFOR
    \end{algorithmic}
    \end{algorithm}
    
    \caption{An algorithm for converting a sequence of poses from a path planning algorithm into a sequence of discrete commands to be executed by the agent. The values used in this script, such as $distance\_step$ are the same as what were used in the training data. This method uses a look ahead algorithm to check several steps ahead when picking a correct orientation. During testing, the agent would swerve too frequently if the look-ahead position were not sufficiently far in the future. }
\end{figure}

Once these commands are generated, the trajectory is executed in the Gibson simulator capturing the RGBD view along every step. This results in a large, supervised learning dataset of varying trajectory lengths that could be used for training the agent to learn the policy. The trained autoencoder is used to generate embeddings of RGBD images at each step in the expert trajectories. The average number of steps per trajectory varies from 41 (\texttt{house2}) to 332 (\texttt{area1}). Each training\,/\,testing example is constructed by taking the embeddings from past 4 steps concatenated with the embedding at the current step. Because larger environments tend to have longer trajectories, 3000 trajectories are generated for \texttt{area1} and \texttt{area2}, 5000 trajectories for \texttt{house17}, and 7000 trajectories for \texttt{house1} and \texttt{house2} to keep the total number of individual steps the same. $80\%$ of the trajectories are used for training and the rest for testing. The policy network is trained for 200 epochs which takes around 90 GPU hours. The average accuracy for the policy model is 0.91.

\subsubsection{Goal Checker}
For each environment, positive training examples are collected by randomly sampling 150$K$ positions in the environment for the panoramic goal images and then sample another position within a 0.1$m$ radius for the current image. Negative examples are collected by randomly sampling 150$K$ positions for the panoramic goal images and then sample another position at least $1m$ away from the current image. A 0.9\,/\,0.1 train\,/\,test split is used. The network is trained for 300 epochs for on average 36 GPU hours. The average training accuracy is 0.95 across all environments.

\subsubsection{Training / Testing Split for Trajectory Data}
Validation of the navigation policy model was done using a test dataset containing 400 trajectories consisting of varying numbers of commands/images captured from the trajectory execution rendering. Upon every epoch during training for both the autoencoder and policy networks, each network is evaluated using a test dataset of 20\% of the overall dataset for each to make sure the network was still learning. 

The training and test data both contained varied length episodes, ranging from 200 to 300 timesteps per episode. Each timestep corresponds with a single action, which can be a turn in either direction, a forward movement, or a terminus. Each of the three possible movements (left, right, forward) were controlled by discretely moving the robot in a particular direction, which in the configuration of PyBullet produced deterministic results. \autoref{fig:area1_demo} shows sample trajectories used in the training set. 

See \autoref{table:environment_statistics} for statistics about each of the datasets generated. Each cell of the table corresponds to the number of training elements captured for each of the environments and datasets used. The Goal Checker had 9 images captured for every training example, 1 for the current view and 8 for the panoramic goal. The autoencoder and navigation policy had one image captured per position. Overall, $3,026,827$ images were used for training.

\begin{table}[ht!]
    \centering
    \resizebox{1.0\linewidth}{!}{
    \begin{tabular}{ccccc}
        \toprule
        \textbf{Environment} & \textbf{Autoencoder} & \textbf{Goal Checker} & \textbf{$\#$ Trajectories}  & \textbf{Navigation Policy} \\
        \midrule
        area1 & 107668 & 268971 & 2998 & 995720 \\
        area2 & 119532 & 239305 & 2998 & 695515 \\
        house1 & 113780 & 233341 & 5582 & 299612 \\
        house2 & 112095 & 232583 & 5599 & 189789 \\
        house17 & 119852 & 239150 & 5600 & 846191 \\
        \bottomrule
    \end{tabular}}
        \caption{Each environment contained different numbers of images for each environment based on the size of the environment and how many valid positions were available for the robot. Each environment's dataset took 100GB of storage, totaling 500GB of data. }
    \label{table:environment_statistics}
\end{table}

\subsection{Comparison Methods}
The proposed navigation pipeline is compared with an RGBD SLAM~\cite{labbe2019rtab} approach and the target-driven deep RL method from~\cite{THOR}. The methods to be examined are described below.

\begin{enumerate}[label=\alph*)]
    \item \textbf{SLAM} is the Real-Time Appearance-Based Mapping (RTABMap) library provided by~\cite{labbe2019rtab}, which is an RGBD Graph-Based SLAM approach based on an incremental appearance-based loop closure detector. The robot is not given the map beforehand in the implementation and builds it up as it moves along.
    \item \textbf{Siamese Actor-Critic (SAC)} is the method proposed by~\cite{THOR}. A scene specific layer is kept for each environment. The network is then trained for each environment to generalize to unseen targets within the same house. A goal-reaching reward of $10.0$ is provided upon task completion and a small penalty of $-0.01$ at each time step. The network is trained on 100 targets with a maximum step size of 10000 for each episode. Each environment is given a budget of 20$M$ frames (steps).
    \item \textbf{Navigation Pipeline (GPS)} is a variant of the proposed navigation pipeline. Instead of using the learned goal checking model, it uses the GPS information and the provided goal coordinate to check whether the agent is at the goal.
    \item \textbf{Navigation Pipeline (no GPS)} is the proposed navigation pipeline with the learned goal checker, without GPS, odometry or goal coordinate provided.
\end{enumerate}

a), b) and c) assume the agent has an idealized GPS and is provided with the \textit{static} goal coordinate as in~\cite{habitat}. As a result, the agent can compute the relative position of the target at each time step and can use this information to check if the goal has been reached.

\subsection{Evaluation Criteria}
The performance of the navigation pipeline is evaluated using 400 randomly sampled start-goal pairs for which a valid path exists. The start and goal locations have never appeared in the training examples. The agent is started at the starting position and provide it with 8 panoramic goal images taken at the goal location. The objective is to navigate to the goal position (no requirements on the robot's final orientation) autonomously with the shortest path possible using only visual input. The success tolerance is a $0.5m$ radius within the target position. Unlike previous works which do not penalize collision through training and allow collision at runtime~\cite{THOR, habitat}, physics are simulated, and the trial is considered a failure when collision occurs.

Like many previous works on navigation benchmarks~\cite{THOR, habitat, splmetric, mishkin2019benchmarking}, three evaluation metrics are used:
\begin{enumerate}
    \item \textbf{Success Rate} is the number of successful trials over the total number of trials.
    \item \textbf{Success Weighted by Path Length} ($SPL$)~\cite{splmetric} metric is shown in Formula~\ref{spl}, where $l_i$ is the length of the shortest path between start and goal position, $p_i$ is the length of the observed path taken by an agent, and $S_i$ is a binary indicator of success in trial $i$. This metric weighs each success by the quality of path and thus is always $\leq$ \textit{Success Rate}.
    \begin{equation} \label{spl}
        SPL = \frac{1}{N} \sum_{i=1}^{N} S_i \frac{l_i}{\max (p_i, l_i)}
    \end{equation}
    \item \textbf{Observed over Optimal Ratio} ($OOR$) measures the average ratio of observed path length over optimal path length for successful trials.
\end{enumerate}

\subsection{Navigation Results}
The proposed navigation pipeline significantly outperforms RGBD SLAM and the state-of-the-art deep RL method in terms of path quality and success rate, as shown in \autoref{table:results}. See \autoref{fig:area1_demo} for several example trajectories generated by the method in the \texttt{area1} environment.

SLAM~\cite{labbe2019rtab} struggles to localize itself using RGBD alone, due to the high complexity of the testing environments. It succeeds only when the start position is close to the goal position. SAC~\cite{THOR} performs much worse than the navigation pipeline due to the sparse reward and limited number of training frames. In THOR~\cite{THOR}, each environment is a single room, and the researchers use synthetic images, but the environments used here can be up to 1031$m^2$ with 40 rooms with real-world images. These environments have higher complexity with more obstacles and the entrances to the rooms can be extremely narrow resulting in a difficult solution. SAC needs millions of frames to converge in the environments provided, which is not practical. THOR~\cite{THOR} claims they can generalize to new targets by evaluating only on 10 targets that are several steps away from the training targets. In the experimentation, the targets can be anywhere on the map. Most of the successes for SAC is when the target location happens to be in a same room as the start location. The proposed method also requires much fewer simulation steps\,/\,training frames ($\sim700K$ compared to 20$M$) and less training time (90 GPU hours compared to 300 GPU hours). While the proposed approach does not appear to be saturated in learning a policy over the largest environment, \textit{area1}, it is unlikely that given current methods of generating data and training that the model would generalize further to larger environments. These models also struggle to generalize between environments when trained over multiple environment datasets.

The proposed method with no GPS achieves similar performance to the variant with GPS. As an ablation study, instead of having a separate goal checking model, a \textit{done} action is generated directly from the policy model. Using a separate goal checking model increases the success rate by $0.2 \sim 0.5$. In the cases where the policy model incorrectly identifies \textit{done}, it either outputs \textit{done} prematurely or passes the goal without terminating. The amount of training data is intentionally kept the same across all environments to evaluate how the performance changes with the complexity of the environment. When the number of rooms is over 30, the proposed navigation pipeline starts to struggle to get to the goal. Despite the reduction in performance due to environmental complexity, the method performs on average $0.556$ and $0.442$ better in success rate than SLAM and SAC, respectively. Given that high navigation accuracy is achieved on smaller environments, the performance in \texttt{area1} and \texttt{area2} will go up if trained on more expert trajectories. A future direction is to analyze the amount of training data needed for a given environmental complexity.

\begin{table}[H]
    \centering {
        \small
        \begin{tabular}{cccccccc}
            \toprule
            \textbf{Environment} & \textbf{Model} & \footnotesize{\textit{Success Rate}} & \textit{SPL} & \textit{OOR} \\
            \midrule
            
            \multirow{4}{*}{\makecell{house2 \\ (66$m^2$, 6 rooms)}}
            & SLAM         & 0.7575 & 0.5208 & 2.726 \\
            & SAC          & 0.8400 & 0.5620 & 3.538 \\
            & Proposed (GPS)        & \textbf{0.9950} & \textbf{0.9810} & 1.066 \\
            & Proposed (no GPS)        & 0.9875 & 0.9724 & \textbf{1.053} \\
            \midrule
            
            \multirow{4}{*}{\makecell{house1 \\ (89$m^2$, 10 rooms)}} 
            & SLAM         & 0.3575 & 0.2380 & 2.809 \\
            & SAC          & 0.5200 & 0.3648 & 2.961 \\
            & Proposed (GPS)        & \textbf{0.9975} & \textbf{0.9811} & \textbf{1.064} \\
            & Proposed (no GPS)        & 0.9225 & 0.8748 & 1.252 \\
            \midrule
            
            \multirow{4}{*}{\makecell{house17 \\ (220$m^2$, 14 rooms)}} 
            & SLAM      & 0.0900 & 0.0642 & 2.354 \\
            & SAC          & 0.2800 & 0.1281 & 5.919 \\
            & Proposed (GPS)        & \textbf{0.9800} & \textbf{0.7853} & \textbf{2.020} \\
            & Proposed (no GPS)        & 0.9150 & 0.7179 & 2.389 \\
            \midrule
            
            \multirow{4}{*}{\makecell{area2 \\ (1031$m^2$, 31 rooms)}} 
            & SLAM         & 0.0700 & 0.0500 & 4.447 \\
            & SAC         & 0.1700 & 0.1073 & 3.997\\
            & Proposed (GPS)        & \textbf{0.7250} & \textbf{0.5536} & \textbf{2.504} \\
            & Proposed (no GPS)        & 0.6625 & 0.4714 & 2.948 \\
            \midrule
            
            \multirow{4}{*}{\makecell{area1 \\ (786$m^2$, 40 rooms)}} 
            & SLAM         & 0.0100 & 0.0002 & 117.1 \\
            & SAC          & 0.0425 & 0.0195 & 6.967 \\
            & Proposed (GPS)           & \textbf{0.6600} & \textbf{0.3954} & \textbf{4.504} \\
            & Proposed (no GPS)       & 0.5750 & 0.2705 & 5.896 \\
            \midrule
            
            \multirow{4}{*}{\makecell{\textit{Average} \\ (483.4$m^2$, 20.2 rooms)}} 
            & SLAM         & 0.2570 & 0.1746 & 25.89 \\
            & SAC          & 0.3705 & 0.2363 & 4.076\\
            & Proposed (GPS)        & \textbf{0.8715} & \textbf{0.7393} & \textbf{2.232} \\
            & Proposed (no GPS)       & 0.8125 & 0.6614 & 2.707 \\
            \bottomrule
        \end{tabular}
    }
    \caption{Different method results over 5 environments, with the best values in bold. For SPL higher values are better. For OOR lower values are better. The proposed method with no GPS achieves similar performance to the variant with GPS. The proposed methods perform best across all environments and metrics.}
    \label{table:results}
\end{table}

\section{Experiment: MineRL Basalt Competition}
The learned visual navigation system is not limited to robotic visual navigation. The paradigm can be applied to other domains where information is limited and the benefit of classification of state and a policy to reproduce actions are clearly defined. A solution like the proposed learned visual navigation method was submitted to the MineRL Basalt competition in 2021 by me and collaborators and won first prize for best performance as well as the most human like agent~\cite{goecks2021combining,minerlretrospective}. This section will show how this method worked and show the similarities between robotic visual navigation and the proposed Minecraft solution. Additional information about the competition and solution is provided in \autoref{app:minerl_basalt}.

\subsection{Problem Setup}

The MineRL Basalt competition was a competition held in 2021 to foster creativity in replicating human-demonstration data via learned agents in a Minecraft simulation~\cite{shah2021basalt}. The BASALT competition environments did not include reward functions, as the goal was to build solutions to potential real-world problems. Humans judged the results to evaluate the effectiveness of each agent of completing a given task. The tasks were instead defined by a human-readable description, which was given both to the competitors and to the site visitors and workers doing the evaluation of the videos that trained agents generate. All competition information was provided at \url{https://minerl.io/basalt/}.

The four tasks in the competition were 1) finding a cave (FindCave), 2) placing a waterfall and taking a picture of it (MakeWaterfall), 3) building an animal pen and containing an animal in it (BuildAnimalPen), and 4) replicating the architecture of a village by building a new house (BuildVillageHouse). Each task was designed to be incrementally more challenging than the last. Each of these tasks required the agent to "throw a snowball" when finished with the task. The competition organizers also provided each participant team with a dataset with 40 to 80 human demonstrations for each task, not all of them completing the task, and the starter codebase to train a behavior cloning agent, to evaluate the agent, and to make the solution submission. 

\subsection{Methodology}

\begin{figure}[!ht]
  \centering
  \includegraphics[width=0.9\linewidth]{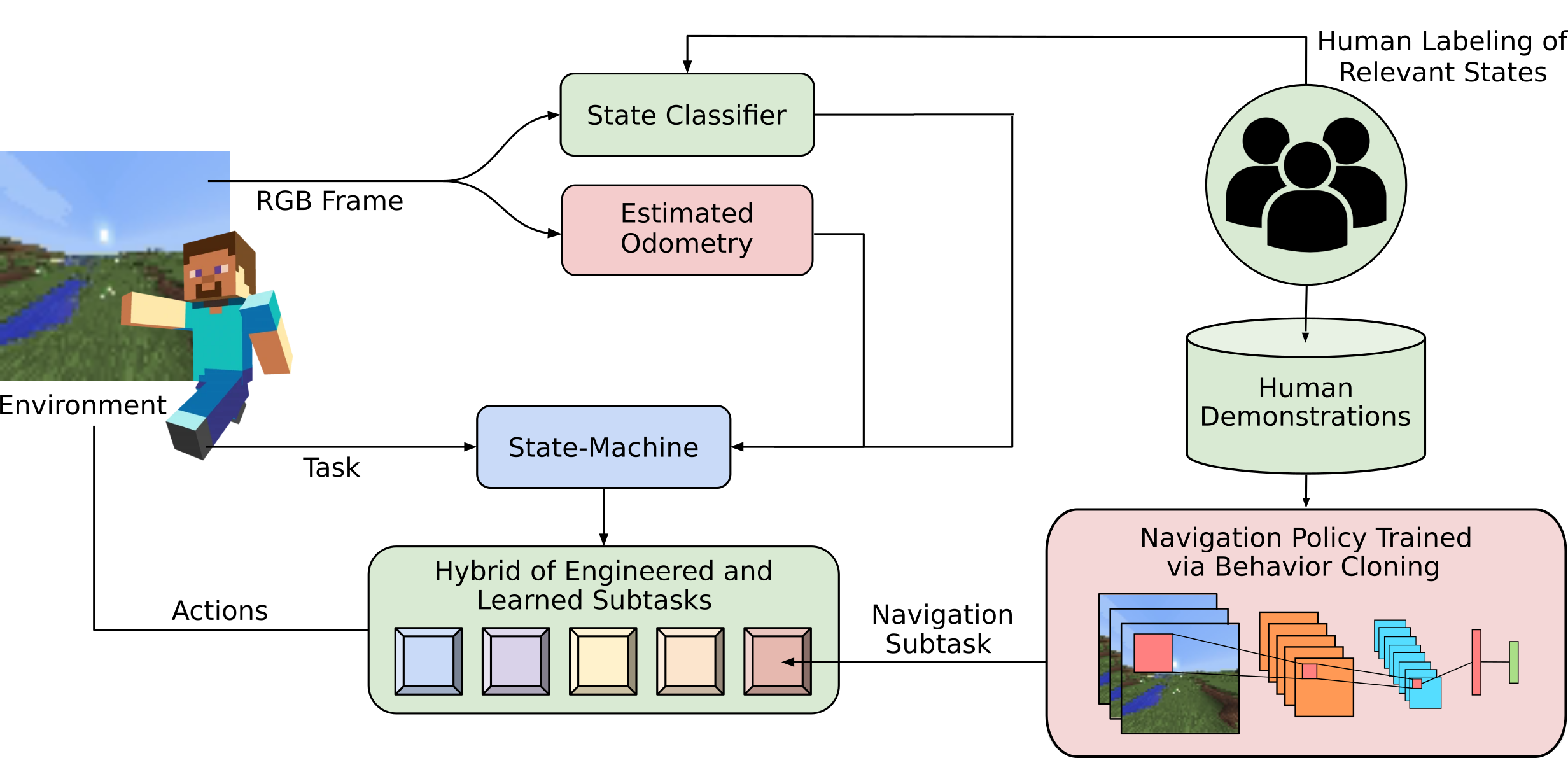}
  \caption{Diagram illustrating the approach. Using data from the available human demonstration dataset, humans provide additional binary labels to image frames to be used to train a state classifier that can detect relevant features in the environment such as caves and mountains. The available human demonstration dataset is also used to train a navigation policy via imitation learning to replicate how humans traverse the environment. A separate odometry module estimates the current agent's position and heading solely based on the action taken by the end. During test time, the agent uses the learned state classifier to provide useful information to an engineered state-machine that controls which subtask the agent should execute at every time-step.}
  \label{fig:minerl_diagram}
\end{figure}

Since no reward signal was given by the competition organizers and compute time was limited, direct deep reinforcement learning approaches were not feasible~\cite{Mnih2013, lillicrap2015continuous,Mnih2016}. With the limited human demonstration dataset, end-to-end behavior cloning also did not result in high-performing policies, because imitation learning requires a substantial number of high-quality data~\cite{Bojarski2016,Nair2017}. Also attempted was to address the tasks using adversarial imitation learning approaches such as Generative Adversarial Imitation Learning (GAIL)~\cite{ho2016generative}, however, the large-observation space and limited compute time also made this approach infeasible.

Hence, to solve the four tasks of the MineRL BASALT competition, a combined machine learning and knowledge engineering approach is used, also known as \emph{hybrid intelligence}~\cite{kamar2016directions,dellermann2019hybrid}. As seen in the main diagram of the approach shown in \autoref{fig:minerl_diagram}, the machine learning part of this method is seen in two different modules: first, a state classifier is learned using additional human feedback to identify relevant states in the environment; second, a navigation subtask is learned separately for each task via imitation learning using the human demonstration dataset provided by the competition. The knowledge engineering part is seen in three different modules: first, given the relevant states classified by the machine learning model and knowledge of the tasks, a state-machine is designed that defines a hierarchy of subtasks and controls which one should be executed at every time-step; second, solutions for the more challenging subtasks were designed that were not able to learn directly from data; and third, an estimated odometry module is engineered that provides additional information to the state-machine and enables the execution of the more complex engineered subtasks.

Like the learned visual navigation pipeline, the state classifier is analogous to the goal checker module where at each step the state classifier is determining the next state of the pipeline. The behavior cloning policy is analogous to the policy model from before, except with more states that are possible due to the more complex action space in Minecraft. This solution does estimate the odometry to enable certain steps in the policy where it is critical that the agent navigate back to a previous state, however the policy itself does not consider this estimated odometry at runtime. 

\subsection{Results}
Four different approaches are considered to solve the four tasks proposed in the MineRL Basalt competition:
\begin{itemize}
    \item \textbf{Hybrid: }the main proposed agent, which combines both learned and engineered modules. The learned modules are the navigation subtask policy (learns how to navigate using the human demonstration data provided by the competition) and the state classifier (learns how to identify relevant states using additional human-labeled data). 
    The engineered modules are the multiple subtasks, hand-designed to solve subtasks that were not able to be learned from data. These engineered modules are the estimated odometry and the state-machine, which uses the output of the state classifier and engineered task structure to select which subtask should be followed at each time-step.
    \item \textbf{Engineered: }almost identical to the Hybrid agent described above, however, the navigation subtask policy that was learned from human demonstrations is now replaced by a hand-designed module that randomly selects movement and camera commands to explore the environment.
    \item \textbf{Behavior Cloning: }end-to-end imitation learning agent that learns solely from the human demonstration data provided during the competition. This agent does not use any other learned or engineered module, which includes the state classifier, the estimated odometry, and the state-machine.
    \item \textbf{Human: }human-generated trajectories provided by the competition. They are neither guaranteed to solve the task nor solve it optimally because they depend on the level of expertise of each human controlling the agent.
\end{itemize}

Each combination of condition (behavior cloning, engineered, hybrid, human) and performance metric (best performer, fastest performer, most human-like performer) is treated as a separate participant of a one-versus-one competition where skill rating is computed using the \textit{TrueSkill}\textsuperscript{TM} Bayesian ranking system~\cite{herbrich2006trueskill}. The main proposed ``Hybrid'' agent, which combines engineered and learned modules, outperforms both pure hand-designed (``Engineered'') and pure learned (``Behavior Cloning'') agents in the ``Best Performer'' category, achieving $5.3 \%$ and $25.6 \%$ higher mean skill rating when compared to the ``Engineered'' and ``Behavior Cloning'' baselines, respectively.
However, when compared to the ``Human'' scores, our main proposed agent achieves $21.7 \%$ lower mean skill rating, illustrating that even the best approach is still not able to outperform a human player with respect to best performing the task. More results and analysis are shown in \autoref{app:minerl_basalt}.

\section{Conclusion}
This chapter proposed a navigation pipeline which does not rely on odometry, map, compass or indoor position at runtime and is purely based on the visual input and a novel 8-image panoramic goal. This method learns from expert trajectories generated using RGBD maps of several real environments. Using robotic simulators with real data and photo-realistic rendering, efficient collection of numerous expert trajectories can be collected. This enables the training of agents in simulation with real-world data, thereby bridging the sim-to-real domain adaptation. Experiments show that the proposed method 1) achieves better performance than state-of-the-art baselines, especially in complex environments with difficult and long-range path solutions; 2) requires fewer training samples and less training time; and 3) can work across different environments given an RGBD map. Additionally, this methodology can be extended to other domains, such as imitation-learned Minecraft agents.

\chapter{Visual Tactile Manipulation}
\label{ch:visual_tactile_manipulation}

Once the agent reaches the object, how can it reason about the object's geometry? What sensory modalities can a robot work with to make an estimate of the object geometry. In previous work, researchers used a single 2.5D image to reconstruct an object~\cite{varley2017shape}. This estimation can be further refined by utilizing tactile sensors on a robotic end-effector to refine a shape estimation better than using depth information alone. By increasing the quality of the shape estimation, it allows for a grasp planner to plan a grasp more accurately for manipulation tasks. This chapter explores the benefits of a shape completion system that leverage visual-tactile information and how the performance is improved over a single-view prediction. 

\section{Introduction}
Robotic grasp planning based on raw sensory data is difficult due to occlusion and incomplete information regarding scene geometry. Often, one sensory modality does not provide enough context to enable reliable planning. For example, a single depth sensor image cannot provide information about occluded regions of an object, and tactile information is incredibly sparse. To solve this, a 3D convolutional neural network is used to enable stable robotic grasp planning by incorporating both tactile and depth information to infer occluded geometries. This multi-modal system uses both tactile and depth information to form a more complete model of the space the robot can interact with and to provide a complete object model for grasp planning.

At runtime, a point cloud of the visible portion of the object is captured, and multiple guarded moves are executed in which the hand is moved towards the object, stopping when contact with the object occurs. The newly acquired tactile information is combined with the original partial view, voxelized, and sent through the CNN to create a hypothesis of the object's geometry.

\begin{figure}[ht!]
    \centering
    \includegraphics[width=\textwidth] {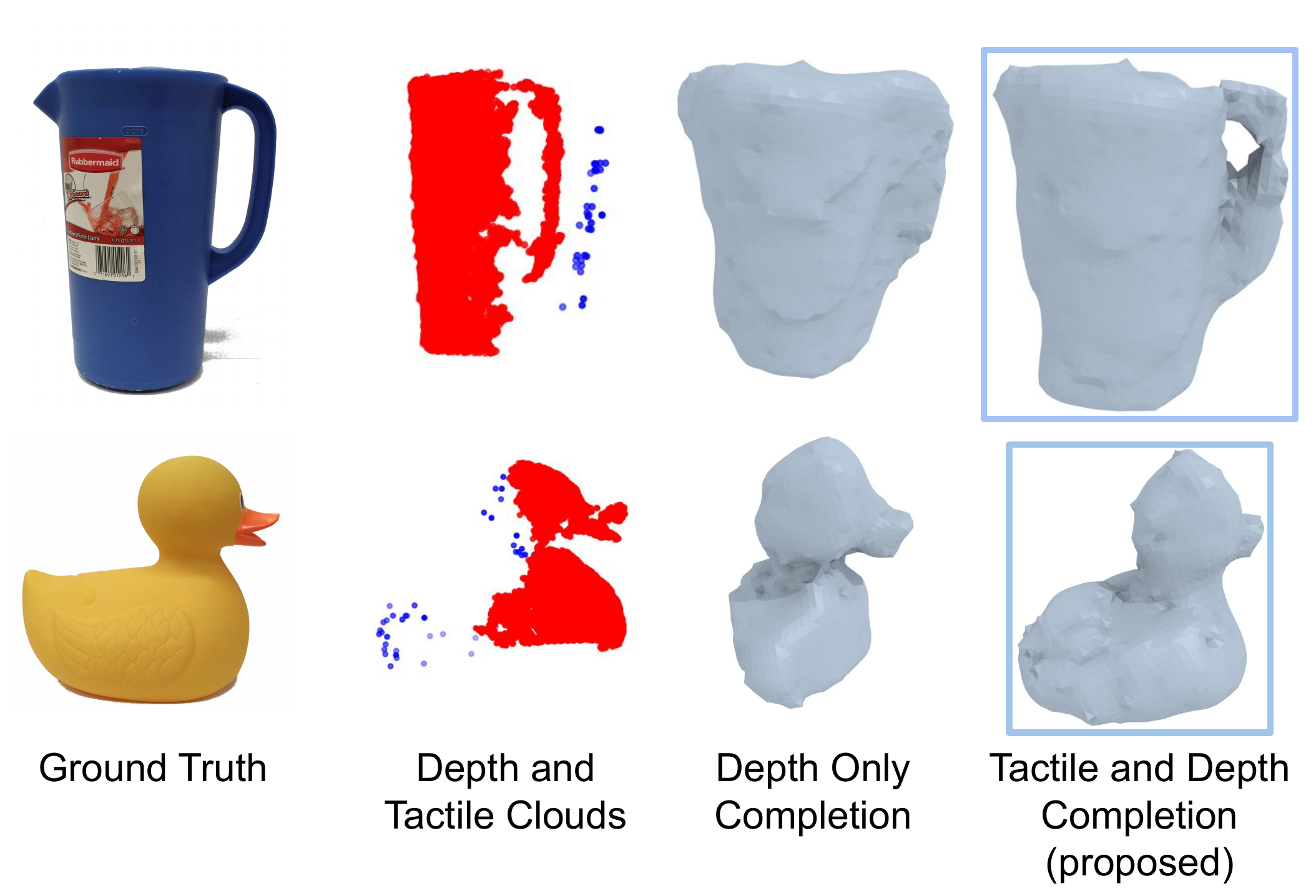}

    \caption{Completion example from tactile and depth data. A few samples of tactile data can significantly improve the system’s ability to reason about 3D geometry. The Depth Only Completion for the pitcher does not capture the handle well, whereas the tactile information gives a better geometric understanding. The additional tactile data allowed the CNN to correctly identify a handle in the completion mesh and similar completion improvement was found for the novel rubber duck not in the training set.}
    \label{fig:visual_tactile_main_figure}
\end{figure}

\begin{figure*}[t]
    \centering
    \includegraphics[width=.95\textwidth] {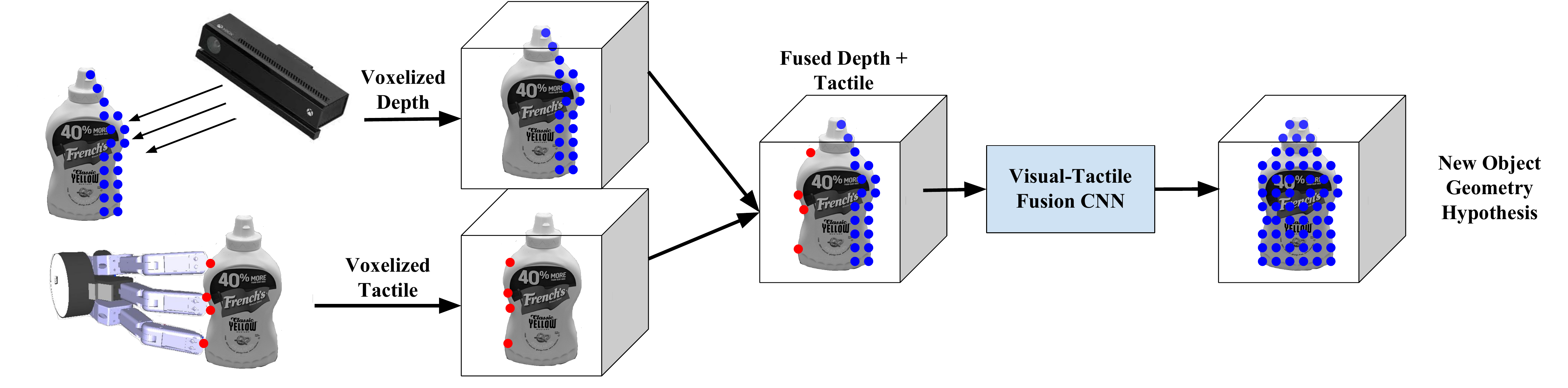}
    \caption{Both tactile and depth information are independently captured and voxelized into $40^3$ grids. These are merged into a shared occupancy map which is fed into a CNN to produce a hypothesis of the object's geometry.}
    \label{fig:visual_tactile_fusion_cnn}
\end{figure*}

Depth information from a single point of view often does not provide enough information to accurately predict object geometry. There is often unresolved uncertainty about the geometry of the occluded regions of the object. To alleviate this uncertainty, tactile information is utilized to generate a new, more accurate hypothesis of the object's 3D geometry, incorporating both visual and tactile information. \autoref{fig:visual_tactile_main_figure} demonstrates an example where the understanding of the object's 3D geometry is significantly improved by the additional sparse tactile data collected via this framework. An overview of the sensory fusion architecture is shown in \autoref{fig:visual_tactile_fusion_cnn}.

This method is differentiated from others~\cite{wang2018gelsighttactile} in that this CNN is acting on both the depth and tactile as input information fed directly into the model rather than using the tactile information to update the output of a CNN not explicitly trained on tactile information. This enables the tactile information to produce non-local changes in the resulting mesh. In many cases, depth information alone is insufficient to differentiate between two potential completions, for example a pitcher vs a rubber duckie. In these cases, the CNN utilizes sparse tactile information to affect the entire completion, not just the regions near the tactile glance. If the tactile sensor senses the occluded portion of a drill, the CNN can turn the entire completion into a drill, not just the local portion of the drill that was touched. 


\section{Visual-Tactile Geometric Reasoning Method}
The framework utilizes a trained CNN to produce a mesh of the target object, incorporating both depth and tactile information in a single input channel. The same architecture as found in~\cite{varley2017shapecompletion_iros} is used. The model was implemented using the Keras~\cite{Keras2015} deep learning library. Each layer used rectified linear units as nonlinearities except the final fully connected (output) layer which used a sigmoid activation to restrict the output to the range $[0,1]$. Cross-entropy error $E(y,y^\prime)$ is used as the cost function with target $y$ and output $y^\prime$:
\[E(y,y^\prime)=-\left( y \log(y^\prime) + (1 - y) \log(1 - y^\prime) \right) \]

\begin{figure}[t]
\centering

	\begin{subfigure}{.23\textwidth}
		\centering
		\includegraphics[width=1\textwidth]{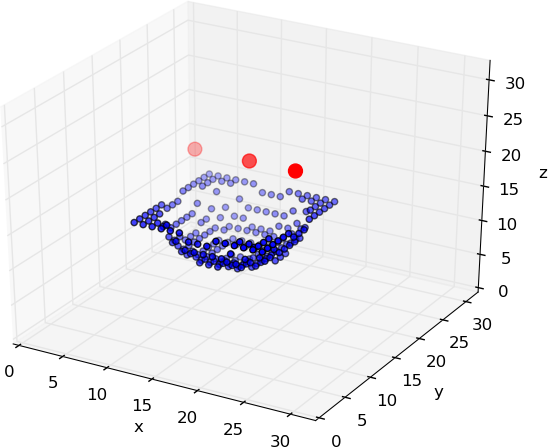}
	\end{subfigure}
	\begin{subfigure}{.23\textwidth}
		\centering
		\includegraphics[width=1\textwidth]{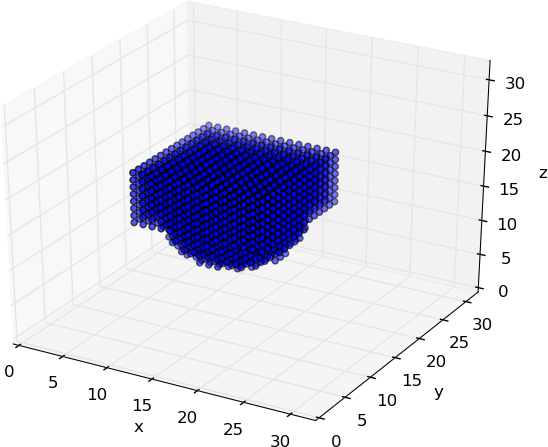}
	\end{subfigure}
	\begin{subfigure}{.24\textwidth}
		\centering
		\includegraphics[width=1\textwidth]{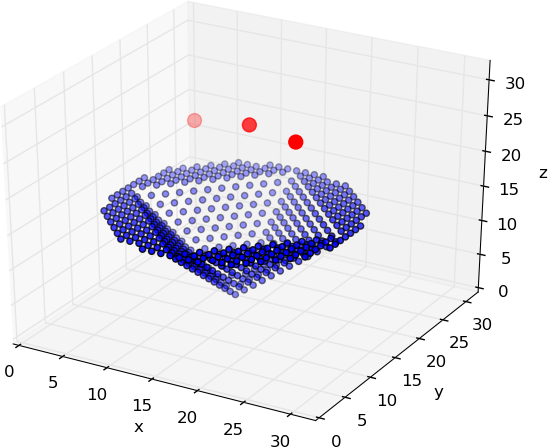}
	\end{subfigure}
	\begin{subfigure}{.24\textwidth}
		\centering
		\includegraphics[width=1\textwidth]{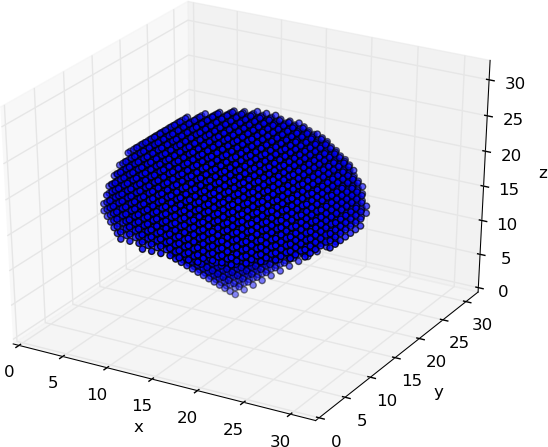}
	\end{subfigure}
	\caption{Example training pair from the geometric shape dataset. For the left, red dots represent tactile readings and blue dots represent the depth image. The blue points on the right are the ground truth 3D geometry.} 
	\label{fig:geo_shape_training_example} 
\end{figure}

\begin{figure}[t]
	\begin{algorithm}[H]
	\caption{Simulated YCB/Grasp Tactile Data Generation}
	\label{alg:TactileDataGenerationUniform}
		\begin{algorithmic}[1]
		\STATE grid\_dim = 40 // resolution of voxel grid \label{line:alg1line2}
		\STATE npts = 40 // num locations to check for contact \label{line:alg1line3}
		\STATE vox\_gt\_cf = align\_gt\_to\_depth\_frame(vox\_gt) \label{line:alg1line4}
		\STATE xs = rand\_ints(start=0, end=grid\_dim-1, size=npts) \label{line:alg1line5}
		\STATE ys = rand\_ints(start=0, end=grid\_dim-1, size=npts) \label{line:alg1line6}
		\STATE tactile\_vox = [] \label{line:alg1line7}
		\FOR{x, y in xs, ys} \label{line:alg1line8}
		    \FOR{z in range(grid\_dim-1, -1, -1)} \label{line:alg1line9}
		        \IF{vox\_gt\_cf[x, y, z] == 1} \label{line:alg1line10}
			        \STATE tactile\_vox.append(x, y, z) \label{line:alg1line11}
			        \STATE continue \label{line:alg1line12}
		        \ENDIF
		    \ENDFOR
		\ENDFOR
		\STATE tactile\_points = vox2point cloud(tactile\_vox) \label{line:alg1line13}
		\RETURN tactile\_points \label{line:alg1line14}
		\end{algorithmic}
	\end{algorithm}
	\caption{The sampling of tactile points in simulation is treated as a two-dimensional problem of trying to recreate a Barrett fingertip capturing tactile data of the object. By sampling 40 random points on the occluded side of the object the robot can capture as much data as captured in 6 guided tactile measurements in a physical experiment. Given the resolution of the voxel grid is $40^3$, the robot is sampling $2.5\%$ of all positions on the occluded side of the object. This sparsity of data worked to advantages both in terms of reducing data required for training and forcing the network to utilize tactile data as much as possible. }
\end{figure}

\noindent This cost function encourages each output to be close to either 0 for unoccupied target voxels or 1 for occupied target voxels. The optimization algorithm Adam~\cite{kingma2014}, which computes adaptive learning rates for each network parameter, is used with default hyperparameters ($\beta_1=0.9$, $\beta_2=0.999$, $\epsilon=10^{-8}$) except for the learning rate, which is set to 0.0001. Weights were initialized following the recommendations of~\cite{he2015} for rectified linear units and~\cite{glorot2010} for the logistic activation layer (batch size=32).


\section{Completion of Simulated Geometric Shapes}
\label{sec:geometric_shapes}

\begin{figure*}[t]

	\centering
	\begin{subfigure}[t]{0.29\textwidth}
		\centering
		\includegraphics[width=1\textwidth,height=33mm]{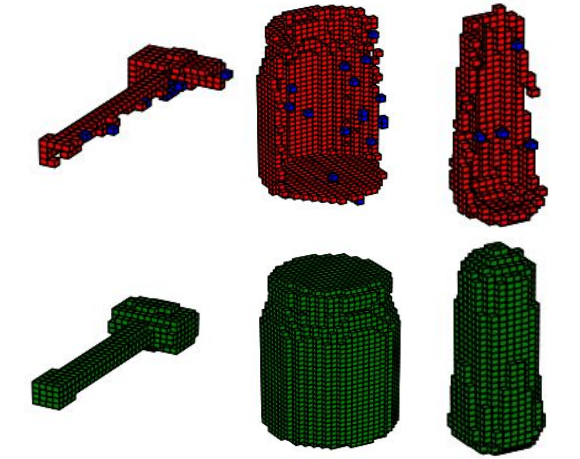}
		\caption{Training Views}
	\end{subfigure}
	\begin{subfigure}[t]{0.29\textwidth}
		\centering
		\includegraphics[width=1\textwidth,height=33mm]{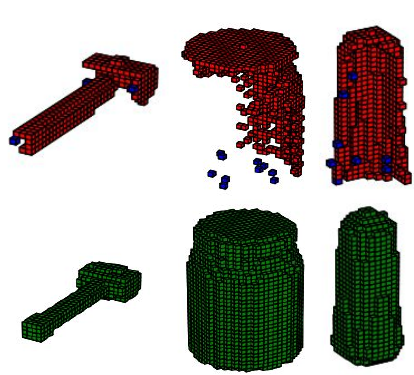}
		\caption{Holdout Views}
	\end{subfigure}
	\begin{subfigure}[t]{0.29\textwidth}
		\centering
		\includegraphics[width=1\textwidth,height=33mm]{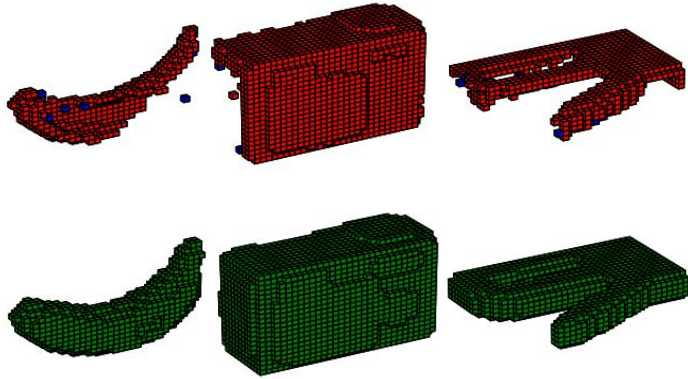}
		\caption{Holdout Meshes}
	\end{subfigure}
    \caption{Several examples from different data splits for evaluating completion quality. The bottom row is ground truth $40^3$ occupancy, and the top row is a voxelization of the depth and tactile information. Training Views were used to train the CNN, and Holdout Views are views of meshes used to train the CNN but from views not used during training. Holdout Meshes are views of meshes that the network never saw during training.}
    \label{fig:datasplits}
\end{figure*}

\begin{figure*}[t]
	\centering
	\begin{subfigure}[t]{0.3\textwidth}
		\centering
		\includegraphics[width=1\textwidth]{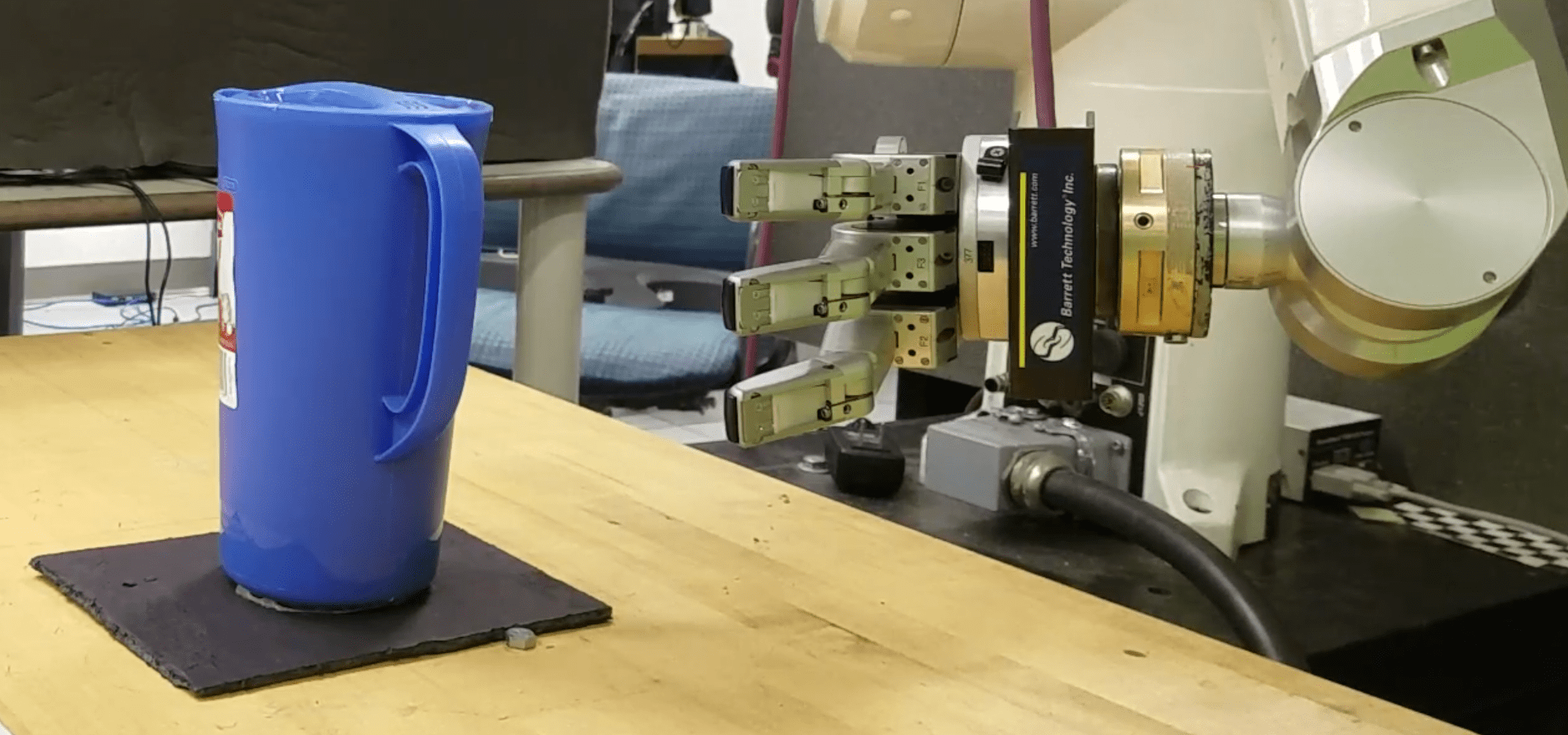}
		\caption{Hand approach}
	\end{subfigure}
	\begin{subfigure}[t]{0.3\textwidth}
		\centering
		\includegraphics[width=1\textwidth]{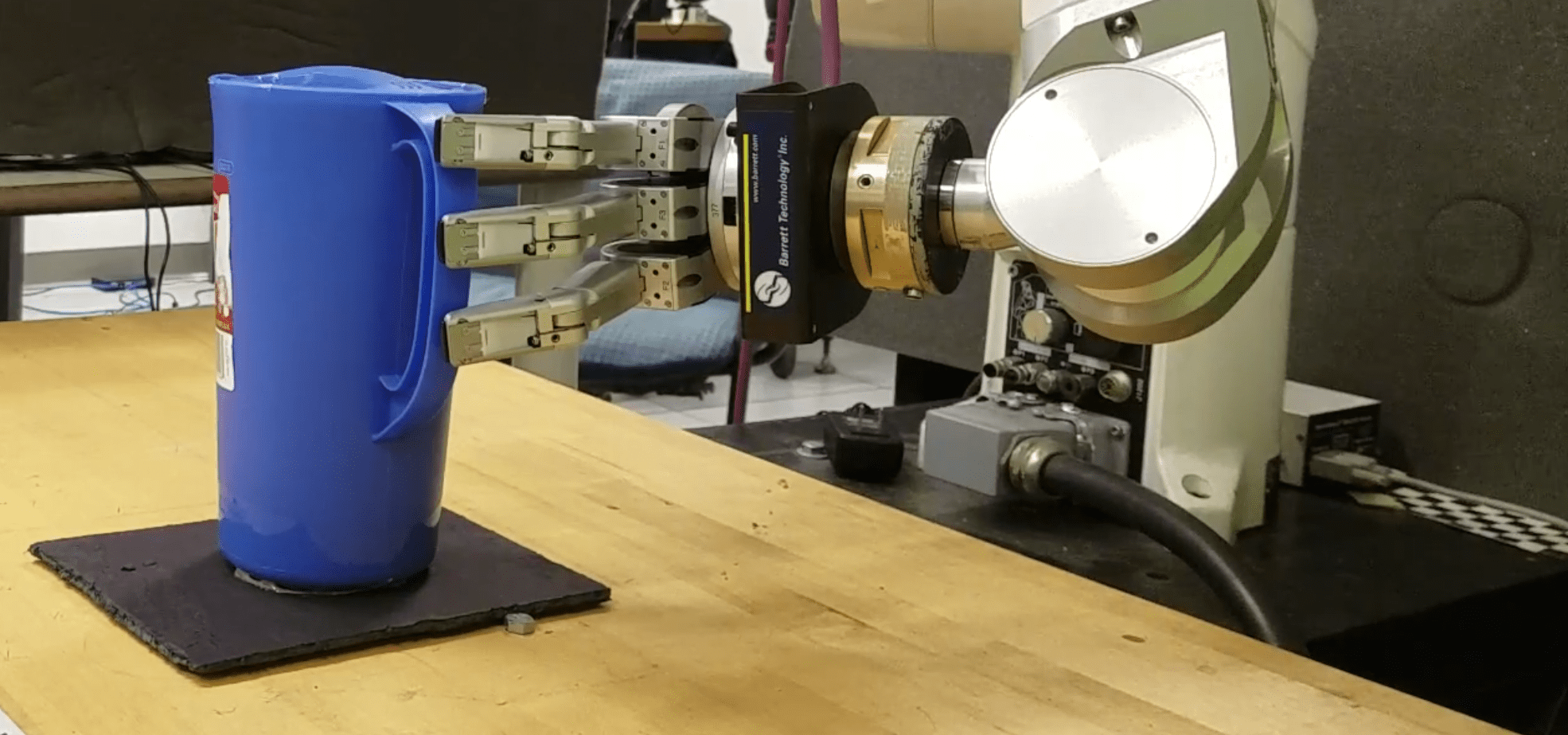}
		\caption{Finger contact}
	\end{subfigure}
	\begin{subfigure}[t]{0.3\textwidth}
		\centering
		\includegraphics[width=1\textwidth]{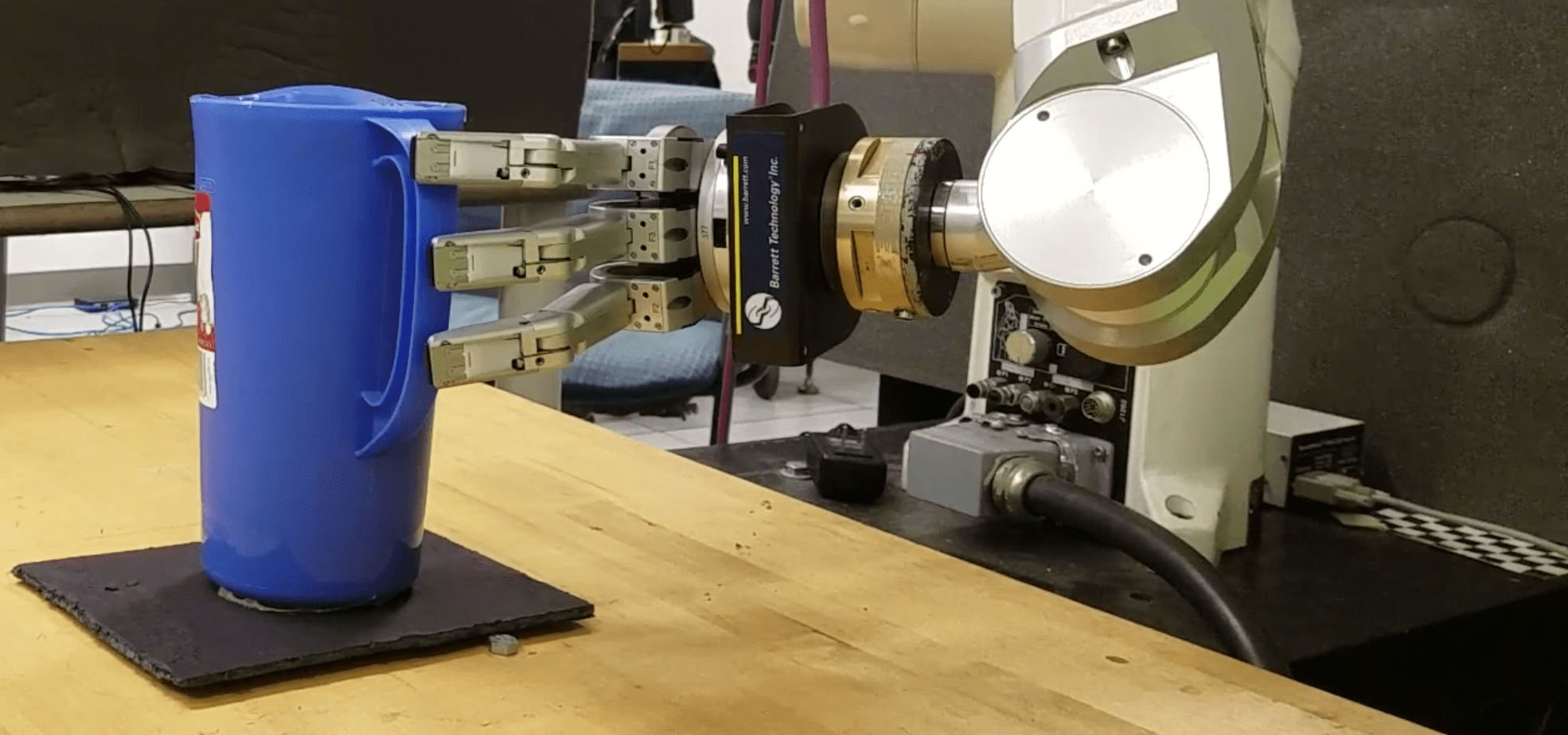}
		\caption{Finger curl}
	\end{subfigure}
	\caption{Barrett hand showing contact with a fixed object. (a) The hand is manually brought to an approach position, (b) approaches the object, and (c) the fingers are curled to contact the object and collect tactile information. This process is repeated 6 times over the occluded surface of the object. } 
	\label{fig:handobjectcontact} 
\end{figure*} 

To evaluate the system's ability to utilize additional tactile sensory information to reason about 3D geometry, initial experimentation was done on a toy geometric shape dataset. This dataset consisted of conjoined half-shapes. Both front and back halves of the objects were randomly chosen to be either a sphere, cube, or diamond of varying sizes. The front and back halves do match in size. An example shape is shown in \autoref{fig:geo_shape_training_example}(b), a half-cube half-sphere. Next, synthetic sensory data was generated for these example shapes and embedded in a $40^3$ voxel grid. Depth information was captured from a fixed camera location, and tactile data was generated by intersecting 3 rays with the object. The rays originated at (13, 20, 40), (20, 20, 40) and (26, 20, 40), and traveled in the $-z$ direction until either contact occurred with the object or the ray left the voxelized volume. Sensory data for a shape is shown in \autoref{fig:geo_shape_training_example}. 

\begin{table}[t]
\centering
\begin{tabular}{|c|c|}
\hline
Input Data                      &  Jaccard similarity   \\ \hline
Tactile Only              &  0.863      \\ \hline
Depth Only                      &  0.890          \\ \hline
Depth \& Tactile                &  \textbf{0.986}        \\ \hline
\end{tabular}
\caption{Jaccard similarity for test set of geometric shapes produced from the same CNN architecture trained with various input data. When trained using both tactile and depth information the CNN can complete the object with a near perfect Jaccard similarity. Depth or tactile alone is not sufficient to reason about object geometry in these problems.}
\label{tab:Shapes_Max_Jaccard}
\end{table}

\begin{figure*}[t]
    \centering
    \includegraphics[width=.95\textwidth]{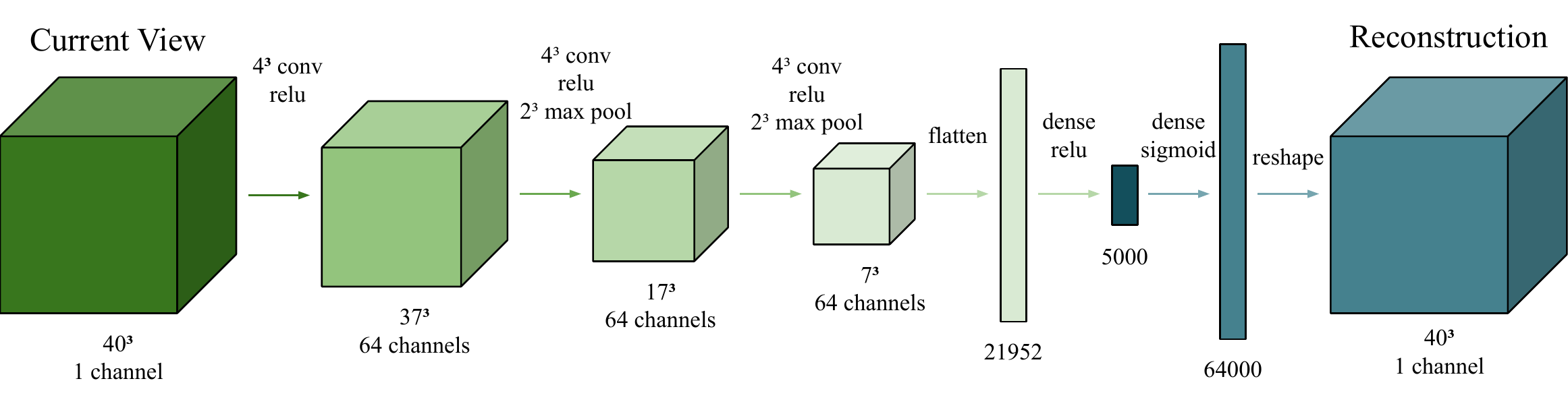}
    \caption{\textbf{Visual-Tactile CNN Architecture} The CNN architecture takes a single $40^3$ voxelization of the input point cloud, embeds it, then performs a dense $64000$ reconstruction of the object. This system works well via convolutional layers abstracting the geometry of the input, and the embedding using information from the entire input region.}
    \label{fig:single_view_architecture}
\end{figure*}

Three networks with the architecture from~\cite{varley2017shapecompletion_iros} were trained on a simulated dataset of geometric shapes (\autoref{fig:geo_shape_training_example}) where the front and back were composed of two differing shapes. Sparse tactile data was generated by randomly sampling voxels along the occluded side of the voxel grid. A network is trained that only utilized tactile information. This performed poorly due to the sparsity of information. A second network was given only the depth information during training and performed better than the tactile-only network did. It still encountered many situations where it did not have enough information to accurately complete the obstructed half of the object. A third network was given depth and tactile information which successfully utilized the tactile information to differentiate between plausible geometries of occluded regions. The network architecture is shown in \autoref{fig:single_view_architecture}.

Results for this geometric test set are shown in \autoref{tab:Shapes_Max_Jaccard}. The Jaccard similarity improved from 0.890 in the depth only network to 0.986 in the depth and tactile network. This task demonstrated that a CNN can be trained to leverage sparse tactile information to decide between multiple object geometry hypotheses. When the object geometry had sharp edges in its occluded region, the system would use tactile information to generate a completion that contained similar sharp edges in the occluded region. This completion is more accurate not just in the observed region of the object but also in the unobserved portion of the object. It was also found that tactile only was not enough information to get a complete understanding of the object's geometry with a Jaccard score of 0.863.



\begin{figure}[hbtp]
\centering
    \includegraphics[width=\textwidth]{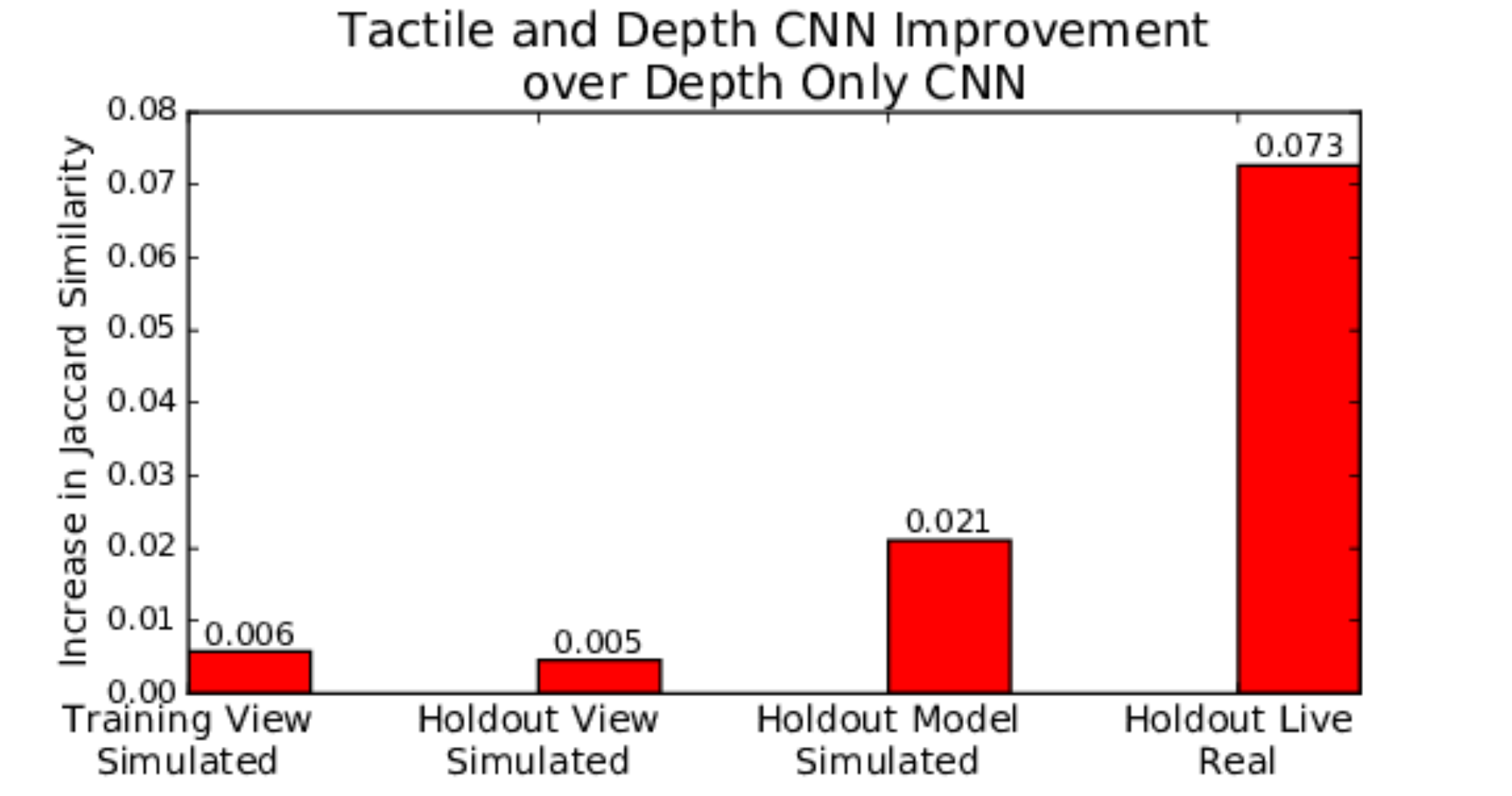}
    \caption{ As the difficulty of the data splits increase, the delta between the \textbf{Depth Only} CNN completion accuracy and the \textbf{Tactile and Depth} CNN completion accuracy increases. The additional tactile information is more useful on more difficult completion problems. }
    \label{fig:jaccard_improvement}
\end{figure}

\begin{table}[t]
	\centering
    \begin{tabular}{|c|c|c|c|c|}
    \hline
    \multicolumn{1}{|c|}{\begin{tabular}[c]{@{}c@{}}\textbf{Completion} \\  \textbf{Method}\end{tabular}} 
    & \multicolumn{1}{c|}{\begin{tabular}[c]{@{}c@{}}\textbf{Train} \\  \textbf{View(Sim)}\end{tabular}} 
    & \multicolumn{1}{c|}{\begin{tabular}[c]{@{}c@{}}\textbf{Holdout} \\  \textbf{View(Sim)}\end{tabular}} 
    & \multicolumn{1}{c|}{\begin{tabular}[c]{@{}c@{}}\textbf{Holdout} \\  \textbf{Model(Sim)}\end{tabular}} 
    & \multicolumn{1}{c|}{\begin{tabular}[c]{@{}c@{}}\textbf{Holdout} \\  \textbf{(Live)}\end{tabular}} \\ 
    \hline
    	Partial        & 0.01          & 0.02          & 0.01          & 0.01          \\ \hline
    	Convex Hull    & 0.50          & 0.51          & 0.46          & 0.43          \\ \hline
    	GPIS           & 0.47          & 0.45          & 0.35          & 0.48          \\ \hline
    	Depth CNN      & 0.68          & 0.65          & 0.65          & 0.37          \\ \hline
    	Depth Tactile  & \textbf{0.69} & \textbf{0.66} & \textbf{0.65} & \textbf{0.64} \\ \hline
    \end{tabular}
    
    \caption{\textbf{Jaccard similarity results} measuring the intersection over union of two voxelized meshes, as described in \autoref{sec:Completion_results}. (Larger is better)}
    \label{tab:Jaccard}
\end{table}

\begin{table}[t]
	\centering
    \begin{tabular}{|c|c|c|c|c|c|c|}
    \hline
    \multicolumn{1}{|c|}{\begin{tabular}[c]{@{}c@{}}\textbf{Completion} \\  \textbf{Method}\end{tabular}} 
    & \multicolumn{1}{c|}{\begin{tabular}[c]{@{}c@{}}\textbf{Train} \\  \textbf{View(Sim)}\end{tabular}} 
    & \multicolumn{1}{c|}{\begin{tabular}[c]{@{}c@{}}\textbf{Holdout} \\  \textbf{View(Sim)}\end{tabular}} 
    & \multicolumn{1}{c|}{\begin{tabular}[c]{@{}c@{}}\textbf{Holdout} \\  \textbf{Model(Sim)}\end{tabular}} 
    & \multicolumn{1}{c|}{\begin{tabular}[c]{@{}c@{}}\textbf{Holdout} \\  \textbf{(Live)}\end{tabular}} \\ 
    \hline
    	Partial       & 7.8           & 7.0           & 7.6           & 11.9          \\ \hline
    	Convex Hull   & 32.7          & 45.1          & 49.1          & 11.6          \\ \hline
    	GPIS          & 59.9          & 79.2          & 118.0         & 17.9          \\ \hline
    	Depth CNN     & 6.5           & 6.9           & 6.5           & 16.5          \\ \hline
    	Depth Tactile & \textbf{5.8}  & \textbf{5.8}  & \textbf{6.2}  & \textbf{7.4}  \\ \hline
    \end{tabular}
    
    \caption{\textbf{Hausdorff distance results} measuring the mean distance in millimeters from points on one mesh to points on another mesh, as described in \autoref{sec:Completion_results}. (Smaller is better)}
    \label{tab:Hausdorff} 
\end{table}

\begin{table}[t]
	\centering
    \begin{tabular}{|c|c|c|c|c|c|c|}
    \hline
    \multicolumn{1}{|c|}{\begin{tabular}[c]{@{}c@{}}\textbf{Completion} \\  \textbf{Method}\end{tabular}} 
    & \multicolumn{1}{c|}{\begin{tabular}[c]{@{}c@{}}\textbf{Train} \\  \textbf{View(Sim)}\end{tabular}} 
    & \multicolumn{1}{c|}{\begin{tabular}[c]{@{}c@{}}\textbf{Holdout} \\  \textbf{View(Sim)}\end{tabular}} 
    & \multicolumn{1}{c|}{\begin{tabular}[c]{@{}c@{}}\textbf{Holdout} \\  \textbf{Model(Sim)}\end{tabular}} 
    & \multicolumn{1}{c|}{\begin{tabular}[c]{@{}c@{}}\textbf{Holdout} \\  \textbf{(Live)}\end{tabular}} \\ 
    \hline
    	Partial       & 19.9mm          & 21.1mm          & 16.6mm          & 18.6mm          \\ \hline
    	Convex Hull   & 13.9mm          & 16.1mm          & 14.1mm          & 10.5mm          \\ \hline
    	GPIS          & 17.1mm          & 16.0mm          & 21.3mm          & 20.8mm          \\ \hline
    	Depth CNN     & 12.1mm          & 13.7mm          & 12.4mm          & 22.9mm          \\ \hline
    	Depth Tactile & \textbf{7.7mm} & \textbf{13.9mm} & \textbf{13.6mm} & \textbf{6.2mm} \\ \hline
    \end{tabular}
    
    \caption{\textbf{Pose error results} from simulated grasping experiments. This is the average L2 difference between planned and realized grasp pose averaged over the 3 fingertips and the palm of the hand, in millimeters. (Smaller is better) }
    \label{tab:sim_grasp_results} 
\end{table}

\begin{table}[t]
	\centering
    \begin{tabular}{|c|c|c|c|c|c|c|}
    \hline
    \multicolumn{1}{|c|}{\begin{tabular}[c]{@{}c@{}}\textbf{Completion} \\  \textbf{Method}\end{tabular}} 
    & \multicolumn{1}{c|}{\begin{tabular}[c]{@{}c@{}}\textbf{Train} \\  \textbf{View(Sim)}\end{tabular}} 
    & \multicolumn{1}{c|}{\begin{tabular}[c]{@{}c@{}}\textbf{Holdout} \\  \textbf{View(Sim)}\end{tabular}} 
    & \multicolumn{1}{c|}{\begin{tabular}[c]{@{}c@{}}\textbf{Holdout} \\  \textbf{Model(Sim)}\end{tabular}} 
    & \multicolumn{1}{c|}{\begin{tabular}[c]{@{}c@{}}\textbf{Holdout} \\  \textbf{(Live)}\end{tabular}} \\ 
    \hline
    	Partial       & 8.19$^{\circ}$          & 6.71$^{\circ}$          & 8.78$^{\circ}$          & 7.67$^{\circ}$          \\ \hline
    	Convex Hull   & 3.53$^{\circ}$          & 4.01$^{\circ}$          & 4.59$^{\circ}$          & 3.77$^{\circ}$          \\ \hline
    	GPIS          & 4.65$^{\circ}$          & 4.79$^{\circ}$          & 4.95$^{\circ}$          & 5.92$^{\circ}$          \\ \hline
    	Depth CNN     & 3.09$^{\circ}$          & 3.56$^{\circ}$          & 4.52$^{\circ}$          & 6.83$^{\circ}$          \\ \hline
    	Depth Tactile & \textbf{2.48$^{\circ}$} & \textbf{3.41$^{\circ}$} & \textbf{4.95$^{\circ}$} & \textbf{2.43$^{\circ}$} \\ \hline
    \end{tabular}
    
    \caption{\textbf{Joint error results} from simulated grasping experiments. This is the mean L2 distance between planned and realized grasps in degrees averaged over the hand's 7 joints. The Depth and Tactile method' smaller error demonstrates a more accurate geometry reconstruction. (Smaller is better)}
    \label{tab:l2_joint_error} 
\end{table}

\begin{table}[t]
\centering
    \begin{tabular}{|c|c|c|c|c|c|}
    \hline
    \multicolumn{1}{|c|}{\begin{tabular}[c]{@{}c@{}}\textbf{Completion} \\  \textbf{Method}\end{tabular}} 
    & \multicolumn{1}{c|}{\begin{tabular}[c]{@{}c@{}}\textbf{Partial}\end{tabular}} 
    & \multicolumn{1}{c|}{\begin{tabular}[c]{@{}c@{}}\textbf{Convex} \\ \textbf{Hull}\end{tabular}} 
    & \multicolumn{1}{c|}{\begin{tabular}[c]{@{}c@{}}\textbf{GPIS}\end{tabular}} 
    & \multicolumn{1}{c|}{\begin{tabular}[c]{@{}c@{}}\textbf{Depth} \\ \textbf{CNN}\end{tabular}} 
    & \multicolumn{1}{c|}{\begin{tabular}[c]{@{}c@{}}\textbf{Depth Tactile}\end{tabular}} 
     \\ \hline
    
        Time (s) & 1.533s & \textbf{0.198s} & 45.536s & 3.308s & 3.391s\\ \hline
    \end{tabular}
    
    \caption{Algorithmic methods like partial and convex hull have faster runtimes than using a neural network, however all methods are significantly faster than the GPIS timings. This means faster planning and execution of a grasp than using alternative methods. }
    \label{tab:live_grasp_timings} 
    
\end{table}

\section{Completion of YCB/Grasp Dataset Objects}

The dataset from~\cite{varley2017shapecompletion_iros} is used to create a new dataset consisting of half a million triplets of oriented voxel grids: depth, tactile, and ground truth. Depth voxels are marked as occupied if visible to the camera. Tactile voxels are marked occupied if tactile contact occurs within the voxel. Ground truth voxels are marked as occupied if the object intersects a given voxel, independent of perspective. The point clouds for the depth information were synthetically rendered in the Gazebo~\cite{koenig2004design} simulator. This dataset consists of 608 meshes from both the Grasp~\cite{kappler2015leveraging} and YCB~\cite{calli2015ycb} datasets. 486 of these meshes were randomly selected and used for a training set and the remaining 122 meshes were kept for a holdout set. 

The synthetic tactile information was generated according to Algorithm~\ref{alg:TactileDataGenerationUniform}. To generate tactile data, the voxelization of the ground truth high resolution mesh (vox\_gt) (Alg.\ref{alg:TactileDataGenerationUniform}:L\ref{line:alg1line2}) was aligned with the captured depth image (Alg.\ref{alg:TactileDataGenerationUniform}:L\ref{line:alg1line4}). 40 random $(x,y)$ points were sampled to generate synthetic tactile data (Alg.\ref{alg:TactileDataGenerationUniform}:L\ref{line:alg1line5}-\ref{line:alg1line6}). For each of these points (Alg.\ref{alg:TactileDataGenerationUniform}:L\ref{line:alg1line7}), a ray was traced in the $-z$, direction and the first occupied voxel was stored as a tactile observation  (Alg.\ref{alg:TactileDataGenerationUniform}:L\ref{line:alg1line11}). Finally this set of tactile observations was converted back to a point cloud (Alg.\ref{alg:TactileDataGenerationUniform}:L\ref{line:alg1line13}). 

Two identical CNNs were trained where one CNN was provided only depth information (\textbf{Depth Only}) and a second was provided both tactile and depth information (\textbf{Tactile and Depth}). During training, performance was evaluated on simulated views of meshes within the training data (\textit{Training Views}), novel simulated views of meshes in the training data (\textit{Holdout Views}), novel simulated views of meshes not in the training data (\textit{Holdout Meshes}), and real non-simulated views of 8 meshes from the YCB dataset (\textit{Holdout Live}). 

The \textit{Holdout Live} examples consist of depth information captured from a real Kinect and tactile information captured from a real Barrett Hand attached to a Staubli Arm. Depth filtering is used to mask out the background of the captured depth cloud. The object was fixed in place during the tactile data collection process. While collecting the tactile data, the arm was manually moved to place the end effector behind the object and 6 exploratory guarded motions were made where the fingers closed towards the object. Each finger stopped independently when contact was made with the object, as shown in \autoref{fig:handobjectcontact}. 

\autoref{fig:jaccard_improvement} demonstrates that the difference between the \textbf{Depth Only} CNN completion and the \textbf{Tactile and Depth} CNN completion becomes larger on more difficult completion problems. The performance of the \textbf{Depth Only} CNN nearly matches the performance of the \textbf{Tactile and Depth} CNN on the training views. Because these views are used during training, the network can generate reasonable completions. Moving from \textit{Holdout Views} to \textit{Holdout Meshes} to \textit{Holdout Live}, the completion problems move further away from the examples experienced during training. As the problems become harder, the \textbf{Tactile and Depth} network outperforms the \textbf{Depth Only} network by a greater margin, as it can utilize the sparse tactile information to differentiate between various completions. This trend shows that the network can make more use of the tactile information when the depth information alone is insufficient to generate a quality completion. Meshes are generated from the output of the combined tactile and depth CNN using a marching cubes algorithm. The density of the rich visual information and the coarse tactile information is preserved by utilizing the post-processing from~\cite{varley2017shapecompletion_iros}. 

\subsection{Mesh Generation}
Algorithm~\ref{alg:reconstruction} shows how the dense partial view and tactile information are merged into a $40^3$ voxel grid. More information about this method is available from~\cite{varley2017shapecompletion_iros}.

\begin{algorithm}
\caption{Visual-Tactile Shape Completion}\label{alg:reconstruction}
\begin{algorithmic}[1]
\STATE //cnn\_out: $40^3$ voxel output from CNN
\STATE //observed\_pc: captured point cloud of object
\STATE d\_ratio~$\gets$~densityRatio(observed\_pc, cnn\_out) \label{line:densityRatio}
\STATE upsampled\_cnn~$\gets$~upsample(cnn\_out, d\_ratio) \label{line:upsample}
\STATE vox~$\gets$~merge(upsampled\_cnn, observed\_pc) \label{line:merge}
\STATE vox\_no\_gap~$\gets$~fillGaps(vox) \label{line:fill_gaps}
\STATE vox\_weighted~$\gets$~CUDA\_QP(vox\_no\_gap) \label{line:cuda_qp}
\STATE mesh~$\gets$~mCubes(vox\_weighted) \label{line:mcubes}
\RETURN mesh
\end{algorithmic}
\end{algorithm}

To merge with the partial view, the output of the CNN is converted to a point cloud, and its density is compared to the density of the partial view point cloud (Alg.\ref{alg:reconstruction}:L\ref{line:densityRatio}). The CNN output is up-sampled by $d\_ratio$ to match the density of the observed point cloud (Alg.\ref{alg:reconstruction}:L\ref{line:upsample}). The upsampled output from the CNN is then merged with the observed point cloud of the partial view, and the combined cloud is voxelized at the new higher resolution of $(40*d\_ratio)^3$ (Alg.\ref{alg:reconstruction}:L\ref{line:merge}). Any gaps in the voxel grid between the upsampled CNN output and the observed partial view cloud are filled (Alg.\ref{alg:reconstruction}:L\ref{line:fill_gaps}). The voxel grid is smoothed using a CUDA implementation of the convex quadratic optimization problem from~\cite{lempitsky2010surface} (Alg.\ref{alg:reconstruction}:L\ref{line:cuda_qp}). The weighted voxel grid is then run through marching cubes (Alg.\ref{alg:reconstruction}:L\ref{line:mcubes}). \\

\begin{figure*}[ht!]

    \centering
    \includegraphics[width=\textwidth]{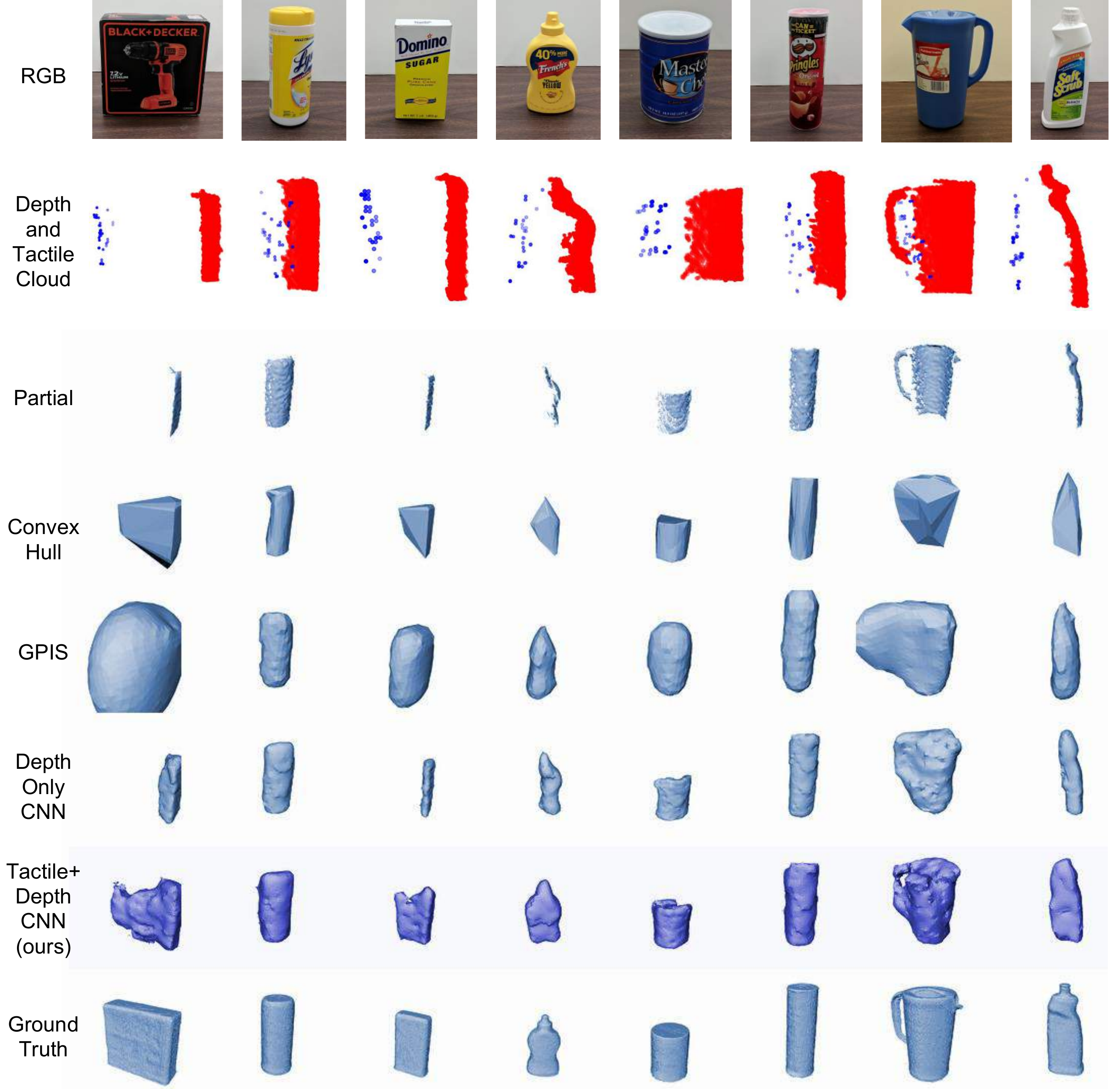}
\caption{The entire \textit{Holdout Live} dataset. These completions were all created from data captured from a real Kinect and a real Barrett Hand attached to a Staubli Arm. The \textbf{Depth and Tactile Clouds} have the points captured from a Kinect in red and points captured from tactile data in blue. Notice many of the \textbf{Depth Only} completions do not extend far enough back but instead look like other objects that were in the training data (ex: cell phone, banana). The \textbf{Depth and Tactile} method outperforms the \textbf{Depth Only}, \textbf{Partial}, and \textbf{Convex Hull} methods in terms of Hausdorff distance and Jaccard similarity. Note that the \textbf{GPIS} completions form large and inaccurate completions for the Black and Decker box and the Rubbermaid Pitcher, whereas the \textbf{Depth and Tactile} method correctly bounds the end of the box and finds the handle of the pitcher. }
\label{fig:live_completions_full_dataset} 
\end{figure*} 

\begin{table}[t]
	\centering
    \begin{tabular}{|c|c|c|c|c|c| }
    \hline
    \multicolumn{1}{|c|}{\begin{tabular}[c]{@{}c@{}}\textbf{Completion} \\  \textbf{Method}\end{tabular}} 
    & \multicolumn{1}{c|}{\begin{tabular}[c]{@{}c@{}}\textbf{Partial}\end{tabular}} 
    & \multicolumn{1}{c|}{\begin{tabular}[c]{@{}c@{}}\textbf{Convex} \\ \textbf{Hull}\end{tabular}} 
    & \multicolumn{1}{c|}{\begin{tabular}[c]{@{}c@{}}\textbf{GPIS}\end{tabular}} 
    & \multicolumn{1}{c|}{\begin{tabular}[c]{@{}c@{}}\textbf{Depth} \\ \textbf{CNN}\end{tabular}} 
    & \multicolumn{1}{c|}{\begin{tabular}[c]{@{}c@{}}\textbf{Depth Tactile}\end{tabular}} 
     \\ \hline
    
        Lift Success (\%) & 62.5\%& 62.5\%& 87.5\%& 75.0\%&\textbf{87.5\%}\\ \hline
        Joint Error ($^{\circ}$) & 6.37$^{\circ}$& 6.05$^{\circ}$& 10.61$^{\circ}$& 5.42$^{\circ}$& \textbf{4.67$^{\circ}$} \\ \hline
        Time (s) & 1.533s & \textbf{0.198s} & 45.536s & 3.308s & 3.391s\\ \hline
    \end{tabular}
    
    \caption{\textbf{Lift Success} is the percentage of successful lift executions. \textbf{Joint Error} is the average error per joint in degrees between the planned and executed grasp joint values. While GPIS and the Depth Tactile method have the same lift success, the Depth Tactile method is 1340\% faster and has 41\% of the joint error, making the process more dependable. (Smaller is better). \textbf{Average time to complete a mesh} using each completion method. While the convex hull completion method is fastest, Depth and Tactile has a superior tradeoff between speed and quality.}
    \label{tab:live_grasp_results} 
\end{table}

\section{Comparison to Other Completion Methods}
\label{sec:Completion_results}

This framework is benchmarked against the following general visual tactile completion methods.\\ 
\indent\textbf{Partial Completion}: The set of points captured from the Kinect is concatenated with the tactile data points. The combined cloud is run through marching cubes, and the resulting mesh is then smoothed using MeshLab's~\cite{cignoni2008meshlab} implementation of Laplacian smoothing. These completions are accurate where the object is directly observed but make no predictions in unobserved areas of the scene. \\
\indent\textbf{Convex Hull Completion}: The set of points captured from the Kinect is concatenated with the tactile data points. The combined cloud is run through QHull to create a convex hull. The hull is then run through MeshLab's implementation of Laplacian smoothing. These completions are reasonably accurate near observed regions. However, a convex hull will fill regions of unobserved space.\\ 
\indent\textbf{Gaussian Process Implicit Surface Completion (GPIS)}: Approximated depth cloud normals were calculated using PCL's KDTree normal estimation. Approximated tactile cloud normals were computed to point towards the camera origin. The depth point cloud was downsampled to size $M$ and appended to the tactile point cloud. A distance offset $d$ is used to add positive and negative observation points along the direction of the surface normal. A sample of the Gaussian process is captured using~\cite{gerardo2014robust} with a $n^3$ voxel grid and a noise parameter $s$ to create meshes from the point cloud. An exhaustive parameter search to determine the values of $M, s, n, d$ is conducted by sampling the Jaccard similarity of GPIS completions where $M=[200, 300, 400]$, $s=[0.001, 0.005]$, $n=[40, 64, 100]$, and $d=[0.005, 0.0005]$. An optimal value of $M=300$ was found to be a good tradeoff between speed and completion quality. Additionally values of $s=0.001$, $d=0.0005$, and $n=100$ were used. 

In prior work~\cite{varley2017shapecompletion_iros}, the Depth Only CNN completion method was compared to both a RANSAC based approach~\cite{papazov2010efficient} and a mirroring approach~\cite{bohg2011mind}. These approaches make assumptions about the visibility of observed points and do not work with data from tactile contacts that occur in unobserved regions of the workspace. 

\subsection{Geometric Comparison Metrics}
Jaccard similarity results are shown in \autoref{tab:Jaccard}. The Jaccard similarity was used to compare $40^3$ CNN outputs with the ground truth. This metric is also used to compare the final resulting meshes from several completion strategies. The completed meshes were voxelized at $80^3$ and compared with the ground truth mesh. The proposed Depth and Tactile method results in higher similarity to the ground truth meshes than do all other described approaches. 

\autoref{tab:Hausdorff} shows the mean values of the symmetric Hausdorff distance for each completion method. The Hausdorff distance metric computes the average distance from the surface of one mesh to the surface of another. A symmetric Hausdorff distance was computed with MeshLab's Hausdorff distance filter in both directions. In this metric, the proposed tactile and depth CNN mesh completions are significantly closer to the ground truth compared to the other approaches' completions. 

Both the partial and Gaussian process completion methods are accurate when close to the observed points but fail to approximate geometry in occluded regions. Through training, the Gaussian Process completion method would often create a large and unruly object if the observed points were only a small portion of the entire object or if no tactile points were observed in simulation. Using a neural network has the added benefit of abstracting object geometries, whereas the alternative completion methods fail to approximate the geometry of objects which do not have points bounding their geometry. 

\subsection{Grasp Comparison in Simulation}

To evaluate the framework's ability to enable grasp planning, the system was evaluated in simulation using the same set of completions. The use of simulation allowed for the quick planning and evaluation of 7900 grasps. GraspIt! was used to plan grasps on all the completions of the objects by uniformly sampling different approach directions. These grasps were then executed not on the completed object but on the ground truth meshes in GraspIt!. To simulate a real-world grasp execution, the completion was removed from GraspIt! and the ground truth object was inserted in its place. Then the hand was placed 20 cm away from the ground truth object along the approach direction of the grasp. The spread angle of the fingers was set, and the hand was moved along the approach direction of the planned grasp either until contact was made or a maximum approach distance was traveled. Then fingers closed to the planned joint values and each finger continued to close until either contact was made with the object, or the joint limits were reached. 

\autoref{tab:sim_grasp_results} shows the average difference between the planned and realized Cartesian fingertip and palm poses. \autoref{tab:l2_joint_error} shows the difference in pose of the end effector between the planned and realized grasps averaged over the 7 joints of the hand. Using the proposed Depth and Tactile method, the end effector ended up closer to its intended location in both joint space and the palm's Cartesian position versus other completion methods' grasps.

\subsection{Live Grasping Results}

To further evaluate the network's efficacy, the grasps were planned and executed on the Holdout Live views using a Staubli arm with a Barrett Hand. The grasps were planned using meshes from the different completion methods described above. For each of the 8 objects, the arm was used once using each completion method. The results are shown in \autoref{fig:live_completions_full_dataset} and \autoref{tab:live_grasp_results}. The proposed Depth and Tactile method enabled an improvement over the other visual-tactile shape completion methods in terms of grasp success rate and resulted in executed grasps closer to the planned grasps, as shown by the lower average joint error (and much faster than GPIS). 
While the success rate of lifting the tactile and depth CNN was equal to the success rate of the Gaussian Process completion, the proposed Depth and Tactile method constructed an object geometry significantly faster, as shown in \autoref{tab:live_grasp_timings}, and had a much lower average joint error, as shown in \autoref{tab:live_grasp_results}. 
After performing lifts on these 8 objects, this visual-tactile completion method performs better in terms of timing, completion quality, and grasp quality for objects the network has not observed. 

\section{Conclusion}
This chapter explored how to leverage visual-tactile information to predict object geometry better than a single-view. A mobile manipulator or stationary robot with both visual information and tactile information would be able to leverage the aforementioned methodology to improve performance without the need to move upon reaching the object. Utilizing even a few tactile contacts was useful for improvement in grasp posturing and completion accuracy. Additionally, this system was experimentally validated on a dataset representative of household and tabletop objects in simulation and in real-world testing. However, not every robot has access to tactile information. In the next chapter, a robot will utilize capturing multiple views of an object to refine its initial object hypothesis. This will offer more rich information than tactile and take advantage of the mobile based. 


\chapter{Two-View Shape Understanding}
\label{ch:two_view_shape_understanding}

Upon reaching an object conveyed through a panoramic target goal, a robotic agent can leverage its mobile based to capture an additional view of the object. This offers uniquely richer data to provide additional information about an object's geometry. However, it can be difficult to register these views due to odometry error and noise. This chapter explores a novel two-view object completion method that allows the robot to use two unregistered views of the object to improve its shape estimation. 

\section{Introduction}
Shape understanding based on single images is difficult. This was explored previously when utilizing visual and tactile information in \autoref{ch:visual_tactile_manipulation} to complete an object but utilizing two images to complete an object is even more challenging. While a single depth sensor can be moved to capture two views of an object, aligning those views is challenging when odometry noise is high. To solve this, a 3D convolutional neural network is used to enable robust shape estimation by leveraging two unregistered views. This means each image of an object is kept in its respective image frame. This methodology can be used to complete objects with only two views. Providing more accurate reconstructions of objects helps to enable a variety of robotic tasks such as manipulation, collision checking, sorting, and cataloging.

\begin{figure*}[t]
    \centering {
        \includegraphics[width=0.95\textwidth]{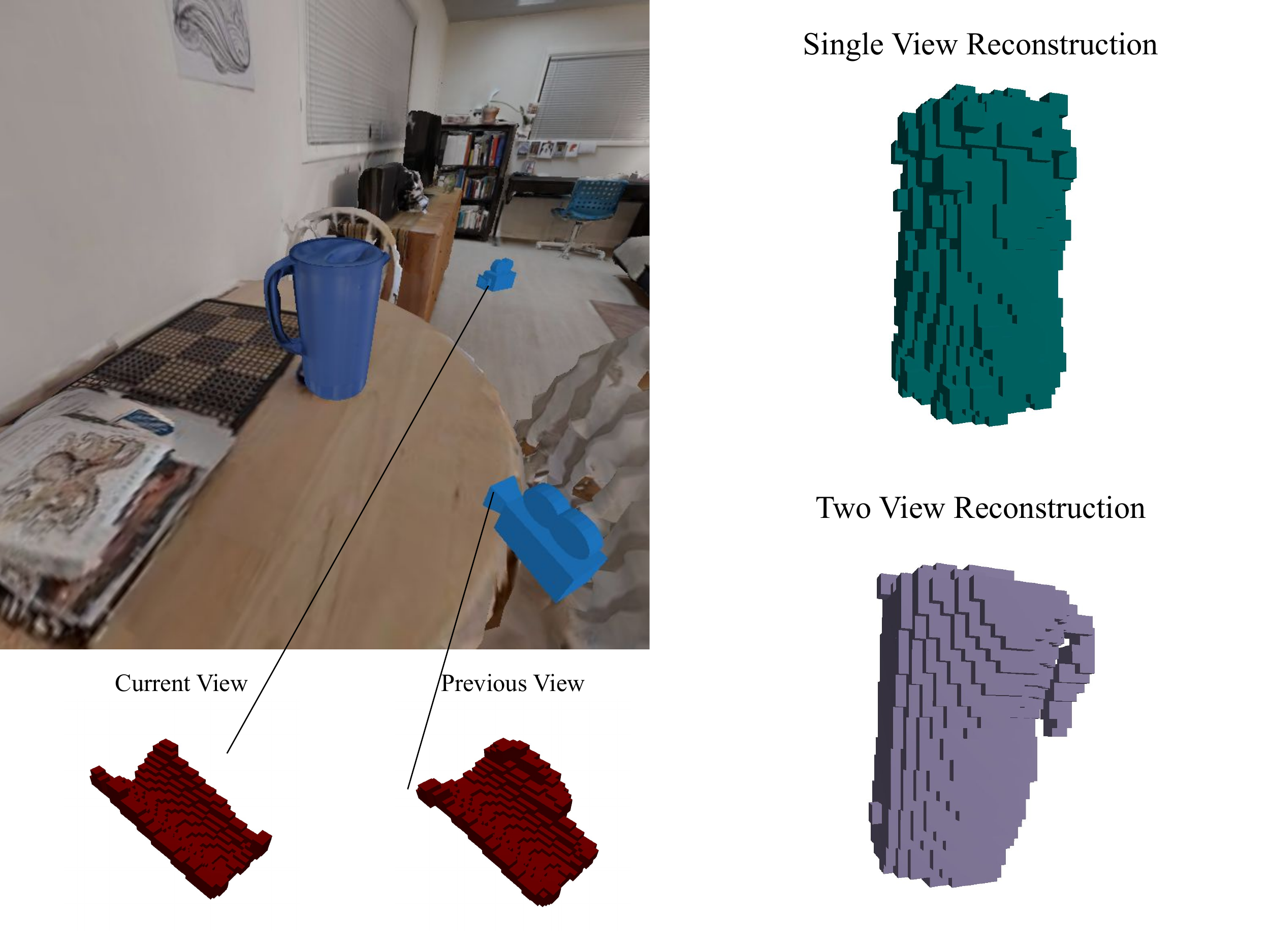}
    }
    \caption{Two views of an object can help refine the prediction. Shown in red are partial views of the target object. A single-view reconstruction of the object is shown in green, and a two-view reconstruction of the object is shown in purple. The two-view prediction correctly captures the handle of the pitcher while the single-view reconstruction misses this critical part of the object geometry, in this case a pitcher from the YCB object dataset~\cite{calli2015ycb}.} \label{fig:two_view_cover_figure}
\end{figure*} 

At runtime, a partial 2.5D image of an object is captured. This first view is passed through a shape completion network to produce a shape estimation of the target object. A second 2.5D view is then captured of the object. These two views are passed into a dual-encoder CNN, called \textbf{two-view split conv}, which encodes each view separately. These encodings are then added together and passed through a CNN decoder to produce a final voxel reconstruction. An example of this reconstruction is shown in \autoref{fig:two_view_cover_figure}. 

Training a neural network to extract registration information from two disparate views allows for a variety of mobile robots to estimate object geometry, such as a drone or two robots within the same environment. Additionally, the utilization of a richer decoder architecture inspired by work from Yang et al.~\cite{Yang18} has improved reconstruction quality over a single-view~\cite{varley2017shape}. An application of this methodology for image reconstruction is shown in Appendix~\ref{app:mnist_reconstruction}.

\section{Methodology}

\begin{figure}[t]
    \centering {
        \includegraphics[width=0.95\linewidth]{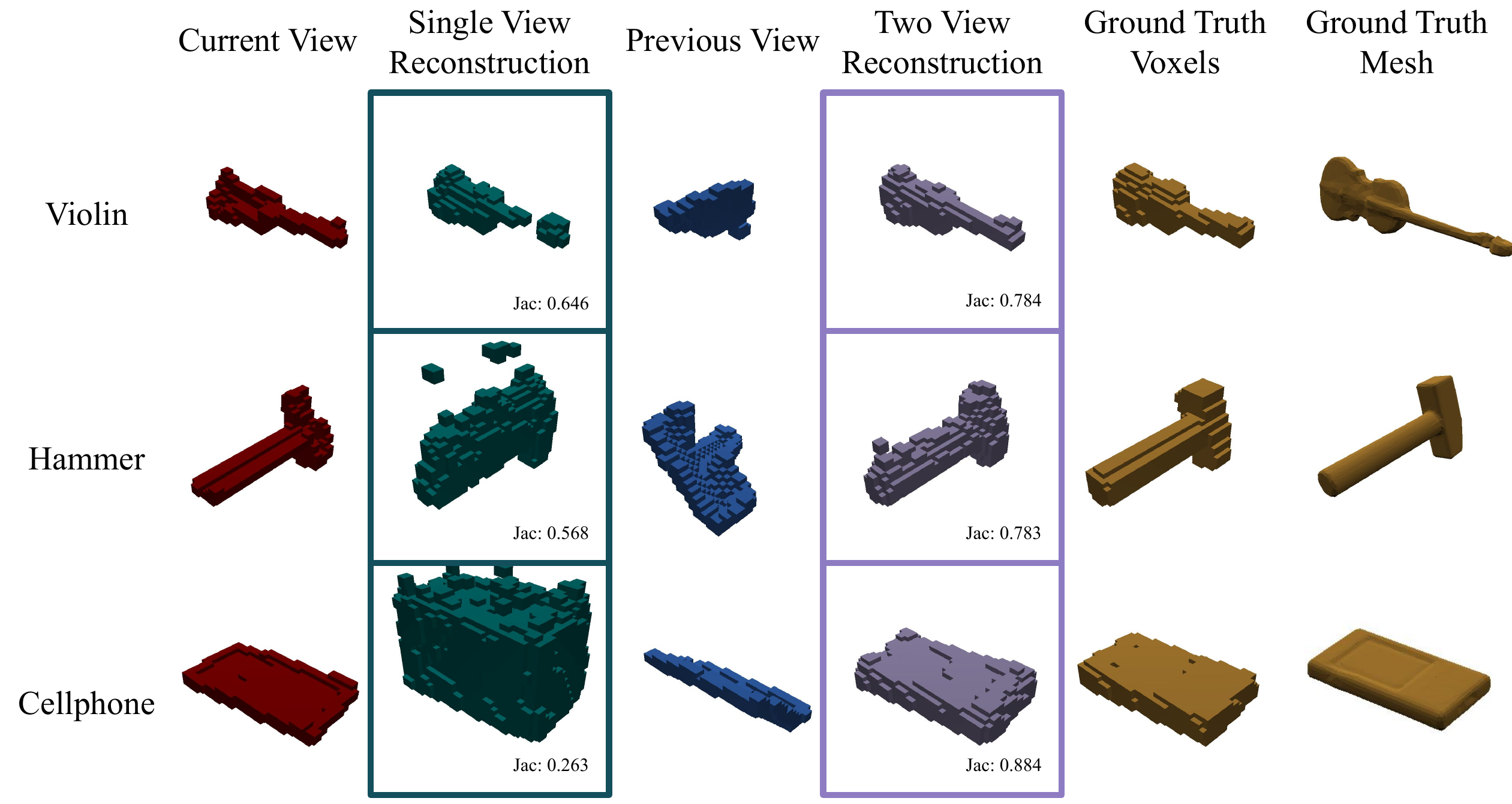}
    }
    \caption{Two-view reconstructions showing that two views are better than one. Current input (red), a single view reconstruction (green), the previous view of the object (blue), the \textbf{two-view split conv} completion (purple), and finally the ground truth mesh (yellow). These meshes were not observed during training for either network. A higher Jaccard score is better. All meshes are from the Grasp dataset~\cite{bohg2014data}. }\label{fig:ch4completion_improvement}

\end{figure}

The CNN architecture in \autoref{ch:visual_tactile_manipulation}~\cite{watkins2019multi} does not address how to combine unregistered information. The tactile information captured on the occluded side of the object was superimposed onto the input. To make sure the mobile robot does not need to localize at runtime, a new CNN architecture is designed that takes two views, the current and previous, that are both kept in their original image frame. As input, it takes two voxelized partial views and outputs a voxelized reconstruction of the original object geometry. This model outputs a voxel grid of occupancy scores. The occupancy scores are thresholded at $0.5$ and perform marching cubes~\cite{lorensen1987marching} to mesh the resultant voxel grid. Example completions and their improvements are shown in \autoref{fig:ch4completion_improvement} which figure demonstrates that a second view can offer significant improvement over a single view. 


\subsubsection{Prediction Refinement}
\label{sec:prediction_refinement}
An additional improvement over previous work is that known empty voxels are marked as empty at the reconstruction step. This ensures full utilization of the input image. This empty check can only occur for the current view, as the previous image is not aligned with the frame the object is being completed in. The point cloud from the previous capture in its image frame and the current view of the object in its image frame are both voxelized into $40^3$ voxel grids. A hypothesis about the object's geometry is generated in the current view image frame and produce a $40^3$ voxel grid of occupancy scores given the two views of the object where $1$ is filled and $0$ is unfilled. Voxel grid certainties are turned into a mesh by thresholding occupancy scores at a decision boundary of $0.5$. Then any voxels that are known to be empty are marked by tracing a ray from the camera vector to points in the voxel grid and any voxels along that line that are not occluded are marked as empty. This fixes any erroneous shapes on the visible side of the object.

\subsubsection{CNN Architecture}
\label{sec:two_view_cnn_architecture}
\begin{figure}[t]
    \centering {
        \includegraphics[width=0.95\linewidth]{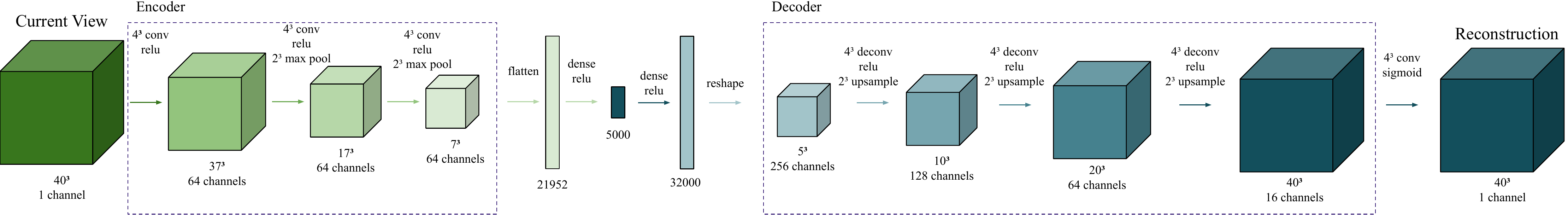}
    }
    \caption{A single-view shape reconstruction architecture that takes advantage of 3D deconvolutional layers to reconstruct the object. Each section of the network is divided into an encoder, embedding, and decoder to be used in other model architectures. This will help control for a variety of different variables when comparing different models. } \label{fig:singleviewdeconvarchitecture}
\end{figure}

The single view architecture used in \autoref{ch:visual_tactile_manipulation}~\cite{watkins2019multi} had two shortcomings. The first was that the reconstruction did not take advantage of deconvolutional layers. The second is that it did not leverage two encoders to incorporate the disparate information coming from multiple views. A proposed improved single-view architecture is shown in \autoref{fig:singleviewdeconvarchitecture} that addresses the first problem. This network architecture is helpful because it defines the beginning of the network as an \textit{encoder}, the middle as the \textit{embedding}, and the end as a \textit{decoder}. These building blocks can be utilized to build better architectures for shape reconstruction. 

\begin{figure}[t]
    \centering {
        \includegraphics[width=0.95\linewidth]{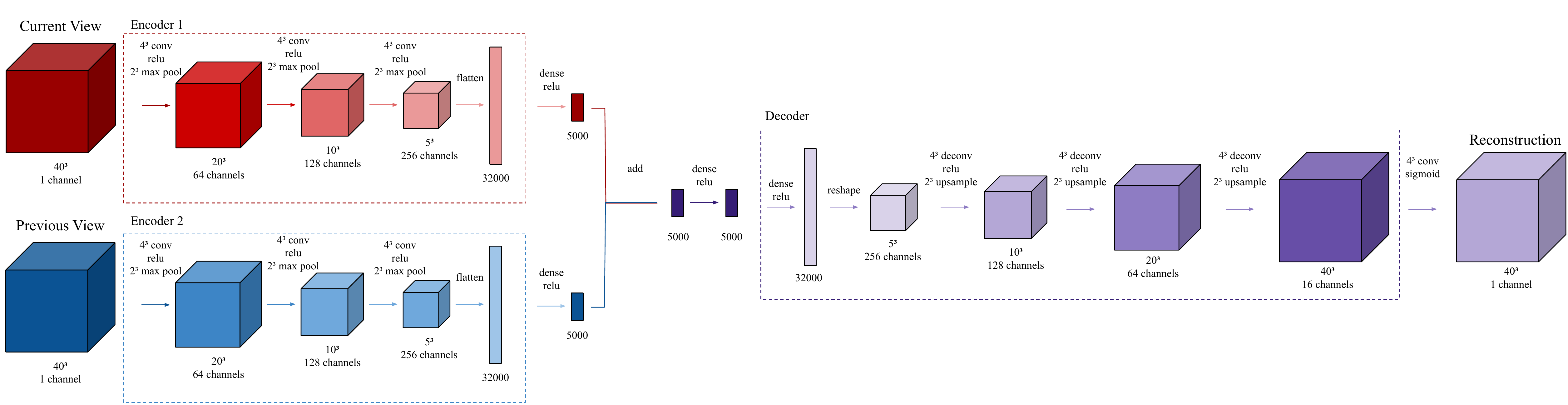}
    }
    \caption{The \textbf{two-view split conv} network architecture takes two unregistered views of an object to produce an accurate reconstruction of an object. To improve on the previous single-view deconvolutional model shown in \autoref{fig:singleviewdeconvarchitecture}, two encoders separately process each view of the object, produce a dense embedding, and are added together to produce a reconstruction of the object. Each view improves the overall reconstruction of the object, but the reconstruction appears in the frame of the current view. } \label{fig:twoviewarchitecture}
\end{figure}

One such better architecture would be to use two encoders for each view and adding their resultant embeddings together. This proposed \textbf{two-view split conv} architecture is shown in \autoref{fig:twoviewarchitecture}. Each encoder takes a $40^3$ voxelized view of the object. Each created by voxelizing a point cloud generated from a 2.5D depth image. All intermediate activation functions are ReLU, and the output activation function is sigmoid. A sigmoid is chosen as it outputs a value between $0$ and $1$.

An encoder is defined as a series of convolutional layers which are then flattened. In the proposed implementation a convolutional layer with kernel size of $4^3$, stride of $1$, and $64$ kernels followed by a max pool of $2^3$ results in an intermediate representation of $20^3\times 64$. Then a convolutional layer with kernel size $4^3$, stride of $1$, and $128$ kernels followed by a max pool of $2^3$ creates a new intermediate of $10^3\times 128$. A final convolutional layer with kernel size $4^3$, stride of $1$, and $256$ kernels followed by a max pool of $2^3$ creates a new intermediate of $5^3\times 256$. This intermediate is flattened to form a vector of size $32000$. Each encoder is followed by a dense layer of size $5000$. 

Two encoders are used, the first being the current view encoder and the second being the previous view encoder. The $5000D$ of both encoders are added together to form a new $5000D$ vector. A second dense layer of size $5000$ processes the addition of these two vectors. These two dense embeddings are added together because it forces the network to allocate a fixed buffer to reconstruction. The alternative would be to concatenate the two vectors, but that will incur additional cost in memory and time to compute. The concatenate operation also implies that each input is equally valuable in reconstruction, however the main input relevant for reconstruction is the current view of the object. The network should learn this importance through training. A final dense layer of size $32000$ is then used.

This embedding is passed into a series of convolutions, which is called a decoder. The new embedding of size $32000$ is reshaped into a vector of size $5^3\times 256$. A convolutional layer with kernel size $4^3$, stride of $1$, and $128$ kernels followed by an upsample 3D of $2^3$ creates a new intermediate of $10^3\times 128$. A convolutional layer with kernel size $4^3$, stride of $1$, and $64$ kernels followed by an upsample 3D of $2^3$ creates a new intermediate of $20^3\times 64$. A convolutional layer with kernel size $4^3$, stride of $1$, and $16$ kernels followed by an upsample 3D of $2^3$ creates a new intermediate of $40^3\times 16$. Finally, a convolutional layer with kernel size $4^3$, stride of $1$, and $1$ kernel creates the final reconstruction of size $40^3\times 1$. This final convolutional layer is the reconstruction of the object with a sigmoid activation function. This model is trained using binary cross entropy loss and the Adam optimizer. 

The model was trained for $151300$ batches with a batch size of $8$ for a total of $10$ hours and $49$ minutes of training time on a NVIDIA $3090$ graphics card. Training was subject to early stopping where if the validation Jaccard similarity did not increase for $5$ epochs, the training would stop. 

\section{Experiments}
To validate that this proposed network works, tests on the ability to address occlusion are required. To validate the \textbf{two-view split conv} model, a series of ablations are performed on the network to validate an improvement in performance. A series of views of different objects needs to be created to train this model and validate it. 

\subsection{Dataset Generation}
\label{sec:fixed_voxel_data_generation}

\begin{figure}
\centering
\begin{subfigure}{.5\textwidth}
  \centering
  \includegraphics[width=.95\linewidth]{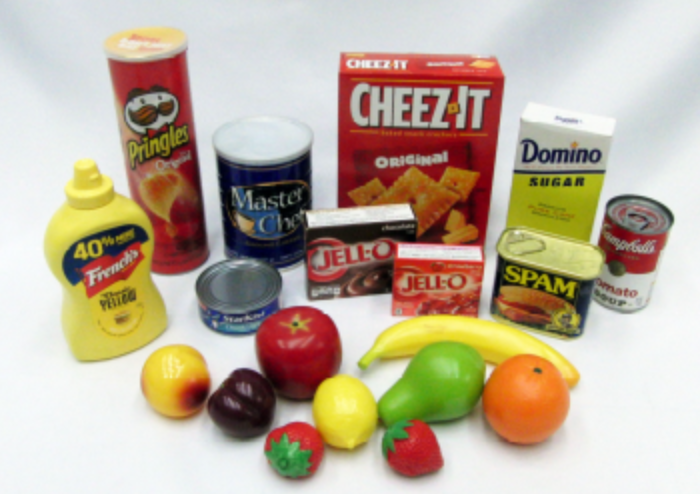}
  \caption{YCB}
\end{subfigure}%
\begin{subfigure}{.5\textwidth}
  \centering
  \includegraphics[width=.95\linewidth]{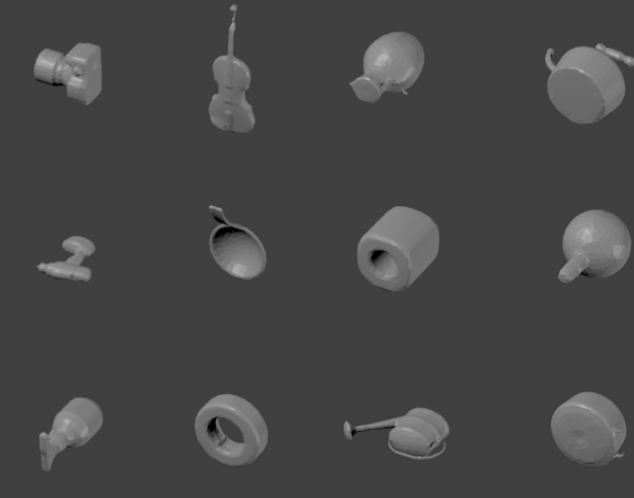}
  \caption{GRASP}
\end{subfigure}
\begin{subfigure}{1.0\textwidth}
    \centering
    \includegraphics[width=0.95\linewidth]{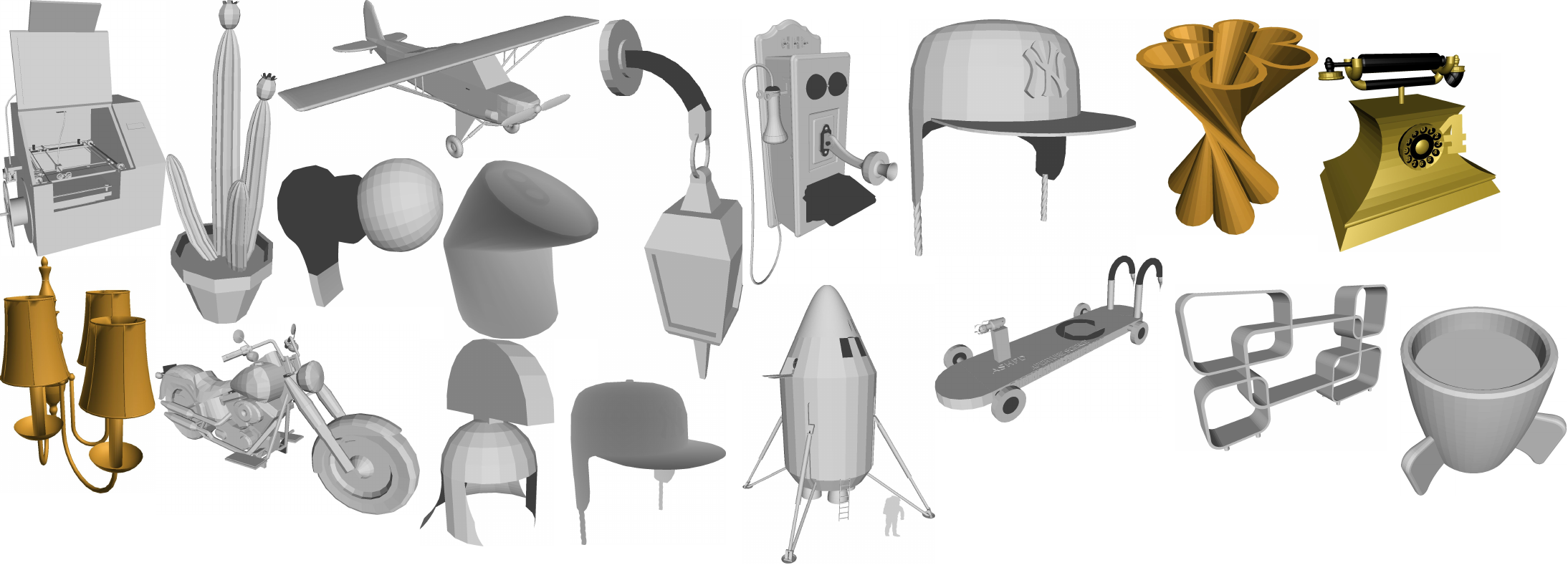}
    \caption{Challenge}
\end{subfigure}
\caption{Objects from both the YCB, GRASP, and challenge datasets. Each object fits within a $0.3m^3$ bounding box and has valid graspable positions using both a Fetch gripper and BarrettHand. These objects are representative of household objects that are useful for validating that a grasp planning solution works.}
\label{fig:all_object_datasets}
\end{figure}

Shape understanding is only made possible through datasets of realistic objects with a variety of different geometries. Having real world data to train a network is more useful than trying to learn purely from synthetic meshes. To that effect, the results in this chapter are created from three datasets: the YCB object dataset~\cite{calli2015ycb}, the GRASP database dataset~\cite{bohg2014data}, and a selection of objects from ShapeNet~\cite{shapenet}. The objects from ShapeNet were specifically selected for their asymmetrical geometry and self-occluding properties. These ShapeNet objects are therefore called the \textit{challenge} dataset. A selection of objects from each dataset are shown in \autoref{fig:all_object_datasets}. 

For each object in this dataset, RGB and depth images were captured using $726$ views per mesh. $726$ views are calculated by enumerating Euler rotation angles. Each of the \textit{roll}, \textit{pitch}, and \textit{yaw} values were calculated over a half-open interval as follows:
\begin{flalign*}
& roll  &{}={}& [0, 0.6, 1.2, ...,2\pi)          & \\
& pitch &{}={}& [-\pi/2, -\pi/2+0.6, ..., \pi/2) & \\
& yaw   &{}={}& [0, 0.6, 1.2, ...,2\pi)          &
\end{flalign*}
\noindent While different rotational values could have been chosen, these provided a representative set of views that a camera would capture in a lab setting. Each view is captured using an OpenGL renderer taken from $0.5m$ away with a camera width and height of $480\times 480$ and a field of view of $45^\circ$. These values were chosen because the Fetch robot uses a PrimeSense camera with resolution $640\times 480$ and field of view of $45^\circ$. The assumption is that at runtime the output image from the Fetch's camera would be cropped to fit the same window size and that the object would be aligned with the center of the image. 

Each of these $726$ views were then voxelized into a $40^3$ voxel grid. In previous work, Varley et al.~\cite{varley2017shape} used a variable voxel resolution by first calculating the bounding box of the ground truth object with the given rotation and then using those bounds to voxelize the input point cloud. To reconstruct the object at runtime the researchers had to voxelize the input cloud and then offset that cloud by a fixed voxel count to ensure the output had enough room for the occluded part of the object. However, this strategy would fail to get a comprehensive completion if not enough room were provided behind the object, such as when only one face of a cube is visible. 

\begin{figure*}[t]
    \centering {
        \includegraphics[width=0.95\textwidth]{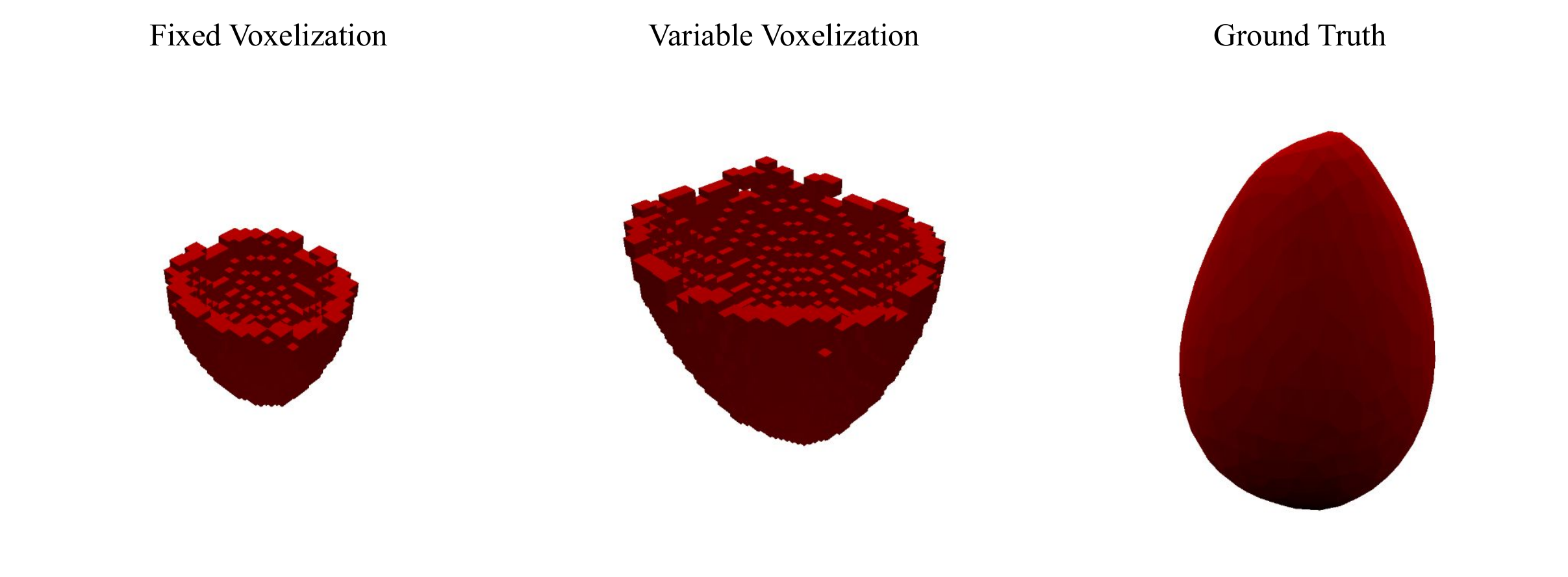}
    }
    \caption{The fixed voxelization strategy provides a constant size voxel and the input does not have any holes. The variable resolution voxelization has holes and does not leave enough room for completing the back side of the mesh, in this case an avocado from the GRASP database dataset. Both images are renders of the voxelized point cloud of the avocado with the camera being the below the voxels with Z up. } \label{fig:variable_versus_fixed_voxelization}
\end{figure*}

To address this, one can center the bounding volume in the $x, y$ of the input point cloud, move the $z$ to be one voxel away from the closest point in the cloud, and use a fixed voxel resolution. This offers some great benefits. The first is that there will always be enough room to complete the object if the object fits within the fixed bounding volume for any rotation. The second is that there is a well-defined function from an input point cloud to the ground truth object. A downside is that there is usually a lot of empty space. While for $40^3$ that is not common, at higher resolutions this voxelization strategy can make it harder for the network to learn object geometry and produce empty outputs. A potential future direction could be addressing a hybrid approach between variable and fixed voxelization. The general approach of the proposed fixed voxelization strategy is as follows:
\lstset{language=Pascal}          
\begin{lstlisting}[frame=single,basicstyle=\tiny]
depth := From Camera
points := depth_to_point_cloud(depth)
voxel_scale := [0.3/(40-2), 0.3/(40-2), 0.3/(40-2)]
xyzmin := [middle(points[:, 0]), middle(points[:, 1]), min(points[:, 2])]
voxel_grid := voxelize(points, xyzmin, voxel_scale)
\end{lstlisting}
\noindent With this voxelized point cloud, there is also a defined bounding box that can be used to voxelize the ground truth mesh. Patrick Min's Binvox~\cite{binvox, nooruddin03} was used to voxelize each mesh according to the rotation of the object. An example of the difference in these voxelization strategies is shown in \autoref{fig:variable_versus_fixed_voxelization}.

An important consideration is the fixed bounding volume and what resolution to choose. The reason a bounding volume of $0.3m^3$ was chosen is that a 3D printer accessible to this work, the Artillery Sidewinder X1, has a build volume of $300mm\times 300mm\times 400mm$. A different build volume could have been chosen by changing the voxel resolution. All models in this chapter use a $40^3$ input and output resolution. This means that each voxel corresponds to a $7.5mm^3$ volume. The YCB object dataset initially comes with $77$ object meshes, the GRASP dataset with $590$ object meshes, and ShapeNet has over $51000$ object meshes. While using all of these meshes would make for a more robust CNN model to reconstruct object geometry, each additional object adds $726$ views to the training dataset which increases the training time. Additionally, the meshes are sometimes too small to be rendered and voxelized for training. To assuage this issue, objects that would not fit inside of a bounding volume of $0.3m^3$ were not used. Additionally, objects that resulted in a ground truth voxelization smaller than $64$ voxels were omitted. $55$ objects in the YCB dataset, $463$ in the GRASP database dataset, and $23$ in the challenge dataset were valid. The challenge dataset is a special case where the objects were resized from their original sizes to fit within the fixed width bounding box. More objects, such as from the Thingi10K~\cite{thingi10k}, could have been used as training data. Ultimately choosing this dataset is about picking objects that are representative of the unseen geometry expected at runtime. The work in this chapter seeks to address completion of objects for household grasping, but in future work can be extended to estimate the geometry of a variety of object geometries. 

Through this process, a total of $392766$ train pairs have been generated. Each of these are then split into four categories: trained object train views, trained object holdout views, holdout object validation, holdout object test. First objects are put into either the train or holdout object categories with an $80/20$ split. The challenge dataset is always kept in the holdout set. The trained object set is then split into the train views and holdout views set with a $90/10$ split. The holdout views set is used during training to evaluate the performance of the model on unseen views. The holdout objects are split into the validation and test sets with a $50/50$ split. The validation set is used to select which model performs best and the test set is used to evaluate the overall performance. These dataset splits were consistent among the testing of all model architectures. 

For model architectures that take multiple views, a selection of random views from the same split were chosen. For example, if a model was trained using $12$ views, the $12$ views were sampled from the train object train views split, evaluated on $12$ views in the holdout views set, evaluated on $12$ views in the holdout models validation set, and tested once on $12$ views from the holdout models test set. Theoretically using more than $726$ views would result in more variety of the training data; however, the results will show that these splits are enough to show generalized performance. 

\subsection{Two-View Ablation}

To evaluate the qualitative performance of the \textbf{two-view split conv} architecture, it is compared to a series of different models with tweaks on the original architecture. The \textbf{two-view split conv} model used here differs from the single-view model used in a previous work~\cite{watkins2019multi}. To ablate two-view architectures, different test cases are evaluated as follows:
\subsubsection{Two-View Same-Views}
The same view of an object is passed into the \textbf{two-view split conv} model architecture to illustrate how much performance is gained from adding additional weights to the network over the single-view model architecture. This test is easily adapted from the existing \textbf{two-view split conv} model architecture by training a model using the same data twice. 
\subsubsection{Two-View Joined}
\begin{figure}[t]
    \centering {
        \includegraphics[width=0.95\linewidth]{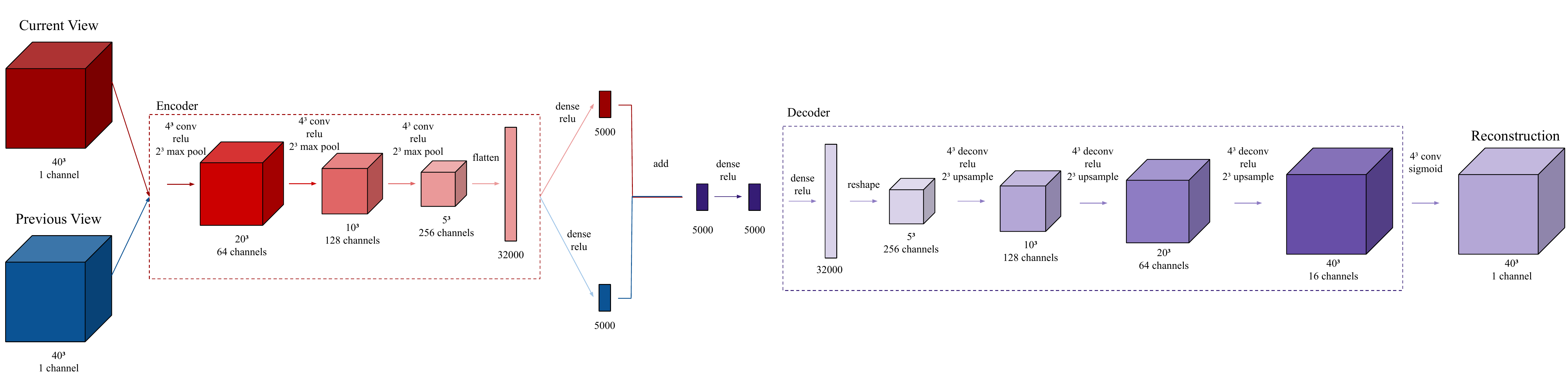}
    }
    \caption{The \textbf{two-view joined} architecture reuses the same encoder for each view. This prevents the network from differentiating the input from each other and making it difficult to complete an object in the current view frame. }\label{fig:two_view_joined_architecture}

\end{figure}

The use of different encoders for the \textbf{two-view split conv} model is not necessarily the most intuitive solution. Additional weights and compute time would be saved if the same encoder were used for each view. The new completion will not be able to differentiate between which view is "current" and which is "previous" as the weights are shared between the two encoders. This is contrary to the "split" model described previously, where each encoder contains its own weights. The architecture for this model is shown in \autoref{fig:two_view_joined_architecture}.

\subsubsection{Two-View Split Dense}
\begin{figure}[t]
    \centering {
        \includegraphics[width=0.95\linewidth]{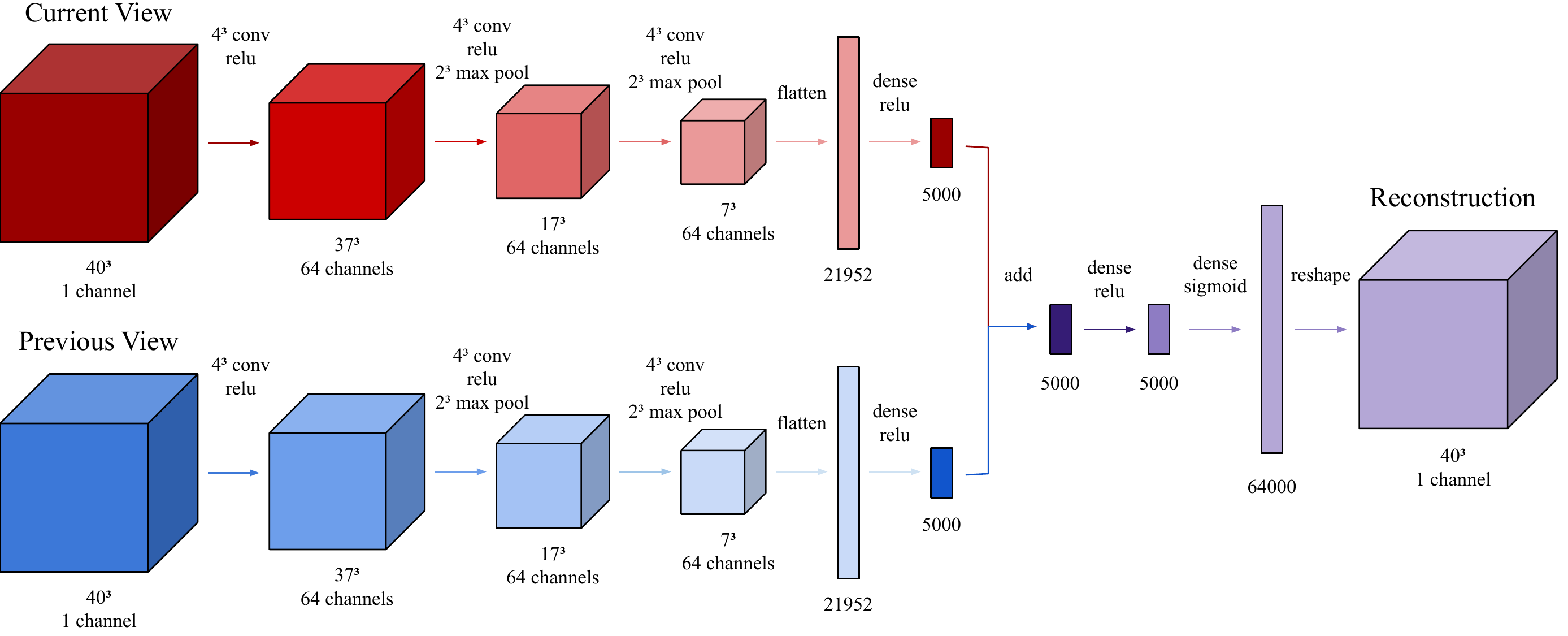}
    }
    \caption{The \textbf{two-view dense} architecture taking two separate unregistered views and reconstructing the object through a dense reconstruction. } \label{fig:twoviewarchitecture_dense}
\end{figure}

In previous work~\cite{varley2017shape}, the encoder was represented by a dense $64000D$ vector that was then reshaped into a $40^3$ voxel grid. To compare the improvement of using deconvolutional layers over a dense reconstruction, an additional model is trained using two views which are then reconstructed using this dense reconstruction layer. The architecture used is shown in \autoref{fig:twoviewarchitecture_dense}.

\subsubsection{Single-View Dense}

To address the more recent work in multi-view reconstruction, it is best to review the work done by Varley et al.~\cite{varley2017shape} The original single-view reconstruction architecture used a CNN which convolved the input into a dense embedding and then reconstructed it using a dense layer reshaped into the $40^3$ completion shape. The point cloud in the image frame is voxelized into a $40^3$ voxel grid. It then generated a hypothesis about the object's geometry and produce a $40^3$ voxel grid of occupancy scores given that initial view of the object where $1$ is filled and $0$ is unfilled. As mentioned previously, the original voxelization strategy used variable resolution voxels. 

\subsection{Evaluation}
A collection of holdout views of training objects is reserved along with a collection of models not seen during training with generated views. Each view is completed, and then compared against the ground truth object for reconstruction quality. For two-view reconstruction methods, a random second view is provided. 

\label{sec:ch5reconstructionqualitymetrics}
The first goal was to validate that getting two views would result in a greater chance of grasping an object. There are three metrics that let us validate the hypothesis without building the entire pipeline: \textbf{Jaccard similarity}, \textbf{Hausdorff distance}, and \textbf{grasp joint error}. 
\begin{enumerate}
    \item \textbf{Jaccard similarity} Jaccard similarity is used to evaluate the similarity between a generated voxel occupancy grid and the ground truth. The Jaccard similarity between sets A and B is given by:
    \[
    J(A, B) = \dfrac{|A\cap B|}{|A\cup B|}
    \]
    The Jaccard similarity has a minimum value of 0 where A and B have no intersection and a maximum value of 1 where A and B are identical~\cite{jaccard}.
    \item \textbf{Hausdorff Quality} The Hausdorff distance is a one-direction metric computed by sampling points on one mesh and computing the distance of each sample point to its closest point on the other mesh. It is useful for determining how closely related two sets of points are~\cite{huttenlocher1993comparing}.
    \item \textbf{Grasp Joint Error} GraspIt!~\cite{miller2004graspit} is used to plan a grasp on the reconstructed mesh, then execute that grasp on the ground truth mesh using a simulated Barrett hand. The average joint error is computed between the planned grasp and the realized grasp in simulation.
    
\end{enumerate}

\section{Results}


\begin{figure}[ht!]
    \centering {
        \includegraphics[width=\linewidth]{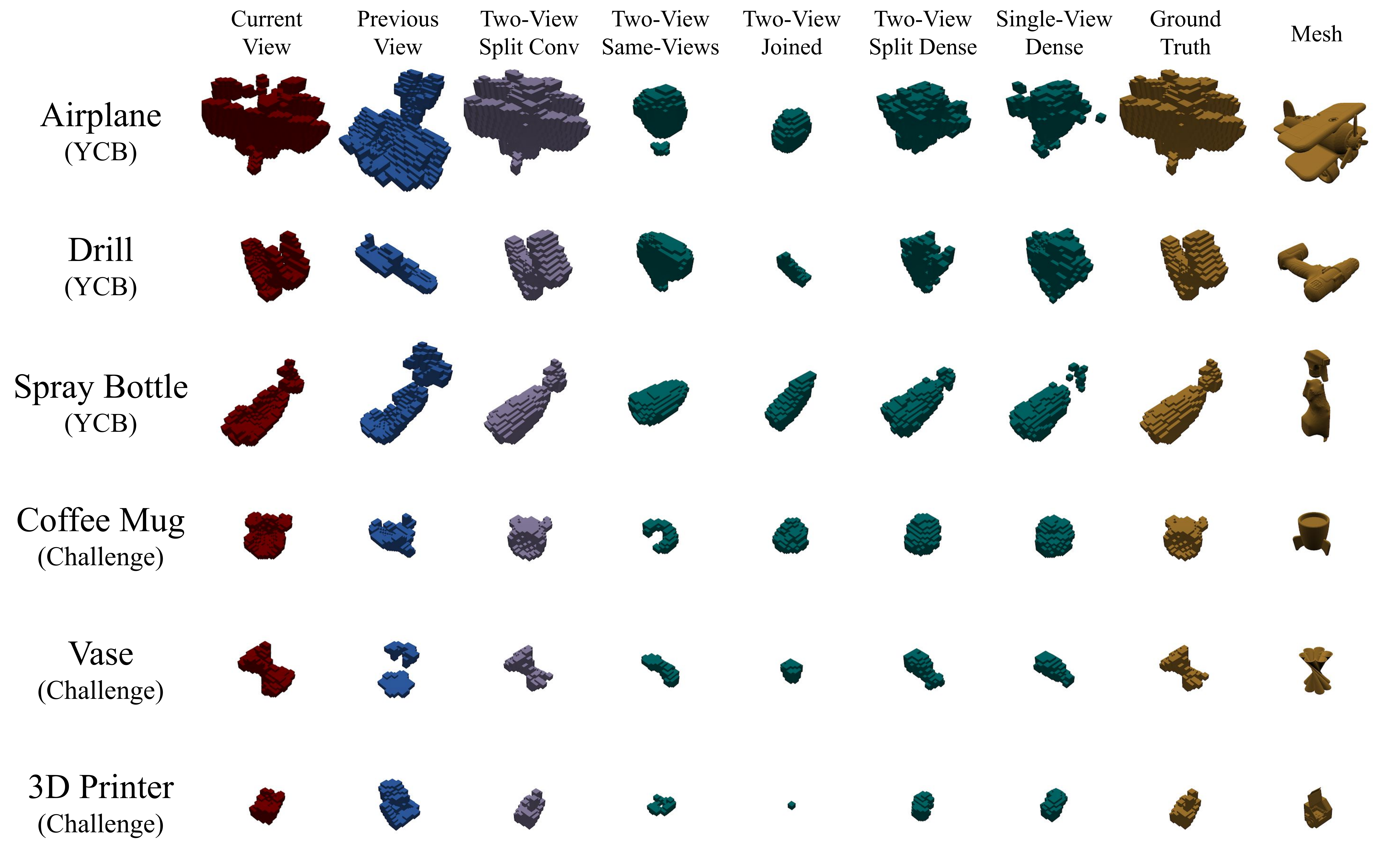}
    }
    \caption{The reconstruction results from holdout meshes in the two-view ablation. Shown in red is the current view of the object and in blue is the previous view of the object. Shown in purple is the completion of the \textbf{two-view split conv} reconstruction. In green are completions from the four ablations: \textbf{same-view two-view}, \textbf{two-view joined}, \textbf{two-view split dense}, and \textbf{single-view dense}. On the right in yellow are the ground truth voxelization and mesh. None of these meshes were observed during training for these models. The \textbf{two-view split conv} result shows a substantial improvement of completion quality over the ablated completions for unseen objects. } 
    \label{fig:two_view_ablation}
\end{figure} 

The ablation provided context into which architectures perform best for two-view completions. Example completions are shown in \autoref{fig:two_view_ablation}. The \textbf{two-view split dense}, \textbf{two-view split conv}, and same-view split models were all able to learn how to reconstruct object geometry. The \textbf{two-view joined} model was unable to learn which frame to complete the object in and would instead predict object geometry that overlaid both inputs over each other. The two-view split dense model had sharper features over the convolutional version. This is likely due to the lack of structural abstraction provided by the addition of convolutional layers. The same-view model did not produce better completions over a single view representation. Some models output no result due to the sparsity of the input. The confidence of these outputs were close to the decision boundary of $0.5$ and therefore did not result in a valid completion. 

\begin{figure}[ht!]
    \centering
    \begin{subfigure}{.5\textwidth}
      \centering
      \includegraphics[width=.95\linewidth]{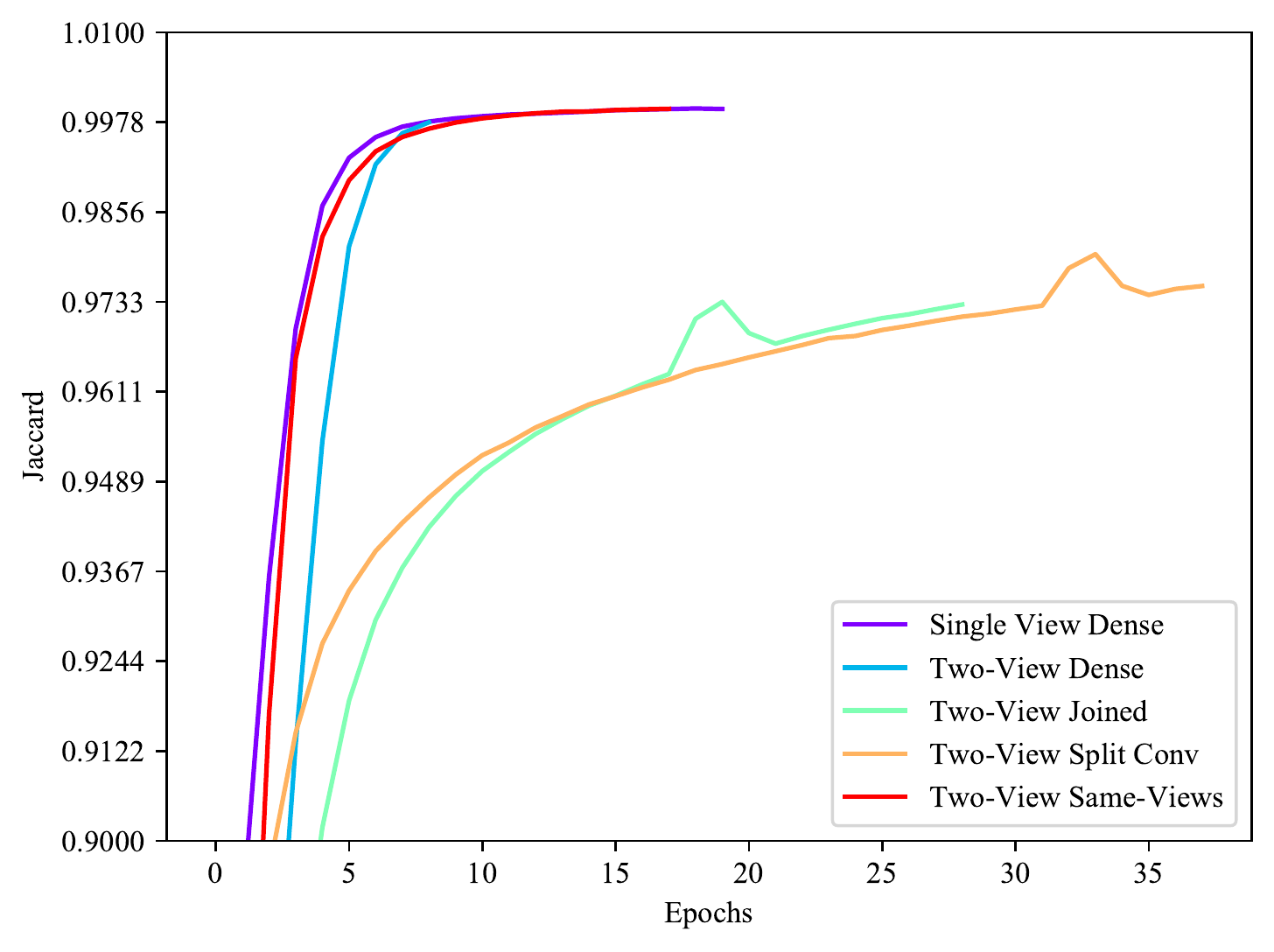}
      \caption{Training Objects Reconstruction Quality Per Epoch}
    \end{subfigure}%
    \begin{subfigure}{.5\textwidth}
      \centering
      \includegraphics[width=.95\linewidth]{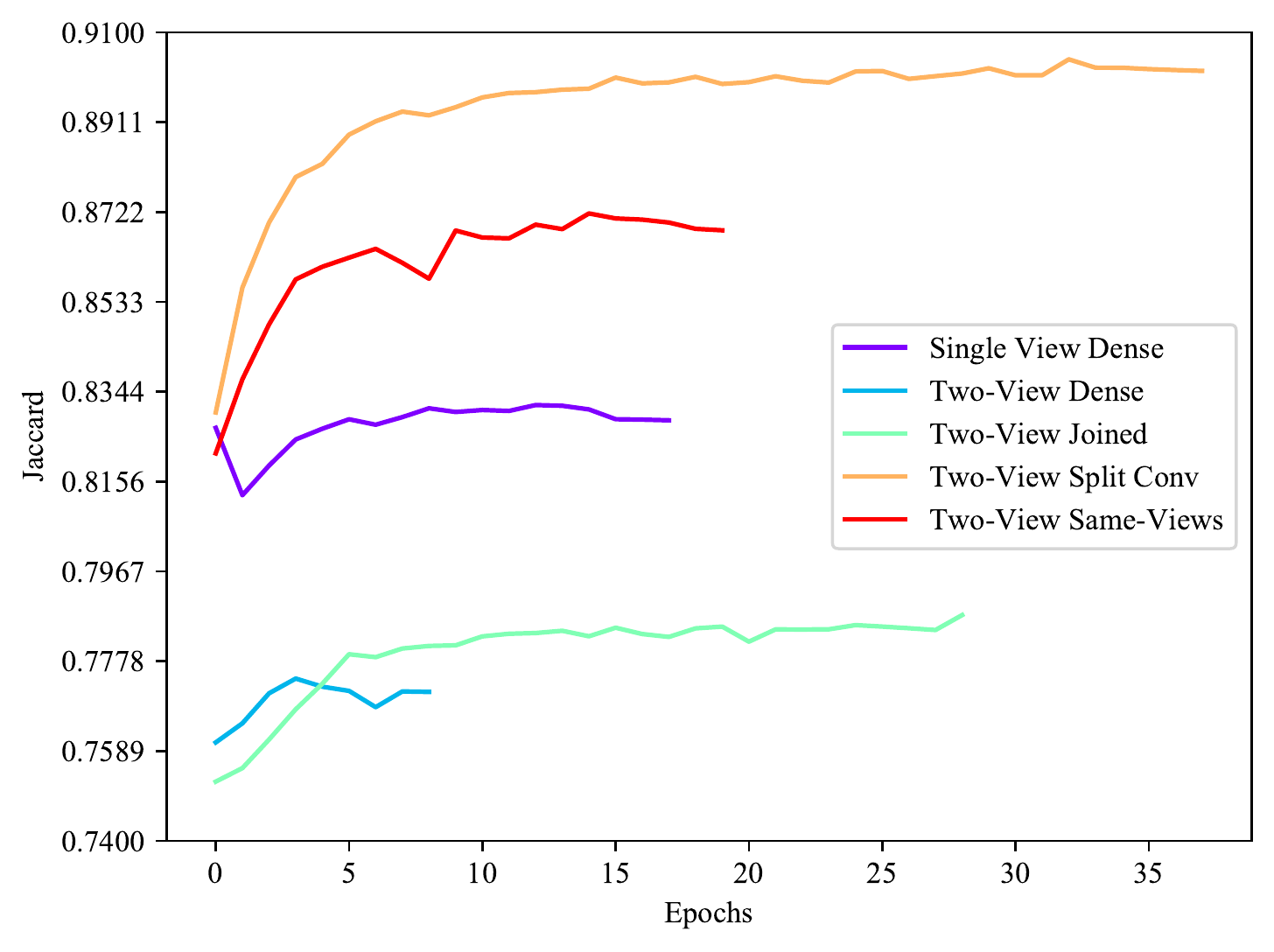}
      \caption{Unseen Objects Reconstruction Quality Per Epoch}
    \end{subfigure}
    \caption{(a) shows the Jaccard similarity over time for observed objects as the networks in the ablation study trained and (b) shows the Jaccard similarity over time for objects in the holdout dataset as the networks in the ablation study trained. The proposed two-view split model performs best at generalizing to unseen objects as shown in (b). A higher Jaccard is better. }
    \label{fig:two_view_training_metrics}
\end{figure}

The two-view ablation training and validation metrics as a function of epochs is shown in \autoref{fig:two_view_training_metrics}. For training data, the \textbf{single-view dense}, \textbf{two-view split dense}, and \textbf{two-view same-views} split models perform significantly better. This is because these models are best at memorization of training data. However, these models fail to generalize to unseen objects well. The \textbf{two-view split conv} based model shows a substantial improvement in Jaccard quality for unseen objects. The \textbf{two-view same-views} model is the second-best performing model, but without a novel second view it fails to achieve the same level of generalization. Each model has different numbers of epochs used to train it. Further testing could be done to further optimize the level of training uniformity across models. 

\begin{table*}[ht!]
    \centering
    \begin{tabular}{|c|c|c|c|}
    \hline
    \multicolumn{1}{|c|}{\begin{tabular}[c]{@{}c@{}}\textbf{Completion} \\  \textbf{Method}\end{tabular}} 
    & \multicolumn{1}{c|}{\begin{tabular}[c]{@{}c@{}}\textbf{Jaccard} \\  \textbf{}\end{tabular}} 
    & \multicolumn{1}{c|}{\begin{tabular}[c]{@{}c@{}}\textbf{Hausdorff} \\  \textbf{}\end{tabular}} 
    & \multicolumn{1}{c|}{\begin{tabular}[c]{@{}c@{}}\textbf{Grasp Joint} \\  \textbf{Error}\end{tabular}} \\ 
    \hline
        Single-View           & 0.782         & 6.573        & $4.52^\circ$      \\ \hline
    	Two-View Same-Views   & 0.802         & 6.423        & $4.36^\circ$      \\ \hline
    	Two-View Split Dense  & 0.807         & 6.416        & $4.32^\circ$      \\ \hline
    	Two-View Joined       & 0.673         & 8.539        & $7.64^\circ$      \\ \hline
    	Two-View Split Conv   & 0.818         & 6.251        & $3.97^\circ$      \\ \hline
    \end{tabular}
    \caption{\textbf{Two-View Completion Results}, measuring the performance of the reconstruction quality of the test set meshes, including challenge meshes, given a \textbf{single-view dense}, \textbf{two-view same-views}, \textbf{two-view split dense}, \textbf{two-view joined}, and \textbf{two-view split conv}, as described in \autoref{sec:ch5reconstructionqualitymetrics}. A higher Jaccard is better. A lower Hausdorff is better. A lower Grasp Joint Error is better.}\label{tab:two_view_ablation_tests}
\end{table*}

Reconstruction results for test shapes are shown in \autoref{tab:two_view_ablation_tests}. They showed a Jaccard reconstruction of 0.818 for the \textbf{two-view split conv} implementation versus 0.782 for \textbf{single-view dense} for unseen objects. This improvement in Jaccard resulted higher joint accuracy with a grasp joint error of $3.97^\circ$ given a second view. The \textbf{two-view same-views} method had demonstrably worse performance than the proposed \textbf{two-view split conv} network, with a grasp joint error of $4,36^\circ$, showing that adding weights did not improve performance. The \textbf{two-view split dense} model 0.807 did slightly better than a same-view model 0.802 in terms of Jaccard quality, but the addition of a convolutional decoder in the \textbf{two-view split conv} model brought the performance above substantially at 0.818. The \textbf{two-view joined} model was unable to learn a proper reconstruction policy and was restricted to a Jaccard quality of 0.673. 

\section{Conclusion}
This chapter presented a novel approach to combining two unregistered views of an object to create a higher accuracy mesh prediction. A \textbf{two-view split} convolutional neural network method was presented that combines two unregistered views of an object. This method is shown to outperform baseline methods or perform comparably. Additionally, a novel approach to creating training data for shape completion systems was provided. This architecture can be utilized for an end-to-end mobile manipulation pipeline where views cannot be registered together.

\chapter{Mobile Manipulation}
\label{ch:mobile_manipulation}

\section{Introduction}

Improvements in mobile manipulation come via improvements in its subcomponents or the glue that holds those components together. A robotic agent can navigate through an environment via some understanding of its sensory information, and then form an understanding of its environment to act upon it. This concert of systems makes research in mobile manipulation a challenge. Errors in odometry, depth sensor noise, error in neural network output, can all propagate further error resulting in low success rates at completing a task. Tasks such as grasping an object can therefore be quite challenging. This thesis is attempting to solve a component of this larger issue while not giving the robotic agent information about its position in space at runtime. The proposed mobile manipulation pipeline also serves as an integration of previous work in robotic visual navigation and multiple-view shape understanding. 

Why would a roboticist care about mobile manipulation in general? Fully autonomous mobile manipulation means that a robotic agent can operate in task environments that are difficult to be explored by a human being or have difficult terrain making sensory information noisy. It has been an important goal with particular focus on such wide-ranging applications as manufacturing, warehousing, construction, and household assistance~\cite{kalashnikov2018qt, mahler2019learning, jacobus2015automated}. Even individuals looking to have a robotic vacuum in their home are interested in mobile manipulation, as a robot vacuum cleaner is planning a series of navigation steps and vacuuming steps in concert to map its environment, detect litter to be cleaned, plan trips back to its charging station, and empty debris it has collected. These tasks individually may be easier to solve, but together they present many errors that can occur at runtime, such as inaccurate mapping, obstacles that make navigation impossible or introduce a stuck condition and running out of battery before recharging can occur. Optimizing for all these corner cases is an interesting problem that has given rise to a lot of interesting research since robotics started looking at mobile manipulation. This can also be described as active perception, to set up a goal based on some current belief to achieve an action~\cite{bajcsy2018revisiting}. 

The problem, as described in the Introduction, is to allow a mobile robot to navigate through an environment to an object, whereby it will gather as much sensory information as required to manipulate an object, all without knowing its position in space with respect to some global coordinate system. This limitation means that while a mobile robot has the luxury of being able to look at an object from multiple views, the agent cannot easily take advantage of its odometry while accommodating noise in the registration of multiple views. This work in mobile manipulation introduces two techniques to address the lack of odometry at runtime: 1) A novel next-best-view prediction method and 2) A novel panoramic goal image prediction for short range navigation. 

\section{Method}

Designing this mobile manipulation pipeline borrows heavily from the previous navigation and multiple-view shape understanding work described in \autoref{ch:learning_visual_navigation} and \autoref{ch:two_view_shape_understanding}. This chapter describes a system that navigates through an environment to find an object and then captures two views of the object, the second view being a novel next-best-view to gain a better shape prediction to plan a grasping task. As shown in \autoref{ch:two_view_shape_understanding}, getting a more reliable prediction of the shape improves grasp success and therefore utilizing a second view, especially in objects with challenging geometry, improves the success rate of the overall system navigating to the object and grasping it. The novelty in this system comes from its ability to use these two views without registering them at runtime or keeping track of the robot's position. 

\begin{figure}[t]
    \centering {
        \includegraphics[width=0.95\linewidth]{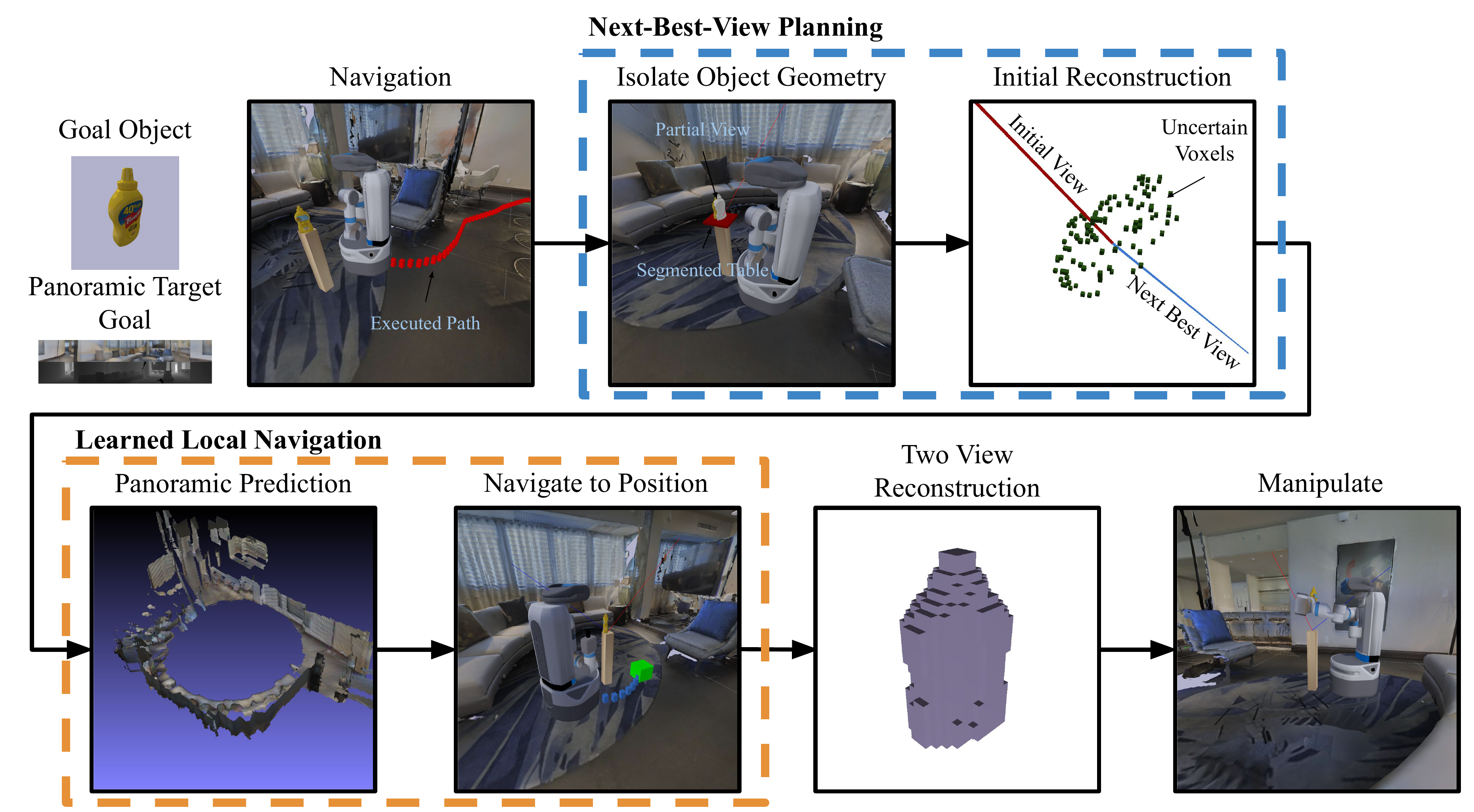}
    }
    \caption{The mobile manipulation system combines navigation to a target goal with shape understanding through a series of discrete stages to decide how to acquire additional information about the object for manipulation. It is a shape completion system that can utilize two unregistered views to get a better model of the target object. } \label{fig:overview2}
\end{figure}

At runtime, the robot is given a panoramic RGBD target goal (8 RGBD images taken at $45\deg$ from each other), such as the one shown in \autoref{fig:overview2}, and an RGBD image of the object of interest. The pipeline then follows the following set of steps to grasp the object:
\begin{enumerate}
    \item The robotic agent navigates to the region described by the panoramic goal
    \item The robotic agent aligns itself with the object and then produces a segmented depth cloud from its RGBD sensor
    \item The robotic agent predicts the geometry of the object using the method described in~\cite{varley2017shapecompletion_iros} 
    \item The robotic agent uses the prediction to determine uncertain portions of the input to determine a next-best-view
    \item The robotic agent captures a panorama of its nearby environment and utilizes it to predict the panoramic goal at the next-best-view location
    \item The robotic agent navigates to this predicted panoramic goal using the learned visual navigation system described in \autoref{ch:learning_visual_navigation}
    \item The robotic agent captures a second view of the object and produces another segmented depth cloud from its RGBD sensor
    \item The robotic agent predicts a shape of the object utilizing the two views kept in their respective image frames using the two-view shape completion method described in \autoref{ch:two_view_shape_understanding}
    \item The robotic agent plans a grasp on this two-view predicted shape and attempts to lift the object
\end{enumerate}
\noindent The pipeline is shown visually in \autoref{fig:overview2}.

The steps required to accomplish the subtask for each step will be defined as well as information the robotic agent produces to be utilized at later stages of the system. 

\subsection{Navigation to the object}
Chapter~\ref{ch:learning_visual_navigation} described a learning-based navigation pipeline which does not rely on odometry, map, compass or indoor position at runtime and is purely based on the visual input and an 8-image panoramic goal. The method learns from the same expert trajectories generated using RGBD maps of real-world environments. Due to the addition of objects in the environment for the robotic agent to manipulate, this addition also needs to be added to the training data of the navigation system. This means the addition of domain randomization to the training of the navigation system by incorporating random objects from both the YCB, GRASP, and challenge datasets. These objects are added during training to ensure that the system can address different adjustments to the environment without degrading success of navigation at runtime. The original navigation system is used to navigate through the environment and upon reaching the goal the agent enters the \textit{Isolate Object Geometry} stage.

\subsubsection{Navigation Training Data}

For mobile manipulation, the agent needs surfaces that objects can be placed on so that it can manipulate these graspable objects and that are within its workspace. Many surfaces in the original Matterport 3D and Stanford 2D-3D-S datasets are uneven and do not allow the robot to capture additional views easily. To remedy this, tables are placed in the environment that would be compatible with the workspace of the robot, in this case a Fetch robot. Each table was placed such that the robot could navigate around it. Each table was sized so that it was large enough to support each object in the dataset but also not so large that the robot would be unable to plan a grasp trajectory while remaining a safe distance away. 


Chapter~\ref{ch:learning_visual_navigation} described a methodology for generating a map to sample expert trajectories on to generate training data for the policy and goal checker. This methodology now needs to be augmented to include these placed tables in the environment. The algorithm crops the bottom-most half of the mesh of the environment and then projects those faces of the mesh onto a 2D plane to generate an occupancy map. Each of the tables are then mapped onto this plane and included as obstacles to the original 2D map generated on the unmodified environment. Trajectories are then generated as described in \autoref{ch:learning_visual_navigation} to train the policy. 

When rendering views for each trajectory or sampled goal check, a random object is placed on each surface to make sure the network can properly navigate with a changed environment. The only objects seen during training are the same objects that the two-view completion model sees during training to make sure training is controlled for object meshes and can rigorously evaluate that the system can navigate around objects not observed during training.

That map must now be updated to include the newly placed tables as obstacles in the environment. To augment the training data for the navigation system, tables are placed throughout the environment which are large enough to fit objects on them, but not so large as to prevent the robot from reaching these objects. These tables are placed such that the robot would not have difficulty navigating around them to find multiple views.

\subsection{Isolate Object Geometry}


Upon reaching the object of interest, the robot needs a way to isolate the point cloud of the object from its environment so that it can predict the object's geometry using shape completion. The robot captures an 8-image RGBD panoramic view of its current surroundings. This panorama is converted into a point cloud. Points within a boundary of $0.3m$ and $0.7m$ from the robot are filtered. The agent then searches for planes in that cloud parallel to the base of the Fetch robot using RANSAC~\cite{fischler1981random}. Once the plane is found, the agent can then segment all points above this plane as the object of interest. This method assumes that the object of interest is within the boundary of $0.3m$ and $0.7m$ and that the surface of the table can be represented as a plane. Any shape segmentation algorithm would work here, and the results show that RANSAC is sufficient for the proposed system. The 8-image RGBD panoramic view of the object is saved for the \textit{Panoramic Prediction} stage. With this partial view, the agent begins the \textit{Completion} stage. 

\subsection{Completion}
\label{sec:completion}
The point cloud generated by the \textit{Isolate Object Geometry} stage is used to perform an initial shape completion of the object. A single-view shape completion CNN is used to predict object geometry using a partial view of an object. This process is like the work described in Chapter~\ref{ch:visual_tactile_manipulation} and previous work by Watkins-Valls et al.~\cite{watkins2019multi}. The network architecture is designed so that it predicts for each voxel in the output a value between $0$ and $1$ where the gradient is the probability of occupancy. In the analysis for Chapter~\ref{ch:visual_tactile_manipulation}, voxels were thresholded where any voxel above $0.5$ were considered filled and then turned that prediction into a mesh. Instead, the agent looks at the predicted values and selects those close to the decision boundary. These uncertain voxels can give insight into where the network was not sure what the object geometry was and therefore the agent can utilize them to determine the next best view. 


This stage can be summarized as 1) voxelizing the isolated point cloud of the object, 2) performing a single-view completion on the voxelized partial, and 3) saving the values of the predicted output for use in the \textit{Next-Best-View} stage. The voxelization of the input is the same as described in \autoref{ch:two_view_shape_understanding}, whereby the input is moved to the $z_{\min}$ and voxelized to the fixed scale all other inputs are aligned to. The resolution of the input and output are kept as described previously, $40^3$.

\subsection{Next-Best-View}
\label{sec:next_best_view}

With this initial completion of the object, the agent can determine where the uncertainty lies in this predicted voxel grid. The principal assumption of this method is that voxels close to $0.5$ are considered uncertain. The CNN architecture used to produce this initial shape hypothesis is generated via a sigmoid function on the output layer and the network is trained used binary cross entropy loss. This means that the network has a fixed output between $0$ and $1$ and it is penalized for outputting values close to $0.5$ during training due to the loss function. This means that voxels whose values are close to the $0.5$ boundary are anomalous and therefore should be viewed to ensure that they have the appropriate amount of information to correctly understand the object's geometry. 


Voxels in this initial hypothesis that have an occupancy score of $0.5 \pm\epsilon$ are considered uncertain as they are close to the decision boundary and a second observation would be helpful to determine their occupancy. $\epsilon$ is the error bounds at $\pm0.025$. A bound of $\epsilon$ was calculated to be $\pm0.025$ by evaluating the completion quality of various bounds and found that the system had the best performance at $\pm0.025$. Ultimately, it is optimal to capture as many of these points as possible in a secondary camera view while still being reachable by the robotic agent. Principle Component Analysis (PCA) solves this objective in a 3D space by taking the smallest component which is orthogonal to a 2D plane that best fits the data. This smallest component will capture the least variance and therefore observe the most voxels. This vector is calculated as shown in Equation~\eqref{eq:pca} where $\mathbf{X}$ is the set of uncertain voxels, $\mathbf{w}$ is the corresponding eigenvector, and $\hat{X}_k$ is the eigenvectors calculated by PCA\@.
\begin{equation}
    \boldsymbol{v_{nbv}} = \arg\min \mathbf{\hat{X}}_k = \arg\min (\mathbf{X} - \sum_{s = 1}^{k - 1} \mathbf{X} \mathbf{w}_{(s)} \mathbf{w}_{(s)}^{\mathsf{T}}\label{eq:pca})
\end{equation}

\noindent
where $\hat{X}_k$ is the eigenvectors calculated by PCA. This next-best-view vector $v_{nbv}$ does not consider the height of the robot. The agent can then extract the $(x, y)$ components of the vector, normalize the vector, and then multiply it by $0.5m$ as an empirically optimal distance from the target object. A $z$ value of $0$ is assigned for the target position for the robotic agent to navigate towards. The target robot position will then be $(x_{target}, y_{target}, 0)$ relative to the current position of the robot, $(0, 0, 0)$. Additionally, a height value $h$ is calculated for the agent to raise or lower its torso to accommodate the next-best-view. The agent calculates this by calculating the optimal head angle $\theta_h$ relative to $\mathbf{v_{nbv}}$ and then calculates the optimal height $h$ to achieve that angle relative to the object. An optimal value $\epsilon$ was calculated to be $\pm0.025$ by evaluating the completion quality of various bounds. With this target position, the agent begins the \textit{Panoramic Prediction} stage. 

\subsubsection{Threshold Calculation}
\label{app:nbvprobability}
To determine the optimal value for $\epsilon$, a series of completions were computed at various values. The completion quality at each value of $\epsilon$ was computed by calculating a next-best-view and using it and a current view to produce a completion. Through experimentation it was found that $0.025$ gave the highest Jaccard completion quality. Centering the uncertainty at $0.5$ was considered optimal and the only benefit derived would be through changing the bounds of what uncertain voxels to consider. If the $\epsilon$ were made too low there would be no voxels for the object to select, so only a lower bound of $0.01$ was used. Overall, $39000$ views were evaluated in the holdout model validation dataset to evaluate which bounds were optimal for this system. If the bounds were made too high, the PCA algorithm applied to the selected voxels would be fitting for more of the reconstruction which is already observed from one view. For a table of completion quality results, see \autoref{tab:nbvprobabilityresults}. 

\begin{table}[t]
    \centering
    \begin{tabular}{|c|c|}
    \hline
    \multicolumn{1}{|c|}{\begin{tabular}[c]{@{}c@{}}\textbf{Next-Best-View} \\  \textbf{Score Epsilon}\end{tabular}} 
    & \multicolumn{1}{c|}{\begin{tabular}[c]{@{}c@{}}\textbf{Jaccard} \\  \textbf{}\end{tabular}} 
    \\
    \hline
        $0.5\pm 0.010$     & 0.751           \\ \hline
        $0.5\pm 0.015$     & 0.782           \\ \hline
        $0.5\pm 0.020$     & 0.824           \\ \hline
        $0.5\pm 0.025$     & 0.852           \\ \hline
        $0.5\pm 0.030$     & 0.837           \\ \hline
        $0.5\pm 0.050$     & 0.791           \\ \hline
        $0.5\pm 0.100$     & 0.758           \\ \hline
    \end{tabular}
    
    \caption{\textbf{NBV Threshold Bounds Completion Quality}, measuring the performance of the reconstruction quality of different holdout meshes given a next-best-view calculated with varying bounds of $\epsilon$. A higher Jaccard is better.}
    \label{tab:nbvprobabilityresults}
\end{table}

\subsection{Panoramic Prediction}
\begin{figure*}[t]
    \centering {
        \includegraphics[width=0.95\textwidth]{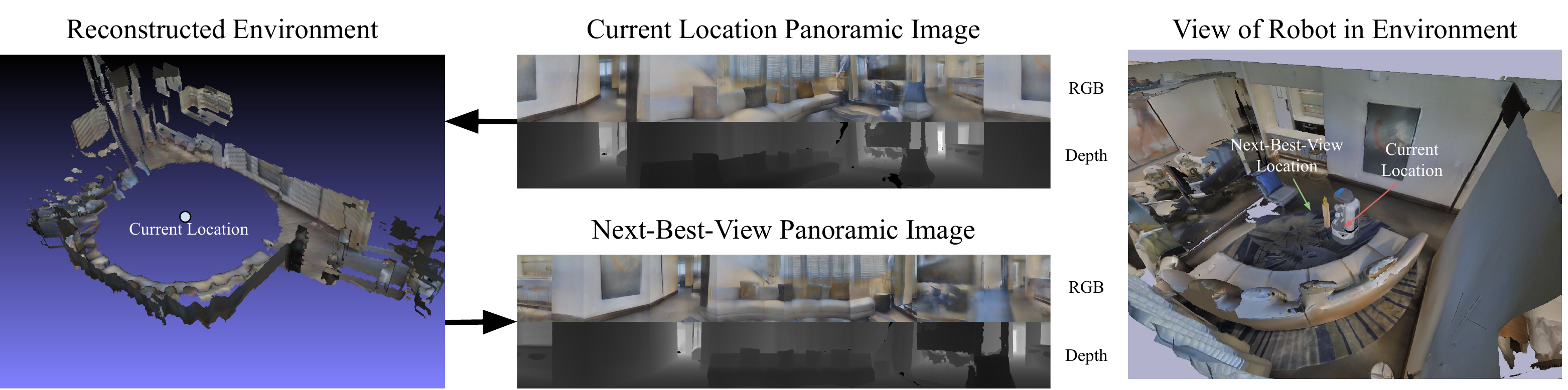}
    }
    \caption{A panoramic image is captured at the current location of the robot, which is then turned into a reconstruction of the nearby environment using Open3D~\cite{open3d}. The target panoramic image is rendered at the Next-Best-View location in this reconstruction to navigate to it without localization. This predicted panorama is shown at the Next-Best-View location here. }\label{fig:predictedpanorama}
\end{figure*}

In the description of the navigation system, it is established that the agent does not have access to its position in the environment at runtime. Now that the agent has a next-best-view vector that is defined in the coordinate space of the object, and subsequently the coordinate space of the robot, the agent needs to convert this into a goal location that the learned navigation system can utilize to get to that next viewpoint. The agent can utilize the panorama captured of the nearby environment from the \textit{Isolate Object Geometry} stage to re-render a view of the environment at the next-best-view location.

The immediate concern with doing a re-render based on partial data is whether these generated views will contain holes. There are two benefits with how the navigation system was trained to deal with this issue. RGB data collected to train the policy and goal checker models is augmented via the domain adaptation module called Goggles from the Gibson simulator. This module is specifically designed to fill holes generated during rendering in the simulator and the network is trained on data exclusively processed via Goggles. This however does not address holes present in the depth information in simulation, but because the environment included holes in the depth map as well and any depth values beyond $3m$ are filtered for a Fetch many of these holes are not relevant or have been addressed during training. 

The agent takes an 8-image RGBD panorama of the environment and utilizes a view reconstruction method. The implementation of this uses the Open3D~\cite{open3d} implementation of Bernardini's ball-pivoting reconstruction paper~\cite{bernardini1999ball}. Using this mesh, the agent can then predict the panorama from the next-best-view target by loading the mesh into a renderer and taking 8 RGBD views at equal $45^\circ$ intervals at this new location. This mesh will have holes in the RGB view. The agent then uses Gibson's Goggles~\cite{igibson} to resolve any missing data in the predicted view. With this predicted panoramic view, the agent can then utilize the learned navigation system to locally navigate to this next view without localizing the agent. When the agent has arrived at this location, the agent starts the \textit{Two-View Completion} stage. An example predicted panorama is shown in \autoref{fig:predictedpanorama}.

\subsection{Two-View Completion}

The two-view completion is designed the same way as described in \autoref{ch:two_view_shape_understanding}. This system utilizes two unregistered views of the object to produce a predicted output shape in the frame of the current view. What was not addressed in the previous description, however, was the benefit that the next-best-view provides. Now that the network is no longer seeing a random second view, but instead an optimized second view, it can better utilize the information to produce a higher quality completion. 

Upon reaching the object for the second time, the robotic agent aligns itself with the height value $h$ and head angle $\theta_h$ calculated in the \textit{Next-Best-View} stage. Once aligned, the agent performs a segmentation of the object from the environment the same way as described in the \textit{Isolate Object Geometry} stage. Both the initial point cloud and the next-best-view point clouds are voxelized into a $40^3$ voxel grid and passed as input into the CNN. The network then outputs a voxel grid of occupancy scores which are thresholded at $0.5$ to determine a final occupancy grid hypothesis. The agent then performs a marching cubes algorithm~\cite{lorensen1987marching} to turn this into a mesh that can be utilized for the \textit{Manipulation} planning stage.

\subsection{Manipulation}

Using this predicted mesh, the agent plans a grasp on the mesh using GraspIt!~\cite{miller2004graspit} to get a series of grasp candidates. Each of these grasp candidates are given an associated volume quality and are filtered for volume quality above 0~\cite{ferrari1992planning}. The agent uses MoveIt~\cite{sucan2013moveit} to plan the pick plan given each grasp and pick the trajectory with the smallest execution time. To generate this pick plan, the agent models the table and a region above the object as obstacles to ensure the trajectory does not disturb the object. If no trajectories or grasps are valid the program terminates.

The reason a mesh is required for this stage is because the grasp planning software, GraspIt!, requires a mesh of the target object. Other solutions for planning manipulation tasks may require additional or different information. This prediction of an object's geometry is enough to reliably plan and execute a grasp on the target object.

\section{Experiments}
\label{sec:Experiments}

To validate that this system reliably navigates to and manipulates objects, the agent needed a series of tests to determine the performance of subcomponents as well as the end-to-end performance. Because this pipeline reuses work from the learning-based visual navigation system and the two-view shape completion methods, their respective datasets can be used to validate that this system extends to the same environments and objects, as well as to unseen objects. 

\subsection{Reconstruction Quality Tests}

\label{sec:reconstructionqualitymetrics}
The first goal is to validate that getting a next-best-view would result in better reconstruction quality. There are three metrics that validate the proposed hypothesis: \textbf{Jaccard similarity}, \textbf{Hausdorff distance}, and \textbf{grasp joint accuracy}.
\begin{enumerate}
    \item \textbf{Jaccard similarity} Jaccard similarity is used to evaluate the similarity between a generated voxel occupancy grid and the ground truth. The Jaccard similarity between sets A and B is given by:
    \[
    J(A, B) = \dfrac{|A\cap B|}{|A\cup B|}
    \]
    The Jaccard similarity has a minimum value of 0 where A and B have no intersection and a maximum value of 1 where A and B are identical~\cite{jaccard}.
    \item \textbf{Hausdorff Quality} The Hausdorff distance is a one-direction metric computed by sampling points on one mesh and computing the distance of each sample point to its closest point on the other mesh. It is useful for determining how closely related two sets of points are~\cite{huttenlocher1993comparing}.
    \item \textbf{Grasp Joint Accuracy} GraspIt!~\cite{miller2004graspit} is used to generate a series of grasp candidates for each predicted mesh and then choose the one with the highest volume quality on that predicted mesh. A simulated BarrettHand is used to execute the grasp within the GraspIt! simulator on the ground truth mesh and calculate the difference between the expected joint values and the realized joint values. 
\end{enumerate}

To evaluate the performance of the proposed next-best-view method, there have five test scenarios. A \textbf{Single View} reconstruction using only the current view of the object that is utilizing the same model architecture as used in Varley et al.~\cite{varley2017shape}. A \textbf{Same View} reconstruction using the proposed two-view architecture, but where the current view is passed in twice. This is to evaluate the performance benefit of using a larger model for shape completion. A \textbf{Two-View (Random)} reconstruction using the proposed two-view architecture but the two views are chosen randomly about the object. A \textbf{Two-View (Opposite)} reconstruction using the proposed two-view architecture where the first view is random, and the second view is chosen by capturing the view opposite to the first. A \textbf{Two-View (Next-Best-View)} reconstruction using the proposed two-view architecture where the first view is random, and the second view is chosen by calculating the next-best-view as described in \autoref{sec:next_best_view}. A \textbf{Three View} reconstruction using a modified version of the two-view architecture where the previous two views are passed into a single encoder and the current view is passed into its own encoder. These three embeddings are then added together and decoded into a reconstruction of the object.

\begin{table*}[t]
\centering
\begin{subtable}{\textwidth}
    \centering
    \begin{tabular}{|c|c|c|c|} 
    \hline
    \multicolumn{1}{|c|}{\begin{tabular}[c]{@{}c@{}}\textbf{Next-Best-View} \\  \textbf{Method}\end{tabular}} 
    & \multicolumn{1}{c|}{\begin{tabular}[c]{@{}c@{}}\textbf{Jaccard} \\  \textbf{}\end{tabular}} 
    & \multicolumn{1}{c|}{\begin{tabular}[c]{@{}c@{}}\textbf{Hausdorff} \\  \textbf{}\end{tabular}} 
    & \multicolumn{1}{c|}{\begin{tabular}[c]{@{}c@{}}\textbf{Grasp Joint} \\  \textbf{Error}\end{tabular}} \\
    \hline
        Single-View     & 0.782         & 6.573        & $4.52^\circ$ \\ \hline
    	Same-View       & 0.802         & 6.423        & $4.36^\circ$ \\ \hline
    	Random          & 0.818         & 6.251        & $4.14^\circ$ \\ \hline
     	Opposite        & 0.826         & 5.421        & $3.85^\circ$ \\ \hline
    	Next-Best-View  & \textbf{0.852} & \textbf{4.912} & $\mathbf{3.28^\circ}$\\ \hline
    \end{tabular}
    \caption{\textbf{YCB \& Grasp Dataset Test Object Results}}\label{tab:reconstruction_quality_tests}
\end{subtable}
\begin{subtable}{\textwidth}
    \centering
    \begin{tabular}{|c|c|c|c|}
    \hline
    \multicolumn{1}{|c|}{\begin{tabular}[c]{@{}c@{}}\textbf{Next-Best-View} \\  \textbf{Method}\end{tabular}} 
    & \multicolumn{1}{c|}{\begin{tabular}[c]{@{}c@{}}\textbf{Jaccard} \\  \textbf{}\end{tabular}} 
    & \multicolumn{1}{c|}{\begin{tabular}[c]{@{}c@{}}\textbf{Hausdorff} \\  \textbf{}\end{tabular}} 
    & \multicolumn{1}{c|}{\begin{tabular}[c]{@{}c@{}}\textbf{Grasp Joint} \\  \textbf{Error}\end{tabular}}
    \\
    \hline
        Single-View  & 0.648          & 7.340 & $5.31^\circ$\\ \hline 
        Same-View            & 0.663          & 7.284 & $5.16^\circ$\\ \hline 
        Random               & 0.753          & 6.418 & $4.85^\circ$\\ \hline 
        Opposite        & 0.831         & 5.924        & $4.53^\circ$ \\ \hline
        Next-Best-View       & \textbf{0.866} & \textbf{5.341} & $\mathbf{4.24^\circ}$\\ \hline 
    \end{tabular}
    \caption{\textbf{Challenge Dataset Results}}\label{tab:reconstruction_quality_tests_challenge}
\end{subtable}
\caption{Measuring the performance of the reconstruction quality of test and challenge meshes given a single-view, same-view, random new view, opposite-view, next-best-view, as described in \autoref{sec:reconstructionqualitymetrics}. None of these meshes were seen during training. A higher Jaccard is better. A lower Hausdorff is better. A lower Grasp Joint Error on the 3 finger BarrettHand is better.}
\end{table*}

\begin{figure}[t]
    \centering {
        \includegraphics[width=0.95\linewidth]{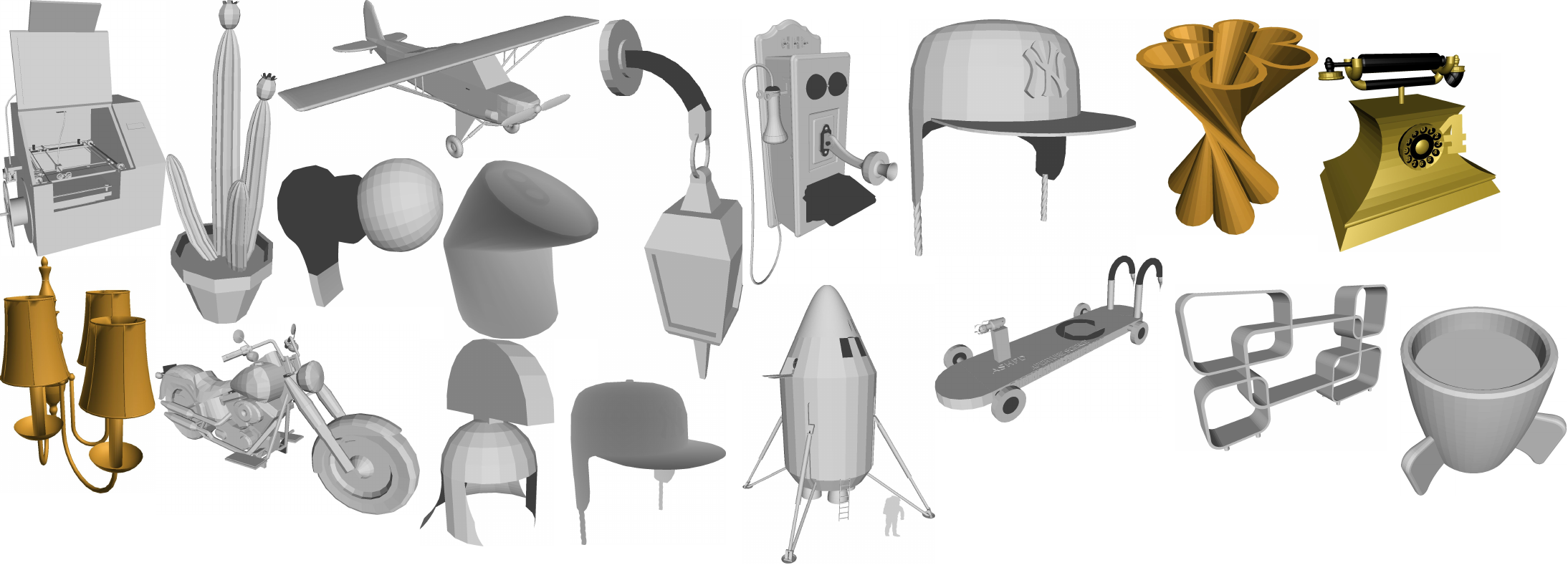}
    }
    \caption{\textbf{Challenge Dataset Meshes} 17 meshes are chosen from the ShapeNet~\cite{chang2015shapenet} dataset that feature self-occlusions, asymmetrical geometry, or deviate significantly from the geometry of the YCB~\cite{calli2015ycb} and Grasp Database~\cite{bohg2014data} datasets. }\label{fig:challenge_dataset}
\end{figure} 

The training and evaluation datasets are identical to those presented in \autoref{ch:two_view_shape_understanding}. All training and evaluation were performed using 590 meshes from the Grasp Database~\cite{bohg2014data} dataset and 28 meshes from the YCB~\cite{calli2015ycb} dataset. 100 meshes were sampled from both datasets to provide 50 validation meshes and 50 test meshes. The validation meshes were used to evaluate the Jaccard quality of each completion whereas the test meshes were used to evaluate the performance of each CNN model. The CNN model that performed best using validation meshes was used for evaluation. A sample of 17 meshes from the Shapenet~\cite{chang2015shapenet} dataset is used and resized to fit within a grip width of $100mm$. These 17 meshes are not observed during training and are chosen to be difficult to complete with only one view, thus this dataset is called the \textit{Challenge} dataset. All meshes from the challenge dataset are shown in \autoref{fig:challenge_dataset}. All views are voxelized using Binvox~\cite{binvox, nooruddin03}.

\begin{figure}[t]
    \centering {
        \includegraphics[width=0.95\linewidth]{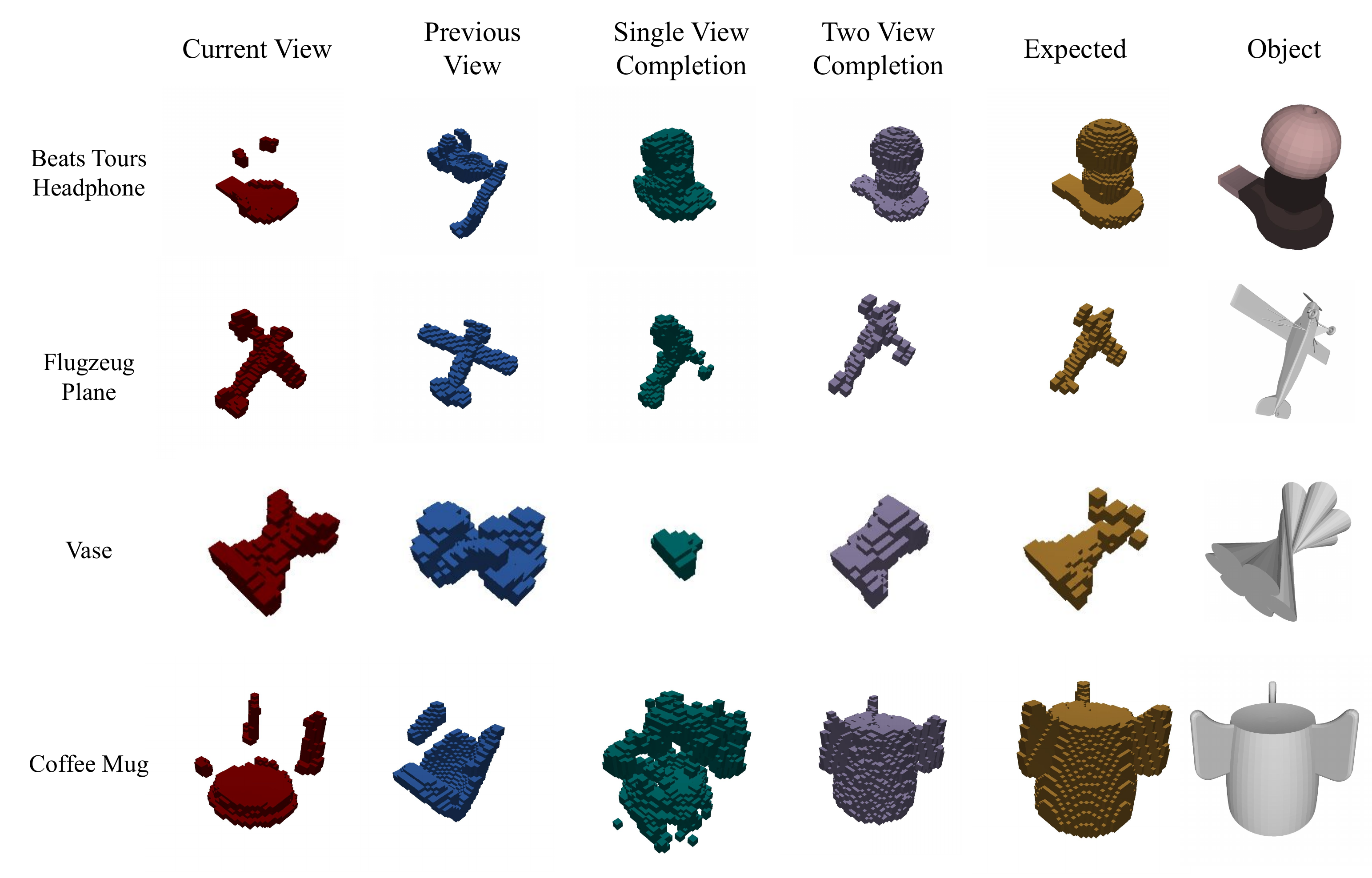}
    }
    \caption{Two view reconstruction inputs showing that the next-best-view is better than a single view. Current input (red), a single view reconstruction (green), the previous view of the object (blue), the two-view completion (purple), and finally the ground truth mesh (yellow). These meshes were not observed during training for either network. All meshes are from the Challenge dataset. }\label{fig:completion_improvement}

\end{figure} 

Test object reconstruction results are shown in \autoref{tab:reconstruction_quality_tests} and challenge reconstruction results are shown in \autoref{tab:reconstruction_quality_tests_challenge}. The most significant result is the Jaccard and Hausdorff for both the test object and the challenge datasets, with the next-best-view method having the best performance in both. They showed a Jaccard reconstruction of $0.866$ for the next-best-view implementation versus $0.648$ for single-view for unseen objects. The same-view method had demonstrably worse performance than the next-best-view algorithm, with a grasp joint error difference of $32.9\%$, showing that adding weights did not improve performance. A grasp joint error of $3.28^\circ$ compares favorably with the single view of $4.52^\circ$. For examples of the next-best-view shape completion improvement using the challenge dataset see \autoref{fig:completion_improvement}. 


\begin{figure}[t]
    \centering {
        \includegraphics[width=0.95\linewidth]{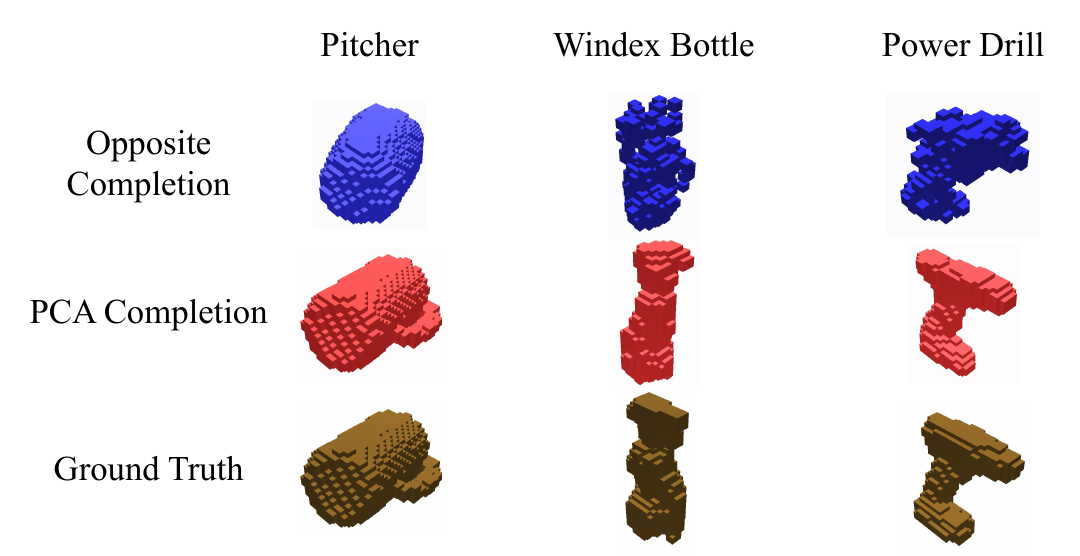}
    }
    \caption{The opposite view is not always enough. Missing an important feature in both the current and the opposite view results in a worse completion and therefore makes it difficult to plan grasps. Utilizing the uncertainty in the single-view completion for next-best-view fixed the missing handle in the reconstruction in the pitcher from the YCB~\cite{calli2015ycb} dataset.}\label{fig:oppositevspca}
\end{figure} 

While three views provided a minor benefit in the Jaccard, Hausdorff, and grasp joint error metrics ($0.868$, $4.782$, and $3.12^\circ$, respectively), the benefit does not warrant the extra effort of capturing a third view in the mobile manipulation step. Capturing the opposite view, although a reasonable strategy, did not outperform the PCA method. At best, the opposite view performs similarly to the PCA method. At worst, opposite-views miss major features of objects. An example of a completion missing the handle of a pitcher is shown in \autoref{fig:oppositevspca}. Without leveraging the uncertain voxels within the initial completion, naively capturing the opposite view is insufficient for completing the object reliably. In testing using the test object dataset, the next-best-view method outperformed the opposite method in all metrics. 

\begin{table}[t]
    \centering
    \begin{tabular}{|c|c|c|}
        \hline
        \textbf{Localization} & \textbf{Jaccard} & \textbf{Hausdorff}  \\
        \hline
        Registered-Views & 0.923 & 3.523 \\
        \hline
        Next-Best-View & 0.852 & 4.912 \\
        \hline
        Noisy-Views & 0.583 & 9.595 \\
        \hline
    \end{tabular}
    \caption{\textbf{Noisy Odometry Results} The results of the two-view system are compared using PCA versus a noisy odometry model. The proposed system outperforms a model using noisy odometry at runtime to align the two views, justifying the use of a model without registration of the two views for shape completion. }
    \label{tab:noisyodometryresults}
\end{table}

This test is to model how even a small amount of noise in odometry can result in negative performance of a method that relies on idealized registration. The proposed system is compared to a version of the single-view model that was trained using two registered views concatenated on the input layer. For training, each view is overlaid using perfect registration. Odometry noise is simulated by adding up to a $5\%$ error during movement around the pedestal at each time step. Both were compared to evaluate how the noise would compare versus the ideal case and versus a model trained without registration. The proposed next-best-view model outperformed the noisy case. When overlaying voxel input with translational error the registered-views model was unable to properly complete the mesh. Results are shown in \autoref{tab:noisyodometryresults}.

\subsection{Navigation Tests}
\begin{table}
    \centering
    \begin{tabular}{|c|c|c|}
      \hline
      \multicolumn{1}{|c|}{\begin{tabular}[c]{@{}c@{}}\textbf{Next-Best-View} \\  \textbf{Navigation}\end{tabular}}
      & \multicolumn{1}{c|}{\begin{tabular}[c]{@{}c@{}}\textbf{SPL} \\  \textbf{}\end{tabular}}
      & \multicolumn{1}{c|}{\begin{tabular}[c]{@{}c@{}}\textbf{Success Rate} \\  \textbf{}\end{tabular}} \\
      \hline
      ROS (map) & 0.923 & 0.943 \\ \hline
      Panorama  & 0.854 & 0.845 \\ \hline
    \end{tabular}

    \caption{\textbf{Long-Range Navigation Results}, measuring the performance of the long-range navigation success and path length of two different methods: ROS Navigation Stack and the learning-based panoramic target navigation, as described in \autoref{sec:longrangenavigationtests}. Higher SPL is better.}\label{tab:longrangenavigationtests}
\end{table}
\begin{table}
    \centering
    \begin{tabular}{|c|c|c|}
      \hline
      \textbf{Navigation Method} & \textbf{SPL} & \textbf{Success Rate}\\
      \hline
      ROS Nav (map) & 0.958 & 0.953 \\ \hline
      True Panorama & 0.874 & 0.897 \\ \hline
      Predicted     & 0.813 & 0.853 \\ \hline
    \end{tabular}
    \caption{\textbf{Next-Best-View Navigation Results}, measuring the performance of the next-best-view navigation success and path length compared using ROS, using the true panorama, or a predicted panorama, as described in \autoref{sec:predicted_view_navigation_tests}. Higher SPL is better.}\label{tab:nextbestviewnavigationresults}
\end{table}

For testing, the agent is placed in the \textit{house1} environment from the Matterport 3D~\cite{Matterport3D} dataset modified with four tables placed in the house. The Matterport 3D dataset features real-world homes scanned and turned into navigable 3D meshes for research. The robot used for navigation and end-to-end testing is the Fetch~\cite{fetch} robot with its parallel jaw gripper. These tables were placed such that the Fetch could navigate around them while still being able to navigate through the environment. Different graspable object meshes are placed on these tables to evaluate grasping performance. 50 holdout meshes are used from the YCB~\cite{calli2015ycb} and GRASP~\cite{bohg2014data} datasets that were randomized on each trial. For each object, the test environment only considers which orientations, out of the maximum 726, the object rested on the table. Out of 36300 orientations 3455 were stable, and thus the agent had 3455 stable table-objects next-best-views. This environment is used for the navigation and end-to-end testing. Tables were placed in the environment that would accommodate the Fetch's grasp workspace. Table heights were varied between 0.65m, 0.7m, 0.75m, and 0.8m. These heights are high enough off the ground to allow the Fetch to raise and lower its torso to view the object from a variety of different vectors. 

\label{sec:longrangenavigationtests}
In \autoref{ch:learning_visual_navigation}, an analysis of the ability of the agent to navigate through an environment using panoramic targets as shown. This analysis is repeated with these tabled environments. The test of long-range navigation is run in this tabled environment with the \textbf{ROS Navigation stack} and the learned navigation system. The \textbf{ROS Navigation stack}~\cite{ros} uses Dijkstra's algorithm to plan paths and is given a point goal and a global map to get theoretically optimal performance with perfect information. The performance of these two agents is evaluated using \textbf{Success Weighted by Path Length (SPL)} and \textbf{Success Rate} of reaching the goal location. 400 holdout trajectories were used in testing for this environment. The long-range learning-based navigation system performed with a success rate of $0.845$ versus $0.943$ for the ROS Navigation stack which navigated using a point goal and map as opposed to a panoramic target goal with no localization. The SPL for the proposed method was 0.854 versus their 0.923 which demonstrates a near expert level path length without a map at runtime. Results are shown in \autoref{tab:longrangenavigationtests}.

\label{sec:predicted_view_navigation_tests}
The agent's ability to navigate locally is validated via two baselines: 1) the \textbf{ROS Navigation Stack} and 2) the learned navigation model using the \textbf{true panorama}. The \textbf{ROS Navigation stack}~\cite{ros} uses Dijkstra's algorithm to plan paths and is given a point goal and a global map to get theoretically optimal performance with perfect information. The \textbf{true panorama} is captured at the next-best-view target location to evaluate how the agent performs when given an ideal panorama. Finally, the performance of the proposed system is evaluated with the predicted panorama at the next-best-view location. The system is evaluated using Success Weighted by Path Length, or \textbf{SPL}, and \textbf{Success Rate} of reaching the goal location. \textbf{SPL}, from Anderson et al.~\cite{anderson2018evaluation}, is shown in Formula~\eqref{spl2}, where $l_i$ is the shortest-path distance from the agent's starting position to the goal in episode $i$, $p_i$ is the length of the path actually taken by the agent in this episode, and $S_i$ is a binary indicator of success in trial $i$. $p_i$ is calculated via the L2 distance between each step in the ground truth trajectory using Dijkstra's algorithm. $l_i$ is calculated via the L2 distance between each step in the executed trajectory. This metric weighs each success by the quality of path and thus is always less than or equal to \textit{Success Rate}.
\begin{equation} \label{spl2}    
    SPL = \frac{1}{N} \sum_{i=1}^{N} S_i \frac{l_i}{\max (p_i, l_i)}
\end{equation}
\autoref{tab:nextbestviewnavigationresults} shows the results for the proposed method in terms of SPL and success rate. The \textbf{ROS Navigation Stack} is used knowing it would outperform the proposed method because it has access to a map of the environment. It made for a good baseline with a success rate of navigating to the goal of 0.953. The predicted and ground truth panoramas were both equally effective at navigating to the target goal with a success rate of 0.853 and 0.897, respectively. The effective navigation to predicted panoramic goals validates the use of the predicted panorama in the proposed system. 

\subsection{End-to-End Mobile Manipulation Testing}
\label{sec:end_to_end_testing}

\begin{table}
    \centering
    \label{tab:longrangenavigationresults}
    \begin{tabular}{|c|c|c|c|c|}
      \hline
      \multicolumn{1}{|c|}{\begin{tabular}[c]{@{}c@{}}\textbf{Navigation} \\  \textbf{Method}\end{tabular}} 
    & \multicolumn{1}{c|}{\begin{tabular}[c]{@{}c@{}}\textbf{Localized} \\  \textbf{}\end{tabular}} 
    & \multicolumn{1}{c|}{\begin{tabular}[c]{@{}c@{}}\textbf{Completion} \\  \textbf{Method}\end{tabular}} 
    & \multicolumn{1}{c|}{\begin{tabular}[c]{@{}c@{}}\textbf{E2ESPL} \\  \textbf{}\end{tabular}}
    & \multicolumn{1}{c|}{\begin{tabular}[c]{@{}c@{}}\textbf{Grasp} \\  \textbf{Success}\end{tabular}}
      \\
      \hline
      ROS Nav (map) & Yes & Two-View & 0.892 & 0.884 \\ \hline
      ROS Nav (map) & Yes & Single-View & 0.771 & 0.781 \\ \hline
      Noisy ROS (map) & Yes & Two-View & 0.562 & 0.531 \\ \hline
      True Panorama & No & Two-View & 0.845 & 0.872 \\ \hline
      Predicted     & No & Two-View & 0.820 & 0.819 \\ \hline
    \end{tabular}
    \caption{\textbf{End-to-End Mobile Manipulation Results}, measuring the performance of the full pipeline success rate and E2ESPL with different methods using 50 novel objects. Higher E2ESPL is better. As described in \autoref{sec:end_to_end_testing}}\label{tab:endtoendresults}
\end{table}

Once the major contributions of the proposed method work in isolation, the performance of the end-to-end mobile manipulation system is evaluated on its ability to navigate to a goal and manipulate the target object. The test trajectories are between 1.5m and 20m in length. For this test, the completion method and navigation method are ablated. The \textbf{ROS Navigation Stack} is used as described in \autoref{sec:predicted_view_navigation_tests}, the \textbf{True Panorama} method, and the proposed \textbf{predicted panorama} method as navigation modules. The \textbf{two-view} and \textbf{single-view} reconstruction methods are used for these tests. The Single-view and ROS Navigation are evaluated together. The two-view method is assessed with every navigation method. Additionally, the performance with the ROS navigation stack is evaluated with a noisy odometry model that has imperfect information at runtime about how the robot moves by 5\%. All the depth information coming from the depth camera has a similar noise model. 400 trials of this system are evaluated against unseen target locations with 50 novel objects placed on tables in the environment. The SPL metric used before defined a binary success of a trial $i$ as $S_i$. If instead it is replaced this with a binary signal of successfully picking up an object $S_p$, it becomes the evaluation of the full end to end pipeline. This new metric is termed End-to-End Success Weighted by Path Length, or $E2ESPL$, that uses this new success value as the following:
\begin{equation} \label{e2espl}    
    E2ESPL = \frac{1}{N} \sum_{i=1}^{N} S_p \frac{l_i}{\max (p_i, l_i)}
\end{equation}
\noindent Where $S_p$ is the success of picking up an object, $l_i$ is the shortest path distance from the start location to the end location of trial $i$, $p_i$ is the length of the path taken by the agent, and $N$ is the number of trials. 

Results for the end-to-end testing are in \autoref{tab:endtoendresults}. The proposed end-to-end localization-free mobile manipulation method was able to reliably navigate to the target positions and grasp the object in $80.7\%$ of tests despite not having access to a map at runtime or the true panorama of the next-best-view target position. The ROS Navigation stack version was able to navigate to and grasp the object $88.4\%$ of the time and the true panorama was able to navigate $87.2\%$ of the time. Additionally, the single-view completion and ROS Navigation stack succeed $78.1\%$ of the time, showing that two-views helps for unseen objects. The proposed model outperformed ROS with a noisy odometry model where ROS was unable to localize itself with noisy sensors. This method failed mostly when colliding with the environment. 

The proposed method reliably navigates to the target positions and grasps the object in $81.9\%$ of tests despite not having access to a map at runtime or the true panorama of the next-best-view target position. When compared to the ideal case, or \textbf{ROS Navigation Stack} with the two-view shape completion, the proposed method performs slightly worse with a $81.9\%$ success rate versus a $88.4\%$ success rate for the ROS method. This is to be expected, as the ROS method has access to the map at runtime and has an idealized odometry while moving around. Once a noise model is introduced to the odometry, the performance takes a substantial hit with a success rate of $53.1\%$ for ROS. This shows the sensitivity of map-based methods to noise at runtime for tracking how much it has moved. The \textbf{True Panorama} method, providing the ground truth panorama to the next-best-view navigation step, performed very well with a success rate of $87.2\%$. This shows that the predicted panorama does hurt overall performance slightly but allows the agent to navigate with relative success through the environment to lift the object. The ROS navigation with single-view had a success rate of $78.1\%$. The single-view method showed a degradation of performance over two-view as it could not estimate the geometry of challenge objects well due to their self-occluding properties. 

\section{Conclusion}

This chapter proposes an end-to-end mobile manipulation system that navigates to and manipulates an object without localization using a novel panoramic prediction method. The chance of object manipulation success is improved using a novel two-view reconstruction architecture. This mobile manipulation system leveraging multiple views is demonstrated to perform competitively against a method with perfect odometry and a map. A next-best-view two-view completion model outperforms single-view reconstructions for unseen objects increasing grasp success with a mobile robot. The learned navigation system can utilize predicted panoramic targets effectively allowing the agent to generate its own goals. 

The combination of navigation and two-view shape completion into a mobile manipulation pipeline is enabled through novel contributions in next-best-view planning and predicted panoramic navigation. The results shown in this chapter demonstrate that whether its ablated or integrated, this pipeline performs comparably to systems with more information. A mobile robot can successfully move around its environment to gather enough understanding of an object's geometry to grasp an object.

\chapter{Multiple View Shape Understanding}
\label{ch:performers}

While capturing two views of an object offers improvement over a single view, what if the robot wanted to capture an arbitrary number of views? This offers uniquely richer data to provide additional information about an object's geometry. However, it can be difficult to register these multiple views due to odometry error and noise. This chapter explores a novel multiple-view object completion method that allows the robot to use multiple unregistered views of the object to improve its shape estimation. 

\section{Introduction}
Shape understanding based on a single image or two images is difficult. This was explored previously in \autoref{ch:two_view_shape_understanding} to complete an object but utilizing multiple images to complete an object is even more challenging. A single depth sensor can be moved to capture multiple views of an object but aligning those views can be challenging. Other works, such as RTABMap~\cite{labbe2019rtab}, attempt to perform visual RGBD slam to register multiple point clouds between each other. This solution can be noisy and introduce error into a machine learned CNN that can be difficult to simulate in training data. To solve this, a 3D convolutional neural network is used to enable robust shape estimation by leveraging multiple unregistered views. This means each image of an object is kept in its respective image frame. This methodology can be used to complete objects with only two views, or up to an arbitrary number of views leveraging a new deep learning architecture known as a performer layer~\cite{performer}. Providing more accurate reconstructions of objects helps to enable a variety of robotic tasks such as manipulation, collision checking, sorting, and cataloging.

\begin{figure*}[t]
    \centering {
        \includegraphics[width=0.95\textwidth]{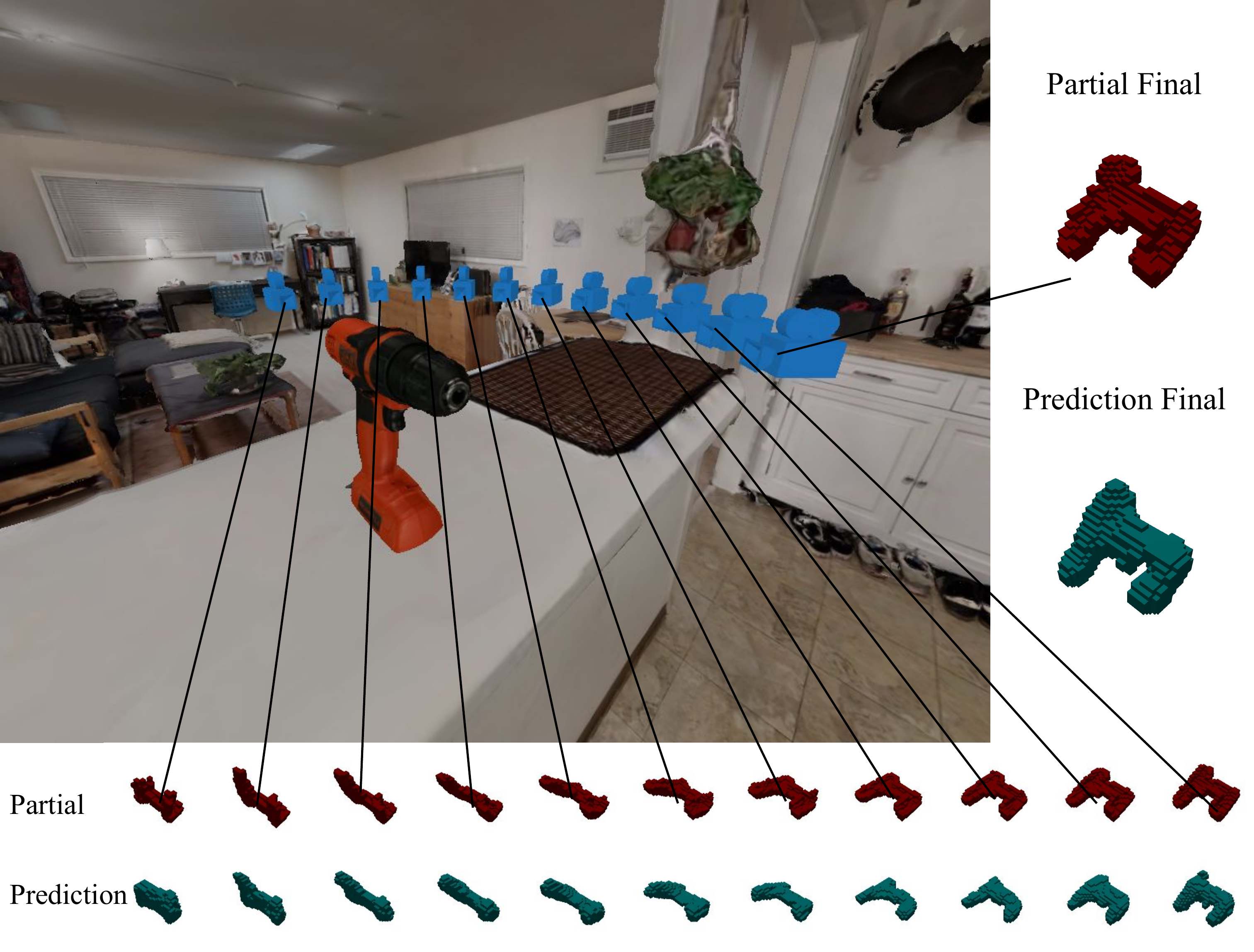}
    }
    \caption{Many views of an object can help refine the prediction. Shown in red are partial views of the target object and shown in green are prediction of each incremental view using a novel performer-based approach to shape completion. The final prediction, shown in the top right, is the culmination of multiple successive views contributing to an overall improved completion of the target object, in this case a drill from the YCB object dataset~\cite{calli2015ycb}. } \label{fig:multiview_cover_figure}
\end{figure*} 

At runtime, a partial 2.5D image of an object is captured. This first view is passed through a shape completion network to produce a shape estimation of the target object. 2.5D views are then captured about the object in a panning motion to create a sweeping snapshot of the object's geometry. For each of these views the network updates its understanding of the object, improving the overall shape estimation. An example of this sweep is shown in \autoref{fig:multiview_cover_figure}. Due to the performer model's ability to leverage multiple views, it also can remember objects that are no longer visible or utilize newly revealed views of objects that were previously hidden. This memory is enabled through using scalable transformers called \textit{Performers}~\cite{performer}. This model allows the current observation of the scene to attend to past observation for its more accurate infilling. The past observations are compressed via compact associative memory approximating modern Hopfield exponential memory, but independent of the number of past observations. 

This work is differentiated from other works, such as Haefner et al.~\cite{haefner2019photometric}, in that it utilizes multiple views of the object without aligning them. Training a neural network to extract registration information from multiple disparate views allows for other mobile robots to estimate object geometry, such as a drone or multiple robots within the same environment. Additionally, the utilization of a richer decoder architecture inspired by work from Yang et al.~\cite{Yang18} has improved reconstruction quality over a single-view~\cite{varley2017shape}. An application of this methodology for image reconstruction is shown in Appendix~\ref{app:mnist_reconstruction}.

\section{Methodology}

To leverage multiple views, a layer architecture that can operate on multiple embeddings is required. There are many potential solutions to this, including LSTM, GRU, and Attention based model architectures. A combined effort from previous shape completion architectures and well-performing layers in time-series architectures through Performers~\cite{performer} allow for a potential solution to this problem. The idea is to take the current view of the object and pass that through its own encoder. Then the network takes all views seen so far and pass them through an additional encoder that only processes previous views. Now there are $N$ tokens where $N$ is the number of views seen so far. Each token is given a position encoding. Each of these views can be used to attend to the input encoding. Once they are attended to, the current embedding from the Performer layer is added to the embedding of the current view from its own encoder, and this aggregate embedding can then be used through the encoder to reconstruct the object geometry. The key component of this architecture is a compact associative memory that is used in the attention module of Performers~\cite{performer}. The output is refined in the same manner as explained in Section~\ref{sec:prediction_refinement}.

\begin{figure}[H]
    \centering {
        \includegraphics[width=\linewidth]{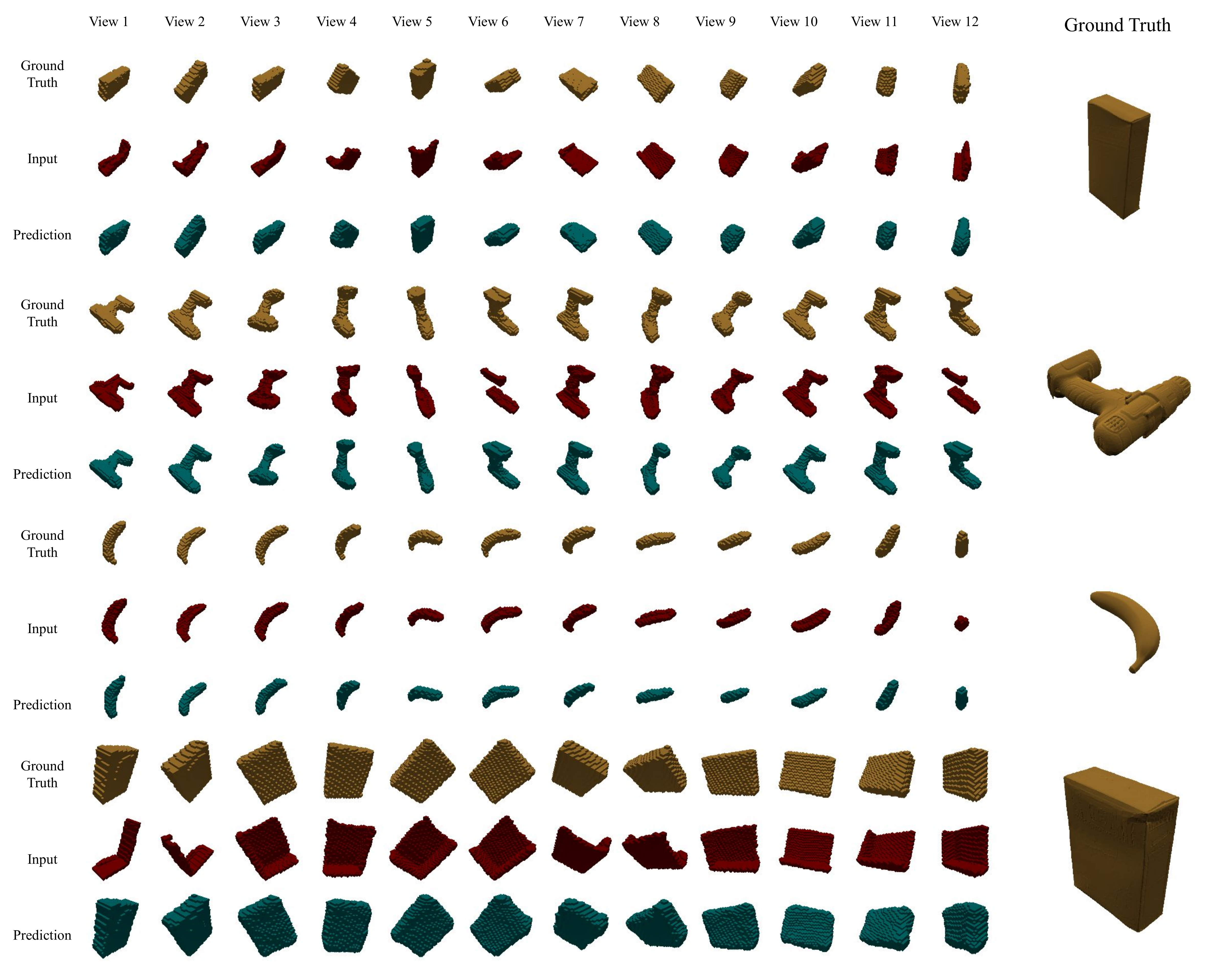}
    }
    \caption{Performer completion examples utilizing multiple views of the input object. The top row, in yellow, are the voxelized ground truth versions of the mesh. The middle row, in red, are partial views of the object captured from various angles. The bottom row, in green, are prediction of the object geometry. This example demonstrates that the prediction is adapted to the current view of the object while utilizing previous information to get a more refined final prediction. All objects shown are part of the YCB object dataset~\cite{calli2015ycb} and not observed during training. }\label{fig:ch5performer_improvement}

\end{figure} 

\begin{figure}[H]
    \centering{
        \includegraphics[width=\textwidth]{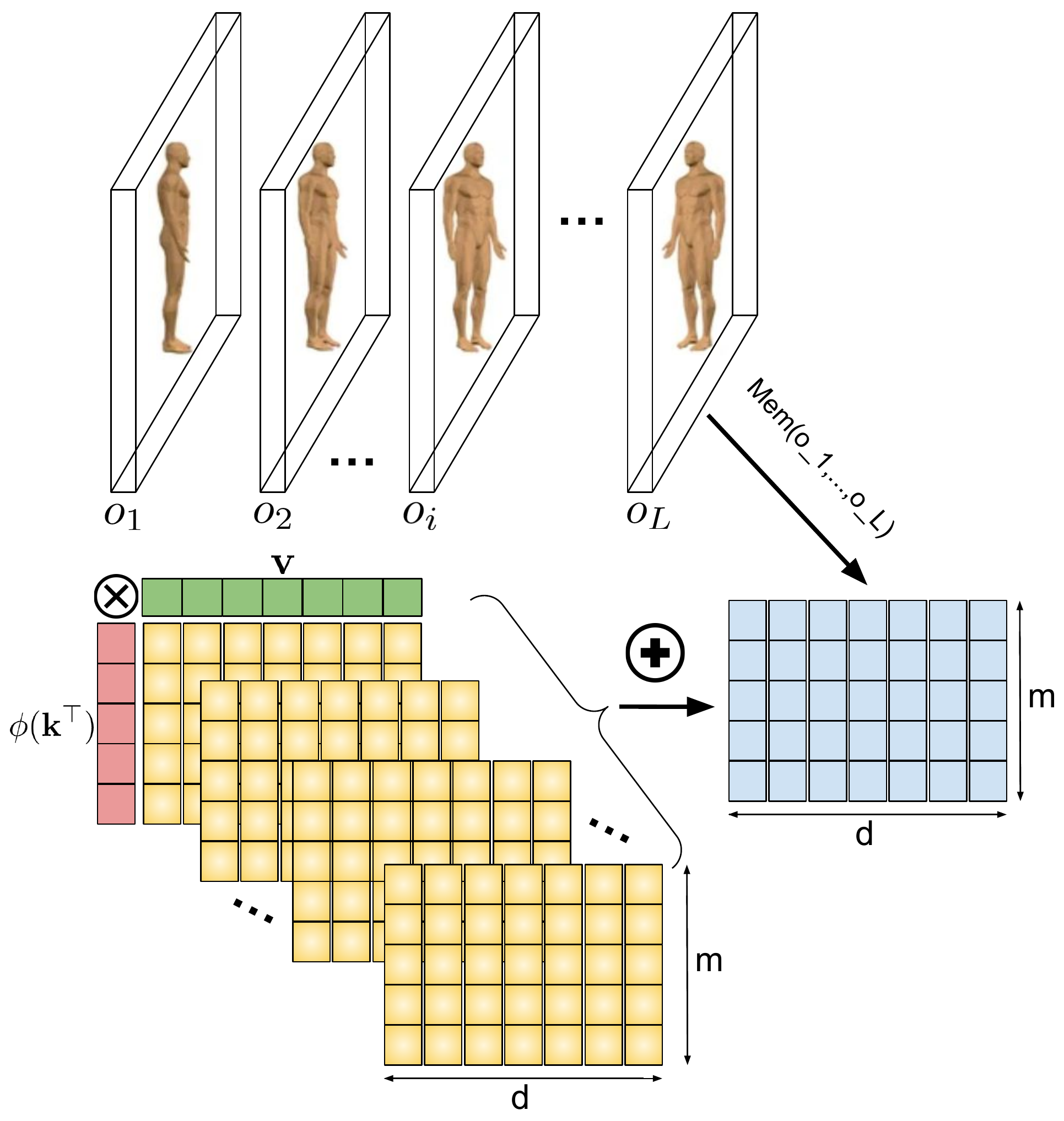}
        \caption{Each row of the resultant latent representation is approximated using the addition of the previous latent vector representations constructed from varying views of a single object.}
        \label{fig:performer_layer_algorithm}
    }
\end{figure}

All the vectors in this section are by default \textbf{row-vectors}. Consider a sequence of observations (image frames) $(o_{1},\dots,o_{L}) \in \mathbb{R}^{40 \times 40 \times 4 \times 1}$, each represented as a voxel grid with occupancy scores. Each observation $o_{i}$ is associated with a latent representation denoted as $\mathbf{v}_{i} \in \mathbb{R}^{d}$ (sometimes called a \textit{value vector}). How much the ith frame attends to jth is quantified by the so-called SoftMax kernel $\mathrm{K}(\mathbf{q}_{i},\mathbf{k}_{j})=\exp(\mathbf{q}_{i}\mathbf{k}_{j}^{\top})$ on two other (learnable) latent encodings corresponding to ith and jth observations, called \textit{query} ($\mathbf{q}_{i}$) and \textit{key} $(\mathbf{k}_{j})$  respectively.
The sequence $(o_{1},\dots,o_{L})$ defines the \textit{memory} $\mathcal{M}$ of the system. A visualization of this is shown in \autoref{fig:performer_layer_algorithm}.

For a newly coming frame $o$, the latent representations of the most relevant frame from the memory are (approximately) retrieved as: 
\begin{equation}
\label{mem-equation}
v(\mathbf{o})=\sum_{j=1}^{L} \frac{\mathrm{K}(\mathbf{q},\mathbf{k}_{j})}{\sum_{l=1}^{L}\mathrm{K}(\mathbf{q},\mathbf{k}_{l})} \mathbf{v}_{j},
\end{equation}
\noindent where $\mathbf{q}$ stands for its corresponding query. This retrieval process can be thought of as a one gradient step (with learning rate $\eta=1$) of the \textit{Hopfield network} with the exponential energy function~\cite{hopfield}. If the keys of the observations are spread well enough, the procedure within a couple of gradient steps converges to the value vector corresponding to the nearest-neighbor of $o$ from $\mathcal{M}$ (with respect to the dot-product similarity in the space where queries/keys live). This is true even for memories of the exponential size; thus, the corresponding memory model is called the \textit{exponential capacity}.

This observation is leveraged in transformers architectures~\cite{vaswani2017attention}, with attention modules equivalent to one-gradient-step Hopfield networks. This approach has a critical caveat though - the memory needs to be explicitly stored. It becomes problematic if a substantial number of observations $L$ is collected since $\mathcal{M}$ grows linearly in $L$. To address this issue, the unbiased linearization is proposed $\mathrm{K}(\mathbf{x},\mathbf{y}) \approx \phi(\mathbf{x})\phi(\mathbf{y})^{\top}$ of the SoftMax kernel, where $\phi:\mathbb{R}^{d_{QK}} \rightarrow \mathbb{R}^{m}$ 

\subsection{CNN Architecture}
\begin{figure}[t]
    \centering {
        \includegraphics[width=0.95\linewidth]{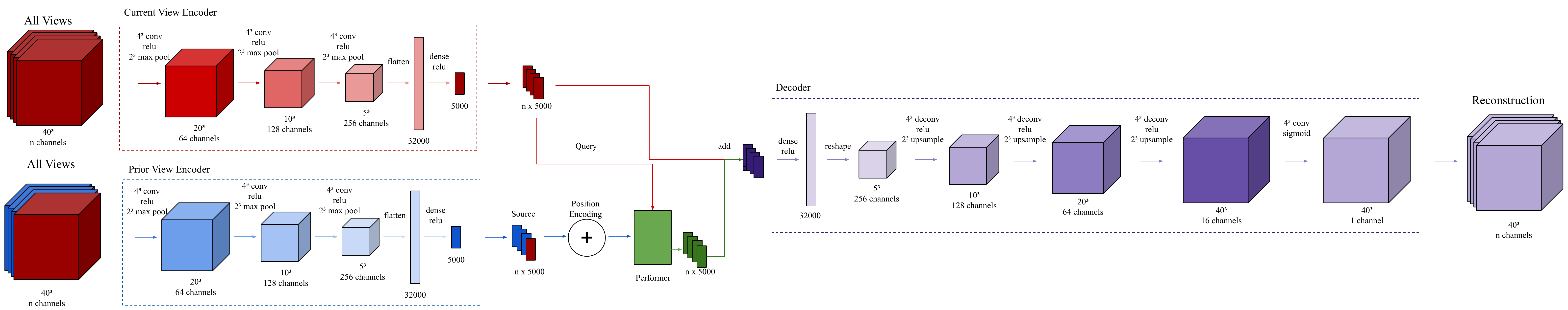}
    }
    \caption{This network takes multiple unregistered views of an object to produce a reconstruction that is better than a single or two views. It leverages a performer layer to attend each input view to the current view of the object to produce a refinement of the input. The number of views is arbitrary as the performer layer can approximate an arbitrary width input buffer. } \label{fig:performerv5architecture}
\end{figure}

Like the two-view architecture in \autoref{ch:two_view_shape_understanding}, the encoder and decoder layers are reused from the single-view architecture. The proposed architecture is shown in \autoref{fig:performerv5architecture}. The network takes $N$ unregistered views as $40^3$ voxel grid inputs. Each created by voxelizing a point cloud generated from a 2.5D depth image. All intermediate activation functions are ReLU, and the output activation function is sigmoid. A sigmoid is chosen as it outputs a value between $0$ and $1$.

An encoder is defined the same way as in \autoref{sec:two_view_cnn_architecture}, as a series of convolutional layers which are then flattened. In the proposed implementation a convolutional layer with kernel size of $4^3$, stride of $1$, and $64$ kernels followed by a max pool of $2^3$ results in an intermediate representation of $20^3\times 64$. Then a convolutional layer with kernel size $4^3$, stride of $1$, and $128$ kernels followed by a max pool of $2^3$ creates a new intermediate of $10^3\times 128$. A final convolutional layer with kernel size $4^3$, stride of $1$, and $256$ kernels followed by a max pool of $2^3$ creates a new intermediate of $5^3\times 256$. This intermediate is flattened to form a vector of size $32000$. Each encoder is followed by a dense layer of size $5000$. 

Two encoders are used, the first being the current view encoder and the second being the previous view encoder. Both encoders see all $N$ views. The previous views encoder's output of $N$ views are each given a position encoding as proposed by Vaswani et al.~\cite{vaswani2017attention}. The position encoded views are now considered the \textit{source}. The current view, passed through the current view encoder, is considered the \textit{query}. A Performer layer with token size $5000$ is introduced with SoftMax kernel transformation and random features of $256$. Additionally, the performer layer has a dropout of $0.5$ and $4$ heads. At timestep $t$ the Performer will output $1$ token, but overall will output $N$ tokens for the number of views. The current view encoding, and the token produced by the Performer are then added together. A final dense layer of size $32000$ processes the changes introduced by adding the two encodings together. 

The network then uses a decoder as defined in \autoref{sec:two_view_cnn_architecture}. The new embedding of size $32000$ is reshaped into a vector of size $5^3\times 256$. A convolutional layer with kernel size $4^3$, stride of $1$, and $128$ kernels followed by an upsample 3D of $2^3$ creates a new intermediate of $10^3\times 128$. A convolutional layer with kernel size $4^3$, stride of $1$, and $64$ kernels followed by an upsample 3D of $2^3$ creates a new intermediate of $20^3\times 64$. A convolutional layer with kernel size $4^3$, stride of $1$, and $16$ kernels followed by an upsample 3D of $2^3$ creates a new intermediate of $40^3\times 16$. Finally, a convolutional layer with kernel size $4^3$, stride of $1$, and $1$ kernel creates the final reconstruction of size $40^3\times 1$. This final convolutional layer is the reconstruction of the object with a sigmoid activation function. This model is trained using binary cross entropy loss and the Adam optimizer. 
The model was trained for $71500$ batches with a batch size of $8$ for a total of $10$ hours and $35$ minutes of training time on a NVIDIA $3090$ graphics card. Each batch contained a sample of $12$ images of the object. Training was subject to early stopping where if the validation Jaccard similarity did not increase for $5$ epochs, the training would stop. 

\section{Experiments}

\subsection{Performer Tests}
Three tests on the performer model are conducted. Each of them uses the same architecture described in \autoref{fig:performerv5architecture}. The training data for the performer model was restricted to only YCB objects as the time to train each model was a large function of the number of meshes and training examples. A similar $80/10/10$ split was employed to the previous data generation technique. 

\subsubsection{Object Hiding}
The object is progressively hidden over time by a sweeping set of voxels that occlude the view. This is evaluating whether the network can continue to output the intended completion after the object has been completely occluded. This is like the concept of object permanence in psychology. The object is hidden from view by an incremental $1$ voxel thick curtain and the voxels associated with the object are removed when occluded by the curtain. 

\subsubsection{Object Reveal}
The object is initially hidden from view by a $1$ voxel thick curtain. The object is revealed incrementally as the curtain is removed. The network will demonstrate that it can incorporate the most recent views even if no view has been provided of the object for the first few steps. 

\subsubsection{Object Panning}
A camera pans over the object capturing a sequence of views of the object. These views are taken from only one dataset, whether it be train, holdout, or test. The views are captured by rotating the camera around the centroid of the object. The network will be able to produce a better completion due to the addition of these views. 

\subsection{Evaluation}
A collection of holdout views of training objects is reserved along with a collection of models not seen during training with generated views. Each view is completed, and then compared against the ground truth object for reconstruction quality. These views were generated using the methodology described in \autoref{sec:fixed_voxel_data_generation}. The dataset used is identical to that in \autoref{ch:two_view_shape_understanding} as well. For the purposes of this chapter only the Jaccard similarity metric was used. The Jaccard similarity between sets A and B is given by:
\[
J(A, B) = \dfrac{|A\cap B|}{|A\cup B|}
\]
The Jaccard similarity has a minimum value of 0 where A and B have no intersection and a maximum value of 1 where A and B are identical~\cite{jaccard}.

\section{Results}

\begin{figure}[htp]

\begin{subfigure}{\textwidth}
\includegraphics[width=\textwidth]{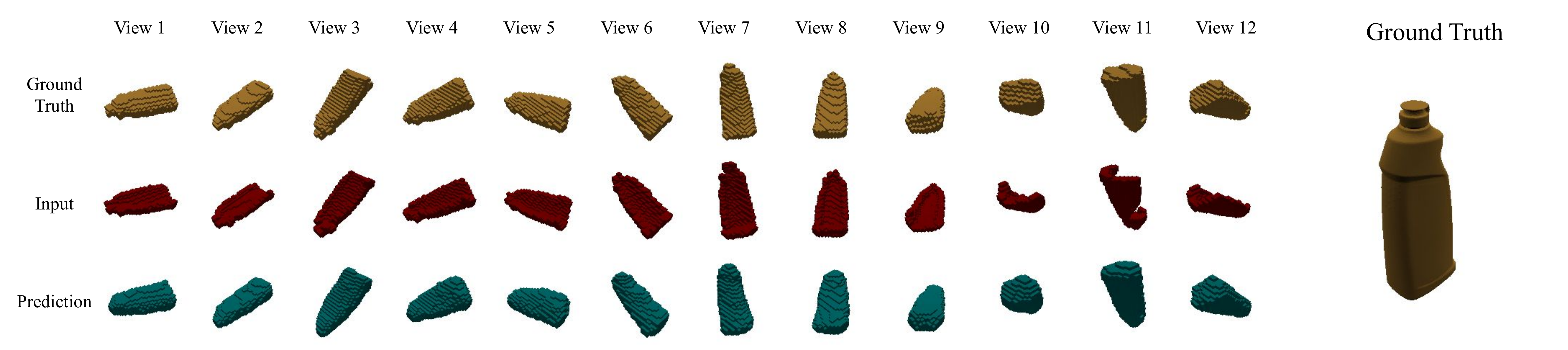}
\caption{Object Panning}
\label{fig:ycb_object_panning_example}
\end{subfigure}
\bigskip

\begin{subfigure}{\textwidth}
\includegraphics[width=\textwidth]{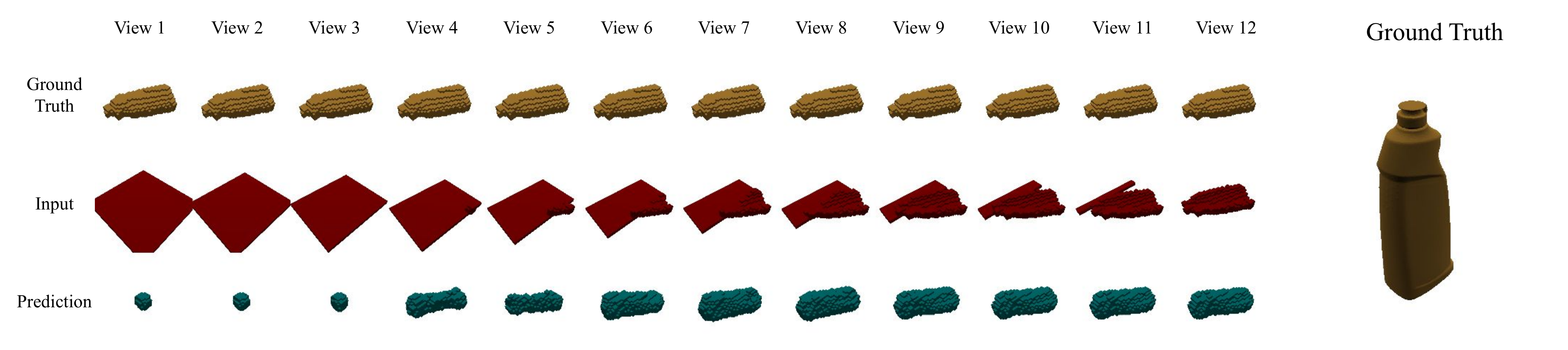}
\caption{Object Reveal}
\label{fig:ycb_object_reveal_example}
\end{subfigure}
\bigskip

\begin{subfigure}{\textwidth}
\includegraphics[width=\textwidth]{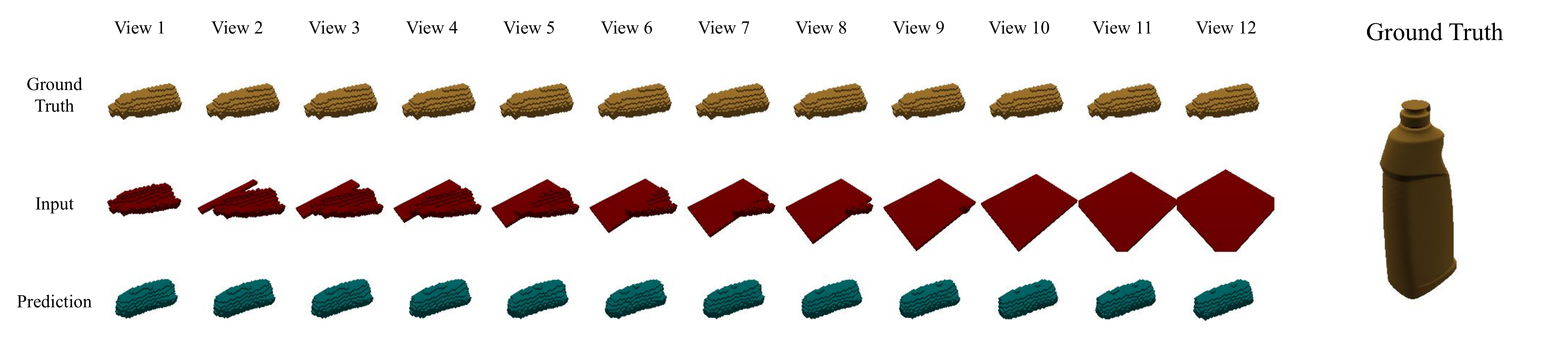}
\caption{Object Hiding}
\label{fig:ycb_object_hiding_example}
\end{subfigure}

\caption{The reconstruction results from the three test conditions for the performer model, object panning, object revealing, and object hiding. The object panning case shows the network can reconstruct the object at any orientation provided considering previous views. The object revealing shows a series of default guesses until parts of the object are revealed in views 6 and 7 at which point it reconstructs the whole object. The object hiding shows the networks ability to remember the object geometry despite it no longer being visible in the input. The input into the network is shown in red, the ground truth is shown in yellow, and the prediction is shown in green. } 
\label{fig:performer_experiment_results}
\end{figure} 

Sample data from the performer tests is shown in \autoref{fig:performer_experiment_results}. The object hiding experiment was able to show that the agent can correctly recall information about an object hidden behind a curtain. The object was fully occluded from view at the 8th image but was still outputting the same reconstruction throughout all demonstrations. In the object reveal case, the network output nothing meaningful until the 6th view where it was able to see part of the object. In the 7th view it got a more complete representation of the object and correctly produced a full completion equivalent to the object hiding case. In the object panning condition the network was able to correctly reconstruct the object in a variety of orientations utilizing information from previous views to further refine the prediction of the object over time. 

\begin{figure}[ht!]
    \centering
    \begin{subfigure}{.5\textwidth}
      \centering
      \includegraphics[width=.95\linewidth]{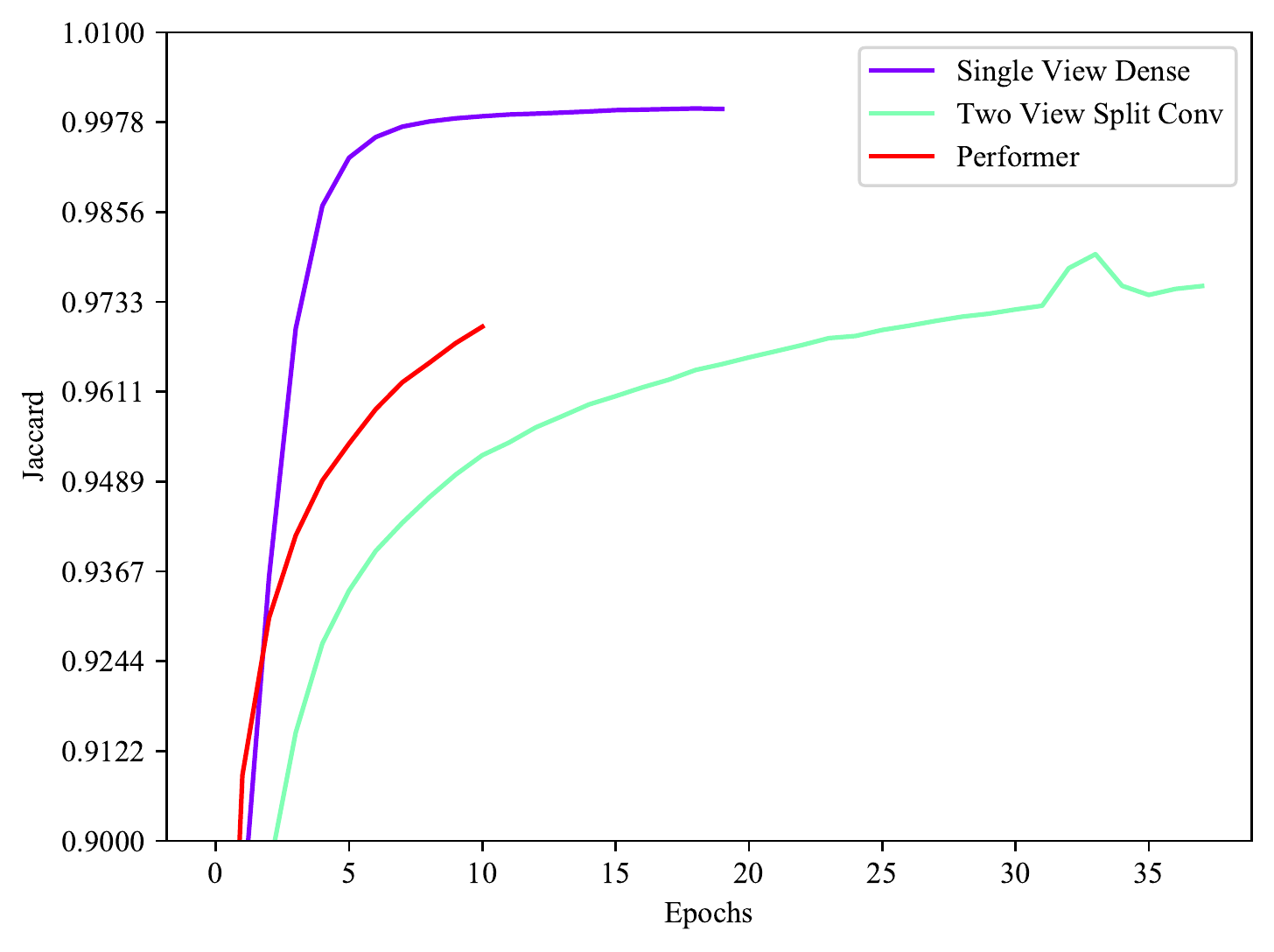}
      \caption{Training Objects Reconstruction Quality Per Epoch}
    \end{subfigure}%
    \begin{subfigure}{.5\textwidth}
      \centering
      \includegraphics[width=.95\linewidth]{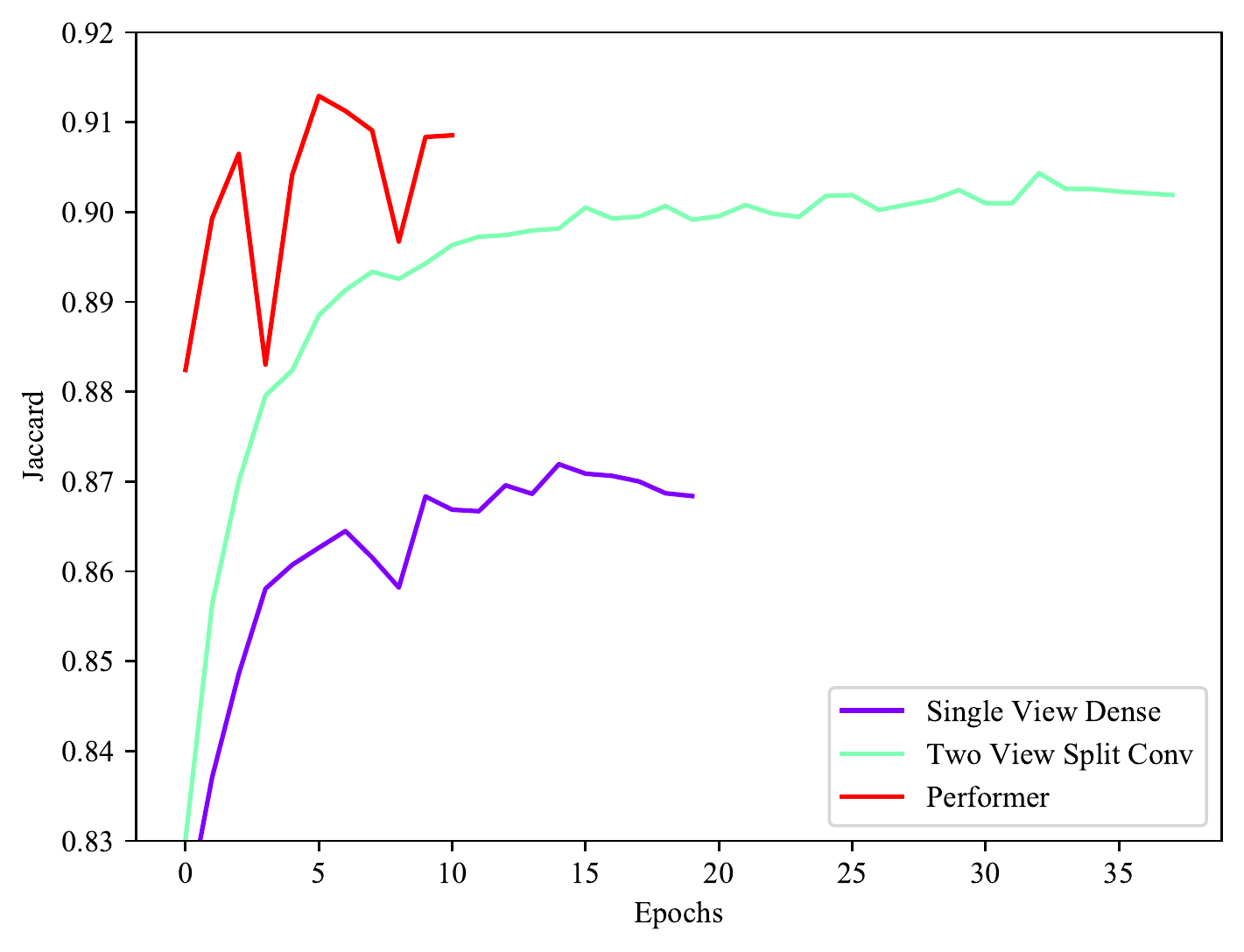}
      \caption{Unseen Objects Reconstruction Quality Per Epoch}
    \end{subfigure}
    \caption{(a) shows the Jaccard similarity over time for observed objects as networks YCB panning trained and (b) shows the Jaccard similarity over time for objects in the holdout dataset as the networks in the YCB panning trained. While the performer model does not achieve train time performance of the single-view dense model, it achieves better performance for unseen objects in validation. The Performer model was also able to converge much faster than the single-view or two-view models. A higher Jaccard is better. }
    \label{fig:performer_training_metrics}
\end{figure}

The performer training and validation metrics as a function of epochs is shown in \autoref{fig:performer_training_metrics}. For training data, the single-view dense model performs significantly better. This is because the single-view model is the best at memorization. However, the single-view model fails to generalize to unseen objects well. The performer-based model shows a substantial improvement in Jaccard quality and converges faster than the two-view and single-view model as a function of epochs. The faster convergence can be attributed to the increased data that each encoder sees per batch versus a two-view or single-view model. 

\begin{table*}[ht!]
    \centering
    \begin{tabular}{|c|c|c|}
    \hline
    \multicolumn{1}{|c|}{\begin{tabular}[c]{@{}c@{}}\textbf{Method Name} \\  \textbf{}\end{tabular}} 
    & \multicolumn{1}{c|}{\begin{tabular}[c]{@{}c@{}}\textbf{Train Jaccard} \\  \textbf{}\end{tabular}} 
    & \multicolumn{1}{c|}{\begin{tabular}[c]{@{}c@{}}\textbf{Test Jaccard} \\  \textbf{}\end{tabular}} \\ 
    \hline
        Single-View          & 0.7472         & 0.7027      \\ \hline
    	Two-View             & 0.7653         & 0.7259      \\ \hline
    	Performer            & 0.8383         & 0.7788      \\ \hline
    \end{tabular}
    \caption{\textbf{YCB Hiding Reconstruction Results}, measuring the performance of the reconstruction quality of different meshes when hiding the views. The single-view serves as a baseline for if one view is provided, and the two-view case considers one view where the object is visible and one where it is not. A higher Jaccard is better. }\label{tab:performer_hiding_results}
\end{table*}

\begin{table*}[ht!]
    \centering
    \begin{tabular}{|c|c|c|}
    \hline
    \multicolumn{1}{|c|}{\begin{tabular}[c]{@{}c@{}}\textbf{Method Name} \\  \textbf{}\end{tabular}} 
    & \multicolumn{1}{c|}{\begin{tabular}[c]{@{}c@{}}\textbf{Train Jaccard} \\  \textbf{}\end{tabular}} 
    & \multicolumn{1}{c|}{\begin{tabular}[c]{@{}c@{}}\textbf{Test Jaccard} \\  \textbf{}\end{tabular}} \\ 
    \hline
        Single-View          & 0.7472        & 0.7027      \\ \hline
    	Two-View             & 0.7653        & 0.7259      \\ \hline
    	Performer            & 0.7226        & 0.6964      \\ \hline
    \end{tabular}
    \caption{\textbf{YCB Reveal Reconstruction Results}, measuring the performance of the reconstruction quality of different meshes when revealing the views. The single-view serves as a baseline for if one view is provided, and the two-view case considers one view where the object is visible and one where it is not. A higher Jaccard is better. }\label{tab:performer_reveal_results}
\end{table*}

\begin{table*}[ht!]
    \centering
    \begin{tabular}{|c|c|c|}
    \hline
    \multicolumn{1}{|c|}{\begin{tabular}[c]{@{}c@{}}\textbf{Method Name} \\  \textbf{}\end{tabular}} 
    & \multicolumn{1}{c|}{\begin{tabular}[c]{@{}c@{}}\textbf{Train Jaccard} \\  \textbf{}\end{tabular}} 
    & \multicolumn{1}{c|}{\begin{tabular}[c]{@{}c@{}}\textbf{Test Jaccard} \\  \textbf{}\end{tabular}} \\ 
    \hline
        Single-View          & 0.847         & 0.7918      \\ \hline
    	Two-View             & 0.8555        & 0.7724      \\ \hline
    	Performer            & 0.9555        & 0.7877      \\ \hline
    \end{tabular}
    \caption{\textbf{YCB Pan Reconstruction Results}, measuring the performance of the reconstruction quality of different meshes given a sequence of views of the object. The single-view serves as a baseline for if one view is provided, and the two-view case considers two views in the sequence of panned views. A higher Jaccard is better. }\label{tab:performer_pan_results}
\end{table*}

The quantitative results for the performer model show an interesting trend. The object hiding results are shown in \autoref{tab:performer_hiding_results}. The results shown that for training data the performer model greater outperforms both the single-view reconstruction and two-view reconstruction. Additionally, for test meshes unobserved during training it still outperforms the single-view and two-view cases. This can be attributed to the significantly higher number of views that the performer model was able to observe during training as opposed to the two other models. The performer reveal results, shown in \autoref{tab:performer_reveal_results}, show that the performer model can incorporate novel views after an object is revealed. The completion quality is lower than the single-view and two-view cases due to the erroneous "ball" that shows up when nothing is visible. Due to the network being trained to always output a mesh, it learned to output an average over the dataset which featured a lot of spherical objects. This can potentially be remedied by training it with both empty views when nothing is visible and the expected object when parts are visible. The performer panning results are shown in \autoref{tab:performer_pan_results}. The results show that the network was able to learn to reconstruct objects during training better than the single-view and two-view models. However, during evaluation of the performer model it operated at a similar level to both the single-view and two-view architectures. This is attributed to the small dataset used to train the model. With additional training data from the GRASP dataset, the network should be able to generalize better to unseen objects. 

\begin{table}[ht!]
	\small
	\centering
	\begin{tabular}{|l|r|r|r|r|r|r|r|r|}
		\hline
		\textbf{Split}         & \multicolumn{1}{l|}{View 1} & \multicolumn{1}{l|}{View 2} & \multicolumn{1}{l|}{View 3} & \multicolumn{1}{l|}{View 4} & \multicolumn{1}{l|}{View 5} & \multicolumn{1}{l|}{View 6} & \multicolumn{1}{l|}{View 7} & \multicolumn{1}{l|}{View 8} \\ \hline
		\textbf{Train}         & 0.9554                      & 0.9552                      & 0.9554                      & 0.9555                      & 0.9556                      & 0.9558                      & 0.9556                      & 0.9557                      \\ \hline
		\textbf{Holdout Views} & 0.9090                      & 0.9107                      & 0.9096                      & 0.9105                      & 0.9106                      & 0.9126                      & 0.9151                      & 0.9197                      \\ \hline
		\textbf{Test}          & 0.7863                      & 0.7864                      & 0.7867                      & 0.7874                      & 0.7871                      & 0.7877                      & 0.7879                      & 0.7881                      \\ \hline
	\end{tabular}
	\caption{Results per view of the YCB panning dataset using the performer reconstruction model. Each view increases the Jaccard quality of the reconstruction. The only exception is the training cases, but this can be attributed to memorization of a small dataset. Higher Jaccard scores are better.}
	\label{tab:performer_pan_over_time_results}
\end{table}

The results shown in \autoref{tab:performer_pan_over_time_results} show how each incremental view impacts the performance of the reconstruction of the object. The network can improve the reconstruction quality with the addition of more views. The train condition was unable to leverage the additional information to produce a better completion which can be attributed to overfitting the input data and memorizing it. In the holdout views and test cases the network was able to improve the reconstruction quality overall by the end. This shows how the network can improve on reconstruction quality leveraging multiple views. 

\section{Conclusion}
This chapter presented a novel approach to leveraging multiple unregistered views of an object to predict the mesh of an object with higher accuracy that a single-view or two-view model would. This method leveraged a novel attention layer called a Performer~\cite{performer} which can approximate arbitrarily long buffers of input to further refine prediction. The network was shown to be able to remember objects that are no longer visible in the input and to leverage information of the object that are captured on a delay. This system can be utilized in a mobile manipulation pipeline where views cannot be registered together.

\clearpage

\phantomsection
\chapter{Discussion and Conclusion}
\label{ch:discussion}

\section{Introduction}
This thesis has introduced methodologies for designing and implementing an end-to-end solution for localization-free robotic mobile manipulation of unseen objects. Throughout this work, distinct design paradigms have come together to provide a novel solution that can be extended in future directions. Existing navigation, shape understanding, and next-best-view planning paradigms were modified and extended to be applied to mobile manipulation. The work presented in this thesis breaks a common convention of assuming localization is mandatory to build a robotic mobile manipulator. Further challenging this assumption is the integration of a novel object shape completion system without impacting the performance of the robotic mobile manipulator. Several spin-off areas of research are documented in this chapter, such as implicit object representation, generalized learned visual-semantic navigation, visual-tactile next-best-view planning, and higher resolution shape understanding.

The many aspects of localization-free mobile manipulation discussed in this dissertation are applicable to a broad range of domains to be considered further. The application discussed in \autoref{ch:learning_visual_navigation} and \autoref{app:minerl_basalt} of a Minecraft agent learning to operate in simulation using human demonstrations is one such example. Further extension to other game environments, drone and aquatic robots, and even local human navigation are all unrealized potential applications of this mobile manipulation paradigm. While this dissertation did not explore forms of manipulation other than grasping, further exploration in these manipulation tasks may reveal yet undiscovered applications of the mobile manipulation subcomponents. 

This concluding chapter concludes the analysis of an end-to-end localization-free mobile manipulation system. The following sections summarize the work thus far, present the contributions of this work, and indicate recommendations for future research in mobile manipulation. 

\section{Summary}
This thesis has presented a solution to mobile manipulation. Each chapter was laid out in the order that an agent would perform each subtask for mobile manipulation. The first stage in this mobile manipulation pipeline is for the agent to navigate throughout its environment. \autoref{ch:learning_visual_navigation} presents a solution to visual ego-centric motion in real-world environments. At each timestep the agent determines a subsequent action to take to navigate successfully to a novel 8-image RGBD panoramic target goal within its environment. The agent uses a binary classifier called a goal checker to determine whether it is done at each time step and spins in place to determine how confident it is in reaching the final goal location. The results in \autoref{ch:learning_visual_navigation} showed that when compared to baseline methods, the learned navigation system outperformed them by a significant margin. Additionally, the agent's performance with respect to an idealized goal checker with access to the agents position in the environment performed similarly to the agent using the trained goal checker. 

Upon reaching a target object, the agent needs to be able to reason about the object's geometry. In \autoref{ch:visual_tactile_manipulation}, an agent was trained to utilize visual and tactile information about an object to refine its understanding. A novel visual-tactile fusion CNN was trained to perform a shape completion utilizing both pieces of information to predict a resultant occupancy grid which is then turned into a mesh. This mesh is then used via GraspIt!~\cite{miller2004graspit} to plan and execute a grasp on the object. Results show that when compared to algorithmic approaches and GPIS~\cite{williams2007gaussian}, the proposed visual-tactile CNN performs best. The tactile information was also enough to improve the completion accuracy over a depth-only model in live testing. The mobile manipulator is now able to grasp the object using sensory information gained from its environment. 

The agent is not restricted to only using tactile information. The agent can also move around its environment to collect additional images of a target object. \autoref{ch:two_view_shape_understanding} describes a process where the agent uses two images of a target object to refine its understanding of the object's geometry. The proposed two-view CNN uses a modified version of the visual-tactile fusion CNN where it encodes two inputs into a dense embedding, adds them together, and then uses a convolutional decoder to reconstruct the object. Results described in \autoref{ch:two_view_shape_understanding} show that the two-view CNN outperforms a series of ablations for random views of the object. 

The agent needs an intelligent way to incorporate multiple views into a mobile manipulation pipeline. In \autoref{ch:mobile_manipulation}, the agent uses a novel next-best-view PCA based method to select views of the object to capture next. The agent then uses a novel panoramic prediction method to predict the view of the environment from that next-best-view to use the learned visual navigation system to navigate there without needing to localize itself at runtime. The two-view CNN performance is enhanced by utilizing a next-best-view over a random view approach used in \autoref{ch:two_view_shape_understanding}. The results show that the agent was able to outperform several baselines. The performance of the local navigation methodology proposed is on par with a baseline method that uses the ROS navigation stack. The overall performance, as measured by the introduced E2ESPL metric, performs favorably when compared with a series of ablations. 

A potential improvement over the proposed two-view completion method is to utilize a Performer~\cite{performer} layer to leverage multiple unregistered views of an object to further refine the predicted mesh geometry. The Performer layer allows the agent to attend the current view with each previous view. The embeddings for each output are added together to produce a series of incrementally better completions. Results show the Performer CNN can improve completion quality over the proposed two-view CNN. Additionally, experiments showed it can remember objects after they are hidden or incorporating views of an object after they are revealed. 

\section{Contributions}
Section~\ref{sec:introduction_contributions} in \autoref{ch:intro} itemizes the contributions of this thesis. This chapter elaborates on these contributions in the areas of robotic mobile manipulation:
\begin{itemize}
    \item \autoref{ch:learning_visual_navigation} demonstrated how to build a learned visual navigation system that trains an agent to navigate through an environment using a series of previous images and a novel panoramic target goal. 
    \item \autoref{ch:visual_tactile_manipulation} presented a learned visual-tactile shape completion method that given the initial and tactile views creates a more accurate reconstruction for robotic manipulation.
    \item \autoref{ch:two_view_shape_understanding} explored a learned two-view shape completion method that given the initial and next-best-view creates a more accurate reconstruction for robotic manipulation. The chapter also showed an ablation study demonstrating the performance benefit of using two views.
    \item \autoref{ch:mobile_manipulation} presented an end-to-end system for mobile manipulation of household graspable objects utilizing novel learning algorithms. Further, it described an algorithm that uses a predicted panoramic goal and reuses our long-range learned image navigation system to navigate to the next-best-view locally. Finally, it showed an algorithm that takes an initial shape completion estimate of the manipulation target that uses voxel grid occupancy thresholding to plan the next-best-view. The work in this chapter was verified via a series of metrics and benchmarks that can be used to evaluate mobile manipulation systems in future work
    \item \autoref{ch:performers} explained the implementation of a learned multiple-view shape completion method that can take an arbitrary number of views to refine its understanding of the object geometry. It showed that a performer-based model can remember geometry seen previously as well as its ability to incorporate additional views into its reconstruction. 
\end{itemize}
\noindent All these contributions culminate in a data-driven robotic mobile manipulation pipeline that can manipulate unseen objects without localizing itself at runtime. 

\section{Current Limitations and Future Work}
The research demonstrated in this dissertation spans several different domains in shape understanding, navigation planning, manipulation, and next-best-view planning. Something shared in all the work described here is that the majority has been validated in simulation alone. This limitation can be attributed in part to the COVID-19 pandemic that has plagued the world since the first case was reported in November of 2019, a grim reality that restricted lab access and has made research more difficult for scientists worldwide. Future work in this field should resume in the real-world as the world recovers from stay-at-home orders.

\subsection{Learned Visual Navigation Future Work}
\autoref{ch:learning_visual_navigation} explored how to build a visual navigation system that utilized advancements in simulators and real-world scanned data to enable a robot to learn to navigate using egocentric motion planning. In 2018, the state-of-the-art simulator for navigation and manipulation research was the Gibson~\cite{GIBSON} simulator with the Matterport3D~\cite{Matterport3D} and Stanford 2D3DS~\cite{stanford2d3ds} datasets. This simulator and the datasets are still useful for conducting robotics research as the simulator easily allowed for grasping experiments in simulation due to the underlying physics simulator being PyBullet~\cite{PYBULLET}. However, Habitat Sim~\cite{habitat} has received a lot of community support for manipulation tasks as well as 3D environment datasets. Given its superior ability to capture over 10000 frames per second when recording visual information, it may be worth evaluating whether an agent can be trained in real time with the Habitat simulator without prefetching all the training data. This means an agent could be trained more thoroughly over the entire environment without worrying about taking up too much storage. Hard drive storage space was a limitation during training of these agents and would benefit from some additional engineering. While this would reduce auditability of the system during training, in the case of a collision with the environment it would allow the agent to quickly replan its trajectory make sure that trajectories go smoothly. Another potential solution to data storage is working on the embedding representation of images and storing those instead of raw images. 

The storage of training data could be remedied using a learned simulator environment. The World Models paper by Ha et al.~\cite{ha2018worldmodels} provides an alternative to simulated data generation from real-world environments. A series of actions of an agent could be captured in the real-world along with the images seen at each time step. These could be used to train a "world model" of the environment that could then be used to train a policy model to navigate through that world. While this approach would potentially reduce the overall time required to scan a real-world home as it would rely only on color, depth, and actions taken, it may result in worse domain adaptation from simulator to real. An additional style transfer model like Goggles from Gibson~\cite{GIBSON} would be needed to make sure the agent could adapt from simulated data to real-world data.

Because the proposed visual navigation system relied heavily on behavioral cloning from expert trajectories captured using a 2D map of the environment, addressing behavioral drift would be a useful next step. Utilizing a methodology such as SQIL~\cite{reddy2019sqil} to introduce out of distribution cases during test runs would improve the accuracy of reaching unseen targets by helping the agent enter within distribution states. This adjustment would be a straightforward change of running the agent through a test sequence that was not seen during training, finding an expert trajectory through that trial, finding the stages where the agent deviates heavily from the intended path, recomputing the trajectory at the deviation, and treating the process as a new training trajectory for the policy model. 

During this research, different model architectures were compared as illustrated in \autoref{ch:learning_visual_navigation}. These alternatives included an LSTM model, longer history buffers, and reinforcement learned models. A potential additional model that was not assssed would be to utilize attention~\cite{vaswani2017attention} to help the agent attend to previous views of the current view to indicate which action should be taken. This method is like the multi-view object completion method shown in \autoref{ch:performers}. Incorporating attention would involve changing the policy model to accept n-views, with n-1 views attending to the current view and then adding the embedding of that attention layer with the embedding of the current view, allowing the network to prioritize information from the current view if it is more helpful that previous views. A self-attention model could also be evaluated here. Utilizing Performers~\cite{performer} would be potentially better as they allow for arbitrary length by approximating a longer fix width buffer, resulting in a policy that can utilize an arbitrarily long history buffer, resulting in an agent that can leverage information from its entire trajectory.

The training data used for this research was semantically labeled at the mesh level. This means that this signal is available in the simulator environment. An RGBD image combined with a per-pixel semantic label may be a rich signal that can be used on the output of the autoencoder to provide the agent with semantic information in the dense embedding used to train the policy model. It may be difficult to appropriately use this information to improve the performance of the navigation agent, but it is as simple as adding the labels to the output of the reconstructed image for each pixel. 

Finally, the major shortcoming of this work is its inability to extend to unseen environments. While it can be difficult to explore an environment and map it simultaneously, this behavioral cloning strategy can be used in other ways to enable an agent to determine how to move around environments it has not observed before. Many home layouts share similarities, where the kitchen may be close to a dining room or bathrooms placed close to areas with high traffic. These kinds of intuitions could be provided to the agent during training to create a policy model with general floor layouts in mind. An additional policy model could then refine this understanding by labeling its environment and providing a probabilistic map of whether to navigate to the next location based on the input images. Additional training data could be generated this way by creating a generative model to produce additional floor layouts based on the real-world homes from the Matterport 3D~\cite{Matterport3D} and Stanford 2D3DS~\cite{stanford2d3ds} datasets. 

\subsection{Visual-Tactile Shape Understanding Future Work}

The shape understanding field has changed since the original inception of the visual-tactile fusion CNN presented in \autoref{ch:visual_tactile_manipulation}. Utilizing two encoders, like the two-view CNN in \autoref{ch:two_view_shape_understanding}, would potentially improve the performance of the model as the network could differentiate which sources of information are contributing to the reconstruction at the architecture level. Due to the time-series nature of collecting multiple tactile inputs, the network could leverage attention~\cite{vaswani2017attention} or performer~\cite{performer} to refine its understanding as it collects more data. This adjustment not only would provide a visually interesting demonstration; it would also provide incremental information about where best to look next. This system could be further improved by utilizing a discriminator to determine whether there is enough information, such as the work in Tandem~\cite{xu2022tandem}. Additionally, newer point cloud based neural network architectures may be beneficial by reducing the memory requirements for the model which would improve resolution of the reconstruction. 

Next-best-view methodology as discussed in \autoref{ch:mobile_manipulation} would be useful for improving the performance of this system. The tactile finger can directly follow the next-best-view vector to collect additional information about the object. Given that next-best-view planning improved performance of the two-view CNN described in \autoref{ch:mobile_manipulation}, it would make sense that a next-best-touch system could improve performance over random touches on the occluded side of the object. 

The empty space of the object during the guarded tactile moves and visibly empty voxels were not considered during reconstruction of the object. A potential improvement over the initial architecture would be to use a ternary voxel representation of empty, occupied, and unknown. Empty voxels would be reserved for voxels observed to be empty and unknown for voxels in occluded or unobserved regions of the space. Providing this information to the CNN would increase the space complexity of the reconstruction but could yield further improvement in reconstruction quality, and therefore improvements in grasp quality and success. 

GraspIt! was used to plan grasps for the visual-tactile fusion system. Graspit! does not take advantage of the uncertainty of the object reconstruction for grasping. A novel grasp quality metric that takes advantage of the uncertainty of the reconstruction would be a straightforward augment of the GraspIt! system without developing a new grasp planner. Because GraspIt! does not have a good integration into the ROS ecosystem that was used for trajectory planning of the arm, a general-purpose grasp planner to plan grasps on meshes using a variety of different grippers would be the preferred solution. This kind of engineering endeavor is challenging and would require an advanced robotics software engineer to implement, but it would be a great service to the field. The disconnect of MoveIt! not having a dedicated grasp planner has made much of this research challenging with many hacked together solutions implemented. 

A final improvement to this visual-tactile shape completion system would be to use the tactile information captured at grasp time to refine the shape estimation and potentially replan the grasp. The agent did not utilize any tactile information captured at the time of grasping the object. With tactile information at grasp time the agent would have full understanding of the bounds of the object and could determine if the grasp would be unstable in that configuration. If unstable it could then replan the grasp and reorient its end effector to successfully pick up the object. 

\subsection{Multi-View Shape Understanding Future Work}

Like the visual-tactile fusion work discussed in \autoref{ch:visual_tactile_manipulation}, the completions in \autoref{ch:two_view_shape_understanding} and \autoref{ch:performers} were restricted to $40^3$. These completions are sufficiently dense for common household objects, with the lack of fidelity becoming problematic as the complexity of objects increases. The reconstruction of objects from the Thingi10K~\cite{thingi10k} would prove challenging as they have small features that are difficult to capture in $40^3$ or are difficult to detect with a depth camera. The major problem with increasing the resolution of input voxel grids is the lack of graphics card memory. GPU memory has improved in consumer grade graphics cards in recent years, but still is not at a level needed for richer representations of objects. The work done by Varley et al.~\cite{varley2017shapecompletion_iros} required at least 11GB of VRAM to train their model with the correct specifications. The two-view and performer models described in \autoref{ch:two_view_shape_understanding} and \autoref{ch:performers} require at least 24GB of VRAM provided by a NVIDIA 3090 GPU. Prices for graphics cards have been exceptionally expensive in the wake of the COVID-19 pandemic, with the cost of a card with at least 11GB of VRAM or greater being \$1000 or more. An NVIDIA A100 GPU, with 80GB of VRAM, would have been an ideal alternative. The cost for this graphics card is currently \$10000 which is out of reach for many researchers. Future reductions in costs of training hardware such as Google's TPU architecture and cloud computing services will remedy this issue allowing for higher resolution 3D convolutional models to increase in size, resulting in higher completion quality and therefore greater chances of grasp success. 

Some testing in higher single-view resolution models of $64^3$ and $128^3$ was performed, but in certain circumstances failed to converge due to the sparsity of the output data. A potential solution to this inconsistency would be to use a signed-distance-field (SDF) representation that would provide a richer output and input of the target object. This change in the representation of the data would require no modification to the architecture of the neural network but instead a modification of the training data used. An SDF representation would mean that training data stored on hard drives would occupy more space, however. In the current implementation of the training data for the shape completion models, the data is stored via run-length-encodings of a binary voxel grid encoded via Binvox~\cite{binvox, nooruddin03}. Data that currently takes kilobytes for $128^3$ would instead occupy several megabytes. With over $400000$ training pairs this would result in an explosion of training data. Computing the data live might be a potential solution to this problem, where the object is rendered and voxelized on demand rather than being precomputed. This approach would be an interesting direction to explore in the future. 

Something that could improve reconstruction quality further would be to utilize sinusoidal activation functions for the dense embeddings over a ReLU activation function. In preliminary testing, the sinusoidal activation function had better reconstruction quality simply by replacing the activation function in the two-view and single-view models evaluated in \autoref{ch:two_view_shape_understanding}. This idea was inspired by the success of SIREN~\cite{siren} in reconstructing images by utilizing Sine in its model. Additionally, this methodology could be incorporated into the convolutional layers or performer layers for unique improvement. 

The final direction for future exploration of this work would be to create an implicit representation of 3D geometry. Newer architectures, called neural radiance fields~\cite{eslami2018neural}, can convert an internal representation of an environment to predict the appearance of the scene from previously unobserved viewpoints. Current research focuses on 2D images or even 2.5D images but has not been explored using 3D convolutions. There may be additional structural information that can be derived from the spatial relationship of voxels that could improve the accuracy of these models. An implicit representation also addresses the issue of low resolution for these voxel grids as it can be refined indefinitely over the volume of the mesh to improve the resolution of the output. 

\subsection{Mobile Manipulation Future Work}

The mobile manipulation pipeline described in \autoref{ch:mobile_manipulation} makes many assumptions that constrain the scope this system is compatible with. The agent is assumed to have access to a scanned version of the environment beforehand. The environment has objects of interest that are graspable and are situated so any next-best-view can be captured. The tables are placed so that the agent can see from below or above by raising its torso. The next-best-view is within $1m$ of the agent's current position so that the predicted panorama has minimal holes for depth information. The agent was only tested in one environment, the $house1$ environment from the Matterport 3D~\cite{Matterport3D} dataset in simulation. Many of these conditions are not feasibly achieved at runtime in the real-world. The lack of testing in multiple environments therefore makes it difficult to assess how generalizable this pipeline can be empirically. 

Future work should address the lack of testing by using the entire Matterport 3D dataset or considering the larger Stanford 2D3DS dataset with its sprawling campus scans. real-world testing would also help to verify that the proposed system can properly extend to a variety of environments as lighting conditions in the real-world can differ dramatically based on the time of day. Artificially placing object surfaces in the target mesh did not allow for testing on arbitrary surfaces. The surfaces in the ground truth mesh are noisy and uneven which makes testing grasping challenging. More software engineering can be put toward addressing these challenges by building a more robust object placement system for collecting training data. Using a newer simulator such as Habitat~\cite{habitat} could also be useful as it includes labeling of manipulatable objects in its environments. 

Future work in next-best-view planning should consider whether a next-best-view is needed. The work in \autoref{ch:mobile_manipulation} assumes that a second view is always advantageous due to a lack of an oracle determining whether the agent has enough information. There are many cases where a single view provides enough information for a grasp plan and having a network that can assess the completion quality of a current completion would be especially useful. A solution would be to design a learned quality discriminator that takes the current view of the object and the prediction of the object and outputs the Jaccard quality score between $0$ and $1$. This network would be able to leverage the existing training data for the shape understanding architectures while providing a much-needed benefit to next-best-view planning. 

The next-best-view algorithm could be further improved by creating a learned next-best-view. The PCA based method does not consider the potential quality gained by that view or by nearby views. Instead, a model could be trained to take both the current view and the current reconstruction and output the predicted Jaccard similarity score for each view of the object as a sequence of points on a sphere. These positions can then be filtered whether they can be captured in the environment and selected for most improvement on the current quality. This predicted next-best-view can be combined with the previously mentioned learned quality discriminator to determine whether that view is good enough. This method would fix the case where the PCA method is too difficult to capture and utilize the entire input when determining a next-best-view rather than relying on voxels close to the decision boundary. 

In this robotic mobile manipulation pipeline, the agent did not consider whether the object would be reachable or graspable at the target location. A future direction would be to use work in reachability aware workspace planning~\cite{akinola2021dynamic} to determine the ability for the agent to grasp from the target location. Many grasps failed during testing due to a lack of checking for reachability at the grasp stage. Other failures were due to strange paths planned by the open motion planning library even when the system was constrained to not collide with the target object. Part of this result is due to the decoupling of grasp planning and trajectory planning. The other part of this problem is it can be difficult to generate paths between arbitrary points quickly. Future work in trajectory planning with open-source trajectory planners would improve the ability for researchers to integrate mobile manipulation systems. 

The work described in \autoref{ch:performers} with multiple view shape understanding was not integrated into the final mobile manipulation system as the next-best-view system did not incorporate multiple unregistered views and could use only the current prediction to plan a next-best-view. A refined next-best-view architecture that is learned from multiple views could improve the reconstruction accuracy of predicted shapes. An example would be a modification of the performer CNN that instead of predicting the object geometry, would predict the vector corresponding to the next-best-view. This system could take each of the unregistered views as input utilizing attention to inform the next best action to take. 

This mobile manipulation system was validated only with grasping of objects. Future work can consider more diverse tasks, such as stacking, pick and place, and door opening. New environments provided by the Habitat simulator aim to provide a sandbox for manipulation task research. Work done in the MineRL Basalt competition in \autoref{app:minerl_basalt} has shown that the navigation system can be extended to video games. This mobile manipulation system could also be extended to game playing where players are tasked with manipulating their environment and moving around using only visual information. This system could also be adapted to a human operator moving through an environment, providing a sequence of instructions for a person to follow based on the visual information from their nearby environment and whether certain objects are dangerous to interact with even if they have not been observed before, such as in strategic military contexts with homes in unmapped territory. 

\section{Learning Mobile Manipulation: A Crucial Step in the Future of Robotics}
Solving robotic mobile manipulation is a crucial step to building a robust system capable of operating outside of the constraint of simulator and laboratory conditions. Mobile manipulation will not be solved dramatically over night, but instead through a series of incremental contributions in subfields like robotic, scene understanding, and manipulation planning that culminate to a novel solution. Missing from the field today is a system that is explainable and augmentable by current researchers. Creating more rich and well documented software enabling robotic mobile manipulation is critical to advancing the field to the exciting point of having robotic helpers in the home. All future solutions will pull from contributions in robotic navigation, scene understanding, and manipulation planning. Contexts such as in-home elder care, home construction, hospital contaminated waste disposal, military drone target acquisition, and ocean waste cleaning, are some of the world's current high-priority concerns where robotic mobile manipulation can assume a central and unique role. Once researchers can solve mobile manipulation, the field will soon be able to build robotic systems capable of addressing humanities most pressing issues. 



\titleformat{\chapter}[display]
{\normalfont\bfseries\filcenter}{}{0pt}{\large\bfseries\filcenter{#1}}  
\titlespacing*{\chapter}
  {0pt}{0pt}{30pt}

\clearpage

\phantomsection
\addcontentsline{toc}{chapter}{References}  

\begin{singlespace}  
	\setlength\bibitemsep{\baselineskip}  
	\printbibliography[title={References}]
\end{singlespace}


\titleformat{\chapter}[display]
{\normalfont\bfseries\filcenter}{}{0pt}{\large\chaptertitlename\ \large\thechapter : \large\bfseries\filcenter{#1}}  
\titlespacing*{\chapter}
  {0pt}{0pt}{30pt}	
  
\titleformat{\section}{\normalfont\bfseries}{\thesection}{1em}{#1}

\titleformat{\subsection}{\normalfont}{\thesubsection}{0em}{\hspace{1em}#1}

\begin{appendices}

\addtocontents{toc}{\protect\renewcommand{\protect\cftchappresnum}{\appendixname\space}}
\addtocontents{toc}{\protect\renewcommand{\protect\cftchapnumwidth}{6em}}


\chapter{MNIST Digit Completion}
\label{app:mnist_reconstruction}

\section{Introduction}
\begin{figure}[ht!]
    \centering
    \includegraphics[width=\textwidth] {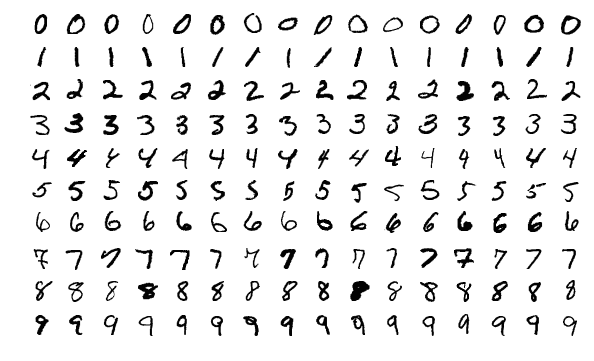}

    \caption{A collection of handwritten digits present in the MNIST dataset. Each of these are 28x28 images with values between 0 and 255. The provide a straightforward method for conducting vision-based research. }
    \label{fig:mnist_digits}
\end{figure}

The MNIST dataset is a collection of handwritten digits that is used to train image processing systems~\cite{lecun2010mnist}. The dataset contains 60000 training images and 10000 testing images. It is a combination of the NIST training dataset and the NIST testing dataset. In the original paper describing the MNIST dataset, the authors validate the datasets effectiveness using a support-vector machine to predict the label of each digit and get an error rate of 0.8\%. Modern approaches for digit classification use neural networks such as fully connected, recurrent, and convolutional neural networks. The simplicity of the MNIST dataset allows researchers to test image reconstruction methods to validate their approach, instead of being restricted to digit classification. Examples of these digits are shown in \autoref{fig:mnist_digits}. 

Through MNIST digits, researchers can easily iterate on model designs and test different architectures quickly. Modeling lots of reconstructions of two-dimensional digits allows a researcher to validate that their methodology can solve an issue via a network architecture. Because the reconstruction of MNIST digits is so simple, it often can be memorized by the intermediate layers of the neural network. While strategies such as regularization can help to address this, it is beyond the scope of these experiments. To reduce the likelihood the network memorized the input, a smaller embedding layer was used in the middle of the network. Shrinking Shrinking the embedding layer forces the network to have a smaller workspace and therefore generalize better assuming this workspace is sufficiently large enough to perform digit reconstruction. The usefulness of this exercise comes from being able to quickly test and validate multiple network architectures. A researcher can easily iterate and manipulate MNIST data to demonstrate the usefulness of different experiments without worrying about acquiring large datasets. 

\section{Methodology}

\subsection{Autoencoder Reconstruction}
\begin{figure*}[ht!]
    \centering {
        \includegraphics[width=\textwidth]{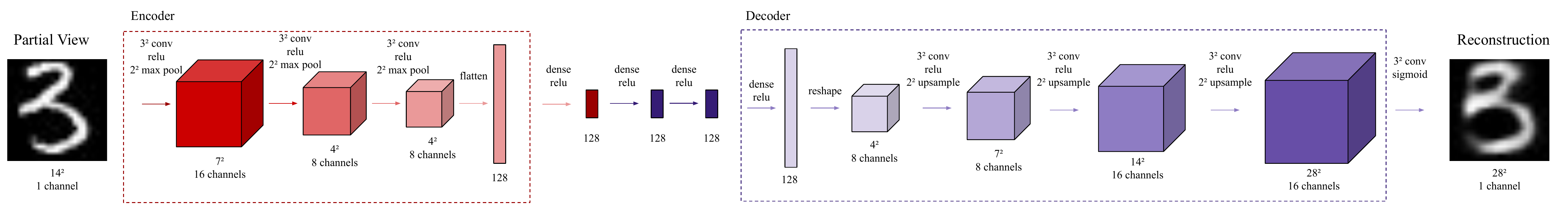}
    }
    \caption{The autoencoder architecture for predicting a ground truth MNIST digit. The network is complex enough to learn to recreate the digit but also simple enough to lay the foundation for further testing. }
    \label{fig:mnist_autoencoder_network}
\end{figure*}

A simple autoencoder architecture can be created by convolving the MNIST digits through a network. As described previously in \autoref{ch:learning_visual_navigation}, an autoencoder is a network that learns to reproduce the input. These networks are useful for reducing the state space of a particular input, or for validating a network architecture. The first step here was to create a network that could recreate MNIST digits accurately. There are many ways of designing an autoencoder, but the most reliable for 2D images is a series of convolutions. Autoencoders are composed of an encoding stage and a decoding stage. The encoder can be represented as a series of convolutions and max pooling. The decoder can be represented as a series of deconvolutions or convolutional transposing layers with upsampling layers in between. All the following models will follow this formula. The architecture initially settled on for this test is shown in \autoref{fig:mnist_autoencoder_network}. 

\begin{figure*}[ht!]
    \centering {
        \includegraphics[width=\textwidth]{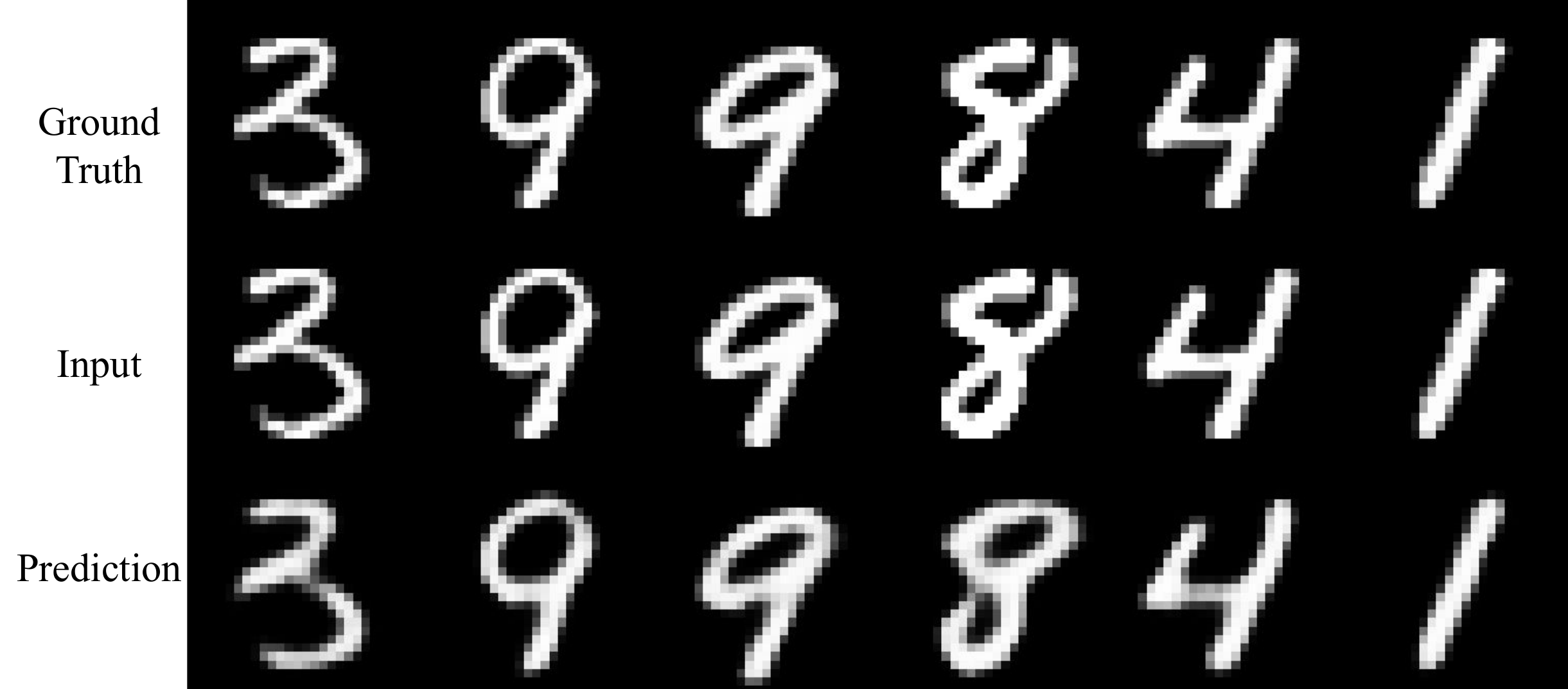}
    }
    \caption{The predictions of the autoencoder network for MNIST can learn a function to reproduce digits reliably. The autoencoder being able to learn digits means that the hypothesis is validated and can work for partial views. }
    \label{fig:mnist_autoencoder_predictions}
\end{figure*}

This network convolves an image down into a dense $128D$ vector. This is a learnable representation, and other values for the dense layer’s width could have been chosen, but for the purposes of exploring the different model architectures $128D$ was enough. Convolutions work well for images because they help to describe geometry well, such as edges and shapes. This will extend to 3D later. A dense layer in the middle is used to make sure that every part of the image can be used at the decoding stage. Convolutions alone do not consider the whole image, but instead use kernels to look at parts of the image. By adding a dense layer, the network learns to reproduce the whole digit using parts. Example reconstructions of the autoencoder are shown in \autoref{fig:mnist_autoencoder_predictions}.

\subsection{Single-View Reconstruction}

\begin{figure*}[ht!]
    \centering {
        \includegraphics[width=0.95\textwidth]{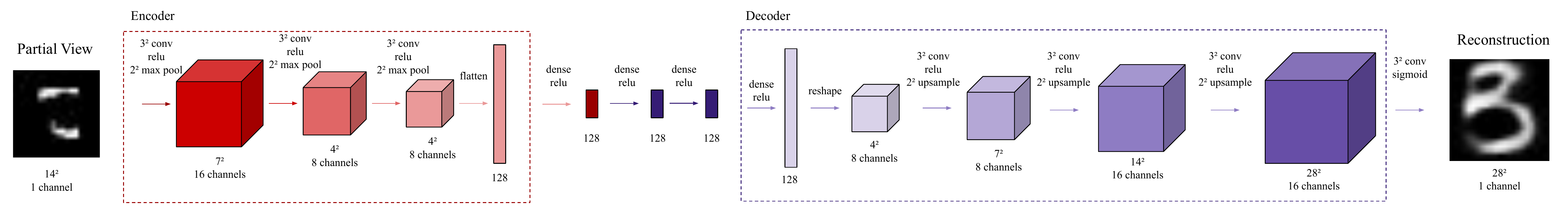}
    }
    \caption{The autoencoder network must be modified slightly to accommodate $14x14$ images. This new network architecture can utilize partial images of MNIST digits for reconstruction. }
    \label{fig:mnist_single_view_architecture}
\end{figure*}

Now that the network can reconstruct the digit using the digit itself, what happens when part of the image is cropped out? The same architecture can be reused in the case of a masked digit, but a small modification is needed when predicting from a $14x14$ partial image. The partial reconstruction from a single image network is shown in \autoref{fig:mnist_single_view_architecture}. The main difference is the first convolutional layer is a smaller size to accommodate the difference in data size. 

\begin{figure}[ht!]
    \centering
    
    \begin{subfigure}{.5\textwidth}
      \centering
      \includegraphics[width=\linewidth]{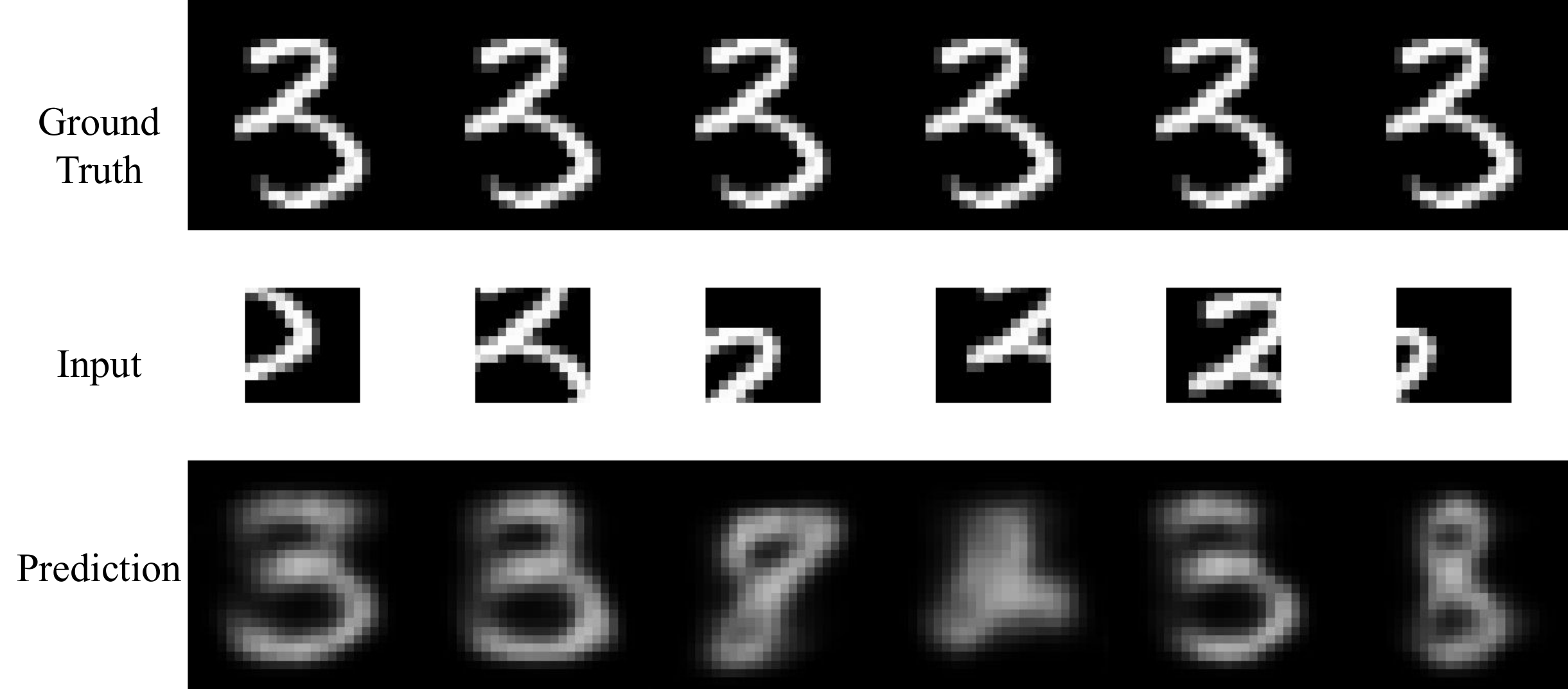}
      \caption{Single-View Partial}
    \end{subfigure}%
    \begin{subfigure}{.5\textwidth}
      \centering
      \includegraphics[width=\linewidth]{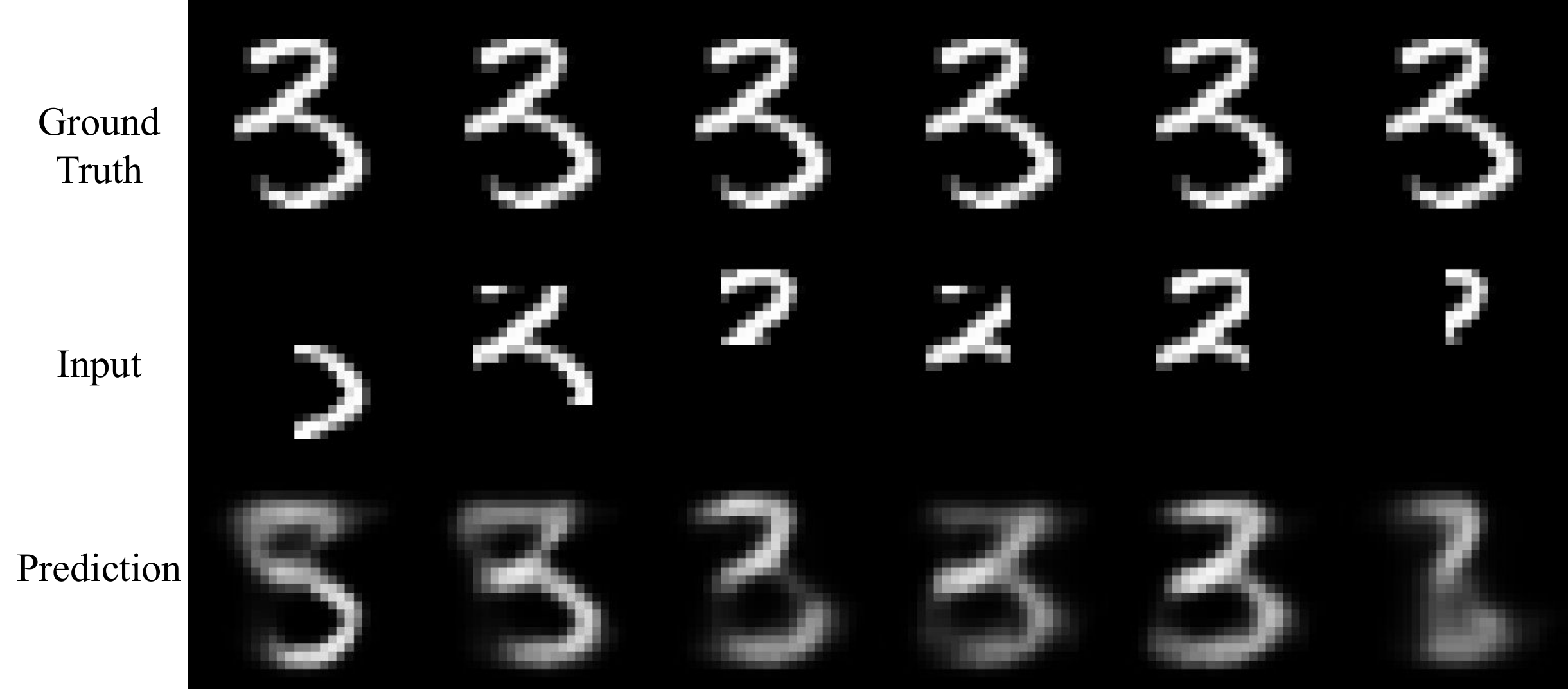}
      \caption{Single-View Masked}
    \end{subfigure}
    
    \caption{(a) reconstruction of digits using $14x14$ partial images and (b) reconstruction of digits using a masked image of size $28x28$. The masked view provides a better reconstruction over partial images due to the context of where the crop took place in the original view. }
    \label{fig:mnist_single_view_reconstructions}
\end{figure}

Each of these models are trained, masked and partial, for 50 epochs viewing each training element in the train dataset, totaling 500000 images. 10 samples per ground truth digit are taken to cover as much of each digit as possible for evaluation. Example reconstructions of single image predictions are shown in \autoref{fig:mnist_single_view_reconstructions}. The single view reconstructions have some false predictions, which helps to show how little information is conveyed in a partial view.

\subsection{Two-View Reconstruction}

\begin{figure*}[ht!]
    \centering {
        \includegraphics[width=\textwidth]{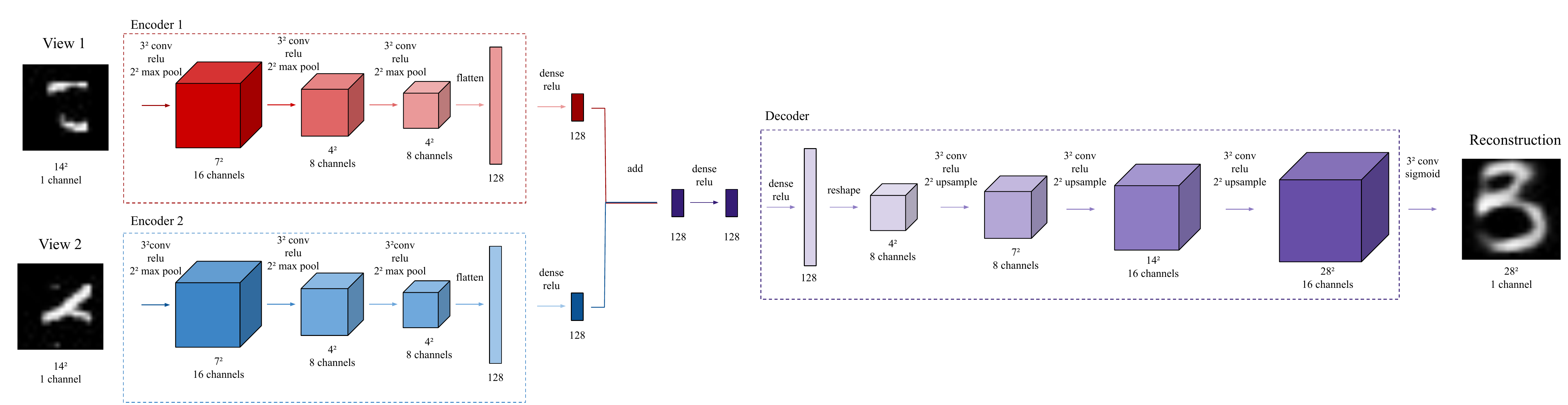}
    }
    \caption{The two-view joined reconstruction network showing the same encoder being used for each input element. Each branch is added together and passed through a dense layer to ensure the entirety of each input image is processed by a dense layer. }
    \label{fig:mnist_two_view_joined_architecture}
\end{figure*}

\begin{figure*}[ht!]
    \centering {
        \includegraphics[width=\textwidth]{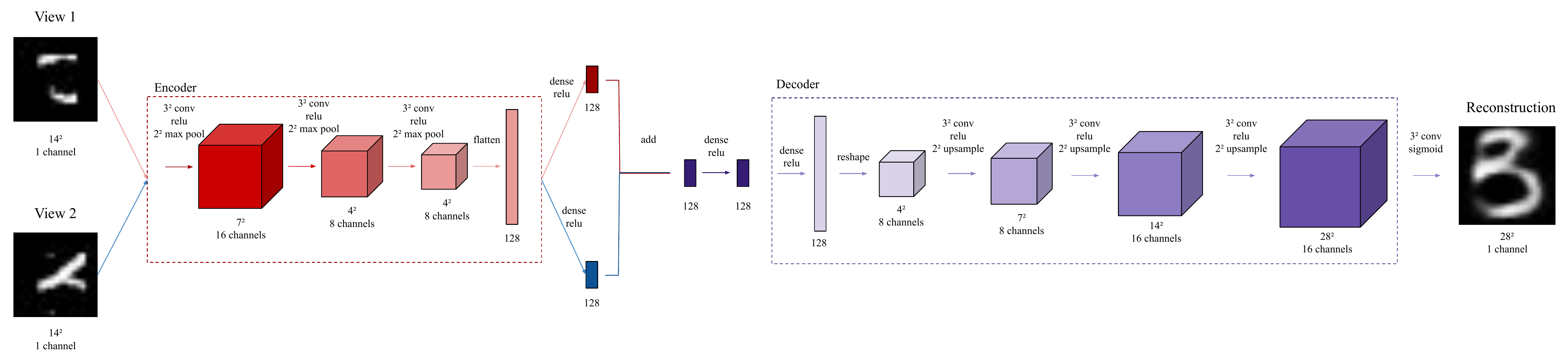}
    }
    \caption{The two-view split reconstruction network showing the two different encoders for each input element. Each branch is added together and passed through a dense layer to ensure the entirety of each input image is processed by a dense layer. }
    \label{fig:mnist_two_view_split_architecture}
\end{figure*}

There are two ways a network could reconstruct the digit using two views: each image could be encoded using the same encoder or each image could be encoded separately. An important consideration is: would it matter the order in which each view was inputted in the network? For a given partial view of the object, at least with how this problem has been defined, the output does not change with the order. This would not be the case in a 3D context as the frame in which an object reconstruction occurs is important. Therefore, using the same encoder for each partial view of the MNIST digit would work as opposed to separate encoders. This approach is validated experimentally as well as logically. The network that uses the same encoder for each input is the \textbf{joined} model and the network that uses two different encoders is the \textbf{split} model. The network architecture for the joined model is shown in \autoref{fig:mnist_two_view_joined_architecture}, and the network architecture for the split model is shown in \autoref{fig:mnist_two_view_split_architecture}. 

\begin{figure}[ht!]
    \centering
    \begin{subfigure}{.5\textwidth}
      \centering
      \includegraphics[width=.95\linewidth]{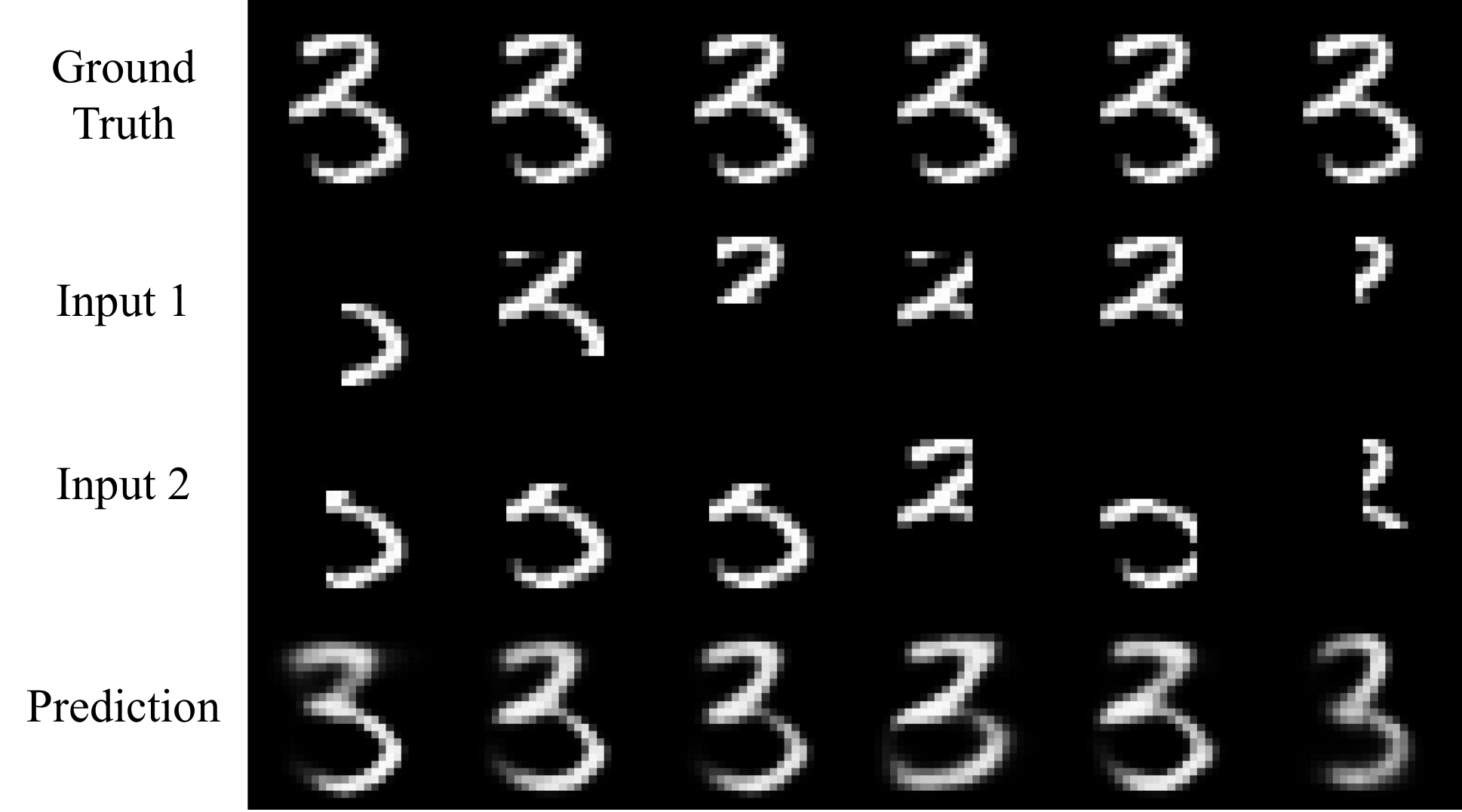}
      \caption{Two-View Masked Joined}
    \end{subfigure}%
    \begin{subfigure}{.5\textwidth}
      \centering
      \includegraphics[width=.95\linewidth]{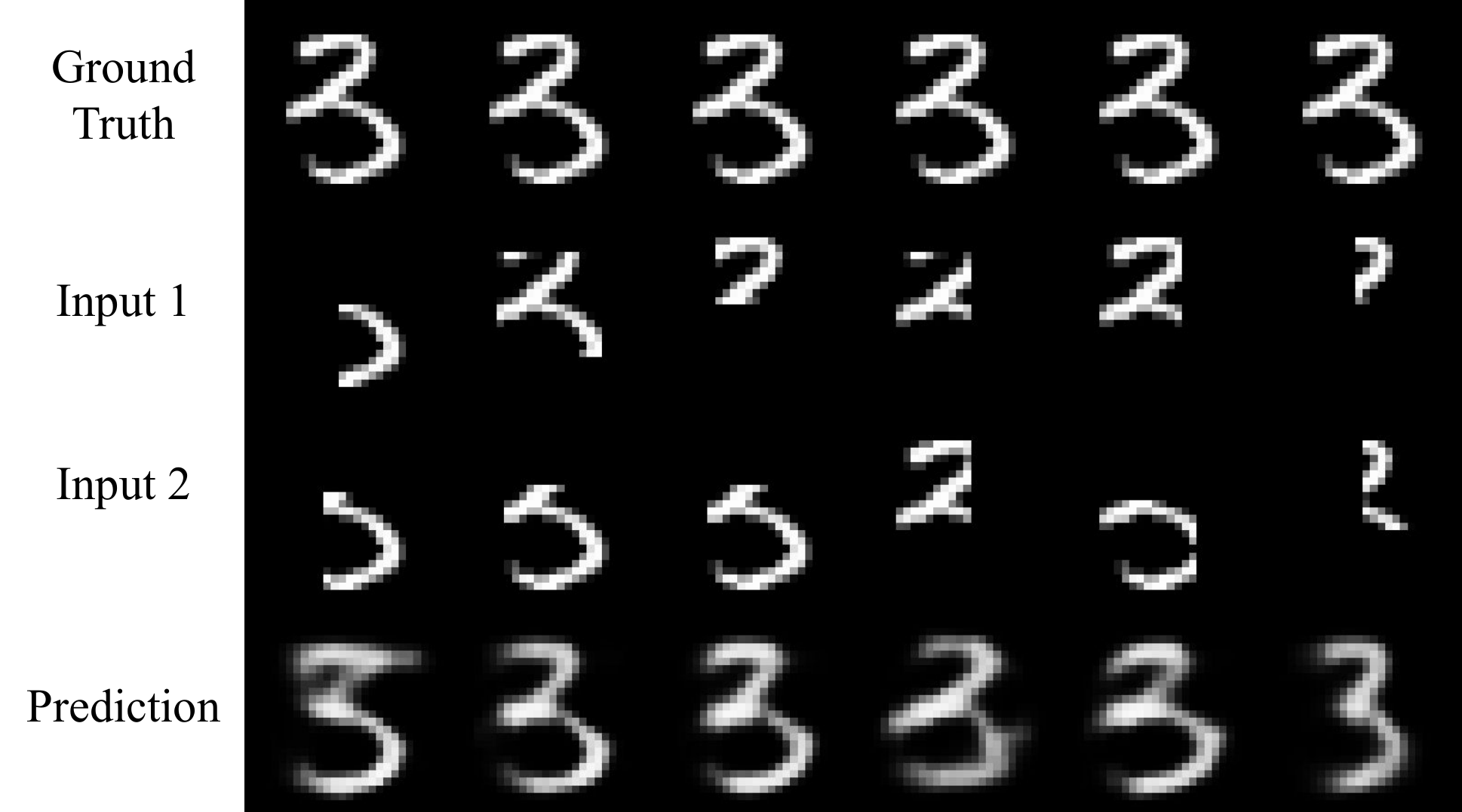}
      \caption{Two-View Masked Split}
    \end{subfigure}
    \bigskip
    
    \begin{subfigure}{.5\textwidth}
      \centering
      \includegraphics[width=.95\linewidth]{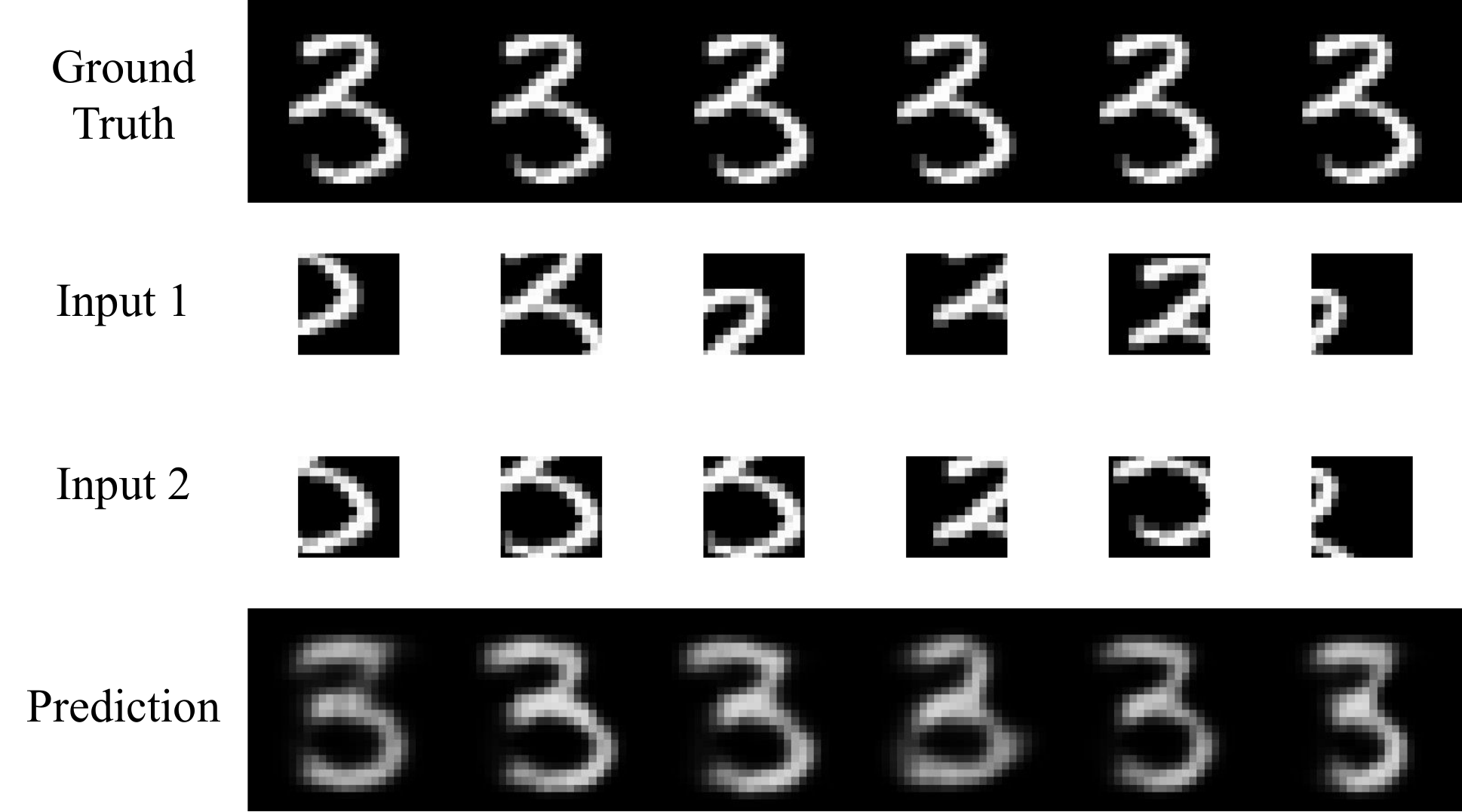}
      \caption{Two-View Partial Joined}
    \end{subfigure}%
    \begin{subfigure}{.5\textwidth}
      \centering
      \includegraphics[width=.95\linewidth]{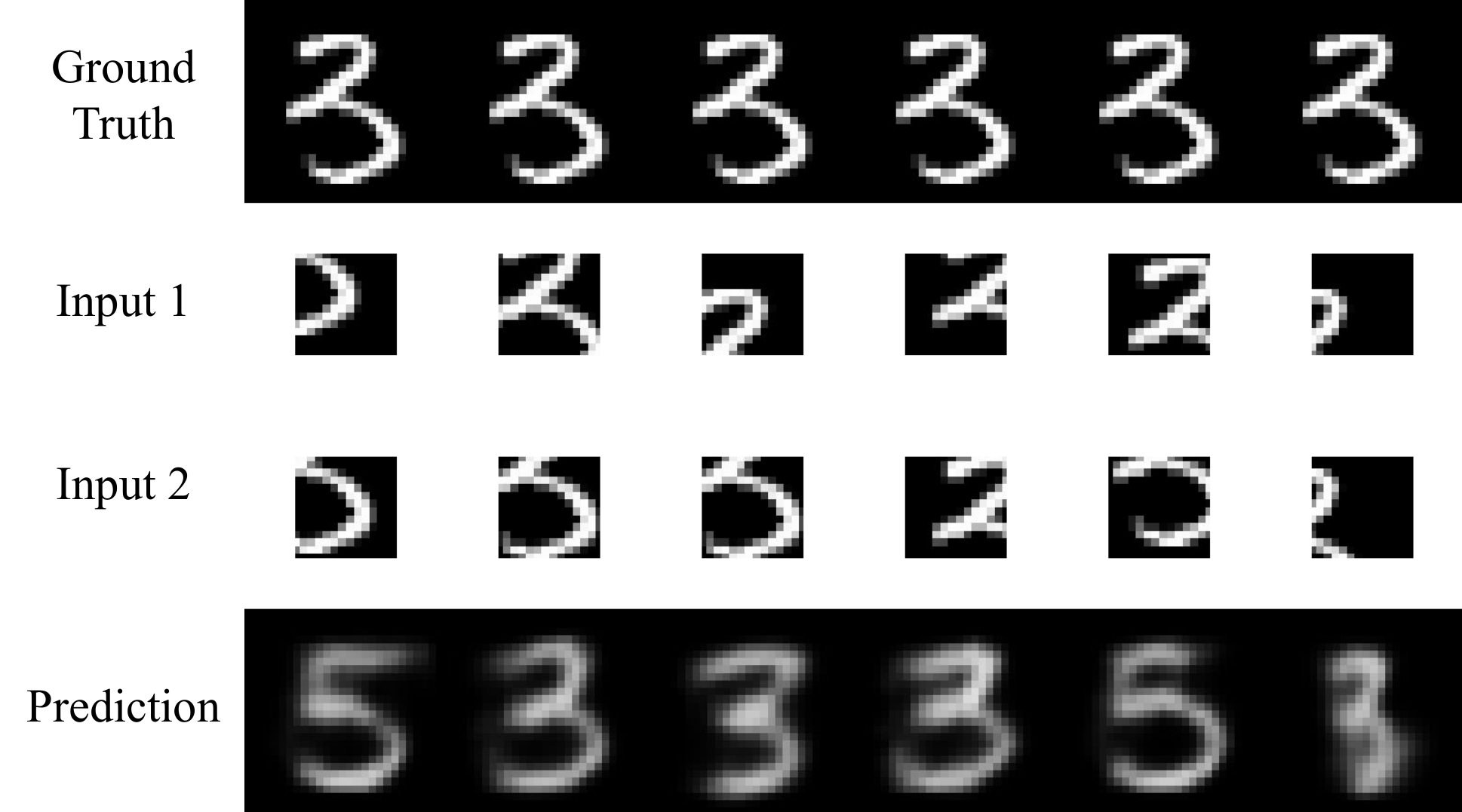}
      \caption{Two-View Partial Joined}
    \end{subfigure}%
    \caption{\textbf{Two-View} reconstruction of digits using partial and masked views, using joined and split architectures. There is trivial difference in the fidelity of the reconstruction of the digits between these models, but they all show marked improvement over the single view reconstruction. }
    \label{fig:mnist_reconstruction_improvement}
\end{figure}

In both the partial and masked cases, two views improved the completion quality. The additional information provided from another view helped inform the reconstruction of digits. Examples of this improvement are shown in \autoref{fig:mnist_reconstruction_improvement}.

\subsection{Multiple-View Reconstruction}

\begin{figure*}[ht!]
    \centering {
        \includegraphics[width=\textwidth]{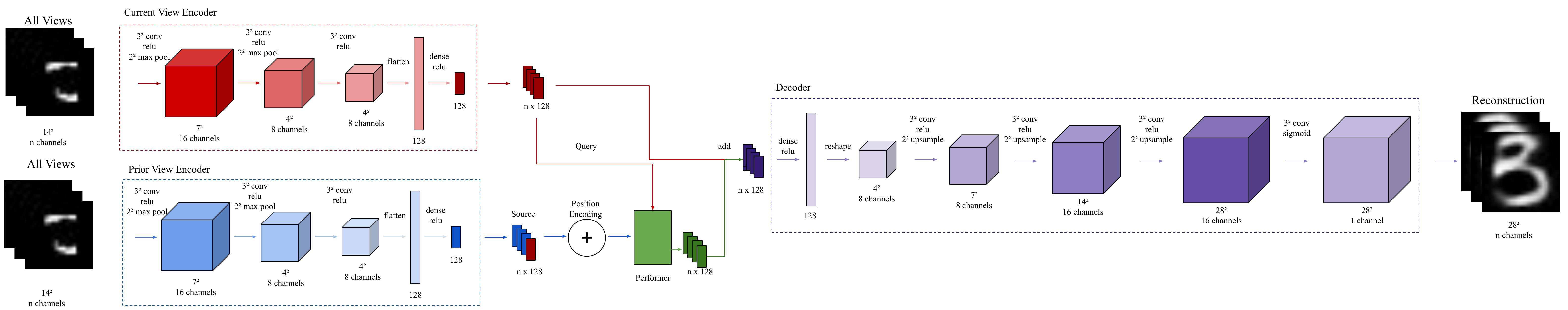}
    }
    \caption{The performer architecture for MNIST showing how to utilize multiple views of a digit to refine the prediction. Each view is attended by the previous views of the digit to refine the overall prediction. }
    \label{fig:mnist_performer_architecture}
\end{figure*}

\begin{figure}[ht!]
    \centering
    
    \begin{subfigure}{\textwidth}
      \centering
      \includegraphics[width=\linewidth]{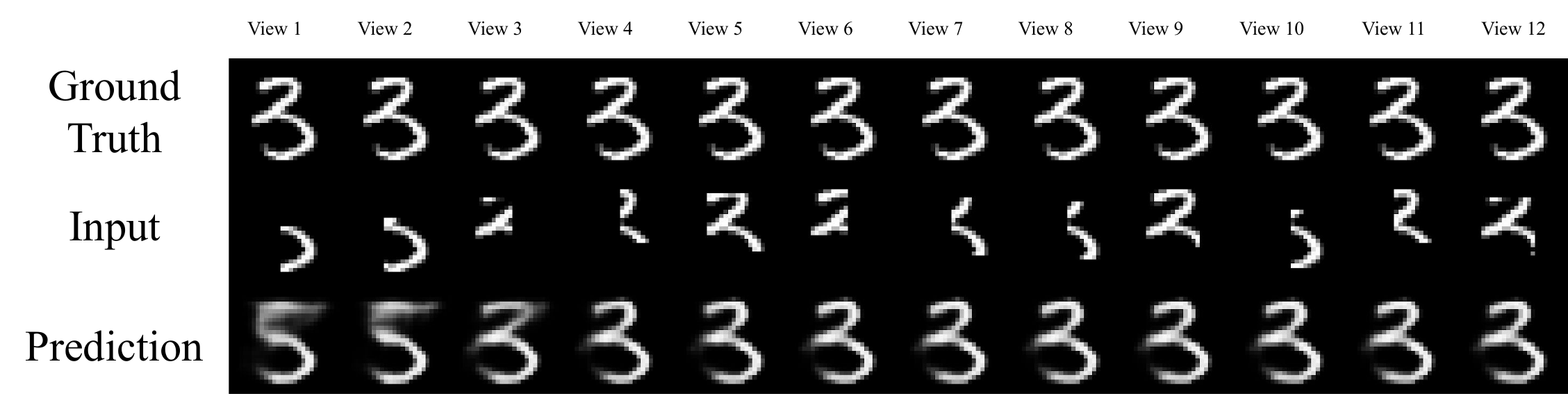}
      \caption{Masked Performer}
    \end{subfigure}
    \bigskip
    
    \begin{subfigure}{\textwidth}
      \centering
      \includegraphics[width=\linewidth]{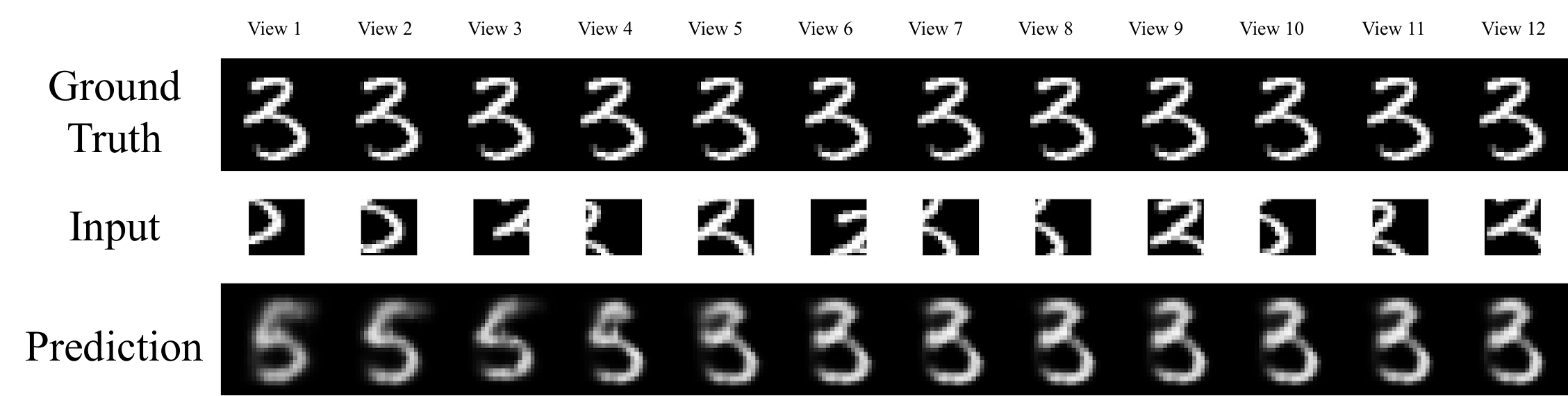}
      \caption{Partial Performer}
    \end{subfigure}%
    
    \caption{\textbf{Performer} reconstruction of digits using partial and masked views. There is is a negligible difference in the fidelity of the reconstruction of the digits between these models, but they all show marked improvement over the single view and two-view reconstruction. The reconstruction improves as the number of views increases. }
    \label{fig:mnist_performer_reconstruction_improvement}
\end{figure}

Previous architectures have shown how to reconstruct an MNIST digit using individual views. This can be extended further by utilizing multiple views of each digit. An attention-based model utilizing a Performer~\cite{performer} layer would achieve this goal by providing a way for the network to leverage time-series data. This network takes two encoders, as described in the two-view model, and creates a set of embeddings. The embeddings are from the current view encoder and previous view encoder. The performer layer, equivalent to an Attention~\cite{vaswani2017attention} layer, takes the previous view tokens as a source and the current view as the query to attend to. The token that comes out of the performer layer is added to the token from the current view encoder to produce a final embedding. This embedding is then passed through a convolutional decoder to produce a final prediction. The architecture is shown in \autoref{fig:mnist_performer_architecture}. This design is expanded upon in \autoref{ch:performers}. Example reconstructions are shown in \autoref{fig:mnist_performer_reconstruction_improvement}.

\section{Experiments}

\subsection{Dataset Modification}

\begin{figure}[ht!]
\centering
\begin{subfigure}{.5\textwidth}
  \centering
  \includegraphics[width=.95\linewidth]{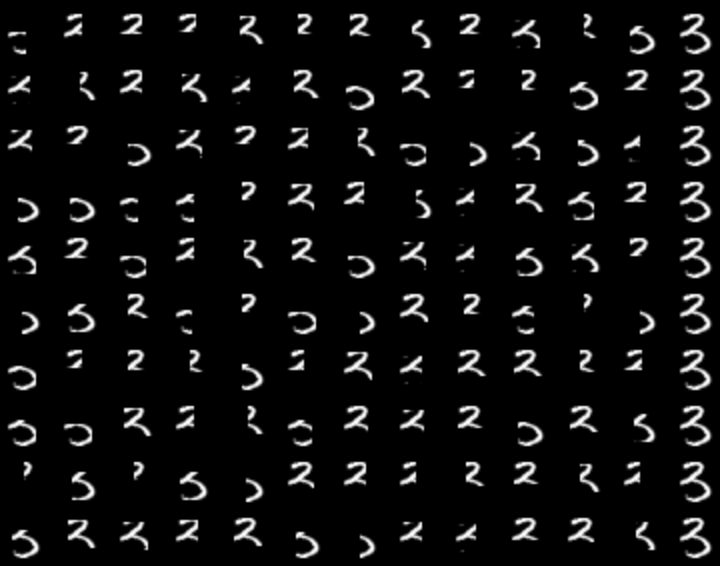}
\end{subfigure}%
\begin{subfigure}{.5\textwidth}
  \centering
  \includegraphics[width=.95\linewidth]{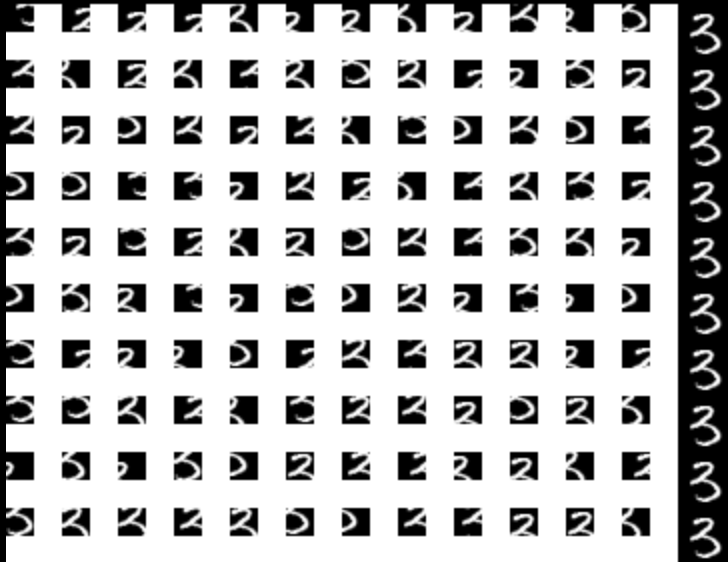}
\end{subfigure}
\caption{MNIST digits cropped either by extracted a $14\times14$ mask or by setting all other values outside the crop to $0$. Each of these datasets are used to train MNIST reconstruction networks. The ground truth for each crop is shown on the rightmost column. }
\label{fig:mnist_partial_masked_examples}
\end{figure}

To train each of the masked and partial MNIST reconstruction networks, each of the 70,000 images are taken from the dataset and a crop of $14\times14$ is computed and paired with the ground truth version of the image. For each image, $12$ crops are generated $10$ times. The total size of the dataset is 600,000 training pairs and 100,000 testing pairs. For partial MNIST reconstruction, a $14\times14$ crop is extracted from the ground truth image and paired with its ground truth image. For masked MNIST reconstruction, a $28\times28$ image is constructed with all zeros and then $14\times14$ crop is placed inside of that empty view. These two different dataset constructions help illustrate how context impacts the performance of digit reconstruction, where the masked view has information about where in the original image that crop was taken. Examples of each dataset are shown in \autoref{fig:mnist_partial_masked_examples}. 

\subsection{Metrics}

To evaluate the performance of each model, an accuracy metric and a loss metric are used. For accuracy, the difference of each pixel prediction from the expected output is computed and then averaged to generate a score. This is implemented by using the TensorFlow accuracy metric. For loss, the binary cross-entropy loss function is used. This loss function helps to ensure the network converges on either $0$ or $1$ during training. For accuracy, a higher value is better. For binary cross-entropy loss a lower value is better.

\section{Results}

To evaluate the relative performance gains that two views have over a single view quantitatively, cosine similarity is used to determine how similar the output image was to the ground truth MNIST digit. \autoref{tab:mnist_reconstruction_results} shows the empirical findings of digit reconstruction accuracy. The best performing model was the Autoencoder, as anticipated, as it had full information of the digit. This reconstruction accuracy sets the baseline for how accurate a reconstruction could be under ideal conditions. The masked networks had a much better reconstruction result as they were given the contextually valuable information of where the view fits into the ground truth. The highest accuracy reconstruction model was the Performer (masked) with an accuracy of 0.8194 versus 0.8107 for the Single-View (masked). This can be explained as the encoder seeing more training data to create a better state space in the embedding resulted in a higher accuracy completion as well as access to more views. The joined method works well when the output data is regularized to a specific region and the input data overlaps well with that data. The partial results echoed the hierarchy found in the masked condition, where the joined model performed with the highest accuracy of 0.8061 versus an accuracy of 0.8010 for the single-view case. 

\begin{table*}[ht!]
    \centering
    \begin{tabular}{|c|c|c|}
    \hline
    \multicolumn{1}{|c|}{\begin{tabular}[c]{@{}c@{}}\textbf{MNIST Reconstruction} \\  \textbf{Method}\end{tabular}} 
    & \multicolumn{1}{c|}{\begin{tabular}[c]{@{}c@{}}\textbf{Accuracy} \\  \textbf{}\end{tabular}} 
    & \multicolumn{1}{c|}{\begin{tabular}[c]{@{}c@{}}\textbf{Loss} \\  \textbf{}\end{tabular}} \\
    \hline
        Autoencoder               & 0.8219        & 0.0997      \\ \hline
        Single-View (masked)      & 0.8107        & 0.1509      \\ \hline
        Two-View Joined (masked)  & 0.8174        & 0.1266      \\ \hline
        Two-View Split (masked)   & 0.8167        & 0.1294      \\ \hline
        Performer (masked)        & 0.8194        & 0.1157      \\ \hline
        Single-View (partial)     & 0.8010        & 0.2029      \\ \hline
        Two-View Joined (partial) & 0.8061        & 0.1770      \\ \hline
        Two-View Split (partial)  & 0.8054        & 0.1803      \\ \hline
        Performer (masked)        & 0.8089        & 0.1652      \\ \hline
    \end{tabular}
    \caption{\textbf{Reconstruction results of MNIST Digits}, measuring the performance of reconstruction of MNIST digits. The performance is evaluated on validation data not seen during training. The autoencoder performs the best as a baseline. The Two-View Joined Masked method had the highest accuracy, followed closely by the Two-View Split Masked. The best performing test model is the performer model. Higher accuracy is better, lower loss is better. }
    \label{tab:mnist_reconstruction_results}
\end{table*}

\section{Conclusion}
MNIST has provided a good basis for a series of network architectures can be tested in 3D for object reconstruction. While the performance of something like manipulation cannot be evaluated on digits, the digits can be qualitatively evaluated to be legible and evaluate their accuracy as images. This chapter showed that image reconstruction provides a good testbed for different reconstruction strategies. It can be further extended to 3D reconstruction by modifying the types of architectures used, but the principles remain similar.

\chapter{Combining Learning from Human Feedback and Knowledge Engineering to Solve Hierarchical Tasks in Minecraft}
\label{app:minerl_basalt}

\section{Introduction}
The solution that won first place and was awarded the most human-like agent in the 2021 Neural Information Processing Systems (NeurIPS) MineRL Benchmark for Agents that Solve Almost-Lifelike Tasks (BASALT) competition\footnote{Official competition webpage: \url{https://www.aicrowd.com/challenges/neurips-2021-minerl-basalt-competition}.}. Most artificial intelligence (AI) and reinforcement learning (RL) challenges involve solving tasks that have a reward function to optimize over. Real-world tasks, however, do not automatically come with a reward function, and defining one from scratch can be quite challenging. Therefore, teaching AI agents to solve complex tasks and learn difficult behaviors without any reward function remains a major challenge for modern AI research. The MineRL BASALT competition is aimed to address this challenge by developing AI agents that can solve complex, almost-lifelike, tasks in the challenging Minecraft environment~\cite{johnson2016malmo} using only human feedback data and no reward function. 

Minecraft is a video game about cultivating resources and building structures using 3D voxel blocks. First created in 2009, Swedish video game developer Mojang studios launch a public beta to mass appeal. As of this writing, the game is the best-selling video game of all time, with over 238 million copies sold and 140 million active users as of 2021. The game is useful for research as it functions as a sandbox and a simulator for many complex tasks that are both quantifiable and can be evaluated qualitatively by human operators. The release of Malmo~\cite{johnson2016malmo}, a platform that enabled AI experimentation in the game of Minecraft, gave researchers the capability to develop learning agents to solve tasks similar or analogous to the ones seen in the real world. The Minecraft environment also served as a platform to collect large human demonstration datasets such as the \textit{MineRL-v0} dataset~\cite{guss2019minerldata} and experiment with large scale imitation learning algorithms~\cite{amiranashvili2020scaling} as a world generator for realistic terrain rendering~\cite{hao2021gancraft}, a sample-efficient reinforcement learning competition environment using human priors (MineRL DIAMOND challenge)~\cite{guss2019minerlcomp}; and now as a platform for a competition on solving human-judged tasks defined by a human-readable description and no pre-defined reward function, the MineRL BASALT competition~\cite{shah2021basalt}.

The MineRL BASALT competition tasks do not contain any reward functions for the four tasks. A human-centered machine learning approach is proposed instead of using traditional RL algorithms~\cite{guss2019minerlcomp}. However, learning complex tasks with high-dimensional state-spaces (\emph{i.e.}  from images) using only end-to-end machine learning algorithms requires copious amounts of high-quality data~\cite{Bojarski2016,Nair2017}. When learning from human feedback, this translates to numerous of either human-collected or human-labeled data. To circumvent this data requirement, combining machine learning with knowledge engineering is used, also known as hybrid intelligence or informed AI~\cite{kamar2016directions,dellermann2019hybrid}. The approach uses human knowledge of the task to break it down into a natural hierarchy of subtasks. Subtask selection is controlled by an engineered state-machine, which relies on estimated agent odometry and the outputs of a learned state classifier. Also used is the competition-provided human demonstration dataset to train a navigation policy subtask via imitation learning to replicate how humans traverse the environment.

This chapter gives a detailed overview of the approach and will show the results from an ablation study to investigate how well the hybrid intelligence approach works compared to using either learning from human demonstrations or engineered solutions alone. 
The two main contributions are:
\begin{itemize}
    \item An architecture that combines knowledge engineering modules with machine learning modules to solve complex hierarchical tasks in Minecraft.
    \item Empirical results on how hybrid intelligence compares to both end-to-end machine learning and pure engineered approaches when solving complex, real-world-like tasks, as judged by human evaluators.
\end{itemize}

\section{Problem Setup}

The 2021 NeurIPS MineRL BASALT competition, ``Learning from Human Feedback in Minecraft'', challenged participants to produce creative solutions to solve four different tasks in Minecraft~\cite{shah2021basalt} using the ``MineRL: Towards AI in Minecraft''\footnote{MineRL webpage: \url{https://minerl.io/}.} simulator~\cite{guss2019minerldata}.
These tasks aimed to mimic real-world tasks, being defined only by a human-readable description and no reward signal returned by the environment.
The official task descriptions for the MineRL BASALT competition\footnote{MineRL BASALT documentation: \url{https://minerl.io/basalt/}.} were the following:
\begin{itemize}
    \item \textit{FindCave}: The agent should search for a cave and terminate the episode when it is inside one.
    \item \textit{MakeWaterfall}: After spawning in a mountainous area, the agent should build a beautiful waterfall and then reposition itself to take a scenic picture of the same waterfall.
    \item \textit{CreateVillageAnimalPen}: After spawning in a village, the agent should build an animal pen containing two of the same kind of animal next to one of the houses in a village.
    \item \textit{BuildVillageHouse}: Using items in its starting inventory, the agent should build a new house in the style of the village, in an appropriate location (e.g\., next to the path through the village) without harming the village in the process.
\end{itemize}

The competition organizers also provided each participant team with a dataset of 40 to 80 human demonstrations for each task, not all completing the task, and the starter codebase to train a behavior cloning baseline. Additionally, the training time for all four tasks together was limited to four days and participants were allowed to collect up to 10 hours of additional human-in-the-loop feedback.

\section{Methods}

\begin{figure}[!ht]
  \centering
  \includegraphics[width=0.9\linewidth]{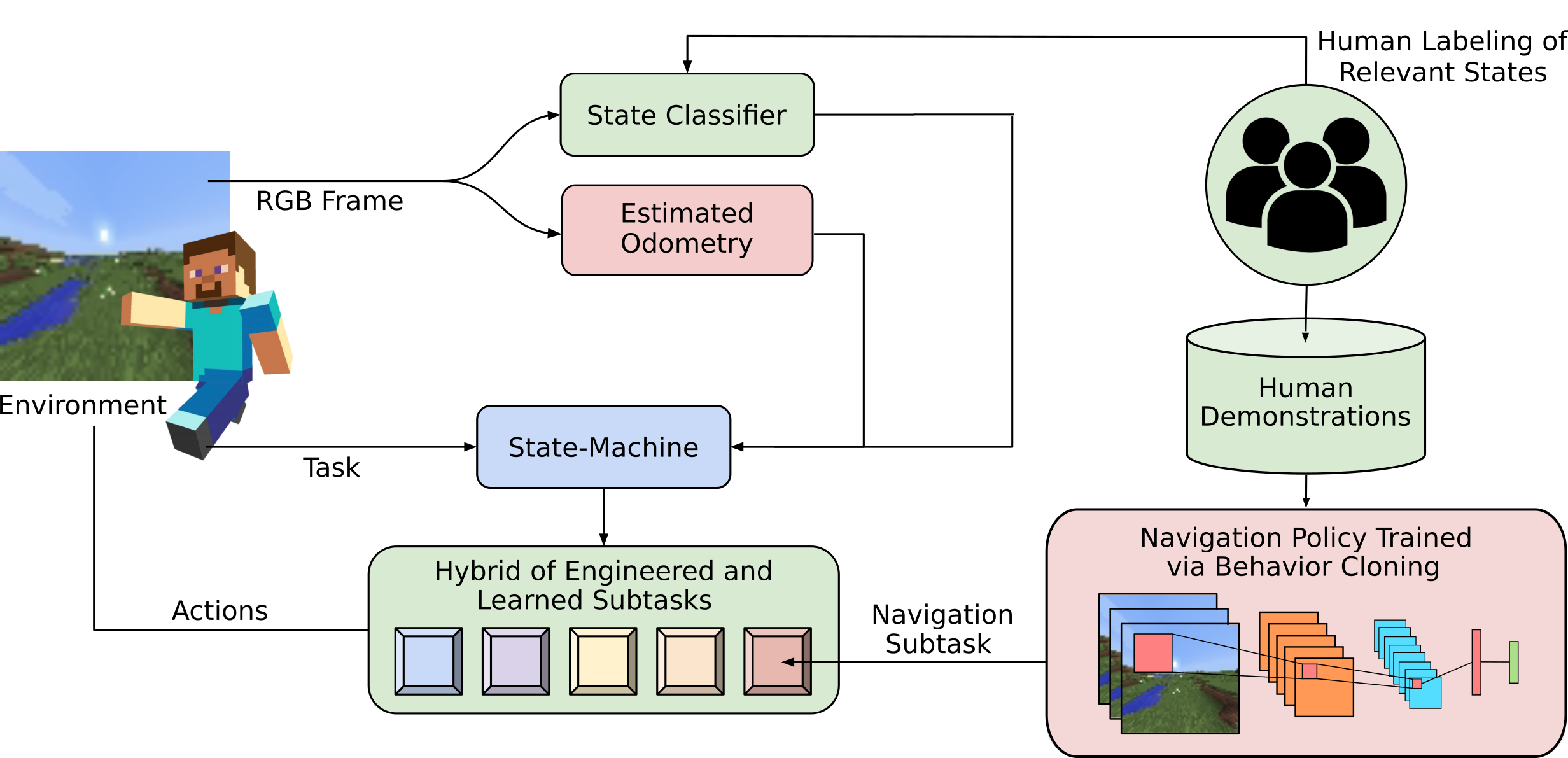}
  \caption{Diagram illustrating the approach. Using data from the available human demonstration dataset, humans provide additional binary labels to image frames to be used to train a state classifier that can detect relevant features in the environment such as caves and mountains. The available human demonstration dataset is also used to train a navigation policy via imitation learning to replicate how humans traverse the environment. A separate odometry module estimates the current agent's position and heading solely based on the action taken by the end. During test time, the agent uses the learned state classifier to provide useful information to an engineered state-machine that controls which subtask the agent should execute at every time-step.}
  \label{fig:diagram}
\end{figure}

Since no reward signal was given by the competition organizers and compute time was limited, direct deep reinforcement learning approaches were not feasible~\cite{Mnih2013, lillicrap2015continuous,Mnih2016}. With the limited human demonstration dataset, end-to-end behavior cloning also did not result in high-performing policies, because imitation learning requires enormous amounts of high-quality data~\cite{Bojarski2016,Nair2017}. Also attempted was to solve the tasks using adversarial imitation learning approaches such as Generative Adversarial Imitation Learning (GAIL)~\cite{ho2016generative}, however, the large-observation space and limited compute time also made this approach infeasible.

Hence, to solve the four tasks of the MineRL BASALT competition, combining machine learning with knowledge engineering is used, also known as \emph{hybrid intelligence}~\cite{kamar2016directions,dellermann2019hybrid}. As seen in the main diagram of the approach shown in Figure~\ref{fig:diagram}, the machine learning part of the method is seen in two different modules: first, a state classifier is trained using additional human feedback to identify relevant states in the environment; second, a navigation subtask is trained separately for each task via imitation learning using the human demonstration dataset provided by the competition.
The knowledge engineering part is seen in three different modules: first, given the relevant states classified by the machine learning model and knowledge of the tasks, a state-machine is designed that defines a hierarchy of subtasks and controls which one should be executed at every time-step; second, solutions are engineered for the more challenging subtasks that were not able to be learned directly from data; and third, an estimated odometry module is engineered that provides additional information to the state-machine and enables the execution of the more complex engineered subtasks.

\subsection{State Classification}

\begin{figure}[!ht]
    \centering
    \begin{tabular}{cccc}
        \subfloat[\textit{has\_cave}]{\includegraphics[width=.18\linewidth]{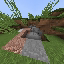}} &
        \subfloat[\textit{inside\_cave}]{\includegraphics[width=.18\linewidth]{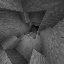}} &
        \subfloat[\textit{danger\_ahead}]{\includegraphics[width=.18\linewidth]{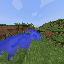}} &
        \subfloat[\textit{has\_mountain}]{\includegraphics[width=.18\linewidth]{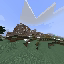}}\\
        \subfloat[\textit{facing\_wall}]{\includegraphics[width=.18\linewidth]{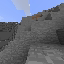}} &
        \subfloat[\textit{at\_the\_top}]{\includegraphics[width=.18\linewidth]{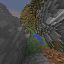}} &
        \subfloat[\textit{good\_waterfall\_view}]{\includegraphics[width=.18\linewidth]{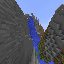}} &
        \subfloat[\textit{good\_pen\_view}]{\includegraphics[width=.18\linewidth]{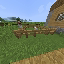}}\\
        \subfloat[\textit{good\_house\_view}]{\includegraphics[width=.18\linewidth]{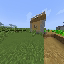}} &
        \subfloat[\textit{has\_animals}]{\includegraphics[width=.18\linewidth]{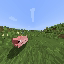}} &
        \subfloat[\textit{has\_open\_space}]{\includegraphics[width=.18\linewidth]{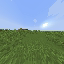}} &
        \subfloat[\textit{animals\_inside\_pen}]{\includegraphics[width=.18\linewidth]{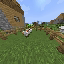}}
    \end{tabular}
    \caption{Illustration of positive classes of states classified using additional human feedback. Humans were given image frames from previously collected human demonstration data and were assigned to give binary labels for each of the illustrated 12 states, plus a null case when no relevant states were identified.}
    \label{fig:state_classifier}
\end{figure}

The approach relies on a state machine that changes the high-level goals depending on the task to be solved. Without having information about the environment's voxel data, the state classifier uses the visual RGB information from the simulator to determine the agent's current state. Due to the low resolution of the simulator of $64\times64\times3$, a classifier is used that labels the whole image rather than parts of the image, such as \emph{You Only Look Once} (YOLO)~\cite{yolo}. Multiple labels can be present on the same image as there were cases with multiple objects or scenes of interest at the same time in the field of view of the agent. These labels are used by the state-machine to decide which subtask should be followed at any time-step.

There are 13 labels for an RGB frame, as illustrated in Figure~\ref{fig:state_classifier} and described below: 
\begin{itemize}
    \item \textit{none}: frame contains no relevant states (54.47 \% of the labels).
    \item \textit{has\_cave}: agent is looking at a cave (1.39 \% of the labels).
    \item \textit{inside\_cave}: agent is inside a cave (1.29 \% of the labels).
    \item \textit{danger\_ahead}: agent is looking at a large body of water (3.83 \% of the labels).
    \item \textit{has\_mountain}: agent has a complete view of a mountain (usually, from far away) (4.38 \% of the labels).
    \item \textit{facing\_wall}: agent is facing a wall that cannot be traversed by jumping only (4.55 \% of the labels).
    \item \textit{at\_the\_top}: agent is at the top of a mountain and looking at a cliff (3.97 \% of the labels).
    \item \textit{good\_waterfall\_view}: agent see water in view (3.16 \% of the labels).
    \item \textit{good\_pen\_view}: agent has framed a pen with animals in view (4.12 \% of the labels).
    \item \textit{good\_house\_view}: agent has framed a house in view (2.58 \% of the labels).
    \item \textit{has\_animals}: frame contains animals (pig, horse, cow, sheep, or chicken) (9.38 \% of the labels).
    \item \textit{has\_open\_space}: agent is looking at an open-space of about 6x6 blocks with no small cliffs or obstacles (flat area to build a small house or pen) (7.33 \% of the labels).
    \item \textit{animals\_inside\_pen}: agent is inside the pen after luring all animals and has them in view (0.81 \% of the labels).
\end{itemize}


The possible labels were defined by a human designer with knowledge of the relevant states to be identified and given to the state-machine to solve all tasks. These labels were also designed to be relevant to all tasks to ease data collection and labelling efforts. For example, the ``has\_open\_space'' label identifies flat areas that are ideal to build pens or houses for both \emph{CreateVillageAnimalPen} and \emph{BuildVillageHouse} tasks. Unknown and other non-relevant states were attached the label ``\emph{none}'' to indicate that no important states were in view.

To train this system, $81,888$ are labeled images using a custom graphical user interface (GUI), as showed in Section~\ref{appendix:label_state_gui}. Once the data was labeled, $80\%$ of images were used for training, $10\%$ were used for validation, and $10\%$ for testing. The model is a convolutional neural network (CNN) classifier with a $64\times64\times3$ input and $13\times1$ output. The architecture of the CNN is modeled after the Deep TAMER (Training Agents Manually via Evaluative Reinforcement)~\cite{Warnell2018} model. The problem of training with an uneven number of labels for each class was mitigated by implementing a weighted sampling scheme that sampled more often classes with lower representation with probability:
\begin{equation}
    P(x_i) = 1 - \frac{N_i}{M},
\end{equation}
where $P(x_i)$ is the probability of sampling class $i$ that contains $N_i$ number of labels out of the total $M$ labels for all classes.

\subsection{Estimated Odometry}

\begin{figure}[!ht]
  \centering
  \includegraphics[width=0.65\linewidth]{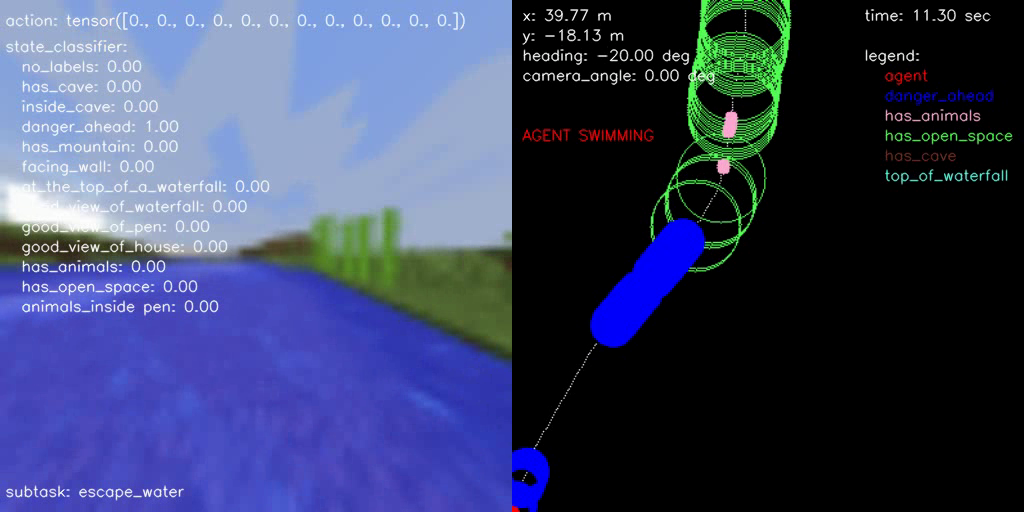}
  \caption{Example of odometry map (right frame) generated in real time from the agent's actions as it traverses the environment. Besides the agent's pose, the classified states from image data (left frame) also have their locations tagged in the map to be used for specific subtasks. For example, when the agent finishes building the pen, it uses the location of previously seen animals to attempt to navigate and lure them to the pen.}
  \label{fig:odometry}
\end{figure}

Some of the engineered subtasks required basic localization of the agent and relevant states of the environment. For example, the agent needs to know the location of previously seen animals to guide them to a pen in the \textit{CreateVillageAnimalPen} task. However, under the rules of the competition, the agent is not allowed to use any additional information from the simulator besides the current view of the agent and the player's inventory. Which means there is no information about the ground truth location of the agent, camera pose, or any explicit terrain information available.

Given these constraints, the agent uses a custom odometry method that took into consideration only the actions from the agent and basic characteristics of the Minecraft simulator. It is known that the simulator runs at $20$ frames per second, which means there is a $0.05$ second interval between each frame. According to the Minecraft Wiki\footnote{Minecraft Wiki - Walking: \url{https://minecraft.fandom.com/wiki/Walking}.}, walking speed is approximately $4.317$ m/s, $5.612$ m/s while sprinting, or $7.127$ m/s when sprinting and jumping at the same time, which translates to approximately $0.216$, $0.281$, or $0.356$ meters per frame when walking, sprinting, or sprinting and jumping, respectively. Assuming the agent is operating in a flat world, starting at position $(0, 0)$ in map coordinates facing north, when the agent executes a move forward action its position is moved $0.216$ meters north to position $(0, 0.216)$. The agent does not have acceleration in MineRL, and their velocity is immediately updated upon keypress. Another limitation is that the agent is not able to reliably detect when it is stuck behind an obstacle, which causes the estimated location to drift even though the agent is not moving in the simulator.

Since the agent already commands camera angles in degrees, the heading angle $\theta$ is simply updated by accumulating the horizontal camera angles commanded by the agent.
More generally, this odometry estimation assumes the agent follows point-mass kinematics:
\begin{align*}
    \dot{x} &= V cos(\theta) \\
    \dot{y} &= V sin(\theta),
\end{align*}
where $V$ is the velocity of the agent, which takes into consideration if the agent is walking, sprinting, or sprinting and jumping.

Using this estimated odometry and the learned state classifier, it is possible to attach a coordinate to each classified state and map key features of the environment so that the agent has access to them for different subtasks and the state-machine. For example, it is possible to keep track of where the agent found water, caves, animals, and areas of open space that can be used to build a pen or a house. Figure~\ref{fig:odometry} shows a sample of the resulting map overlaid with the classified states' location and current odometry readings.

\subsection{Learning and Engineering Subtasks and the State-Machine}

One of the main complexities in solving the proposed four tasks is that most required the agent to have certain levels of perception capabilities, memory, and reasoning over long-term dependencies in a hierarchical manner. For example, the \textit{CreateVillageAnimalPen} task required the agent to first build a pen nearby an existing village, which requires identifying what a village is, then indicating an effective location to build a pen such as a flat terrain. Once the pen was built, the agent had to search for at least two of the same animal type in the nearby vicinity using $64 \times 64$ resolution images as input. Return animals to the pen required coordination to combine different blocks and place them adjacently to each other in a closed shape. After the animals were found, the agent had to lure them with the specific food type they eat, walk them back to the pen the agent initially built, leave the pen, lock the animals inside, then take a picture of the pen with the animals inside.

Reasoning over these long-term dependencies in hierarchical tasks is one of the main challenges of end-to-end learning-based approaches~\cite{vezhnevets2017feudal}. Conversely, reactive policies such as the one required to navigate with certain boundaries and avoid obstacles have been learned directly from demonstration data or agent-generated trajectories~\cite{giusti2015machine,Bojarski2016}. Human knowledge of the tasks is used to decompose these complex tasks in multiple subtasks, which are either reactive policies learned from data or directly engineered, and a state-machine that selects the most appropriate one to be followed at every time-step. The subtask that performs task-specific navigation is learned from the provided human demonstration dataset. For example, searching for the best spot to place a waterfall in the \textit{MakeWaterfall} task requires navigation. Subtasks with little demonstration data available are engineered in combination with the learned state classifier. Throwing a snowball while inside the cave to signal the end of the episode can be engineered using human demonstration data. 

Once the complex tasks are decomposed into multiple small subtasks, a state-machine is engineered in combination with the learned state classifier to select the best subtask to be followed at every time-step. Each of these engineered subtasks was implemented by a human designer who hard-coded a sequence of actions to be taken using the same interface available to the agent. In addition to these subtasks, the human designer also implemented a safety-critical subtask allowing the agent to escape a body of water whenever the state classifier detects that the agent is swimming. Section~\ref{appendix:state_machine} describes in detail the sequence of subtasks followed by the state-machine for each task.

\subsection{Evaluation Methods}

Four different approaches are evaluated to solve the four tasks proposed in the Minecraft competition:
\begin{itemize}
    \item \textbf{Hybrid: }the main proposed agent, which combines both learned and engineered modules. The learned modules are the navigation subtask policy (learns how to navigate using the human demonstration data provided by the competition) and the state classifier (learns how to identify relevant states using additional human-labeled data). 
    The engineered modules are the multiple subtasks, hand-designed to solve subtasks that were not able to be learned from data. These engineered modules are the estimated odometry and the state-machine, which uses the output of the state classifier and engineered task structure to select which subtask should be followed at each time-step.
    \item \textbf{Engineered: }almost identical to the Hybrid agent described above, however, the navigation subtask policy that was learned from human demonstrations is now replaced by a hand-designed module that randomly selects movement and camera commands to explore the environment.
    \item \textbf{Behavior Cloning: }end-to-end imitation learning agent that learns solely from the human demonstration data provided during the competition. This agent does not use any other learned or engineered module, which includes the state classifier, the estimated odometry, and the state-machine.
    \item \textbf{Human: }human-generated trajectories provided by the competition. They are neither guaranteed to solve the task nor solve it optimally because they depend on the level of expertise of each human controlling the agent.
\end{itemize}

To collect human evaluations for each of the four baselines in a head-to-head comparison, a web application\footnote{Custom MineRL BASALT evaluation webpage: \url{https://kairosminerl.herokuapp.com/}.} is used in a manner like how the teams were evaluated during the official MineRL BASALT competition, as seen in Section~\ref{appendix:evaluation_interface}. In this case, each participant was asked to see two videos of different agents performing the same task then to answer three questions:
\begin{enumerate}
    \item \textit{Which agent best completed the task?}
    \item \textit{Which agent was the fastest completing the task?}
    \item \textit{Which agent had a more human-like behavior?}
\end{enumerate}
\noindent
For each question, the participants were given three answers: ``Agent 1'', ``Agent 2'', or ``None''.

The demonstration database had videos of all four types of agents (Behavior Cloning, Engineered, Hybrid, and Human) performing all four tasks (FindCave, MakeWaterfall, CreateVillageAnimalPen, and BuildVillageHouse). There were $10$ videos of each agent type solving all four tasks, for a total of $160$ videos in the database. Task, agent type, and videos were uniformly sampled from the database at each time a new evaluation form was generated and presented to the human evaluator. A total of 268 evaluations were collected (pairwise comparison where a human evaluator judged which agent was the best, fastest, and more human-like performing the tasks) from 7 different human evaluators.

All agent-generated videos were scaled from the original $64 \times 64$ image resolution returned by the environment to $512 \times 512$ image resolution to make the videos clearer for the human evaluators. The videos of the ''Human'' agent type were randomly selected from the video demonstrations provided by the MineRL BASALT competition and scaled to $512 \times 512$ image resolution to match the agent-generated videos. All videos were generated and saved at $20$ frames per second to match the sampling rate of the Minecraft simulator used by both agents and humans.

\section{State Classifier Labeling GUI}\label{appendix:label_state_gui}

\begin{figure}[!ht]
  \centering
  \includegraphics[width=0.75\linewidth]{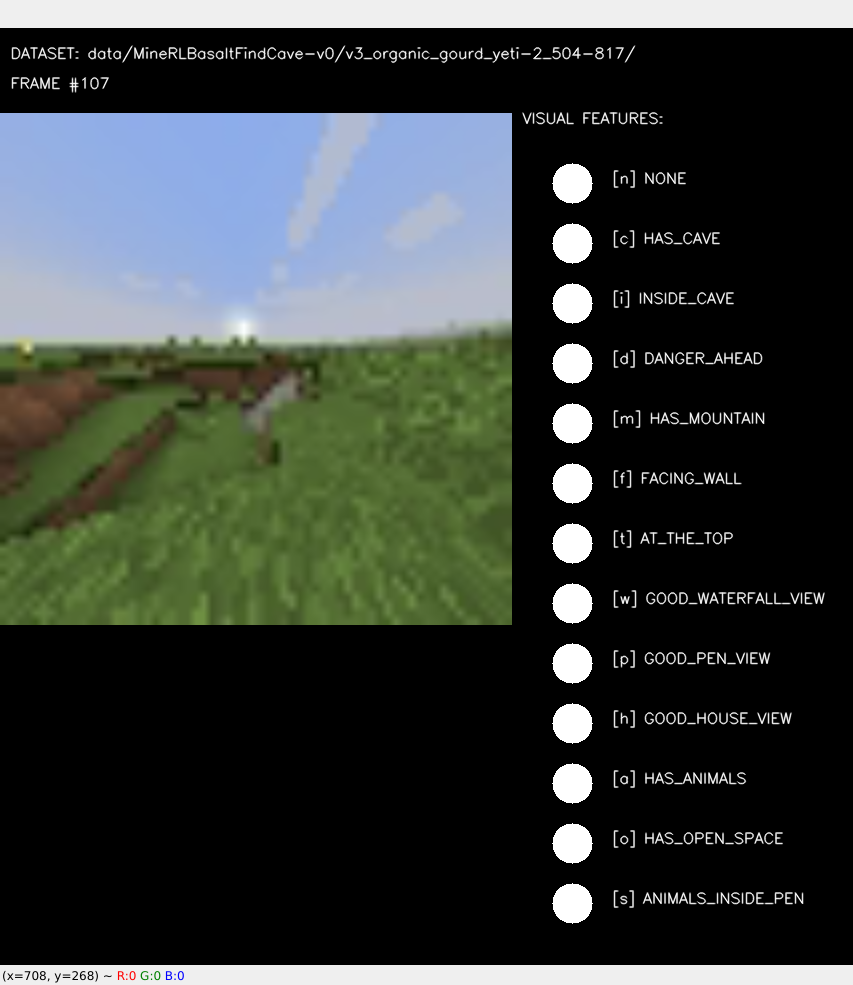}
  \caption{Custom GUI to relabel human dataset provided by the competition to train a classifier to identify relevant states for the state-machine. This figure illustrates the ease of use for labeling multiple data by a human operator. }
  \label{fig:label_state_gui}
\end{figure}

The labeling process of relevant states to the state-machine uses both mouse clicks and keyboard presses and takes place in a custom GUI, as seen in Figure~\ref{fig:label_state_gui}. On the top left of the GUI, users can double-check which dataset and frame number they are labeling. Below that, the GUI displays the RGB frame to be labeled (center left) and the options for labels (center right, same labels for all tasks). To label a frame, the user can simply press the keyboard key corresponding to the desired label (shown in brackets, for example, $[c]$ for $has\_cave$), or click in the white circles in front of the label, which will then turn green, indicating that the label was selected. Frames will automatically advance when a key is pressed. If the users only use the mouse to select the labels, they will still need to press the keyboard key to advance to the next frame (any key for a label that was already selected clicking).

\section{State-Machine Definition for each Task}\label{appendix:state_machine}

The sequence of subtasks used by the state-machine for each task is defined as follows:
\begin{itemize}
    \item \textit{FindCave}:
        \begin{enumerate}
            \item Use navigation policy to traverse the environment and search for caves.
            \item If the state classifier detects the agent is inside a cave, throw a snowball to signal that the task was completed (end of episode).
        \end{enumerate}
    \item \textit{MakeWaterfall}:
        \begin{enumerate}
            \item Use navigation policy to traverse the environment and search for a place in the mountains to make a waterfall.
            \item If the state classifier detects the agent is at the top of a mountain, build additional blocks to give additional height to the waterfall.
            \item Once additional blocks are built, look down and place the waterfall by equipping and using the bucket item filled with water.
            \item After the waterfall is built, keep moving forward to move away from it.
            \item Once the agent has moved away from the waterfall, turn around and throw a snowball to signal that a picture was taken, and the task was completed (end of episode).
        \end{enumerate}
    \item \textit{CreateVillageAnimalPen}:
        \begin{enumerate}
            \item Use navigation policy to traverse the environment and search for a place to build a pen.
            \item If the state classifier detects an open-space, build the pen. The subtask to build the pen directly repeats the actions taken by a human while building the pen, as observed in provided demonstration dataset.
            \item Once the pen is built, use the estimated odometry map to navigate to the closest animal location. If no animals were seen before, use navigation policy to traverse the environment and search for animals.
            \item At the closest animal location, equip food to attract attention of the animals and lure them.
            \item Using the estimated odometry map, move back to where the pen was built while animals are following the agent.
            \item Once inside the pen together with the animals, move away from pen, turn around and throw a snowball to signal that the task was completed (end of episode).
        \end{enumerate}
    \item \textit{BuildVillageHouse}:
        \begin{enumerate}
            \item Use navigation policy to traverse the environment and search for a place to build a house.
            \item If the state classifier detects an open-space, build the house. The subtask to build the house directly repeats the actions taken by a human while building the house, as observed in provided demonstration dataset.
            \item Once the house is built, move away from it, turn around and throw a snowball to signal that the task was completed (end of episode).
        \end{enumerate}
\end{itemize}

\section{Human Evaluation Interface}\label{appendix:evaluation_interface}

\begin{figure}[!ht]
  \centering
  \includegraphics[width=0.95\linewidth]{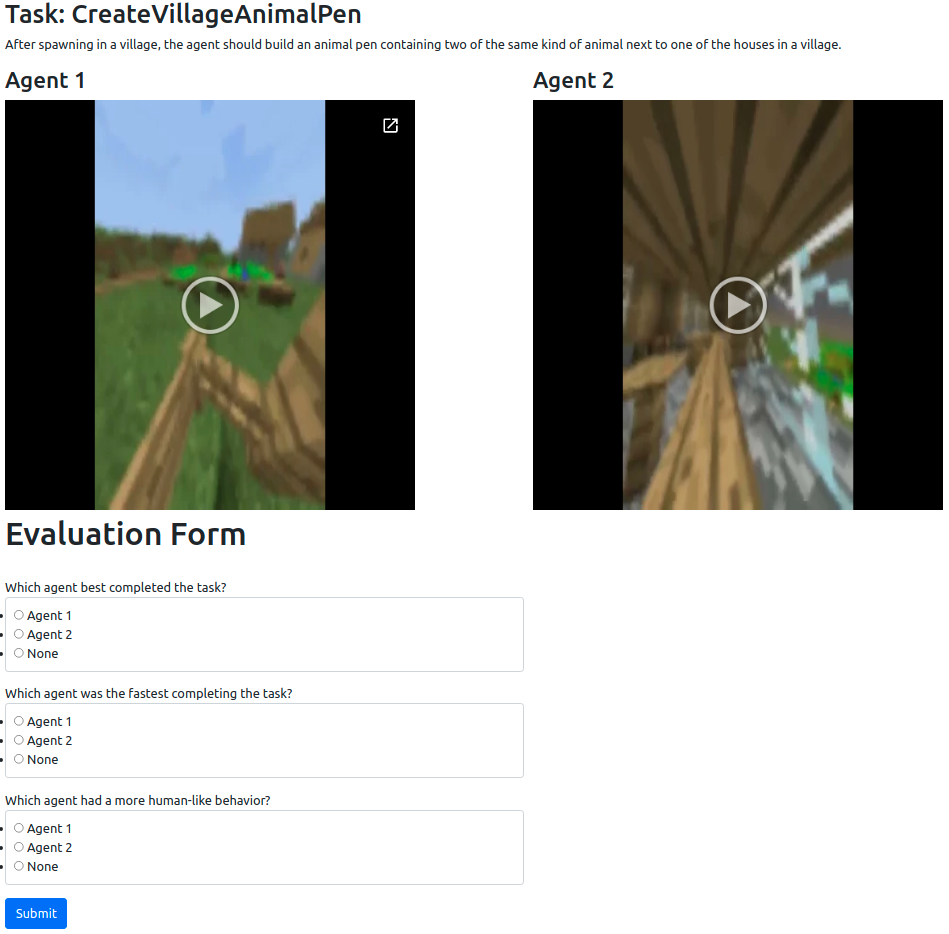}
  \caption{Web evaluation form used to collect additional human evaluation data to evaluate the multiple agent conditions presented. This figure illustrates the ease of collecting multiple evaluations by a human operator. }
  \label{fig:evaluation_interface}
\end{figure}

Figure~\ref{fig:evaluation_interface} shows a sample of the web evaluation form available at \url{https://kairosminerl.herokuapp.com/} that was used to collect human evaluations for each of the four baselines in a head-to-head comparison, like how the teams were evaluated during the official MineRL BASALT competition.
Each participant was asked to see two videos of different agents performing the same task then answer three questions with respect to the agent's performance.

\section{\textit{TrueSkill}\textsuperscript{TM} Score per Match}\label{appendix:trueskill}

\begin{figure}[!ht]
  \centering
    \begin{tabular}{ccc}
        \subfloat[Best Performer]{\includegraphics[width=0.31\linewidth]{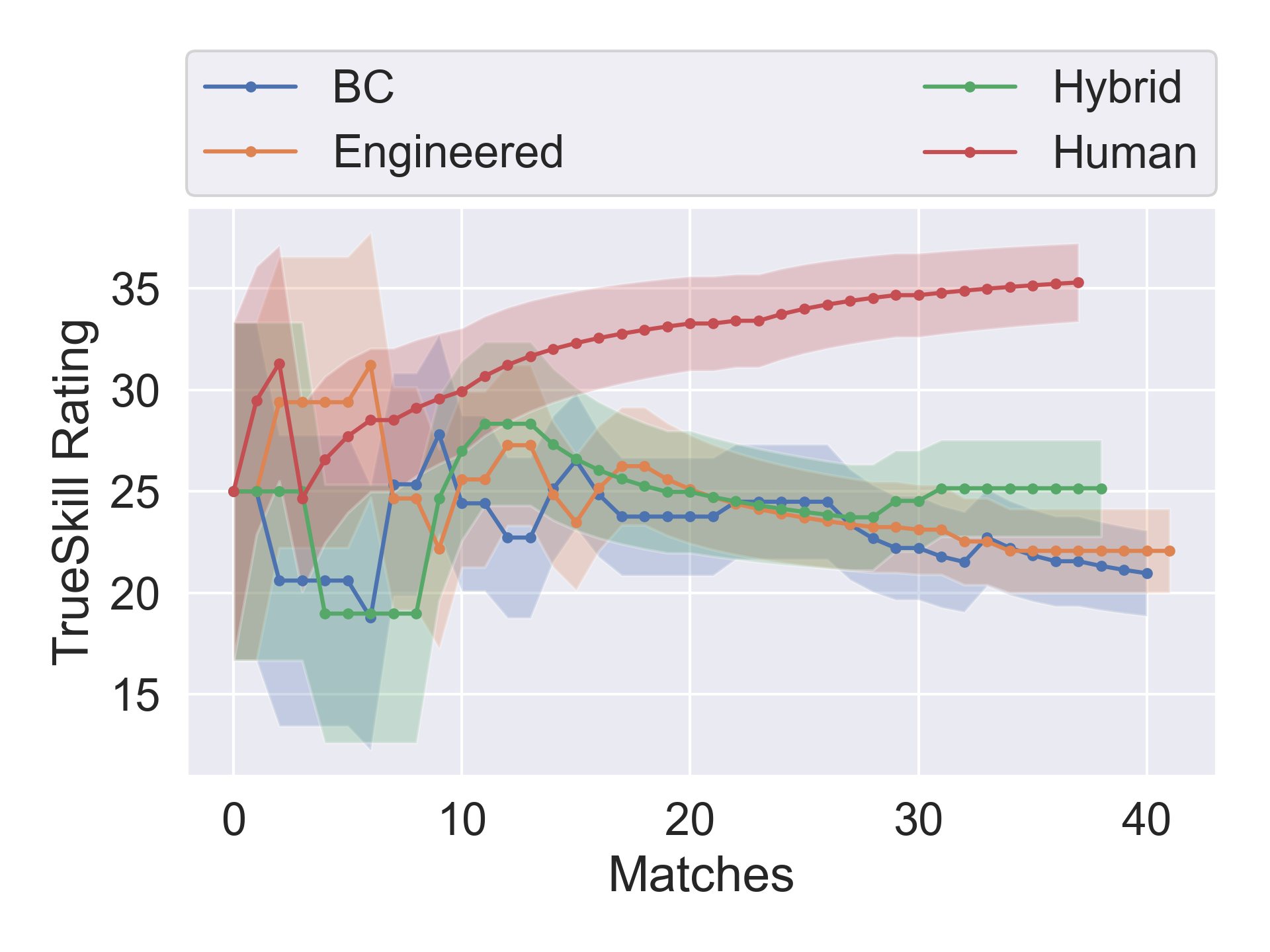}} &
        \subfloat[Fastest Performer]{\includegraphics[width=0.31\linewidth]{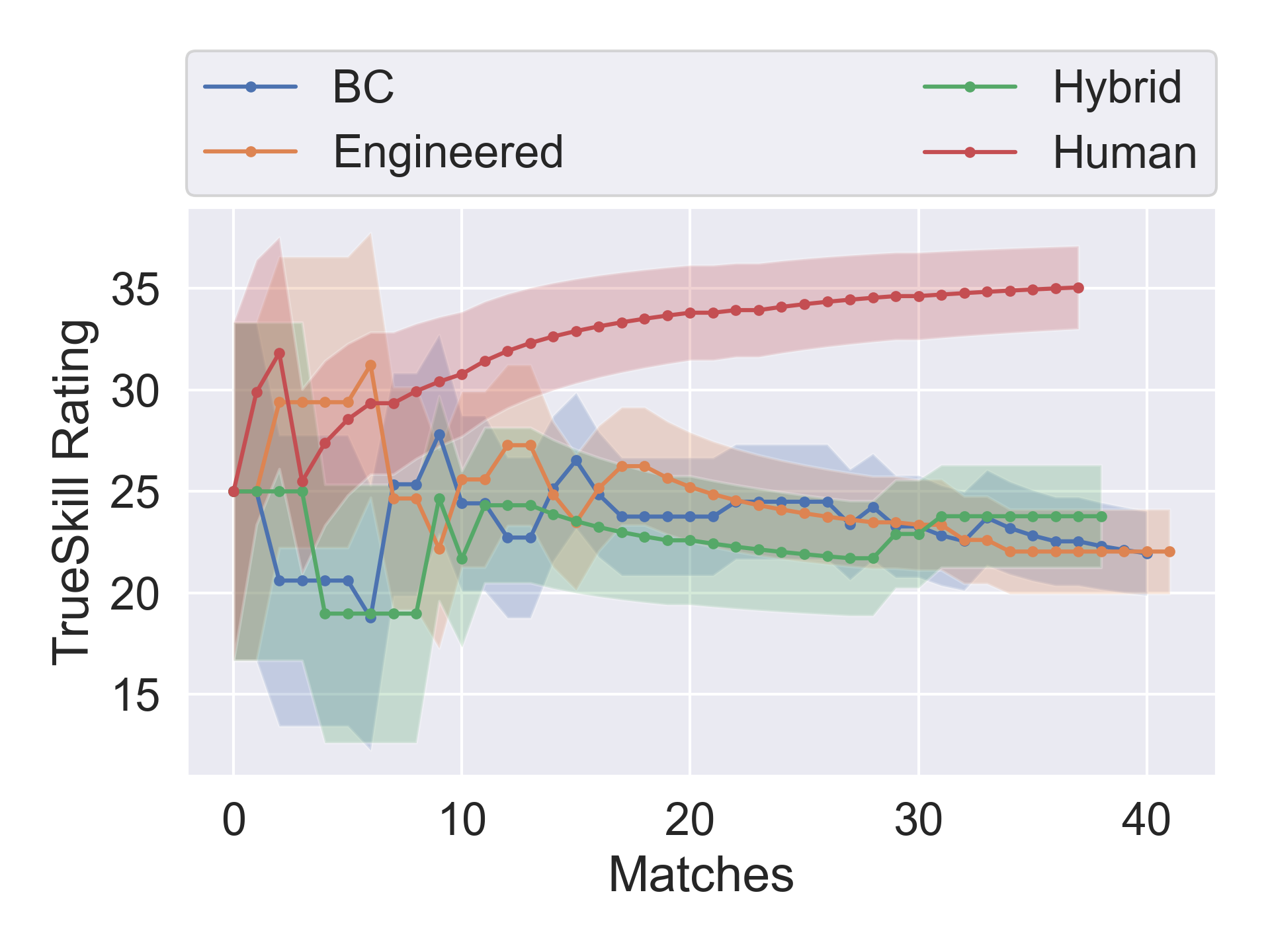}} &
        \subfloat[More Human-like Behavior]{\includegraphics[width=0.31\linewidth]{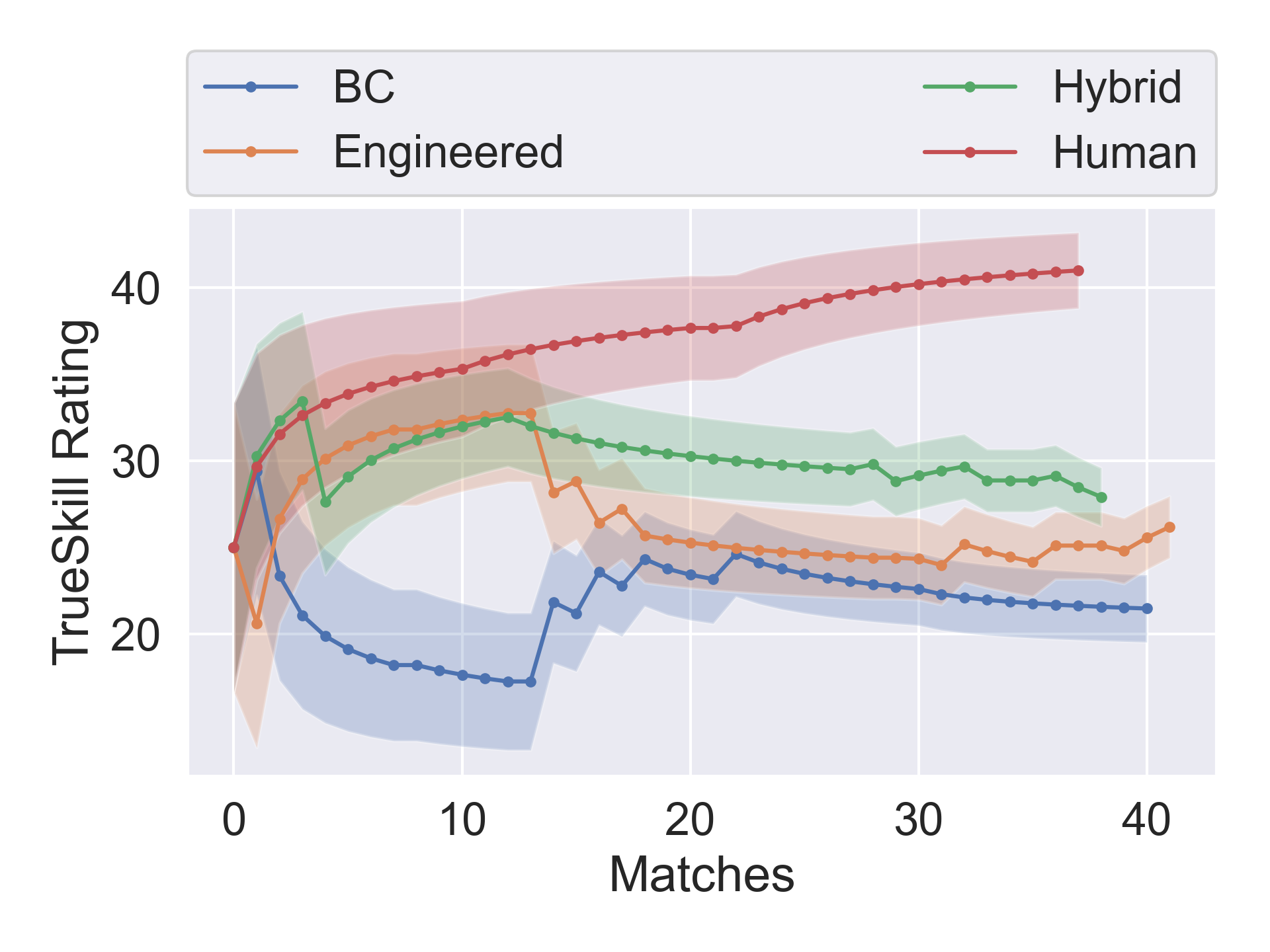}}
    \end{tabular}
  \caption{\textit{TrueSkill}\textsuperscript{TM}\cite{herbrich2006trueskill} scores computed from human evaluations separately for each performance metric and for each agent type performing the \textit{FindCave} task.}
  \label{fig:cave_trueskill}
\end{figure}

\begin{figure}[!ht]
  \centering
    \begin{tabular}{ccc}
        \subfloat[Best Performer]{\includegraphics[width=0.31\linewidth]{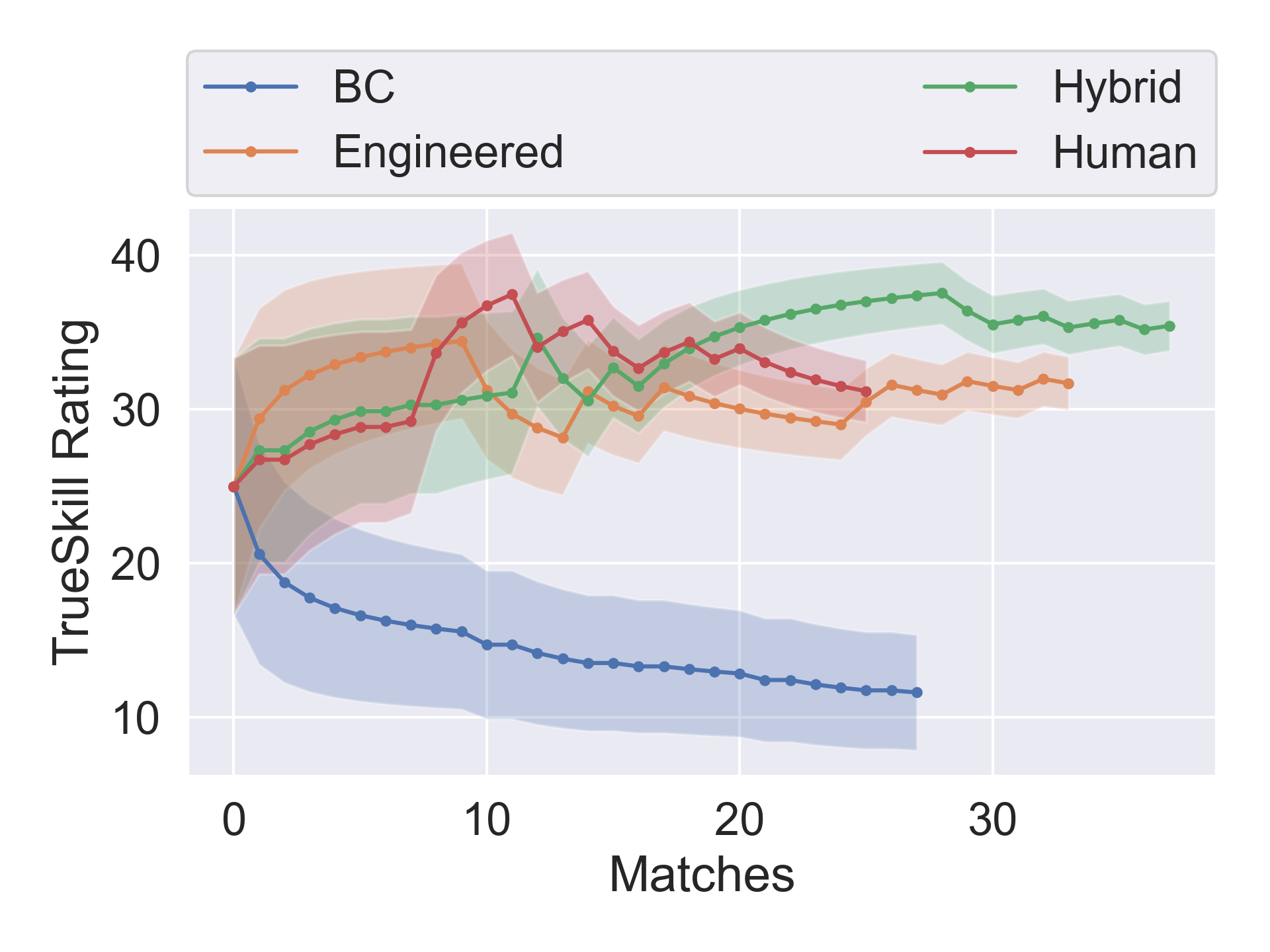}} &
        \subfloat[Fastest Performer]{\includegraphics[width=0.31\linewidth]{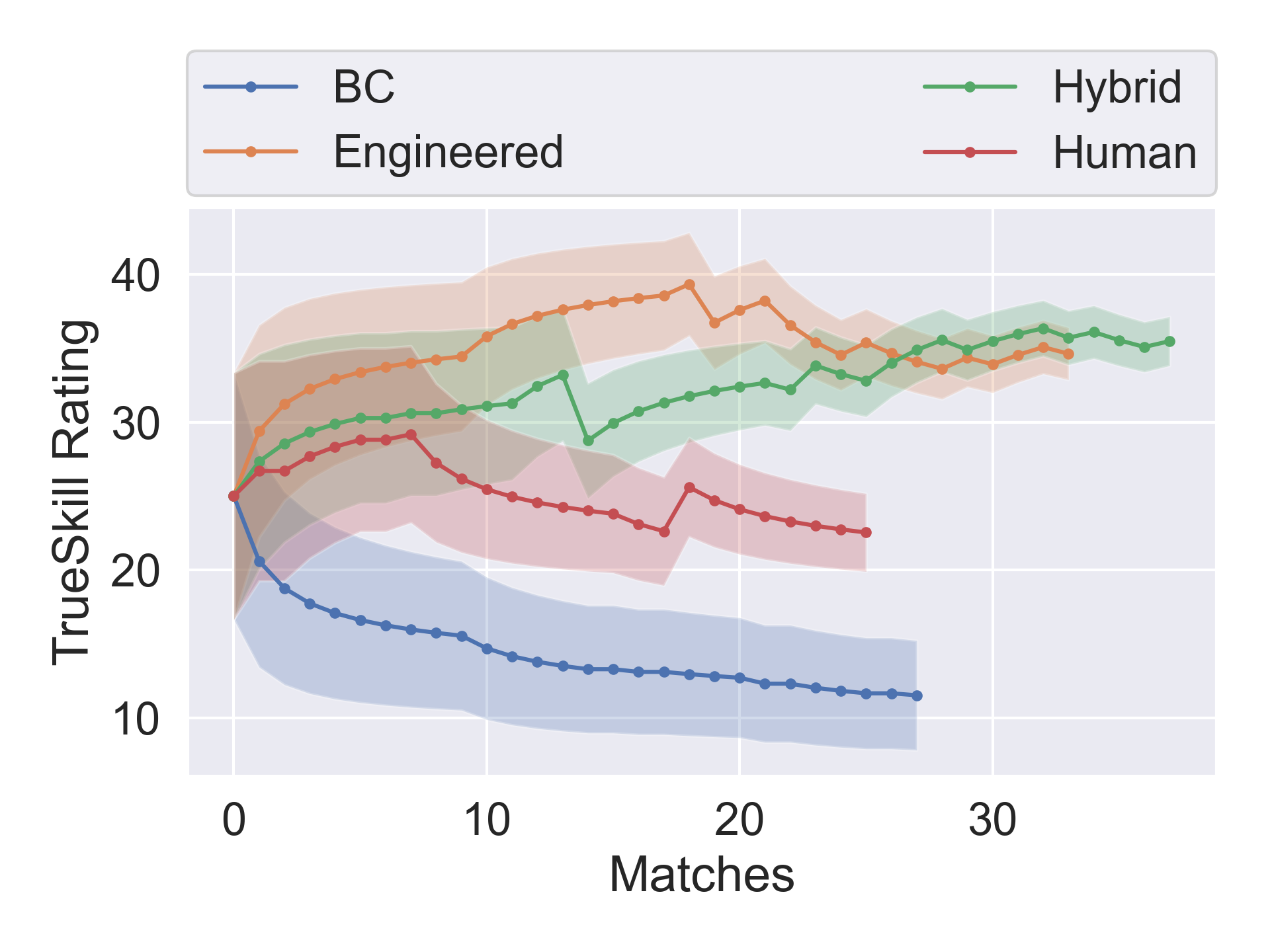}} &
        \subfloat[More Human-like Behavior]{\includegraphics[width=0.31\linewidth]{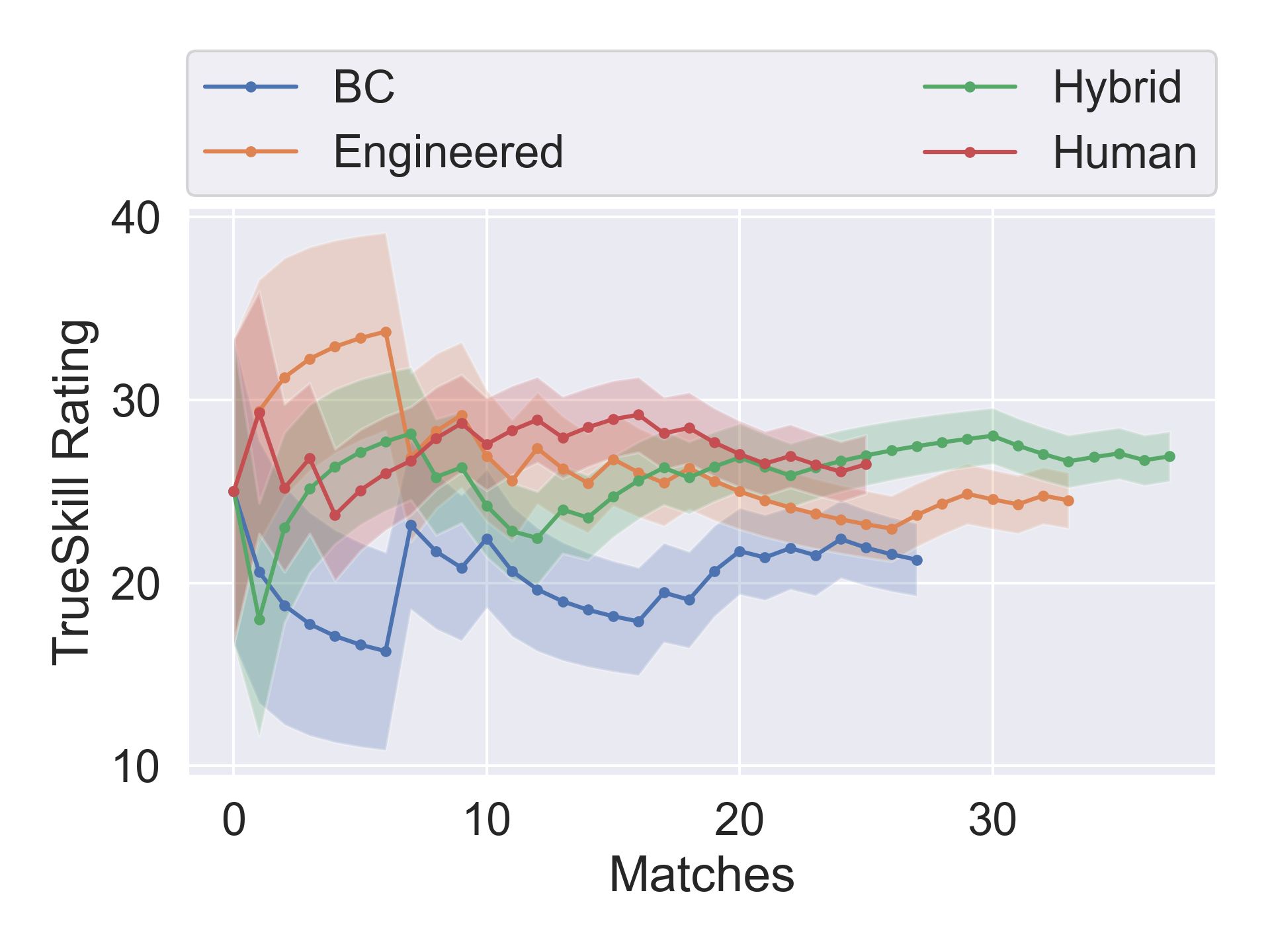}}
    \end{tabular}
  \caption{\textit{TrueSkill}\textsuperscript{TM}\cite{herbrich2006trueskill} scores computed from human evaluations separately for each performance metric and for each agent type performing the \textit{MakeWaterfall} task.}
  \label{fig:waterfall_trueskill}
\end{figure}

\begin{figure}[!ht]
  \centering
    \begin{tabular}{ccc}
        \subfloat[Best Performer]{\includegraphics[width=0.31\linewidth]{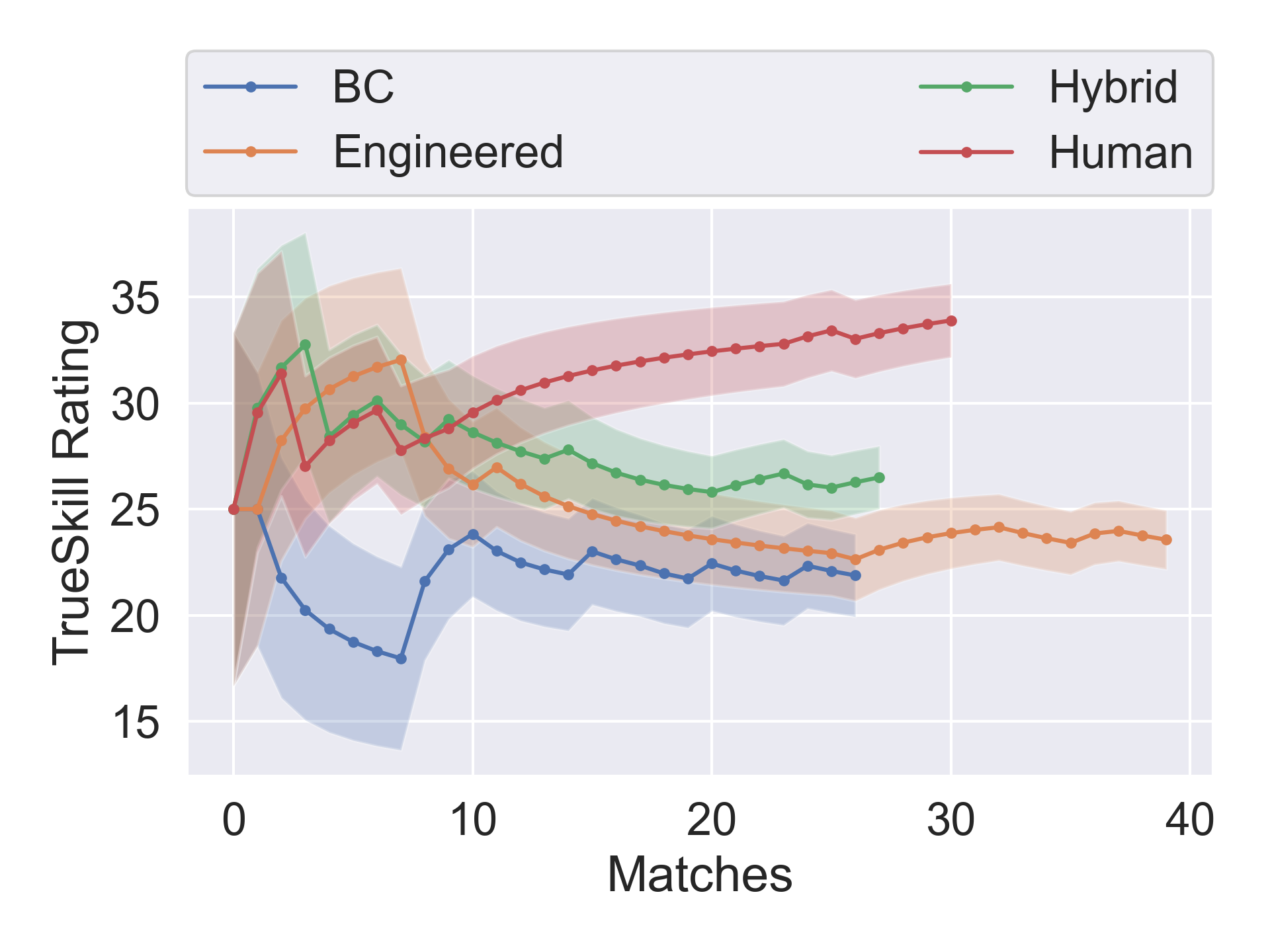}} &
        \subfloat[Fastest Performer]{\includegraphics[width=0.31\linewidth]{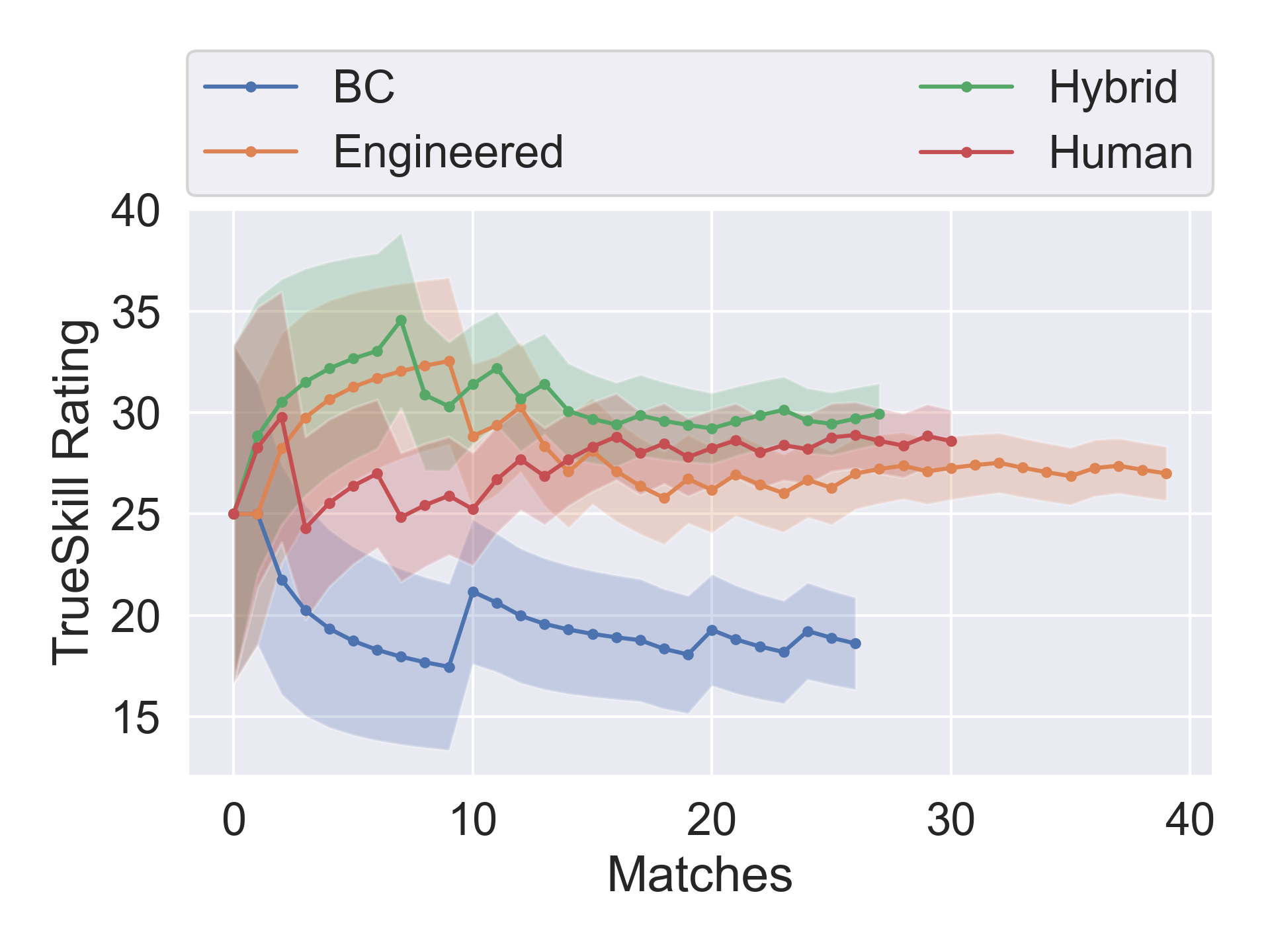}} &
        \subfloat[More Human-like Behavior]{\includegraphics[width=0.31\linewidth]{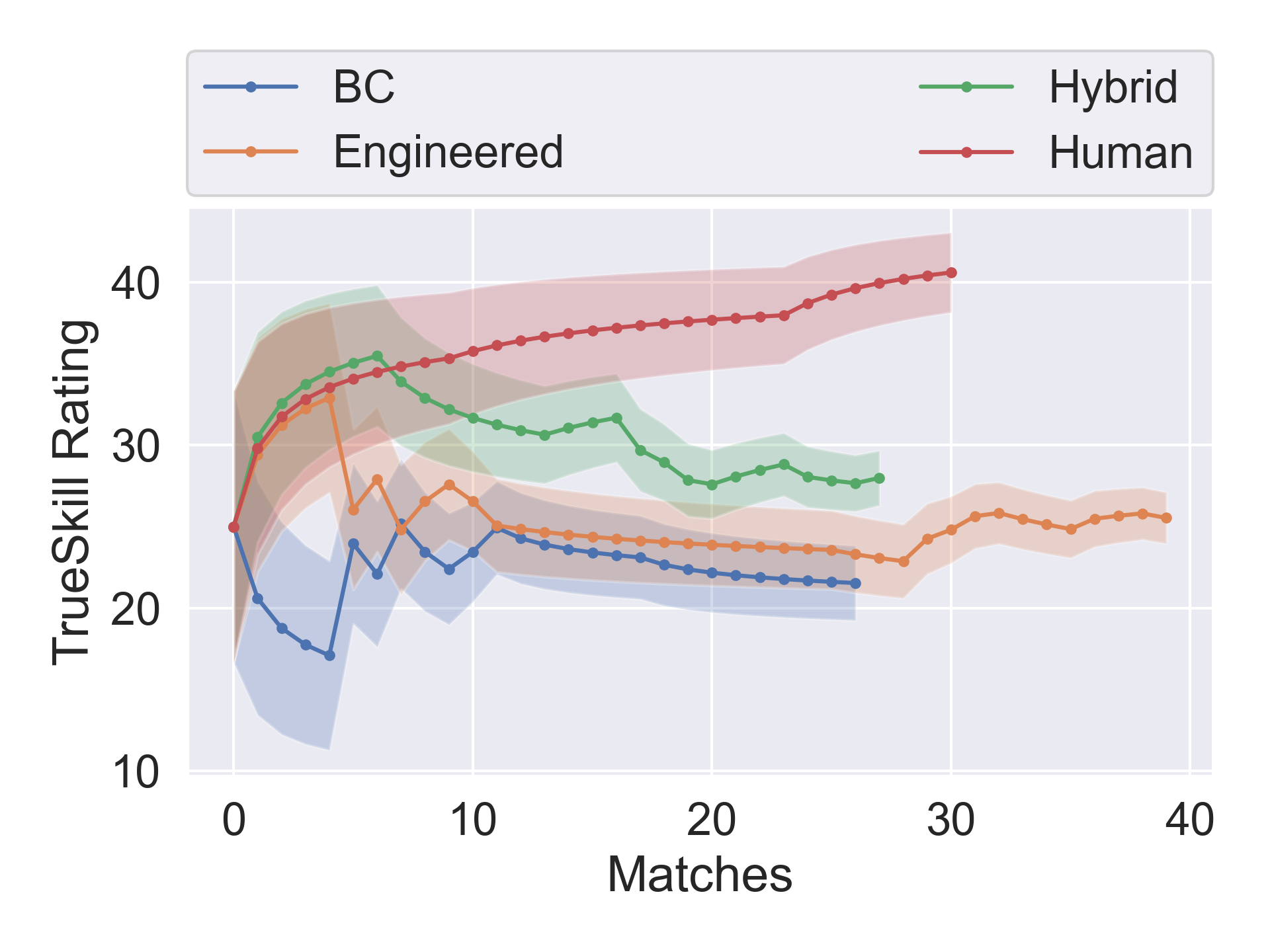}}
    \end{tabular}
  \caption{\textit{TrueSkill}\textsuperscript{TM}\cite{herbrich2006trueskill} scores computed from human evaluations separately for each performance metric and for each agent type performing the \textit{CreateVillageAnimalPen} task.}
  \label{fig:pen_trueskill}
\end{figure}

\begin{figure}[!ht]
  \centering
    \begin{tabular}{ccc}
        \subfloat[Best Performer]{\includegraphics[width=0.31\linewidth]{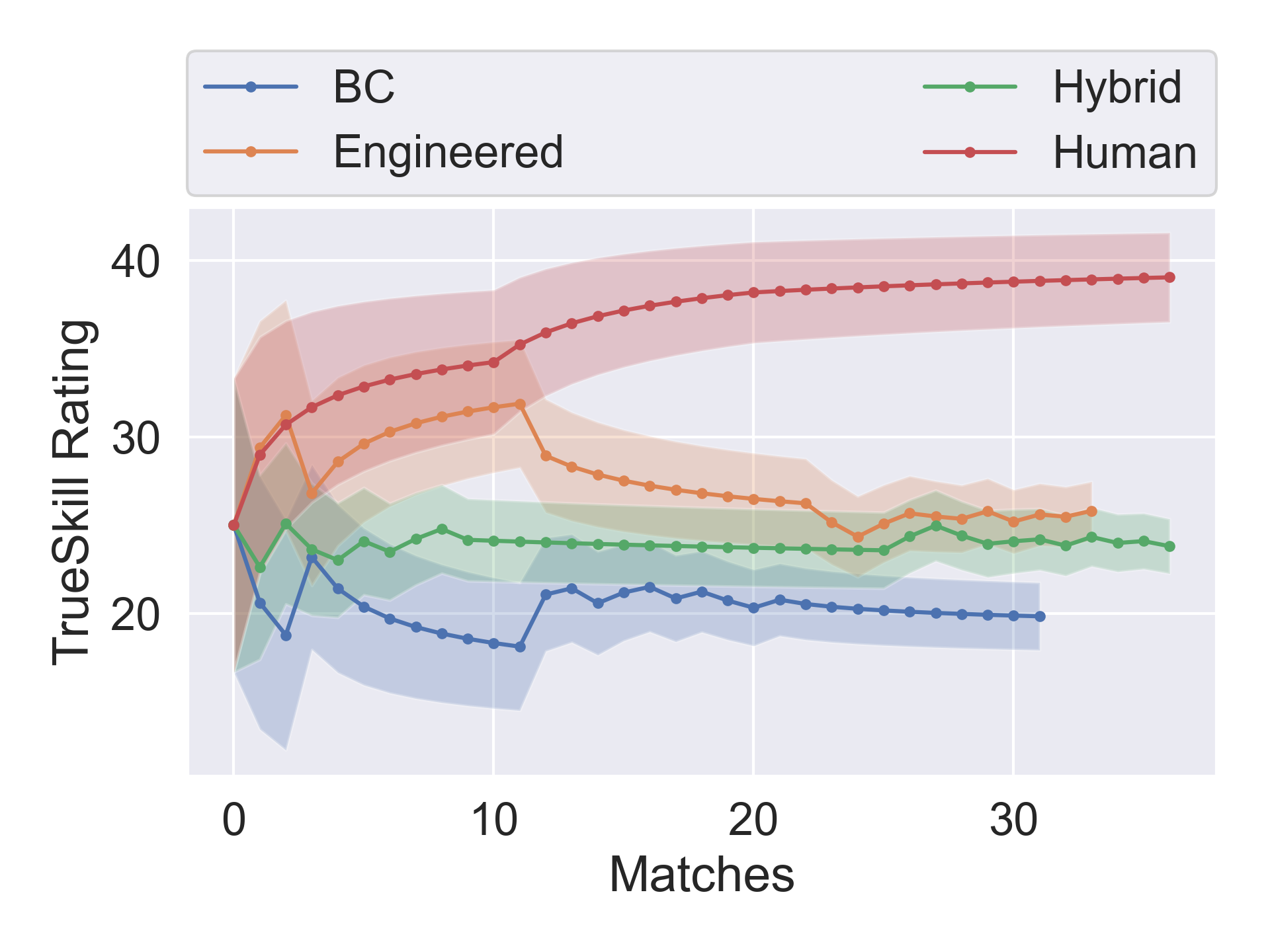}} &
        \subfloat[Fastest Performer]{\includegraphics[width=0.31\linewidth]{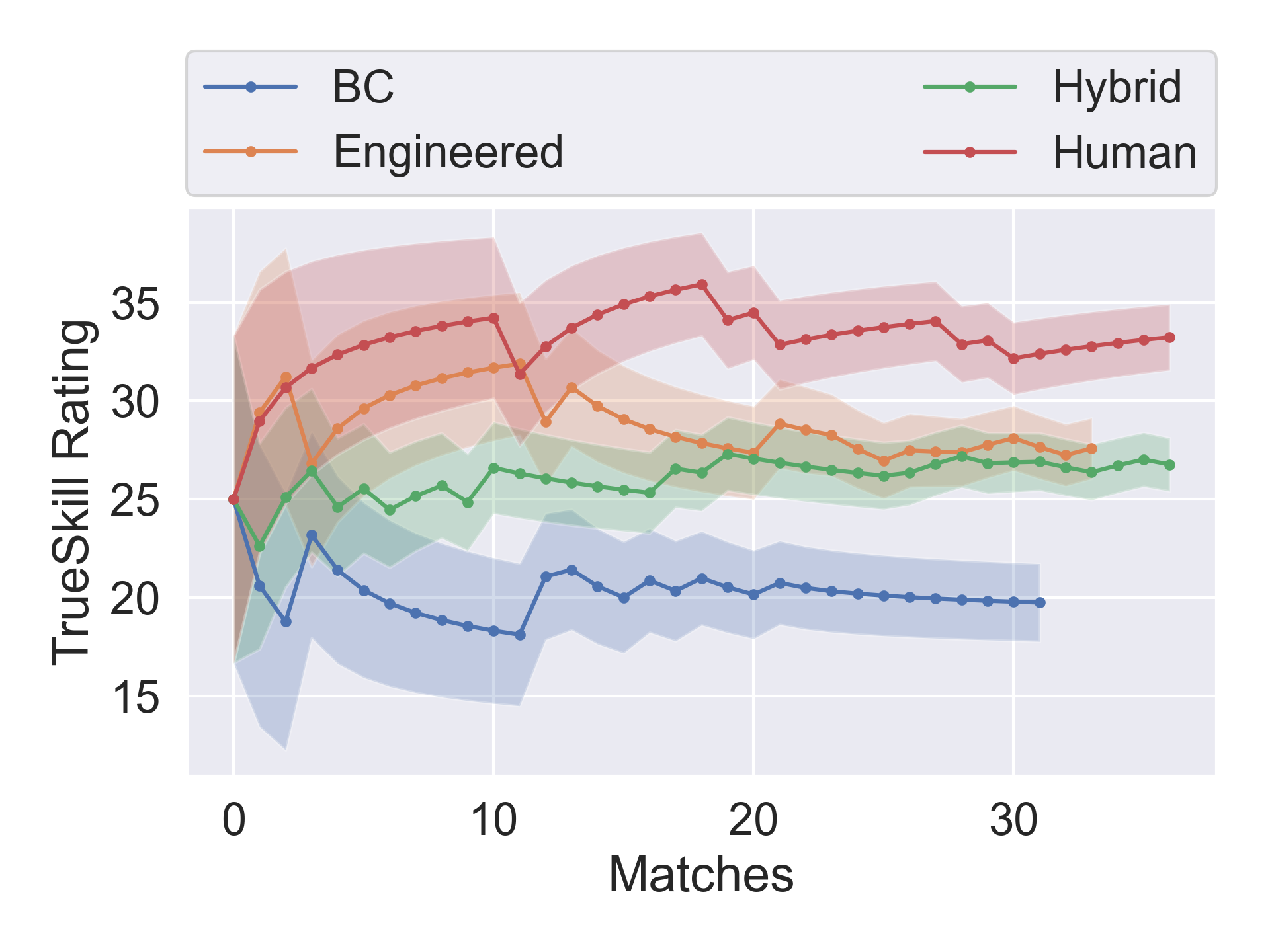}} &
        \subfloat[More Human-like Behavior]{\includegraphics[width=0.31\linewidth]{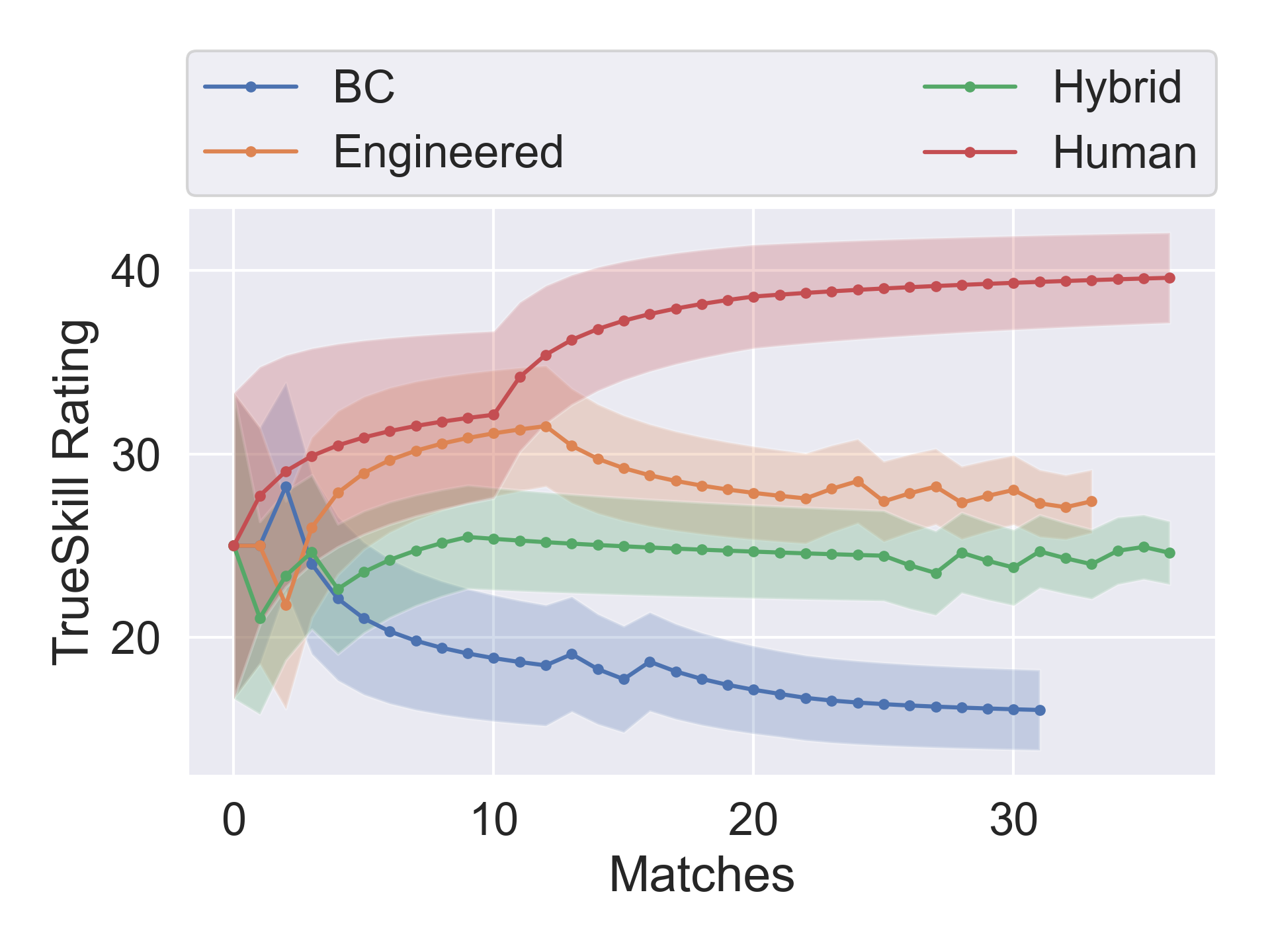}}
    \end{tabular}
  \caption{\textit{TrueSkill}\textsuperscript{TM}\cite{herbrich2006trueskill} scores computed from human evaluations separately for each performance metric and for each agent type performing the \textit{BuildVillageHouse} task.}
  \label{fig:house_trueskill}
\end{figure}

Figures~\ref{fig:cave_trueskill},~\ref{fig:waterfall_trueskill},~\ref{fig:pen_trueskill}, and~\ref{fig:house_trueskill} show the evolution of the \textit{TrueSkill}\textsuperscript{TM} scores after each match (one-to-one comparison between different agent types) for each performance metric when the agents are solving the \textit{FindCave}, \textit{MakeWaterfall}, \textit{CreateVillageAnimalPen}, and \textit{BuildVillageHouse} tasks, respectively.
The bold line represents the mean estimated skill rating and shaded area the standard deviation of the estimation.

\section{Pairwise Comparison per Performance Metric and Task}\label{appendix:barplot}

\begin{figure}[!ht]
  \centering
    \begin{tabular}{ccc}
        \subfloat[BC vs Engineered]{\includegraphics[width=0.3\linewidth]{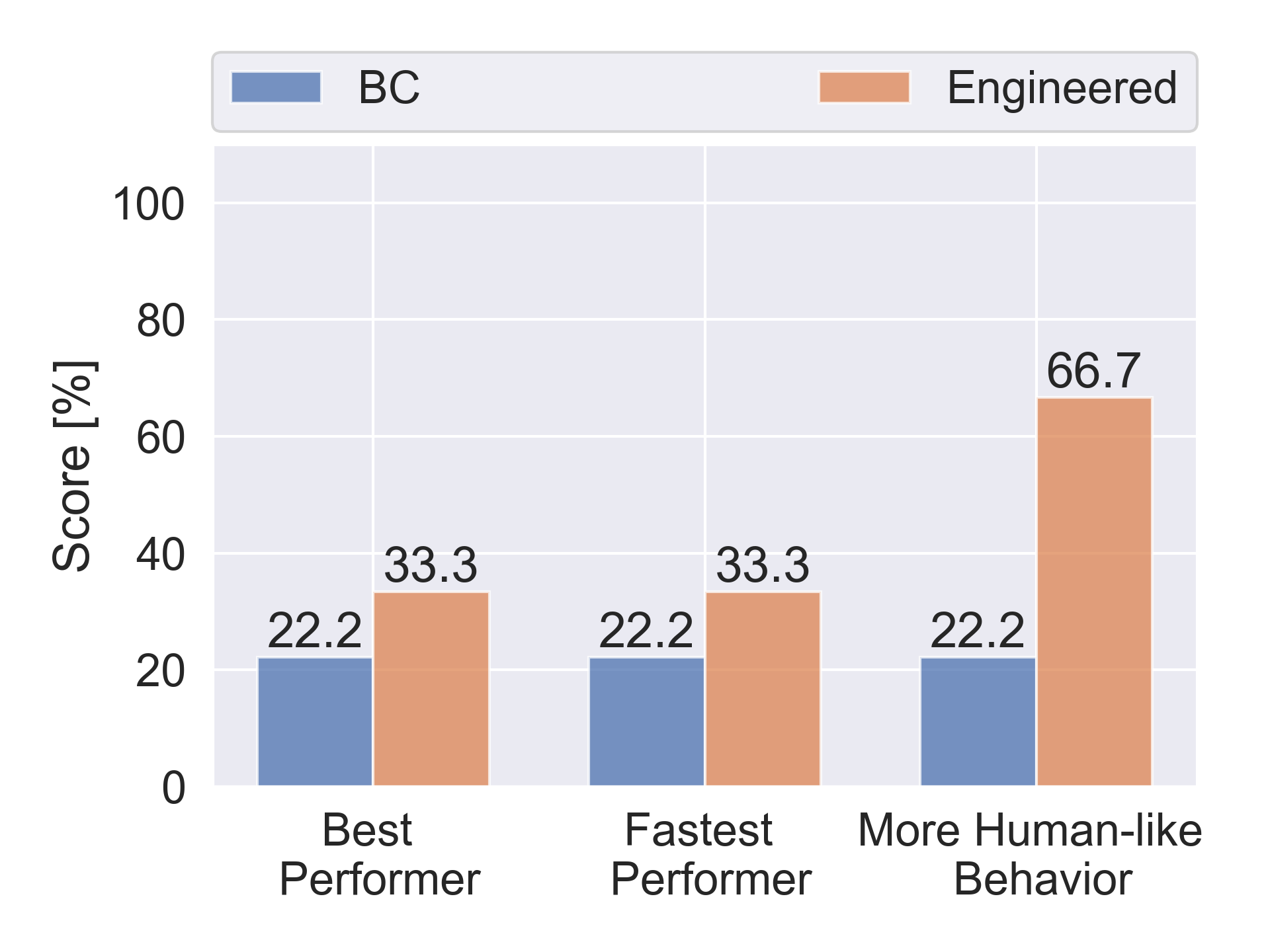}} &
        \subfloat[BC vs Human]{\includegraphics[width=0.3\linewidth]{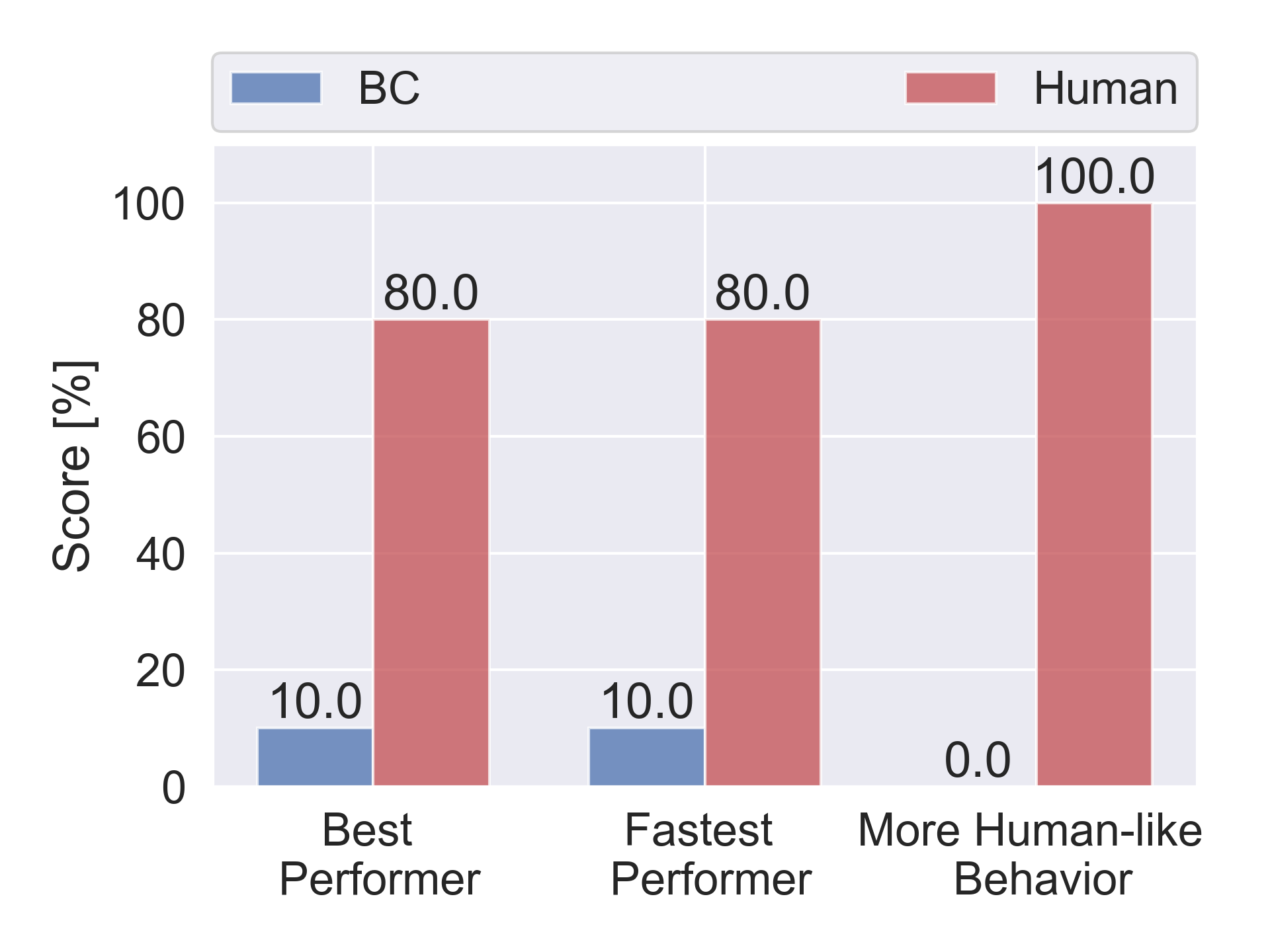}} &
        \subfloat[BC vs Hybrid]{\includegraphics[width=0.3\linewidth]{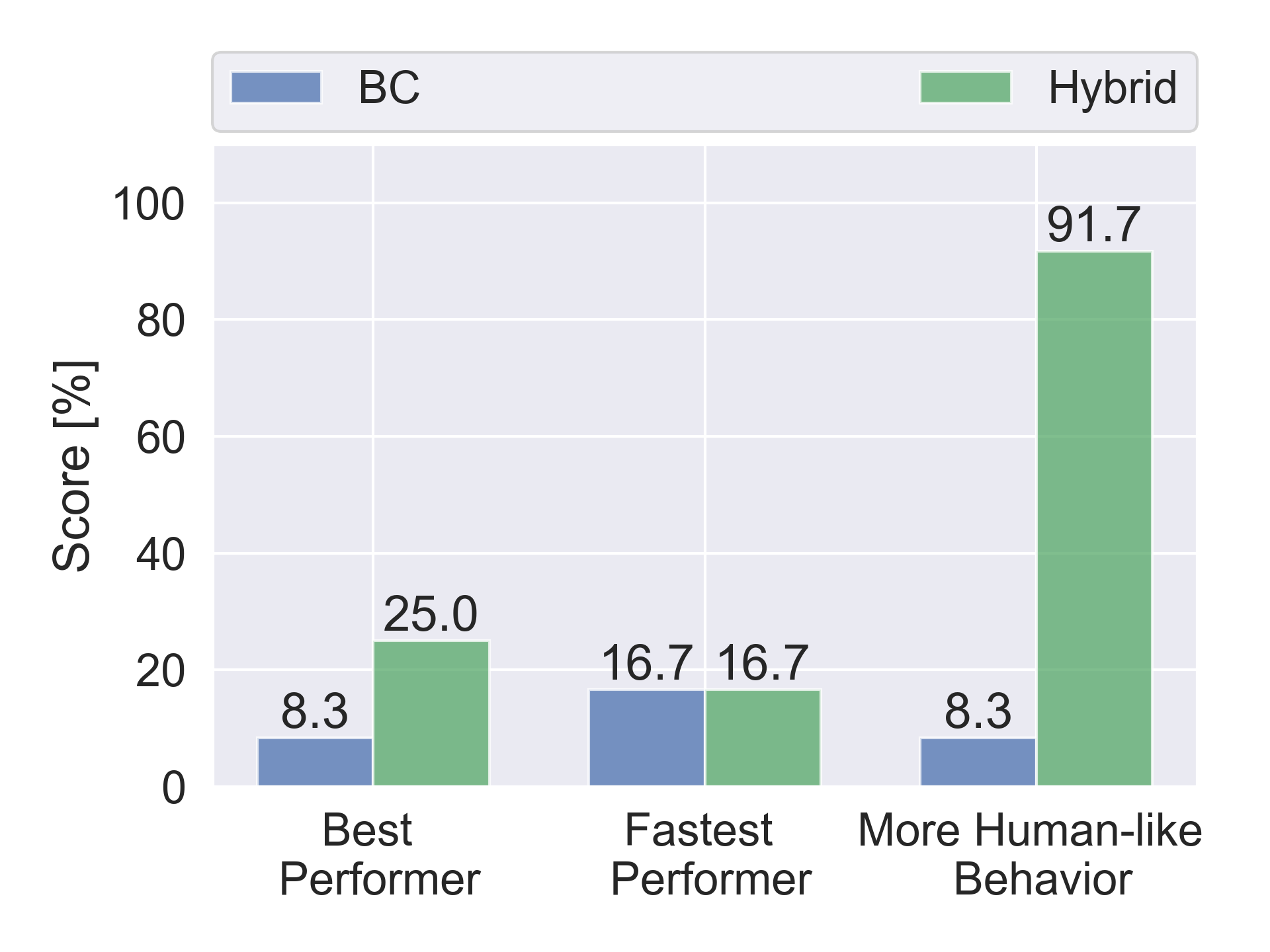}}\\
        \subfloat[Engineered vs Hybrid]{\includegraphics[width=0.3\linewidth]{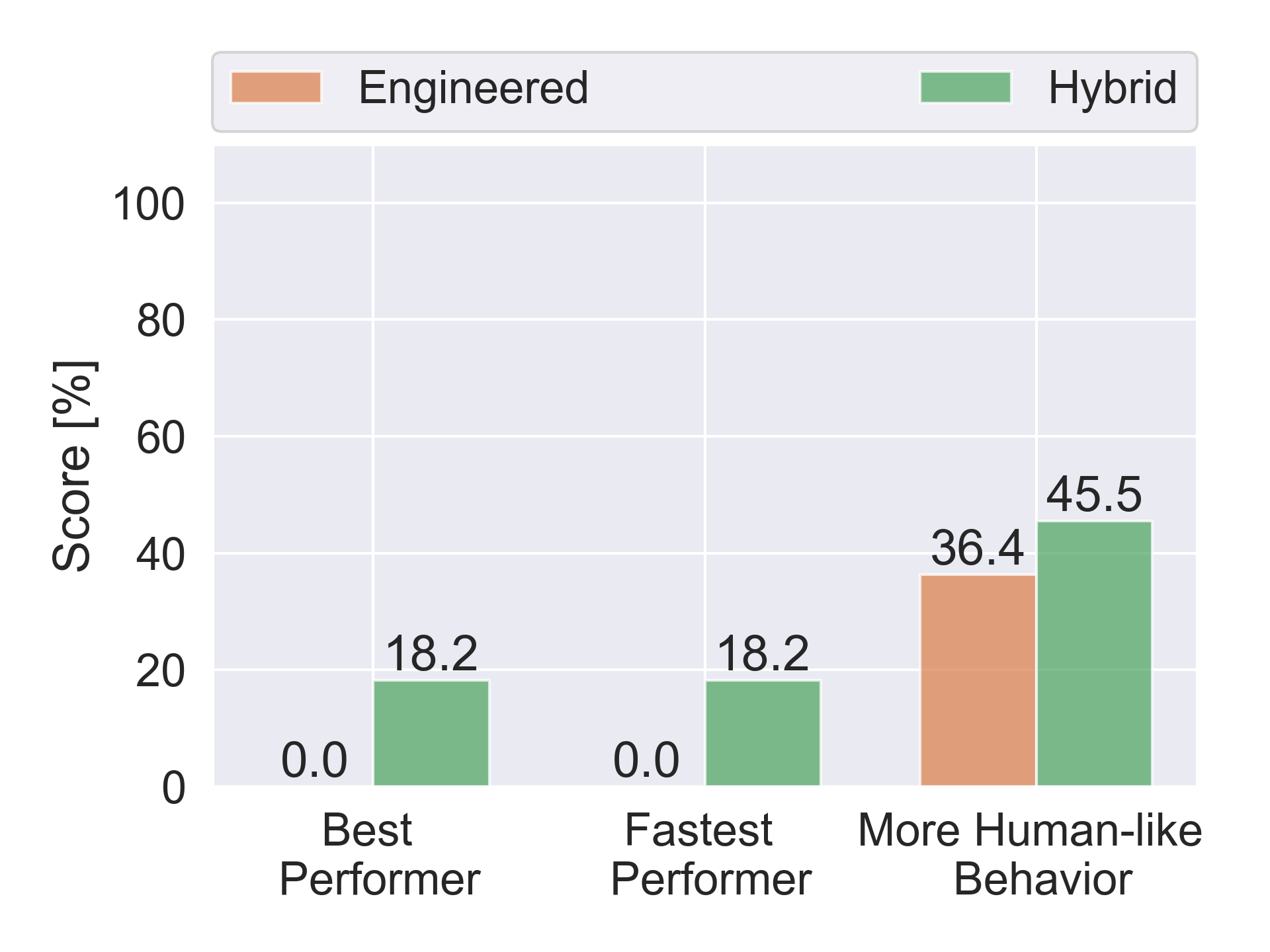}} &
        \subfloat[Human vs Engineered]{\includegraphics[width=0.3\linewidth]{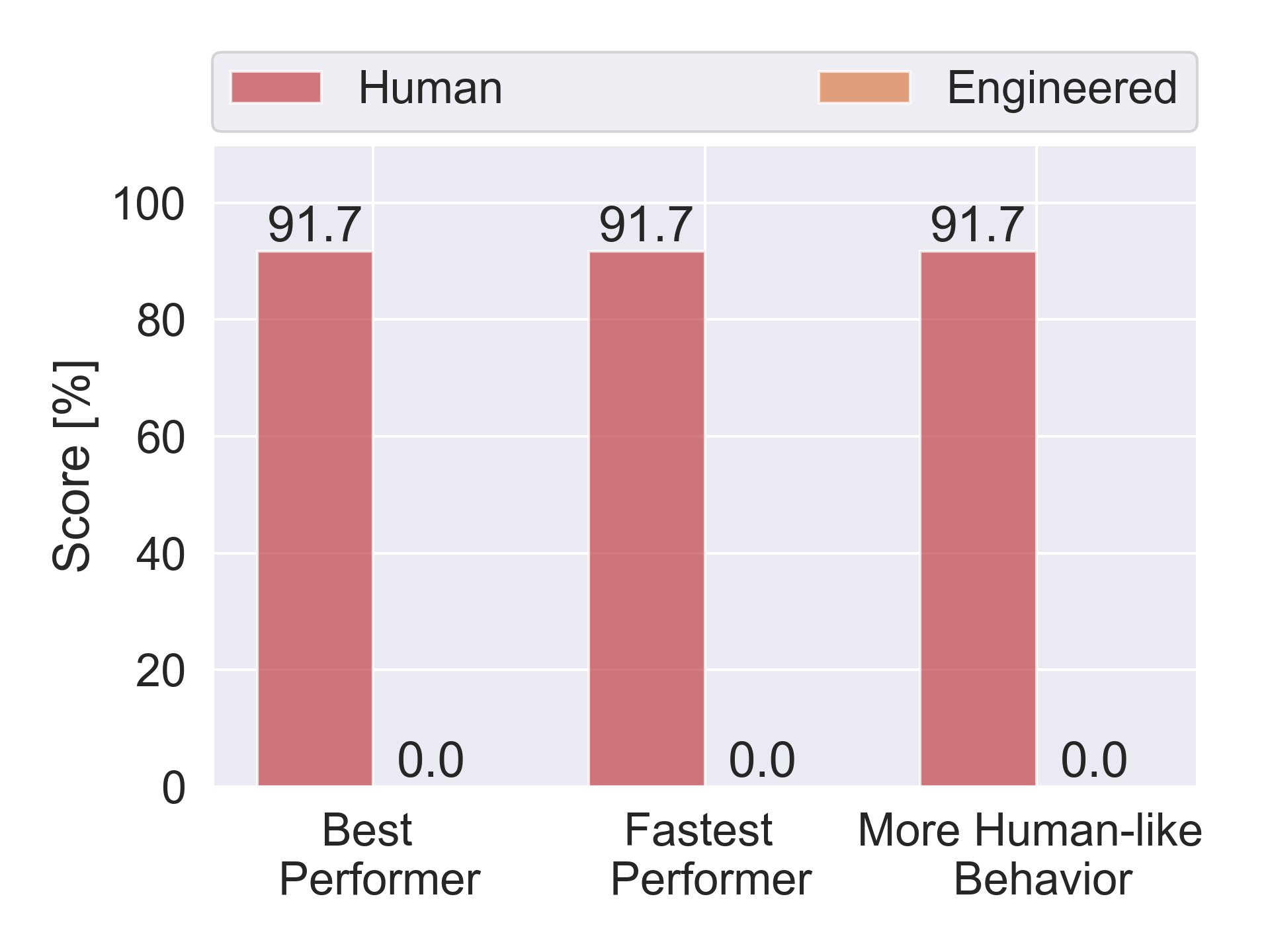}} &
        \subfloat[Human vs Hybrid]{\includegraphics[width=0.3\linewidth]{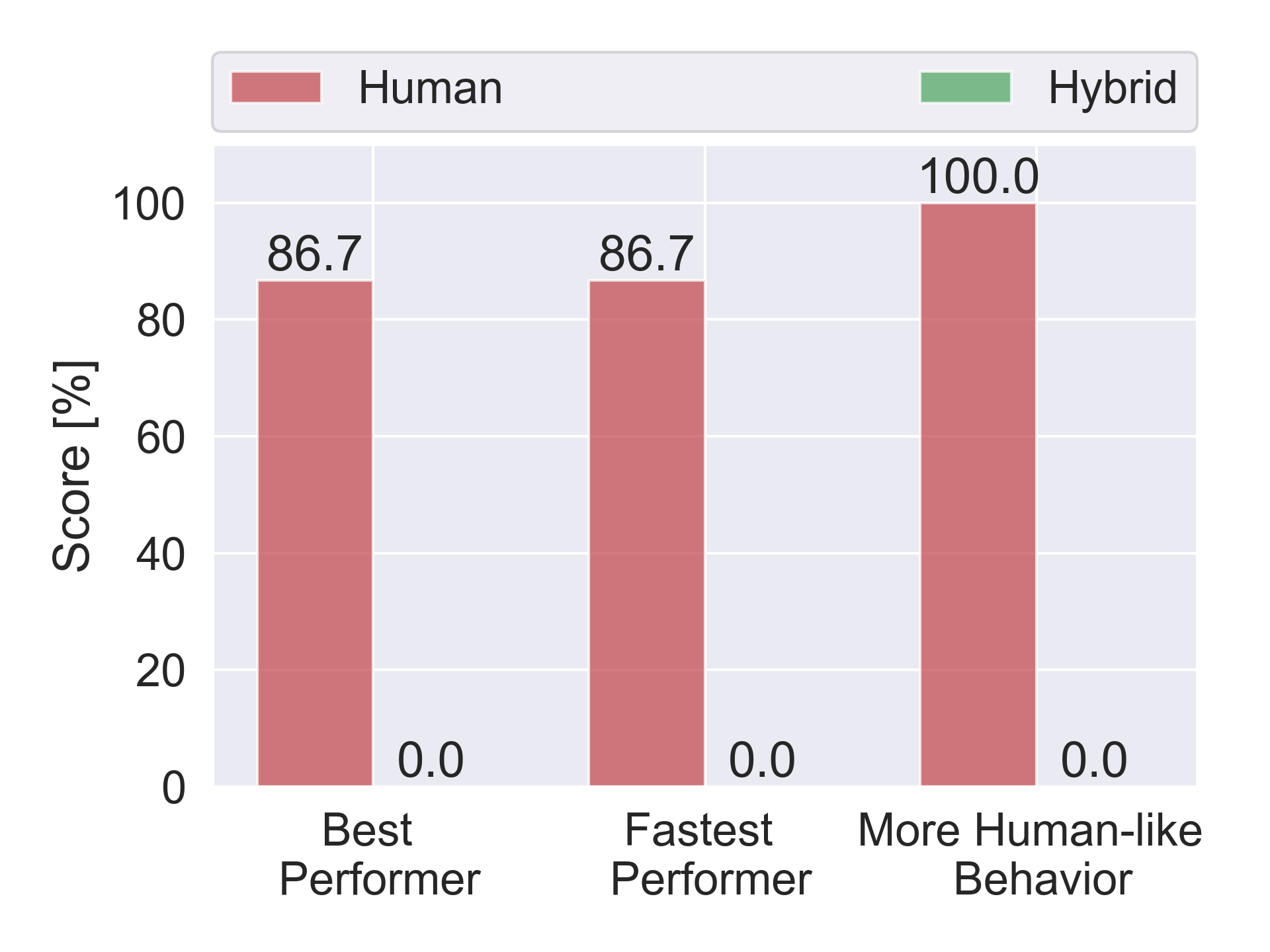}}
    \end{tabular}
  \caption{Pairwise comparison displaying the normalized scores computed from human evaluations separately for each performance metric on all possible head-to-head comparisons for all agent type performing the \textit{FindCave} task.}
  \label{fig:cave_barplot}
\end{figure}

\begin{figure}[!ht]
  \centering
    \begin{tabular}{ccc}
        \subfloat[BC vs Engineered]{\includegraphics[width=0.3\linewidth]{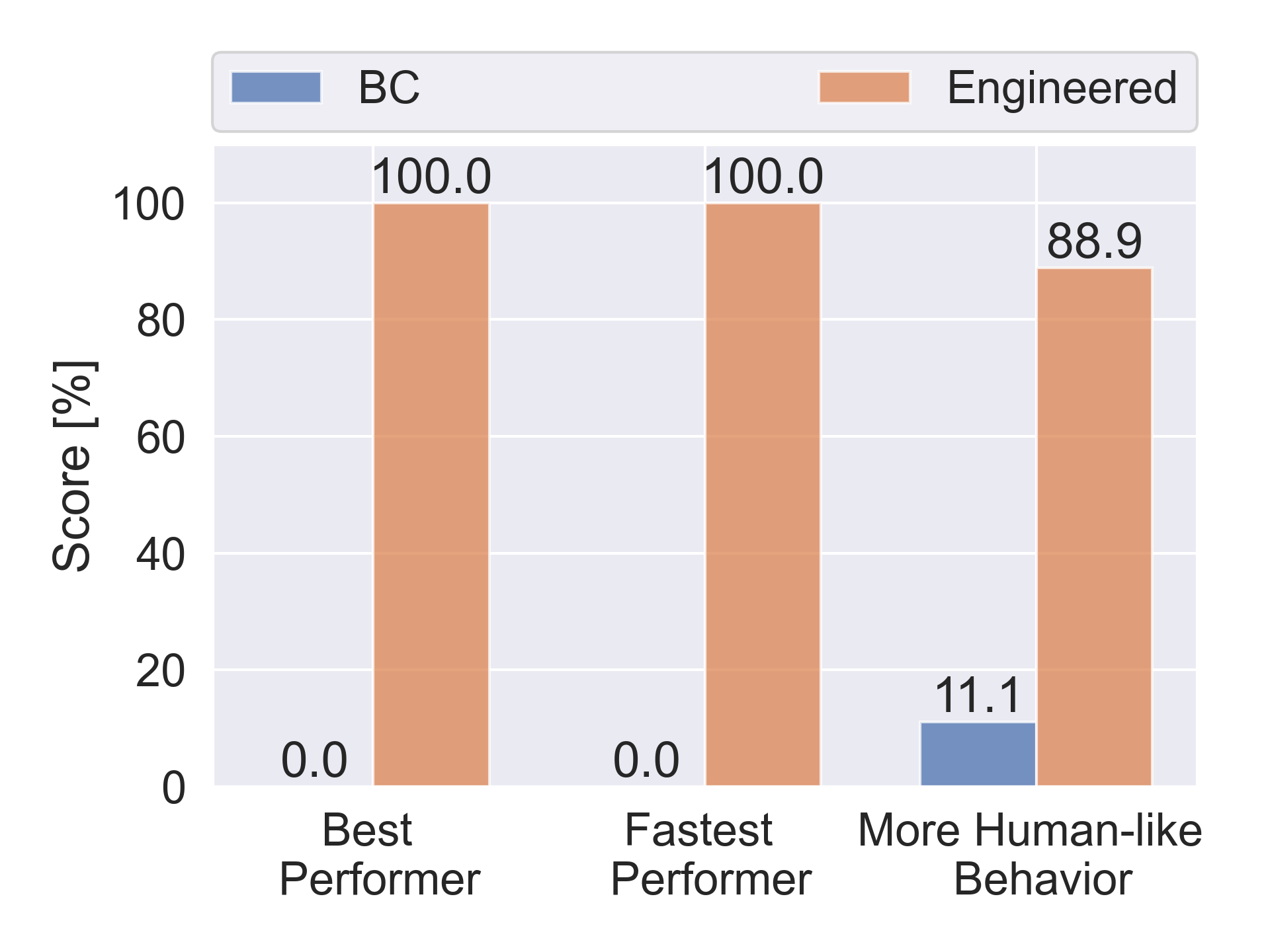}} &
        \subfloat[BC vs Human]{\includegraphics[width=0.3\linewidth]{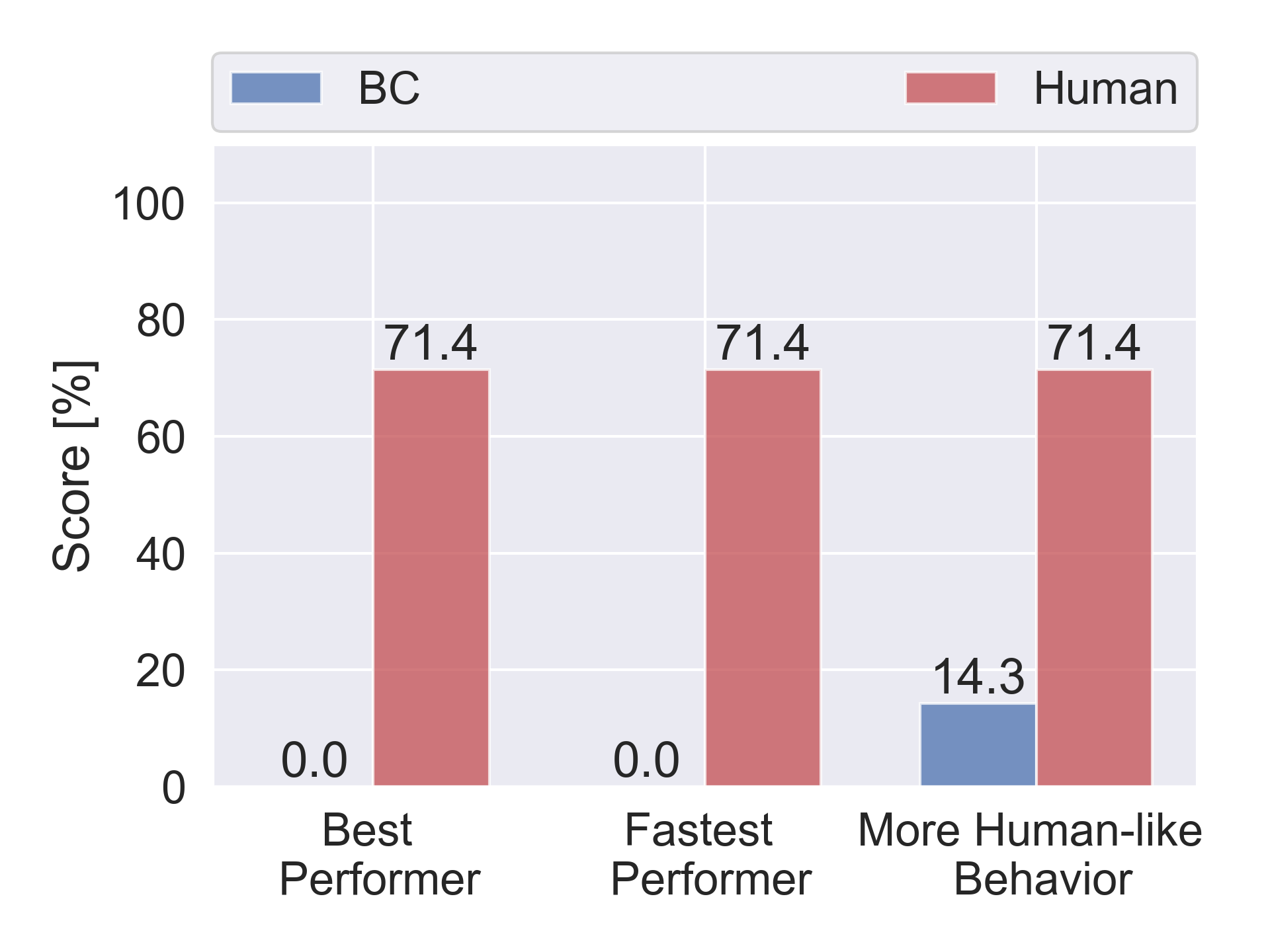}} &
        \subfloat[BC vs Hybrid]{\includegraphics[width=0.3\linewidth]{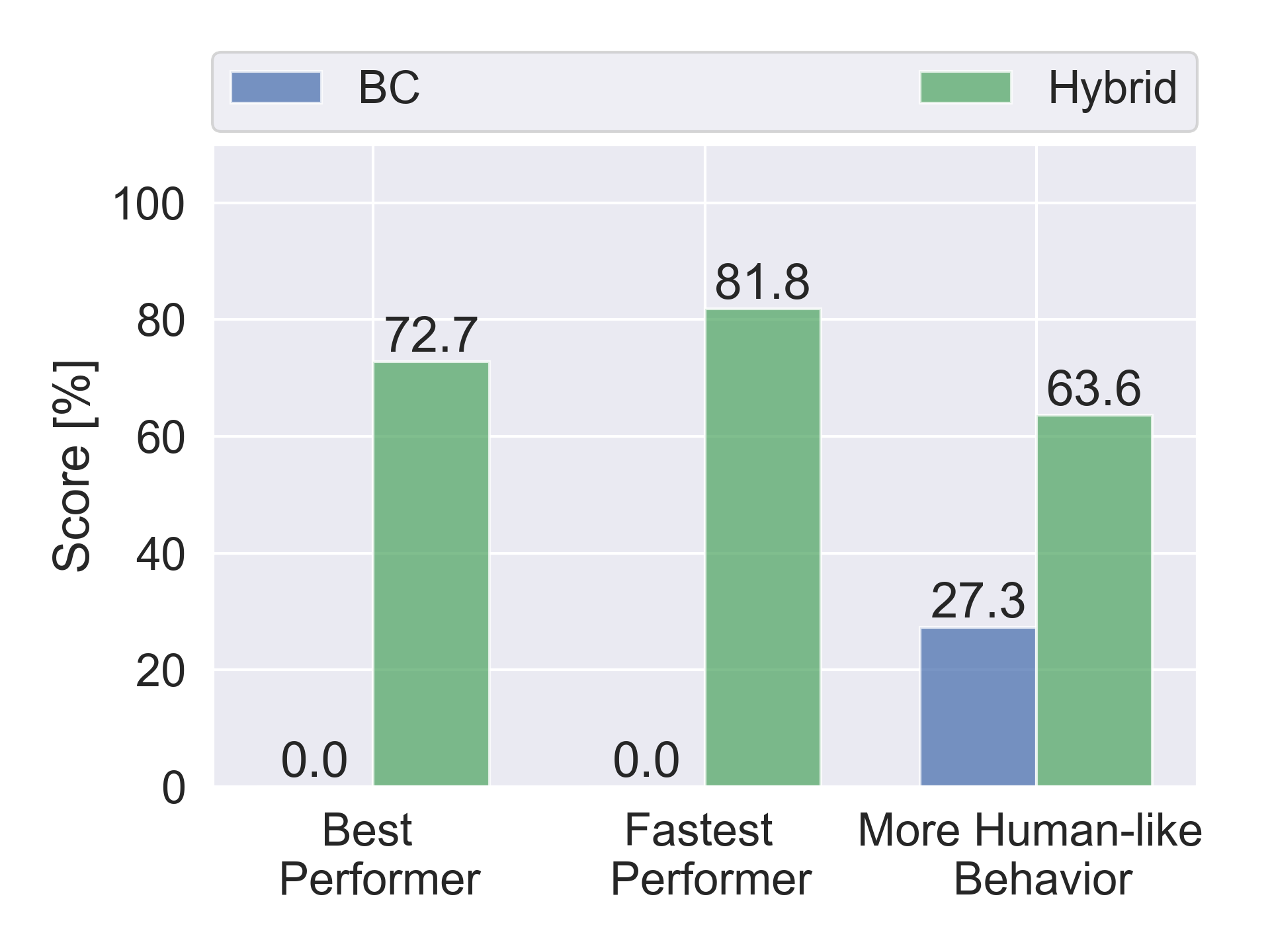}}\\
        \subfloat[Engineered vs Hybrid]{\includegraphics[width=0.3\linewidth]{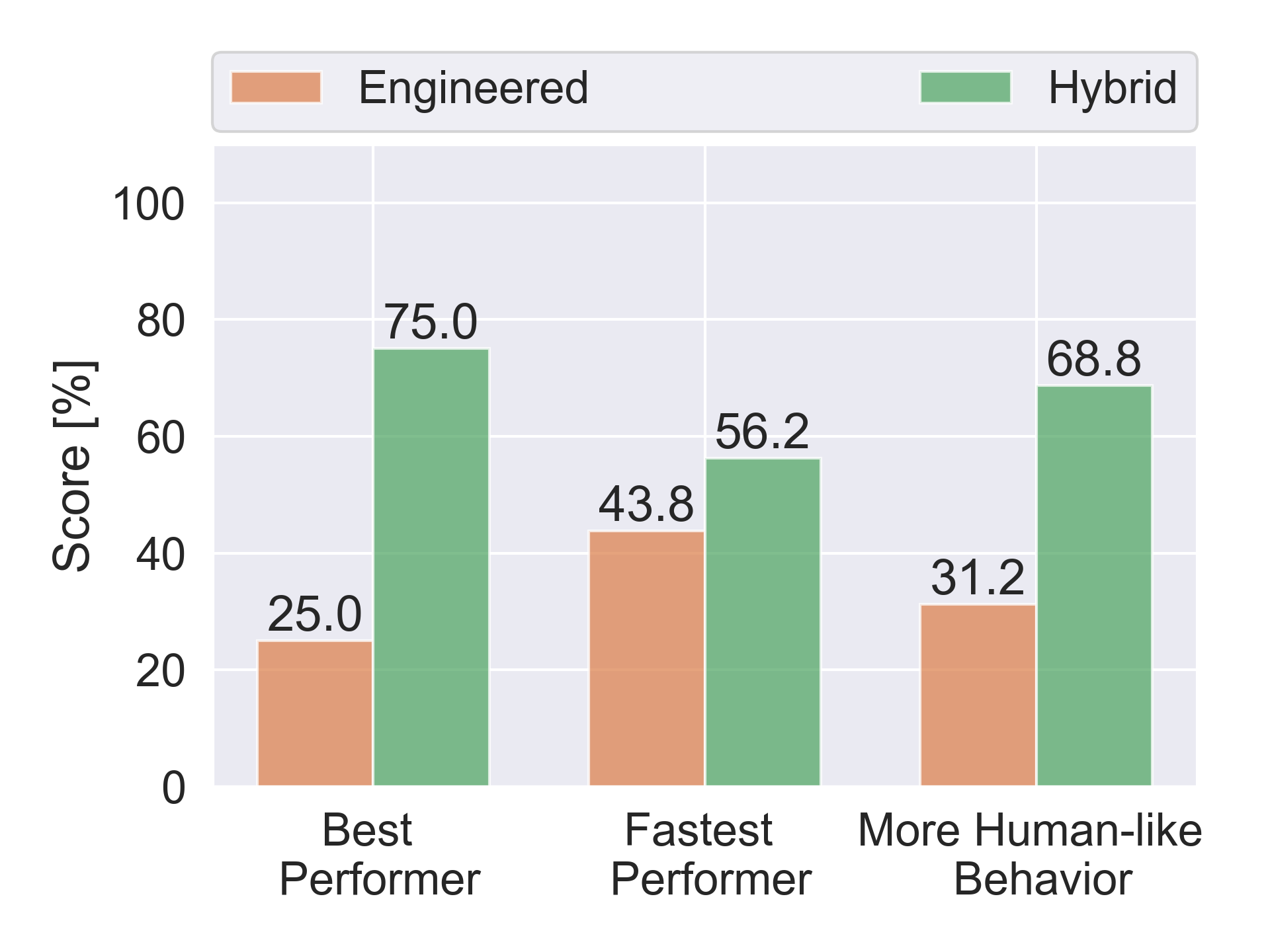}} &
        \subfloat[Human vs Engineered]{\includegraphics[width=0.3\linewidth]{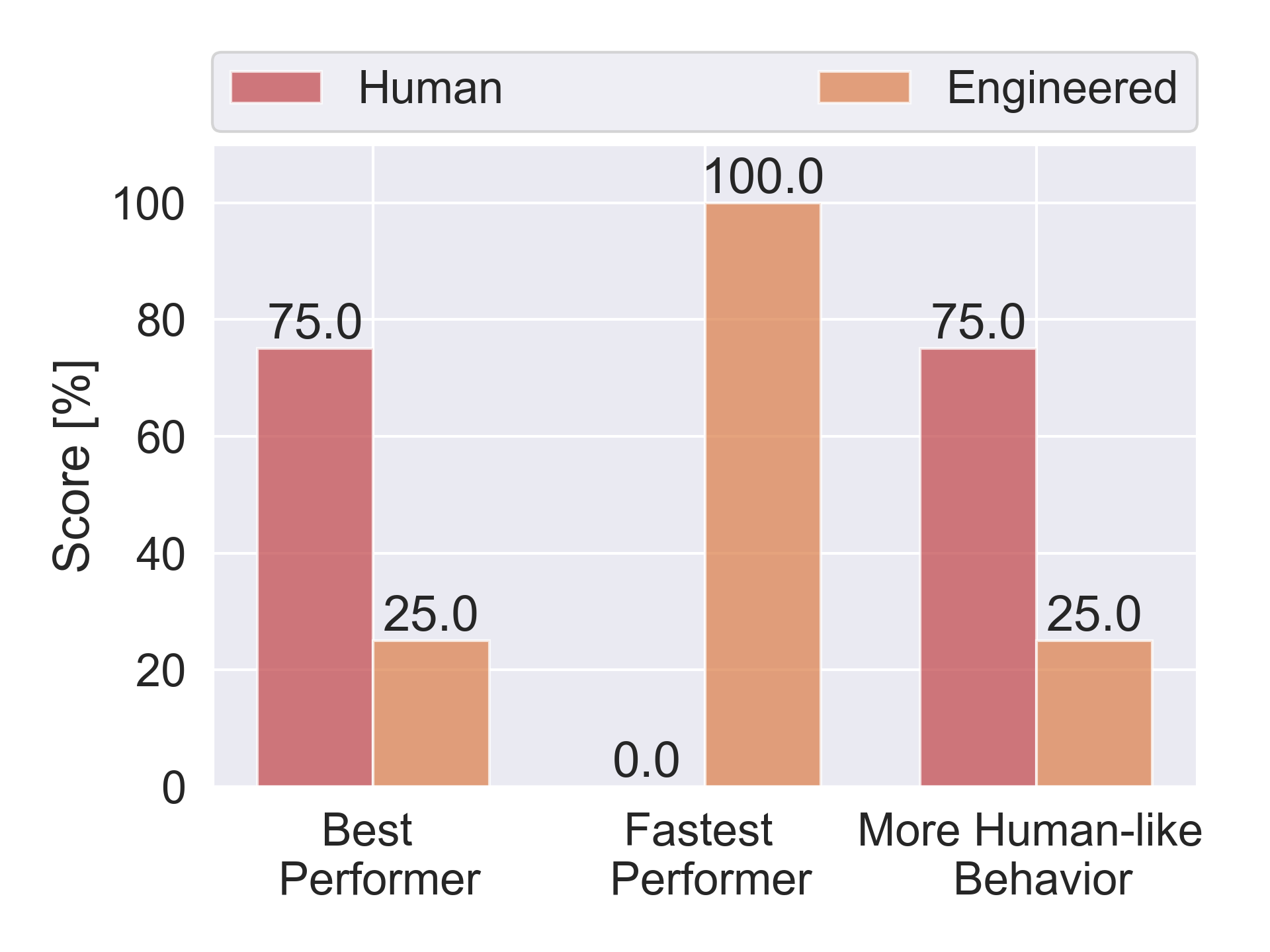}} &
        \subfloat[Human vs Hybrid]{\includegraphics[width=0.3\linewidth]{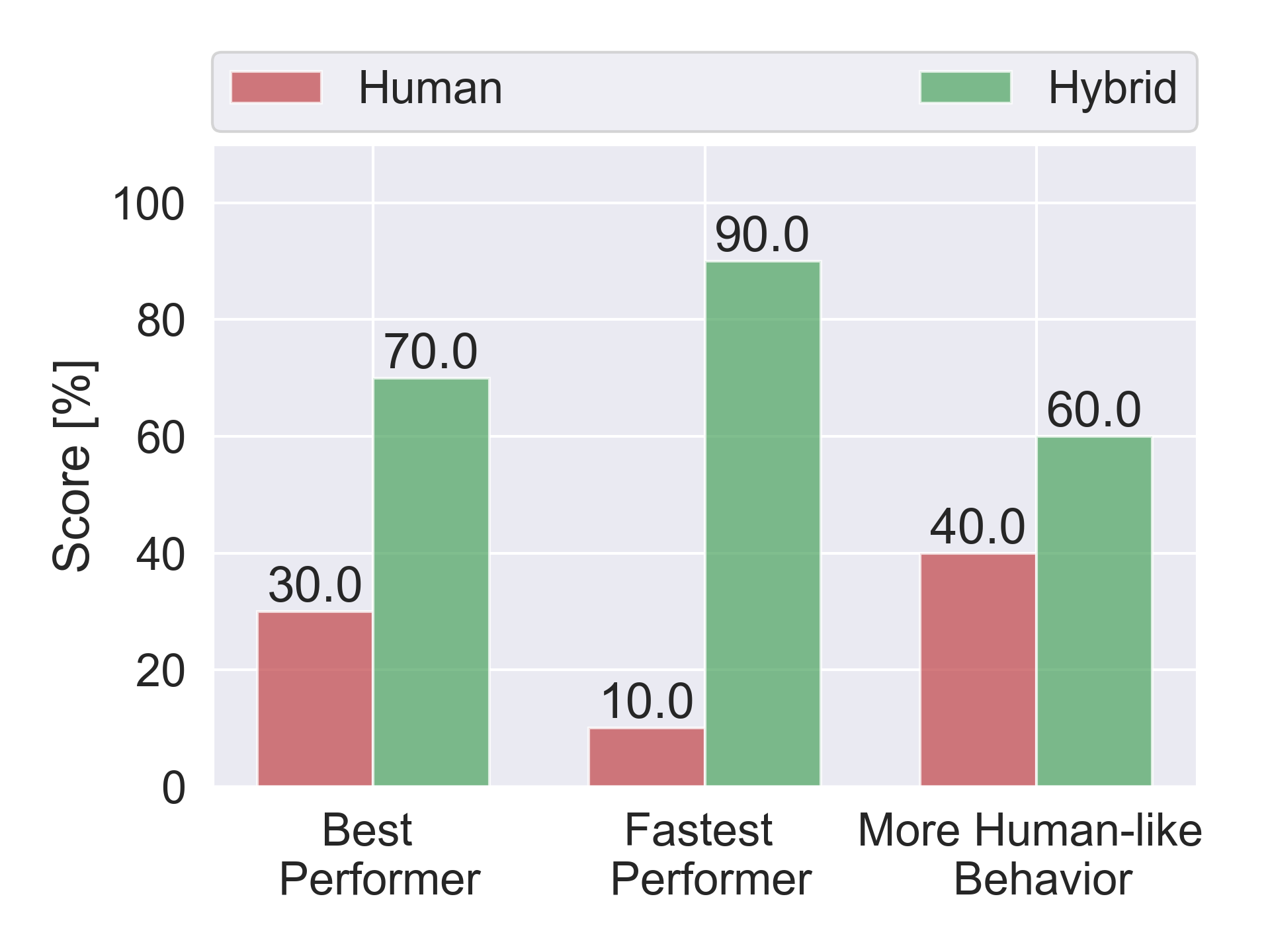}}
    \end{tabular}
  \caption{Pairwise comparison displaying the normalized scores computed from human evaluations separately for each performance metric on all possible head-to-head comparisons for all agent type performing the \textit{MakeWaterfall} task.}
  \label{fig:waterfall_barplot}
\end{figure}

\begin{figure}[!ht]
  \centering
    \begin{tabular}{ccc}
        \subfloat[BC vs Engineered]{\includegraphics[width=0.3\linewidth]{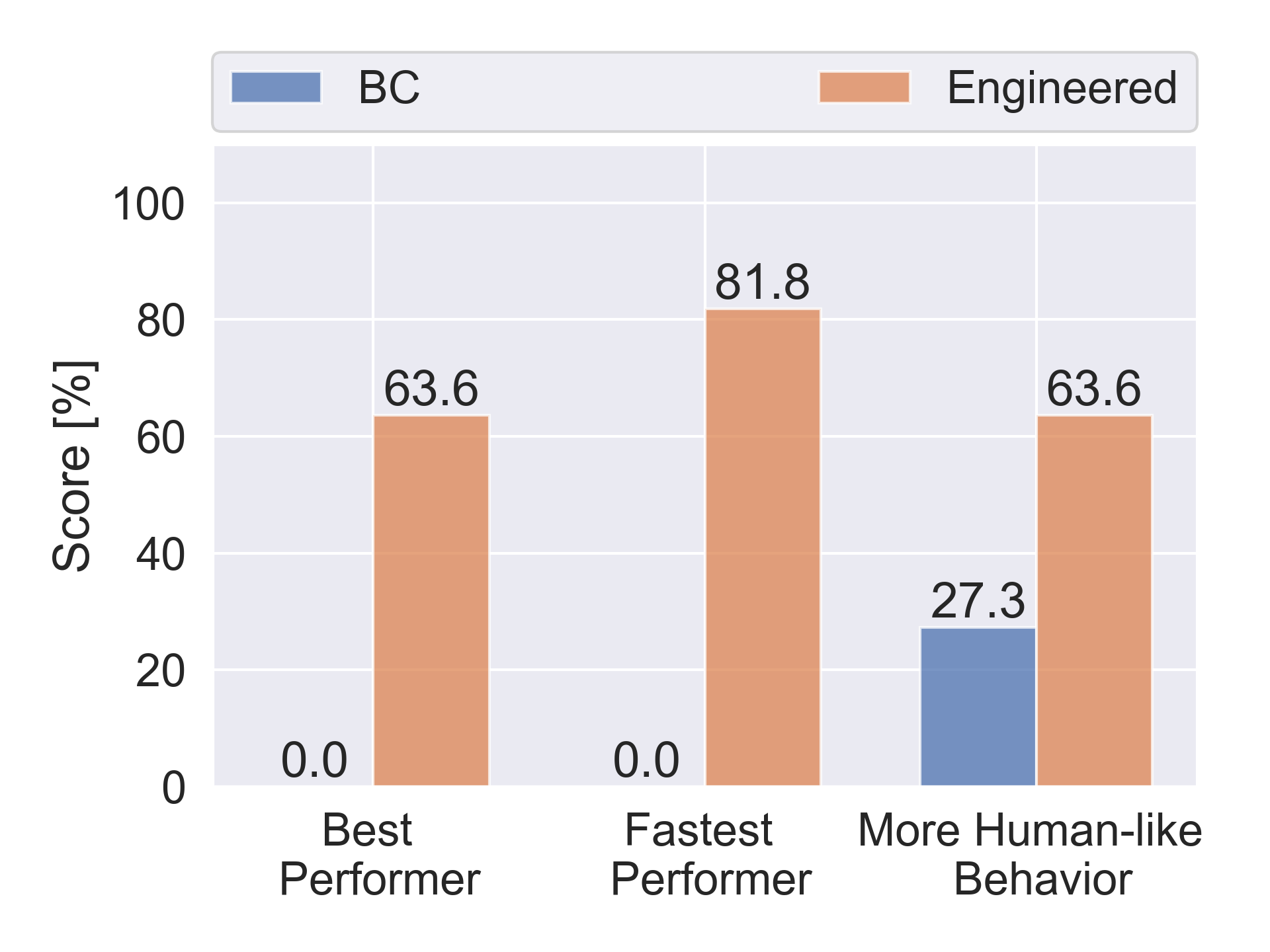}} &
        \subfloat[BC vs Human]{\includegraphics[width=0.3\linewidth]{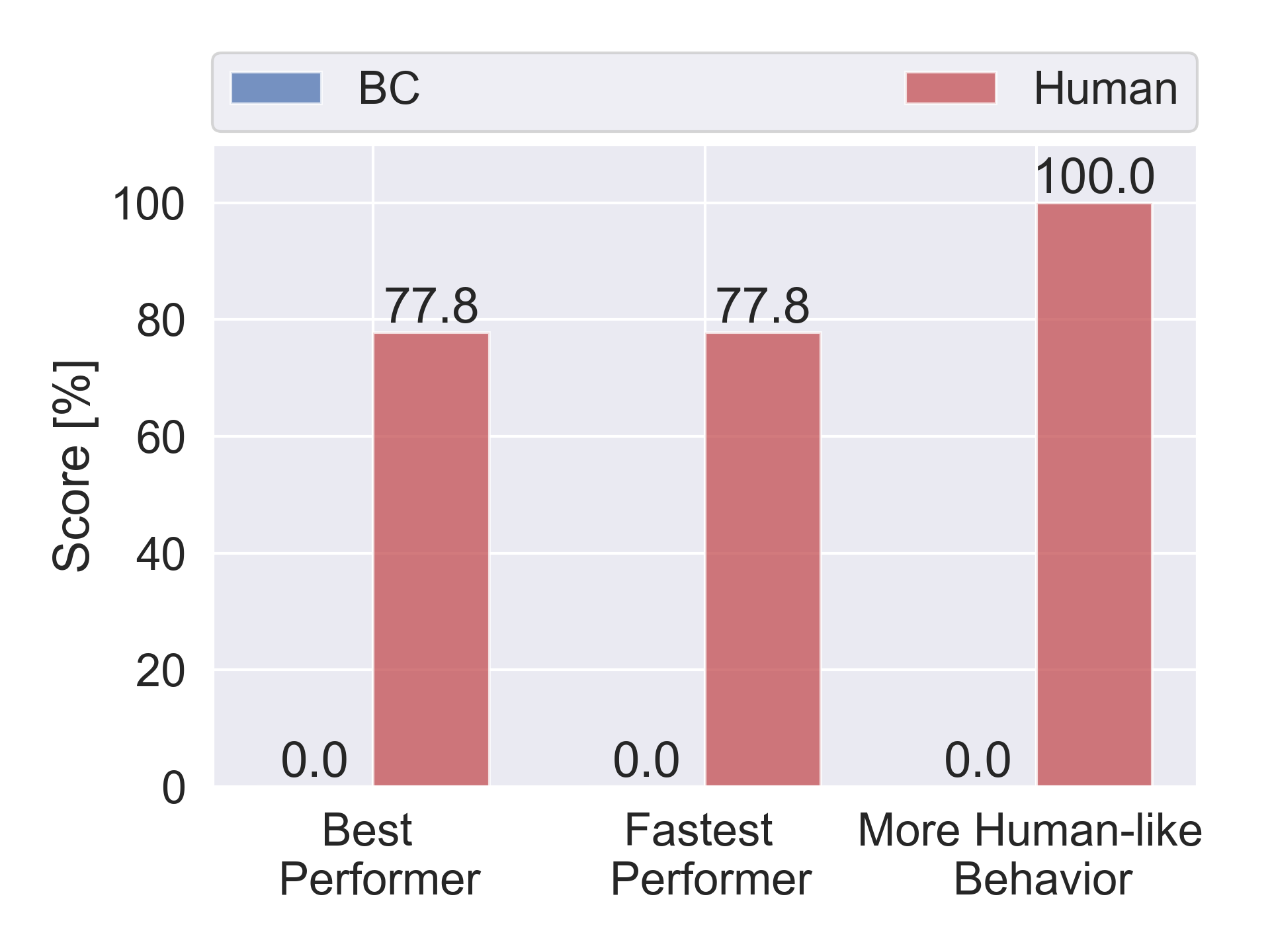}} &
        \subfloat[BC vs Hybrid]{\includegraphics[width=0.3\linewidth]{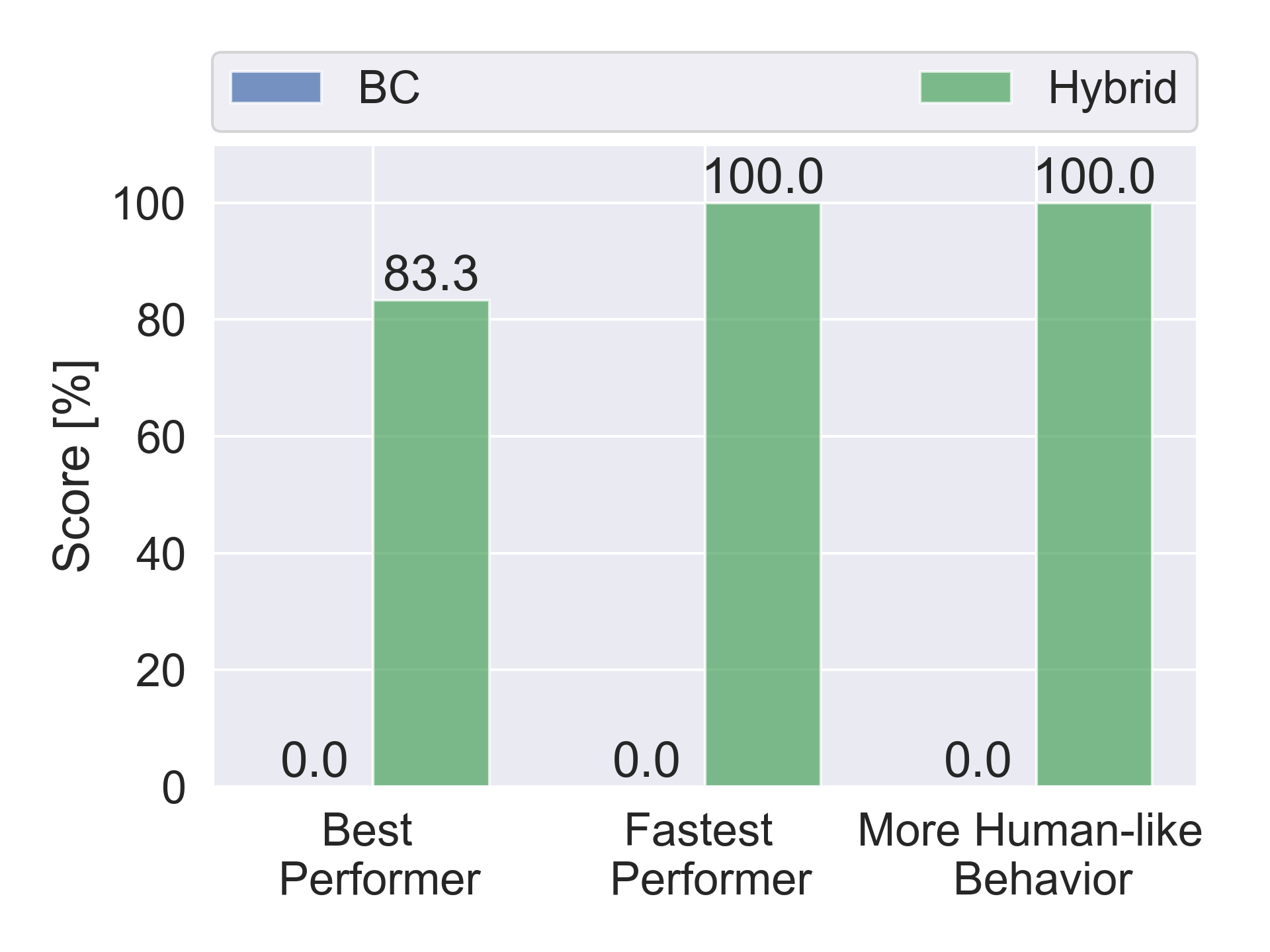}}\\
        \subfloat[Engineered vs Hybrid]{\includegraphics[width=0.3\linewidth]{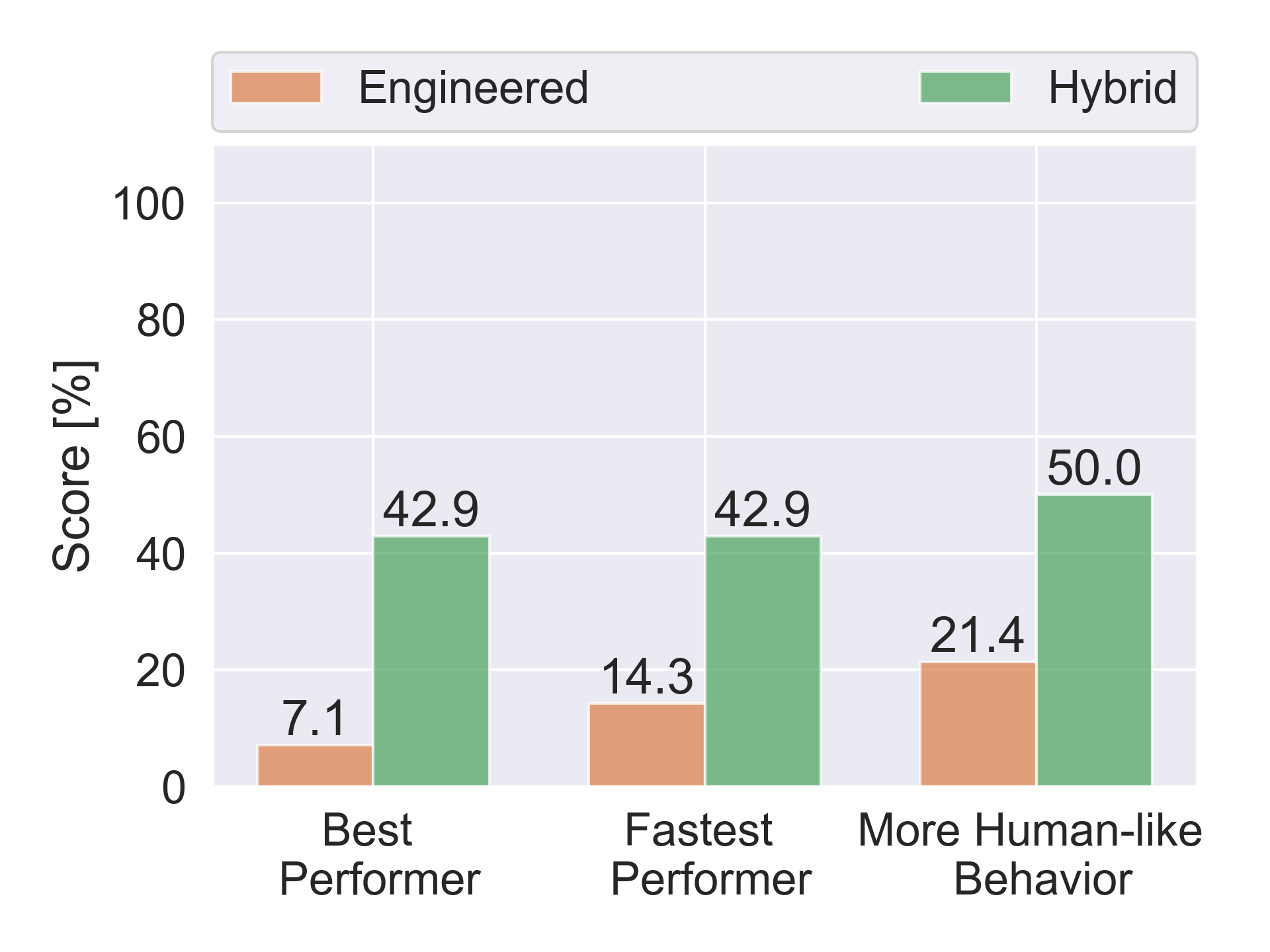}} &
        \subfloat[Human vs Engineered]{\includegraphics[width=0.3\linewidth]{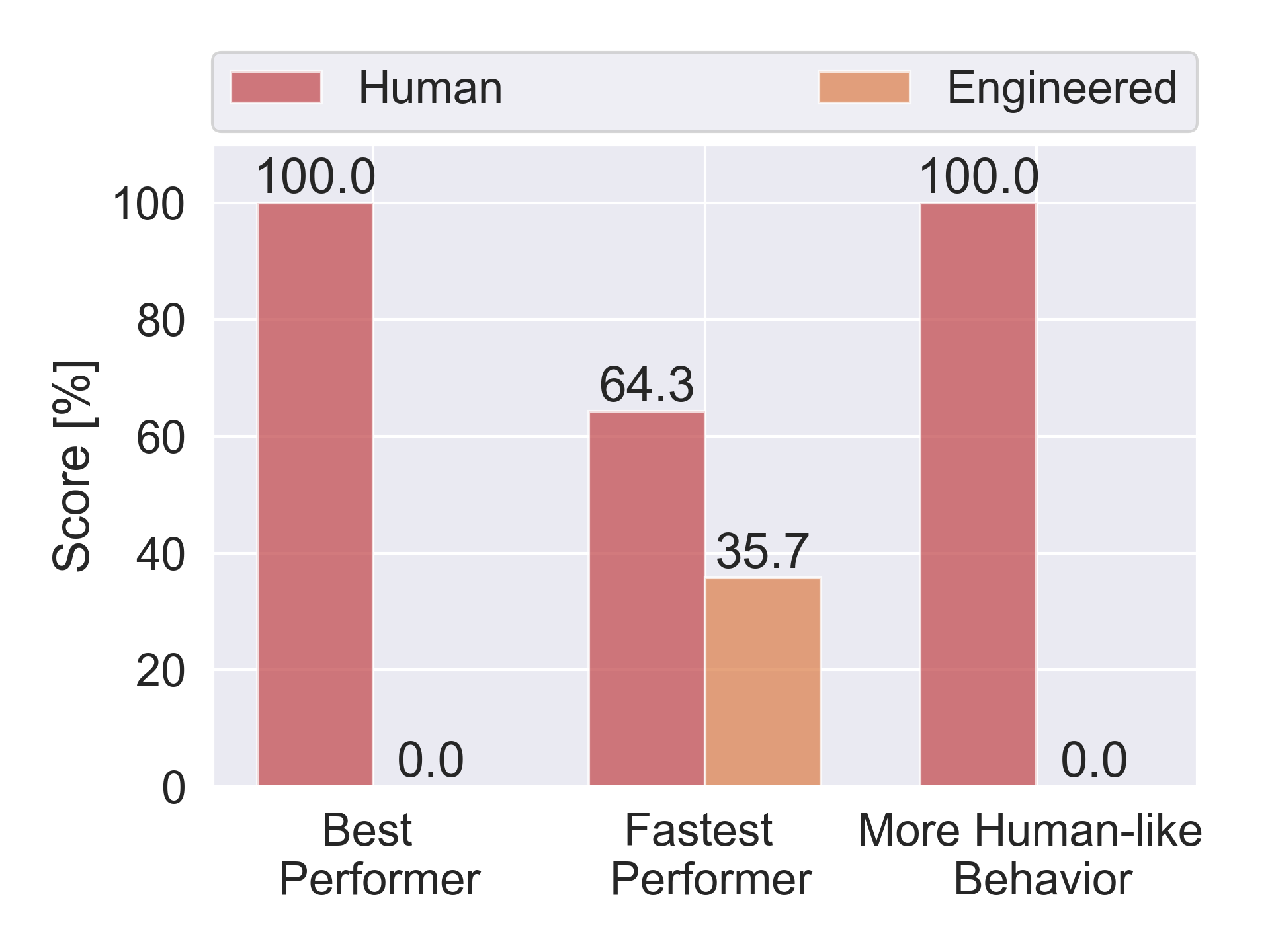}} &
        \subfloat[Human vs Hybrid]{\includegraphics[width=0.3\linewidth]{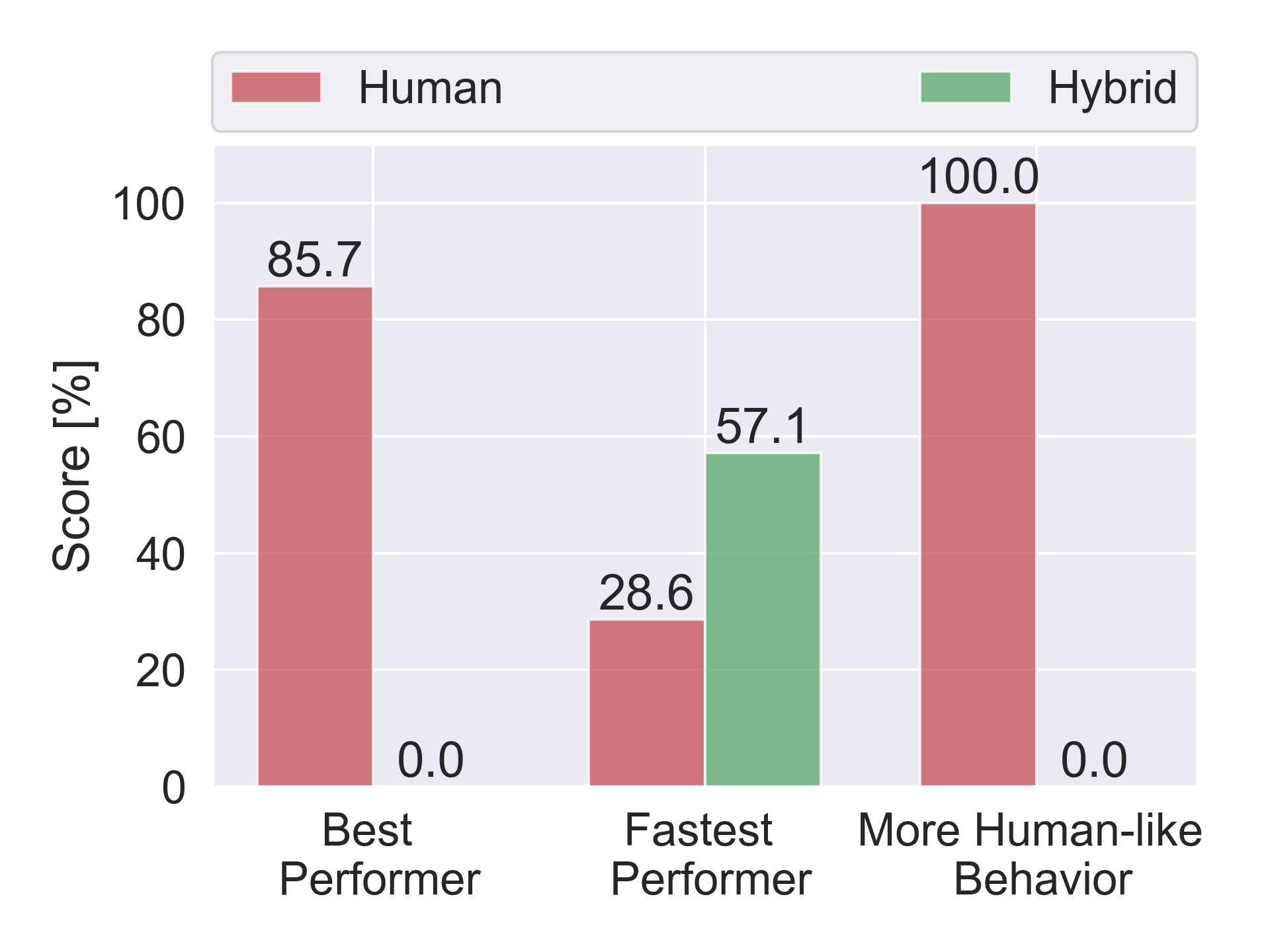}}
    \end{tabular}
  \caption{Pairwise comparison displaying the normalized scores computed from human evaluations separately for each performance metric on all possible head-to-head comparisons for all agent type performing the \textit{CreateVillageAnimalPen} task.}
  \label{fig:pen_barplot}
\end{figure}

\begin{figure}[!ht]
  \centering
    \begin{tabular}{ccc}
        \subfloat[BC vs Engineered]{\includegraphics[width=0.3\linewidth]{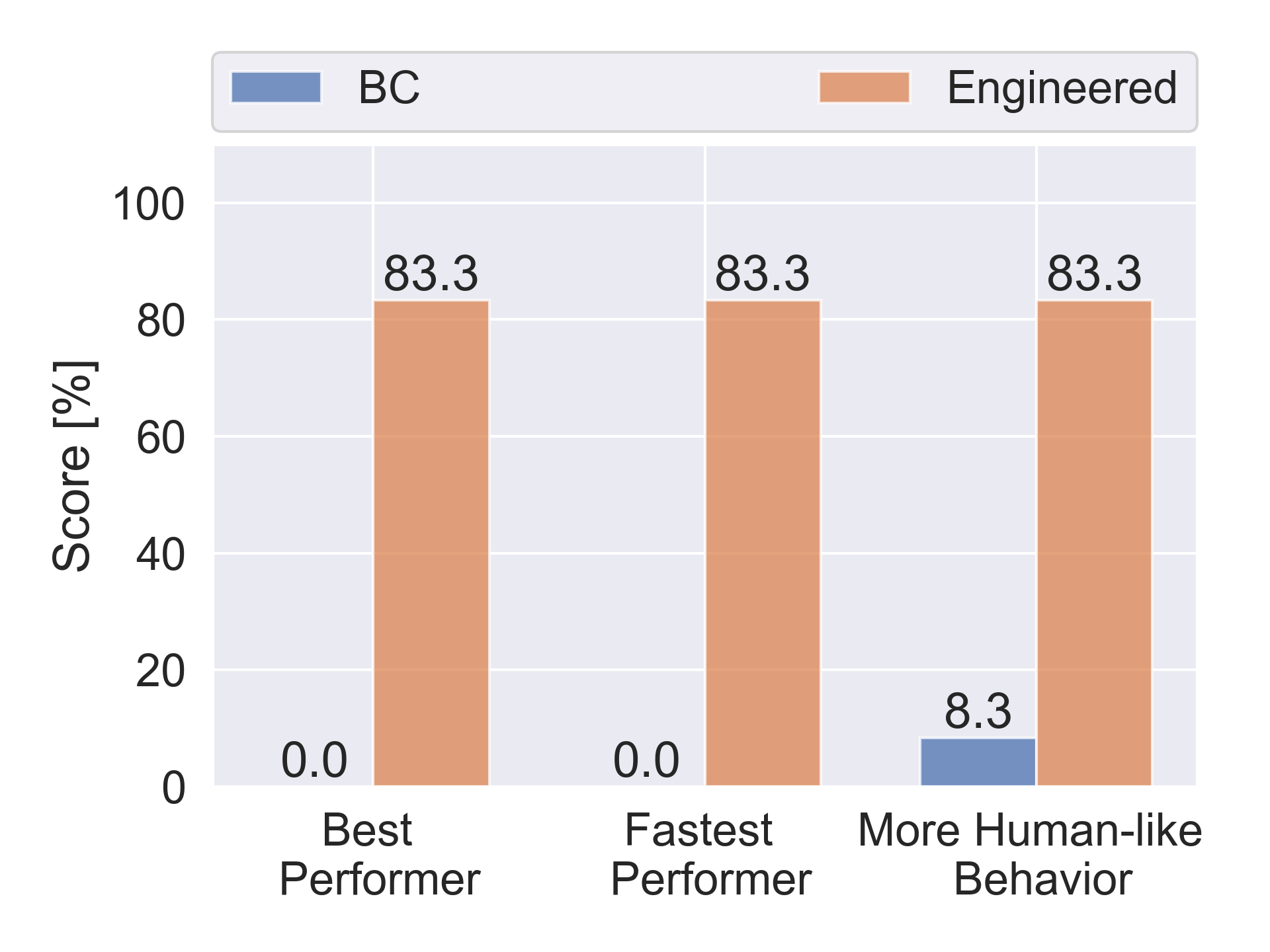}} &
        \subfloat[BC vs Human]{\includegraphics[width=0.3\linewidth]{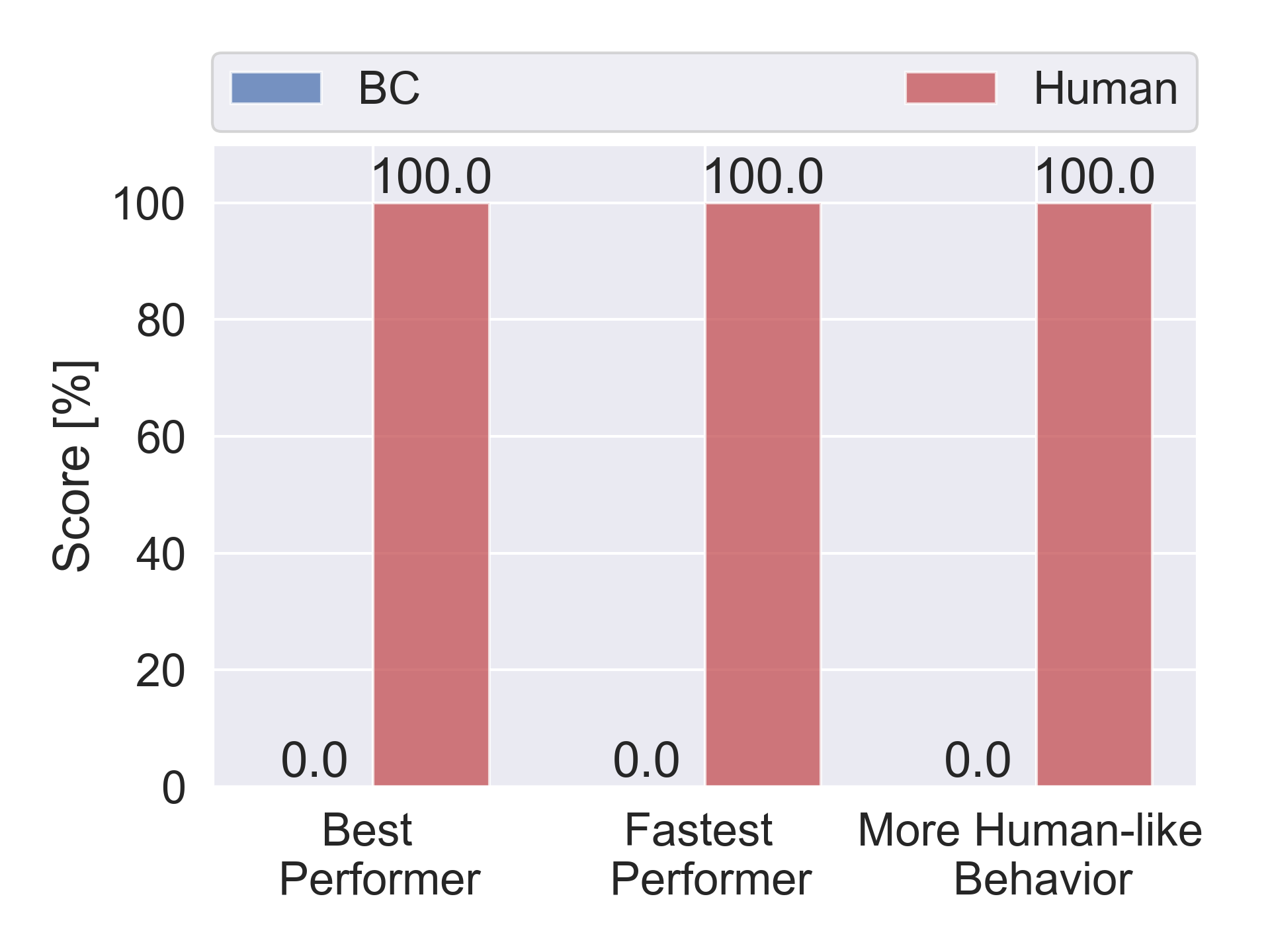}} &
        \subfloat[BC vs Hybrid]{\includegraphics[width=0.3\linewidth]{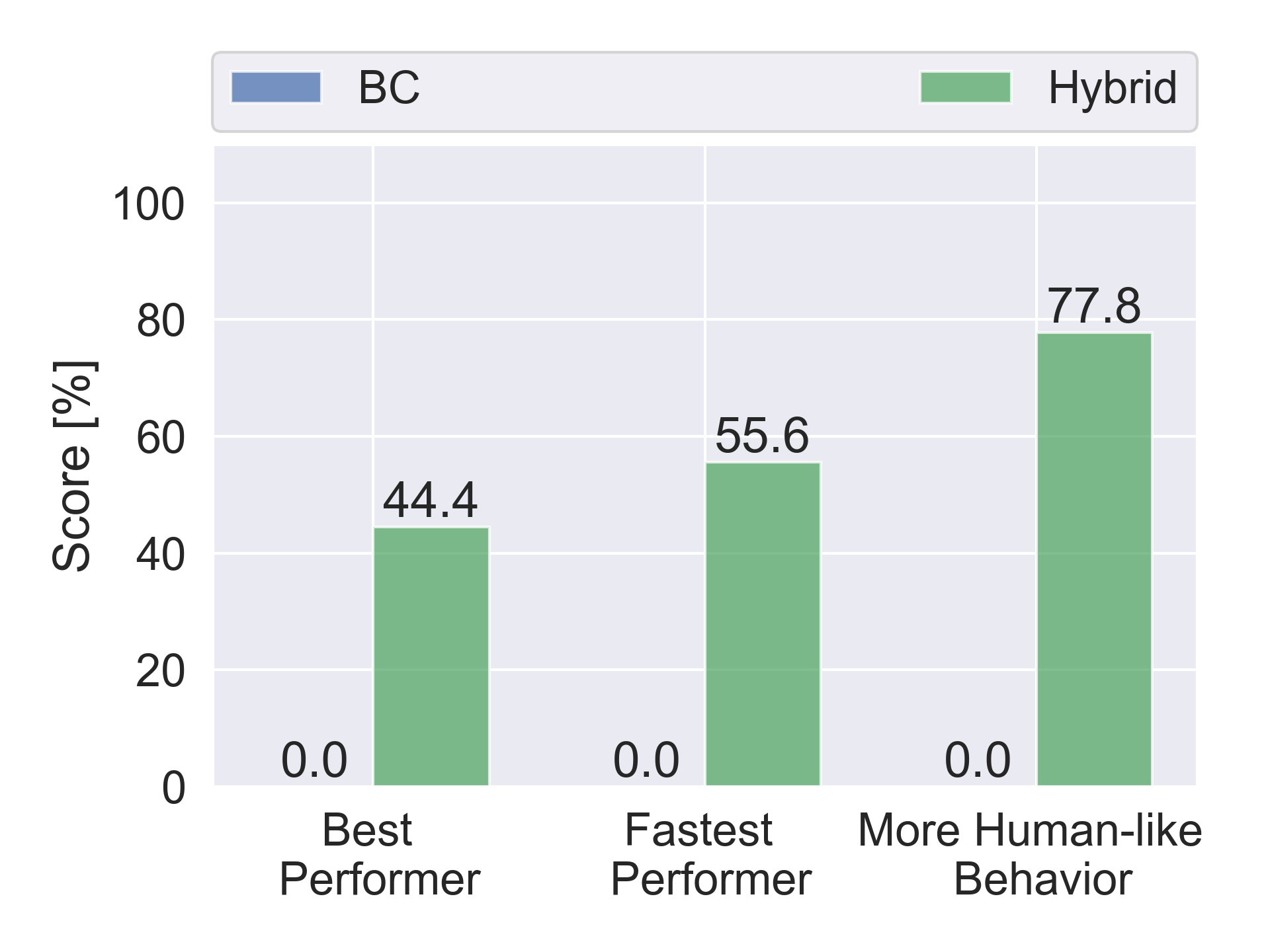}}\\
        \subfloat[Engineered vs Hybrid]{\includegraphics[width=0.3\linewidth]{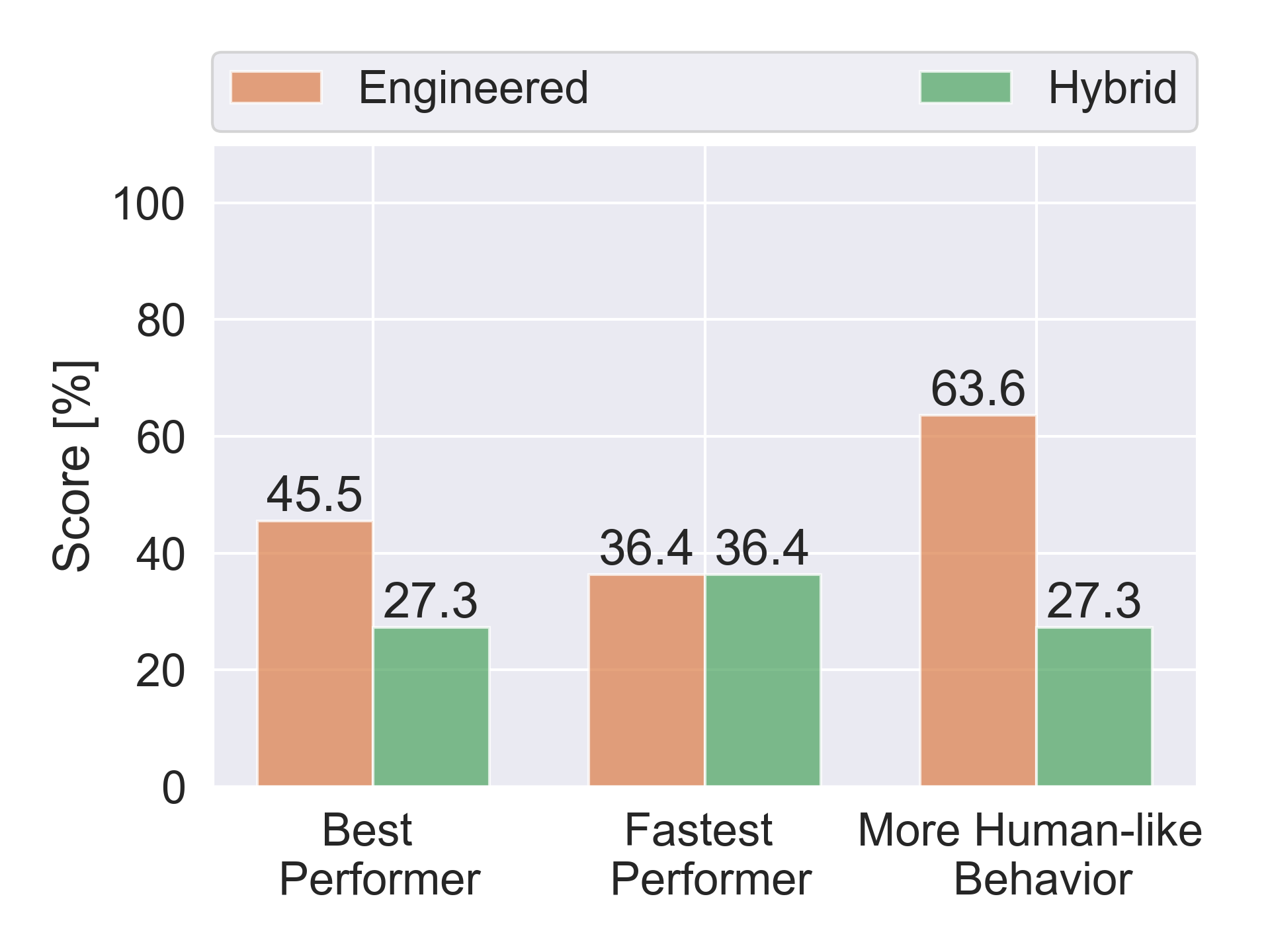}} &
        \subfloat[Human vs Engineered]{\includegraphics[width=0.3\linewidth]{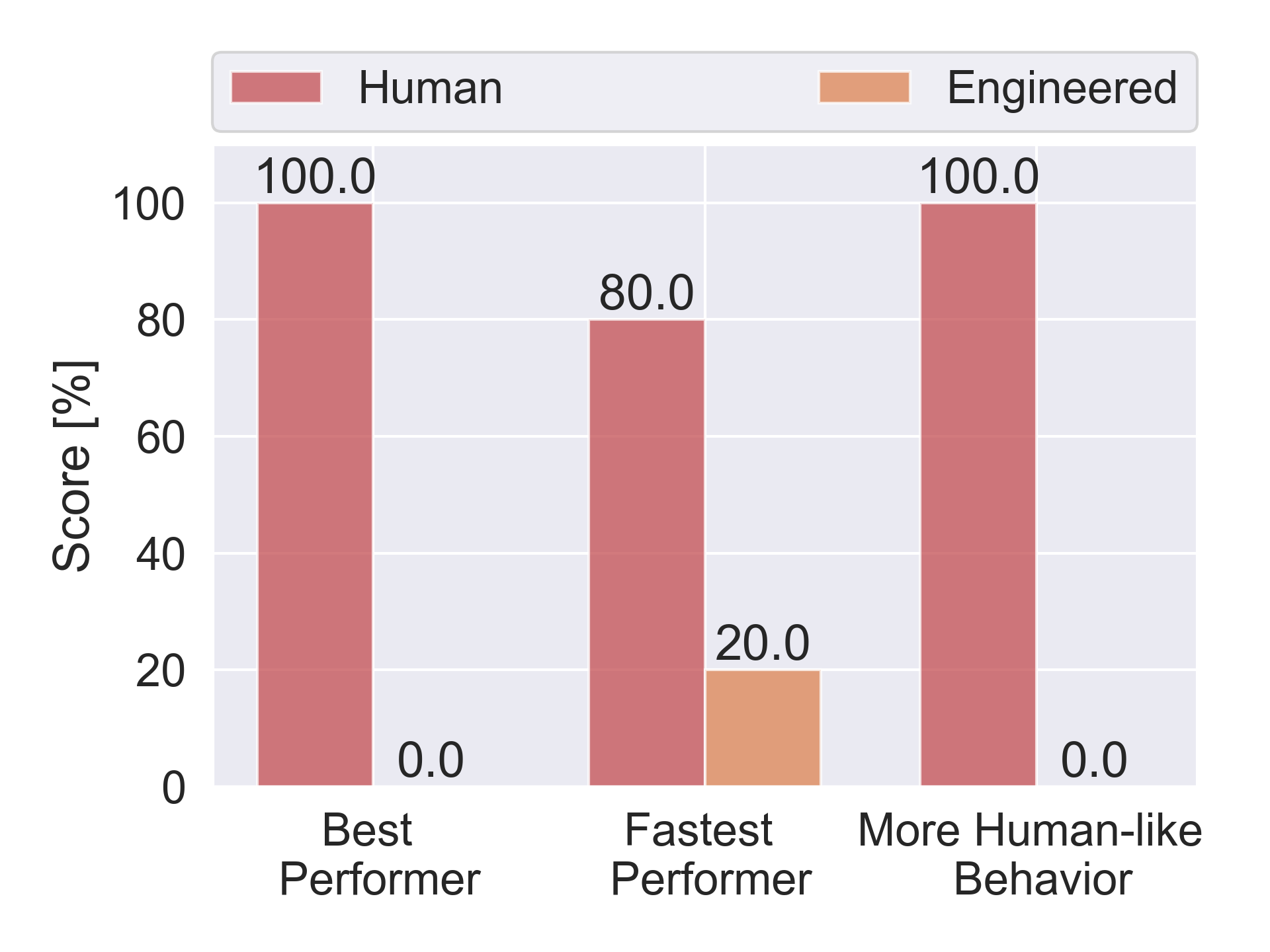}} &
        \subfloat[Human vs Hybrid]{\includegraphics[width=0.3\linewidth]{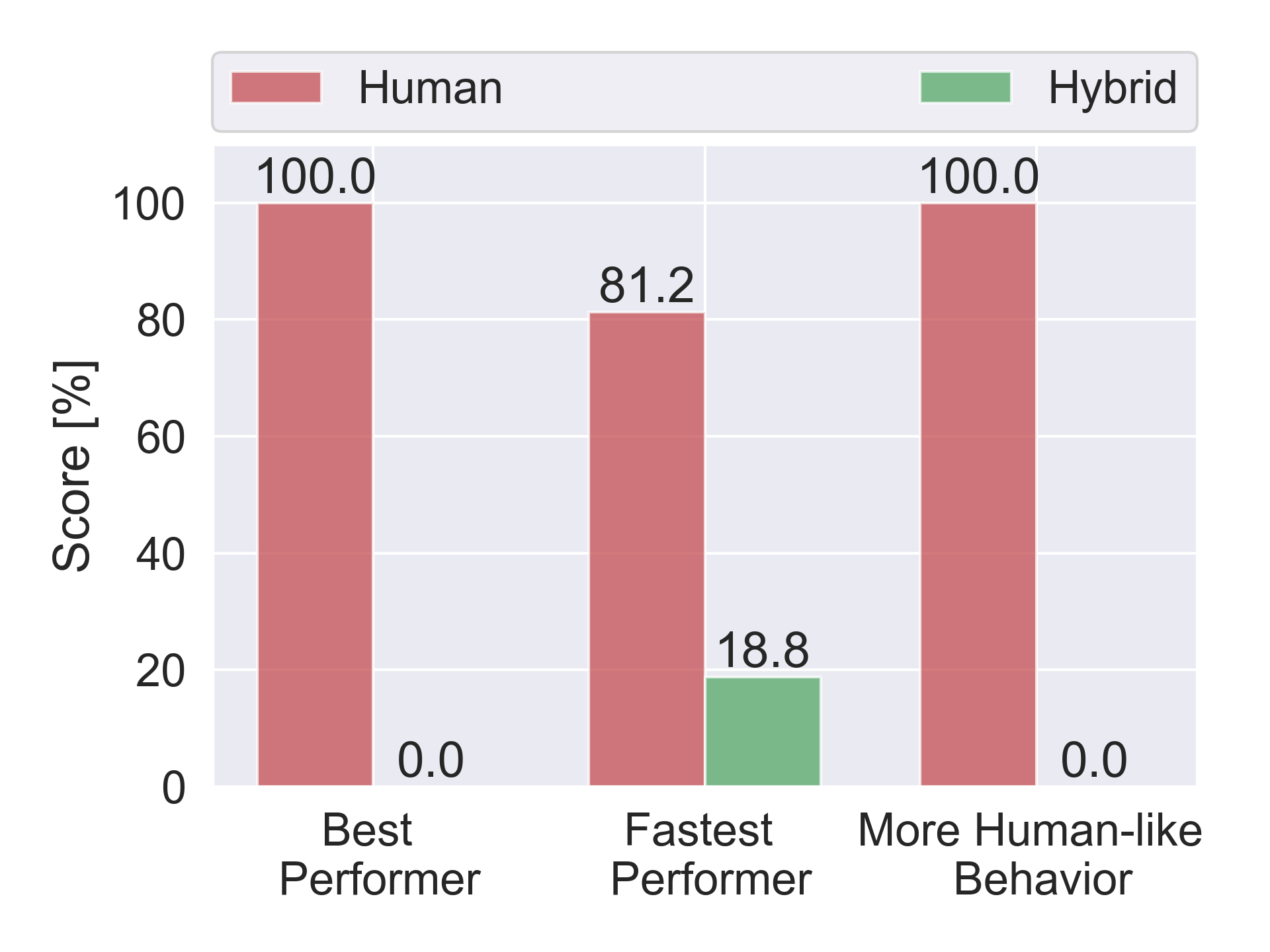}}
    \end{tabular}
  \caption{Pairwise comparison displaying the normalized scores computed from human evaluations separately for each performance metric on all possible head-to-head comparisons for all agent type performing the \textit{BuildVillageHouse} task.}
  \label{fig:house_barplot}
\end{figure}

Figures~\ref{fig:cave_barplot},~\ref{fig:waterfall_barplot},~\ref{fig:pen_barplot}, and~\ref{fig:house_barplot} show bar plots with the individual pairwise comparisons compiled from the human evaluations for the \textit{FindCave}, \textit{MakeWaterfall}, \textit{CreateVillageAnimalPen}, and \textit{BuildVillageHouse} tasks, respectively. Each bar represents the percentage of the time a given condition was selected as a winner for each performance metric by the human evaluator when they were presented with a video of the agent performance solving the task for each analyzed condition. For example, when analyzing Figure~\ref{fig:cave_barplot}(a), the human evaluator was presented with a video of the ``Behavior Cloning'' agent and another from the ``Engineered'' agent, they selected the ``Engineered'' agent as the best performer $33.3 \%$ and the ``Behavior Cloning'' agent $22.2 \%$ of the time. The remaining accounts for the ``None'' answer to the questionnaire selected when none of the agents were judged to have solved the task.

When directly comparing the ``Behavior Cloning'' baseline to the main proposed ``Hybrid'' method for all tasks, as shown in Figures~\ref{fig:cave_barplot},~\ref{fig:waterfall_barplot},~\ref{fig:pen_barplot}, and~\ref{fig:house_barplot} (c) plots, the proposed hybrid intelligence agent always matches or outperforms the pure learned baseline. This is like the case comparing the ``Engineered'' agent to the ``Hybrid'' agent, where the proposed hybrid method outperforms the fully engineered approach in all tasks except the \textit{BuildVillageHouse} task, as seen in Figure~\ref{fig:house_barplot}. The human players always outperform the hybrid agent with exception to the \textit{MakeWaterfall} task, where the ``Hybrid'' agent is judged to better solve the task $70 \%$ of the time, to solve it faster $90 \%$ of the time, and even present a more human-like behavior $60 \%$ of the time. The ``Hybrid'' agent performing better can be attributed to the fact that the human players were not always able or willing to solve the task as described in the prompt.

\section{Samples of Hybrid Agent Solving the Tasks}\label{appendix:frames}

\begin{figure}[!ht]
    \centering
    \begin{tabular}{cc}
        \subfloat[]{\includegraphics[width=0.4\linewidth]{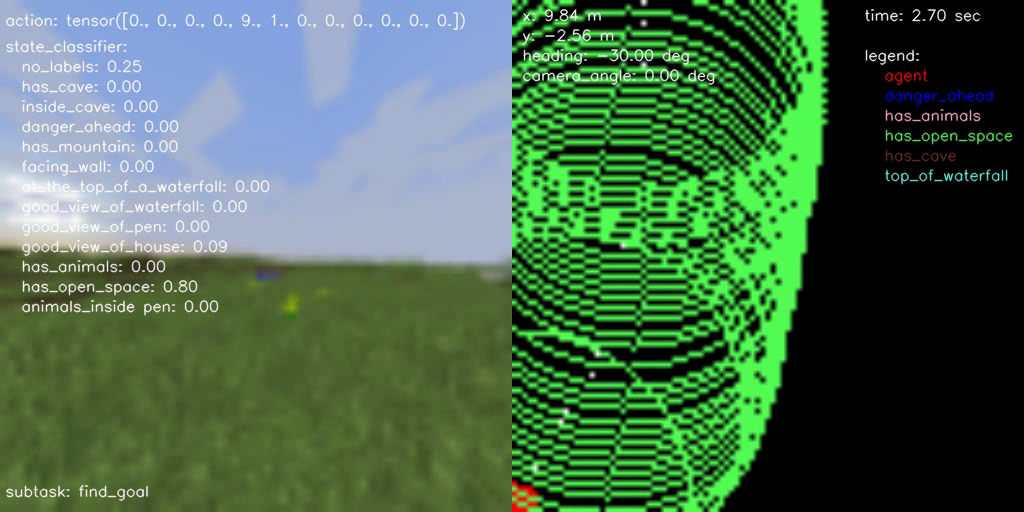}} &
        \subfloat[]{\includegraphics[width=0.4\linewidth]{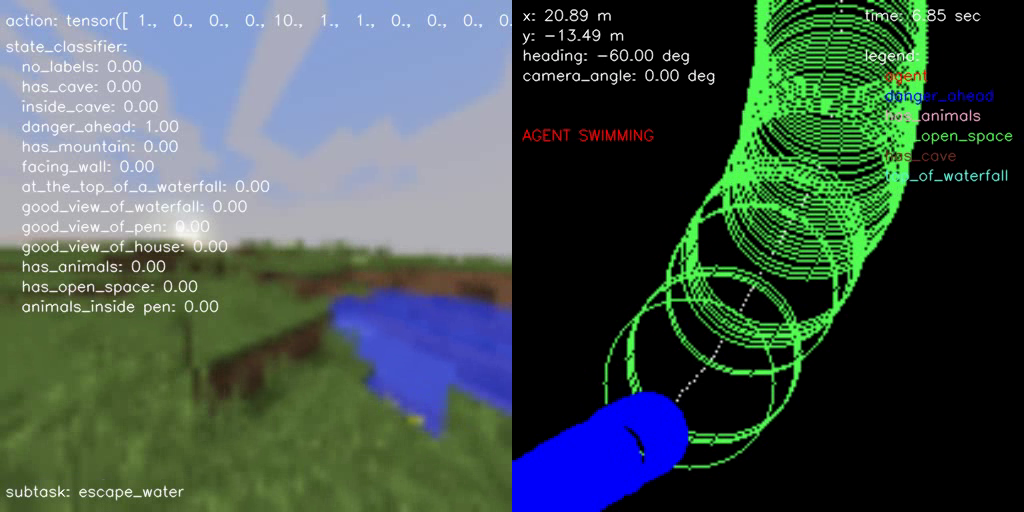}}\\
        \subfloat[]{\includegraphics[width=0.4\linewidth]{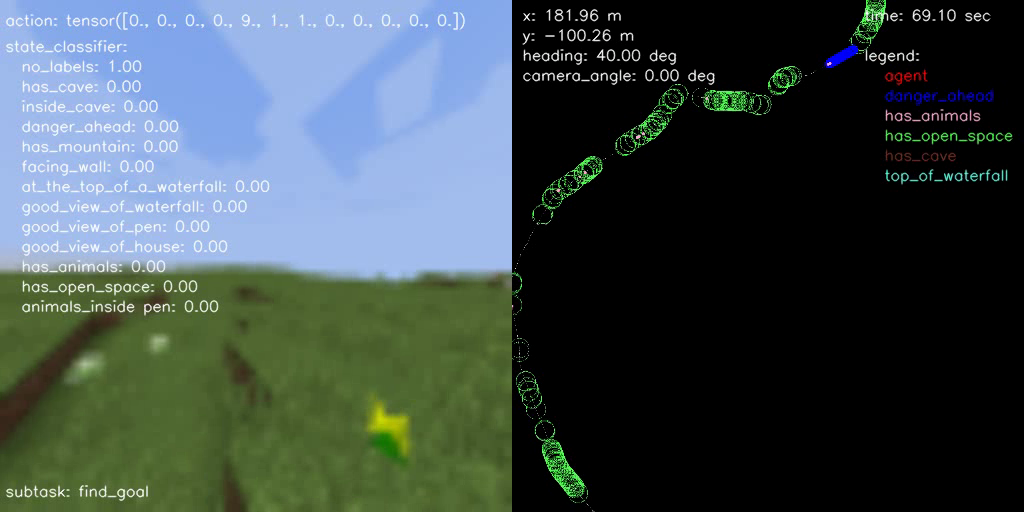}} &
        \subfloat[]{\includegraphics[width=0.4\linewidth]{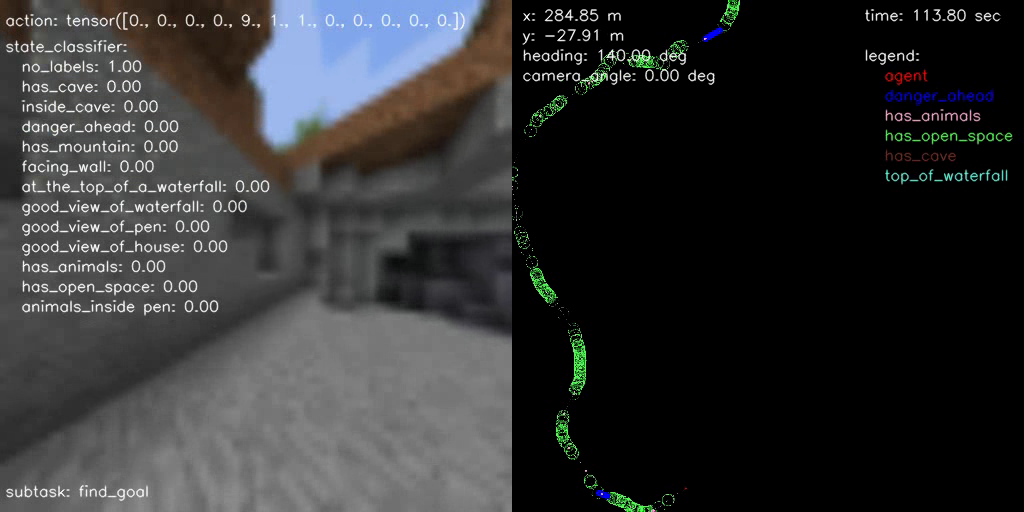}}
    \end{tabular}
    \caption{Sequence of frames of the hybrid agent solving the \textit{FindCave} task (complete video available at \url{https://youtu.be/MR8q3Xre_XY}).}
    \label{fig:bestcave_frames}
\end{figure}

\begin{figure}[!ht]
  \centering
    \begin{tabular}{cc}
        \subfloat[]{\includegraphics[width=0.4\linewidth]{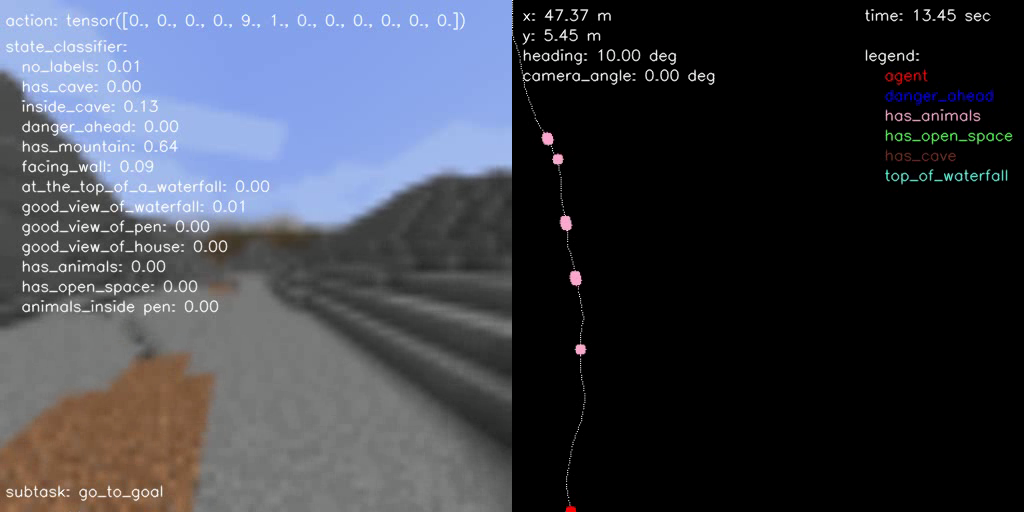}} &
        \subfloat[]{\includegraphics[width=0.4\linewidth]{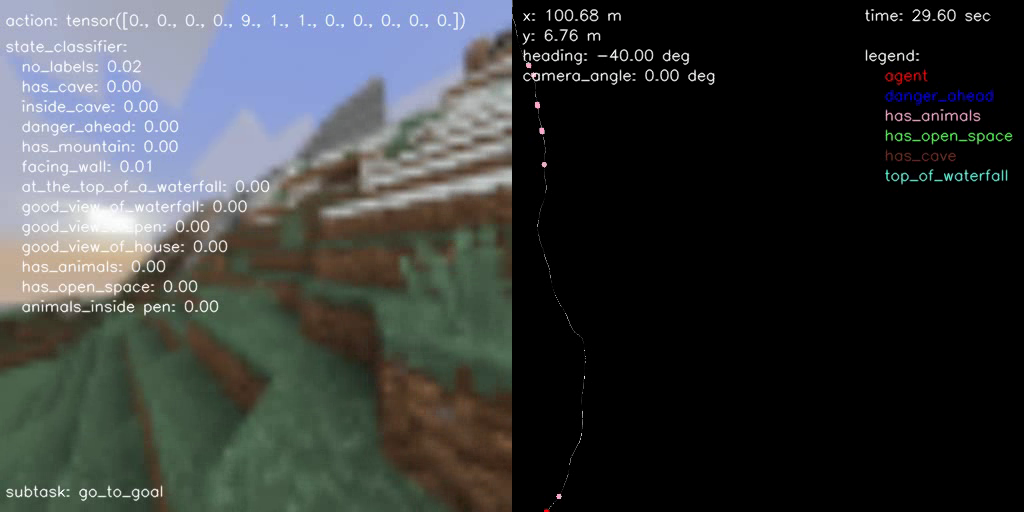}}\\
        \subfloat[]{\includegraphics[width=0.4\linewidth]{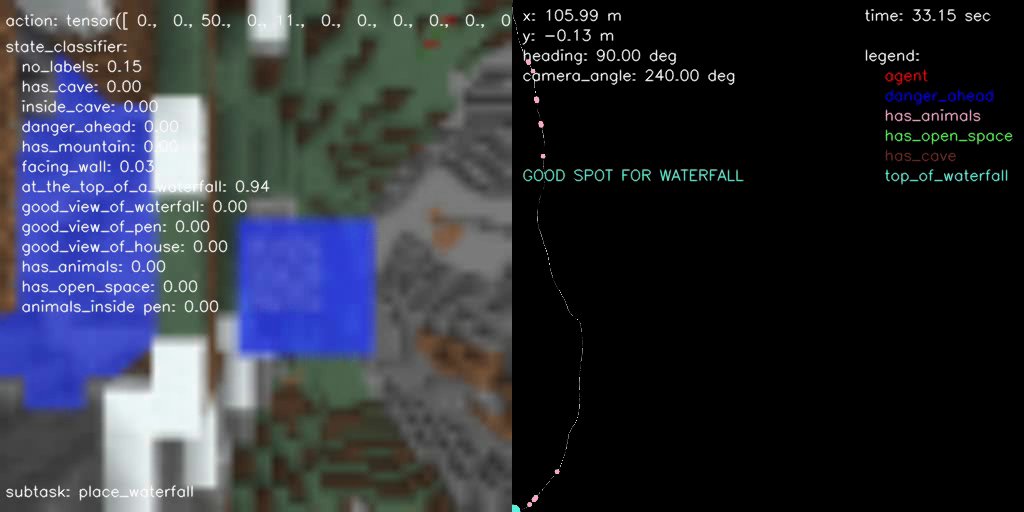}} &
        \subfloat[]{\includegraphics[width=0.4\linewidth]{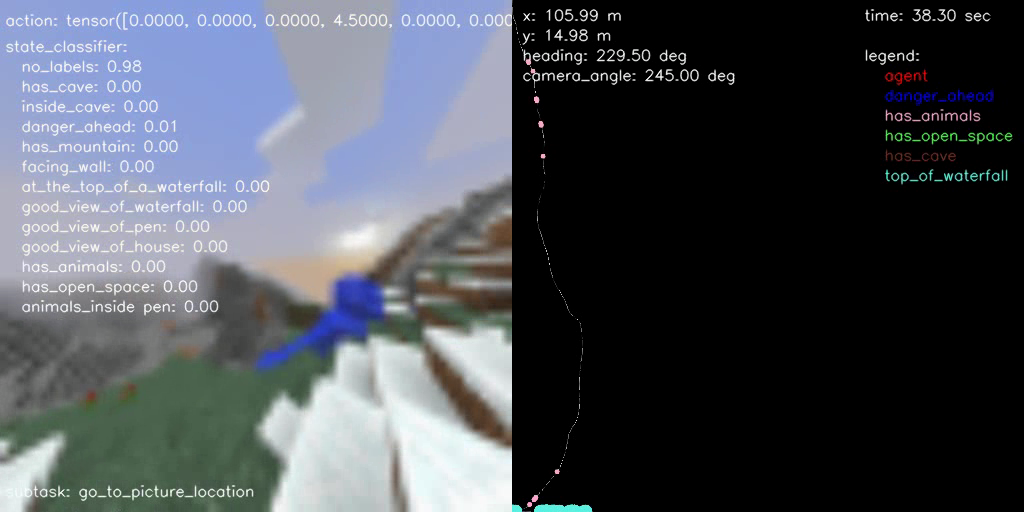}}
    \end{tabular}
  \caption{Sequence of frames of the hybrid agent solving the \textit{MakeWaterfall} task (complete video available at \url{https://youtu.be/eXp1urKXIPQ}).}
  \label{fig:bestwaterfall_frames}
\end{figure}

\begin{figure}[!ht]
  \centering
    \begin{tabular}{cc}
        \subfloat[]{\includegraphics[width=0.4\linewidth]{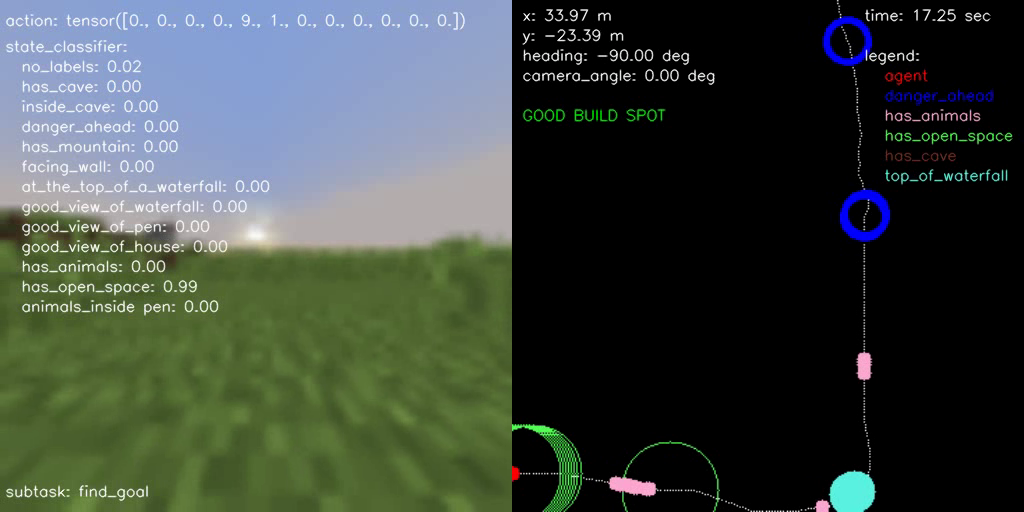}} &
        \subfloat[]{\includegraphics[width=0.4\linewidth]{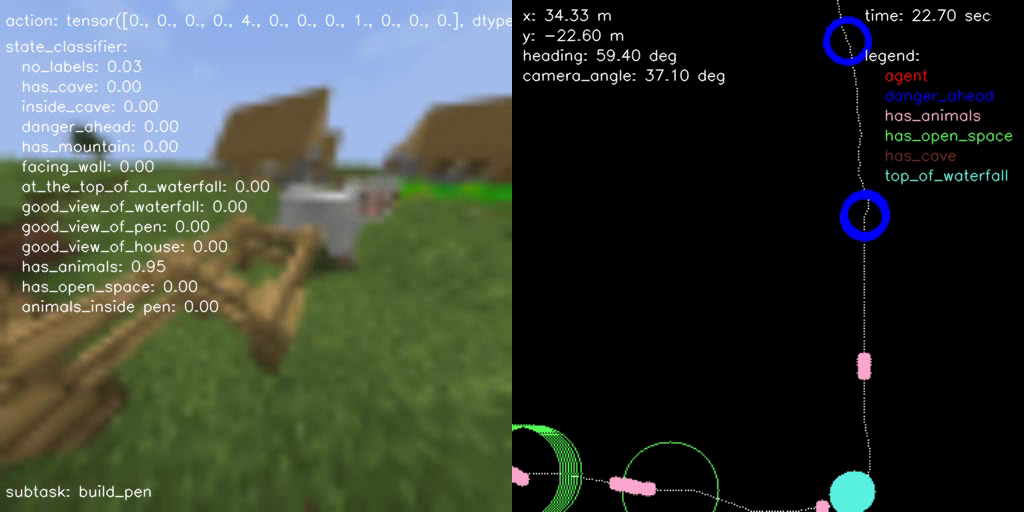}}\\
        \subfloat[]{\includegraphics[width=0.4\linewidth]{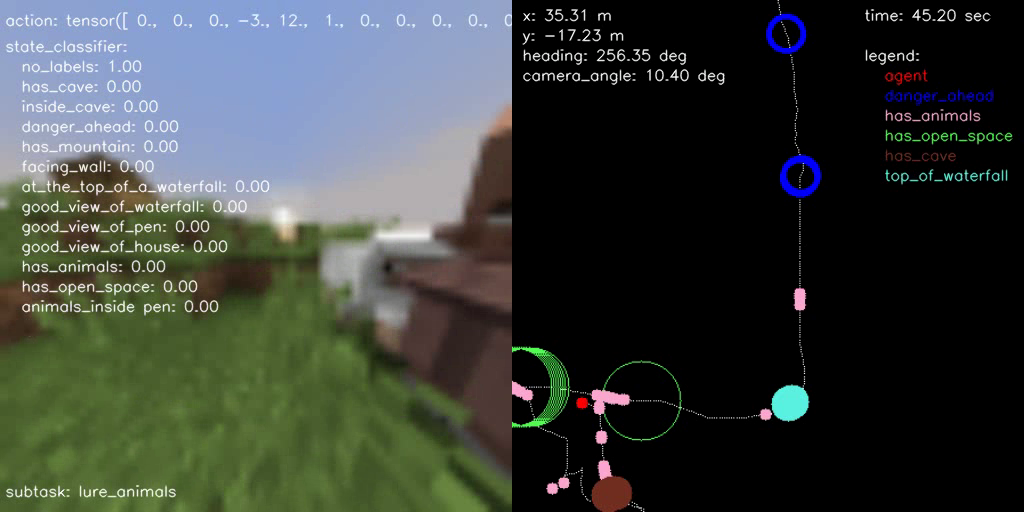}} &
        \subfloat[]{\includegraphics[width=0.4\linewidth]{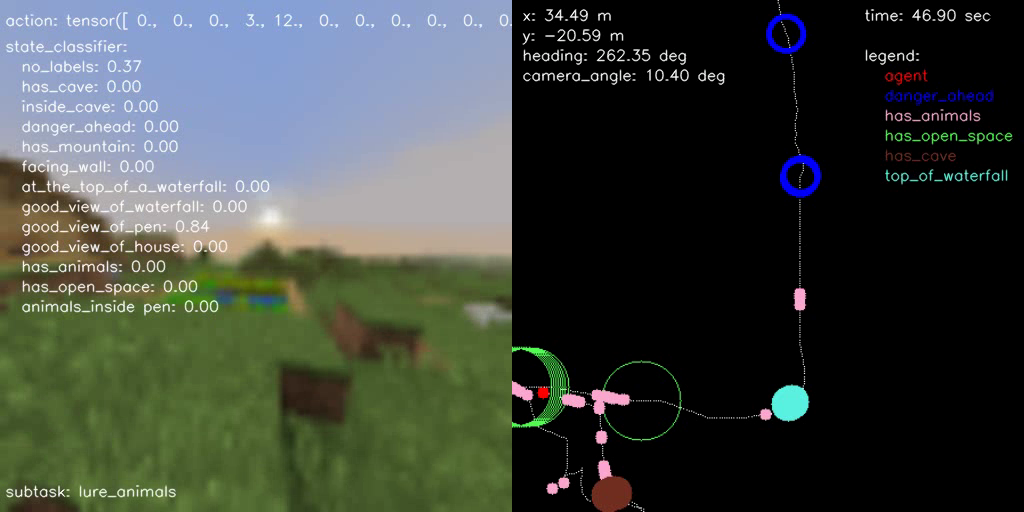}}
    \end{tabular}
  \caption{Sequence of frames of the hybrid agent solving the \textit{CreateVillageAnimalPen} task (complete video available at \url{https://youtu.be/b8xDMxEZmAE}).}
  \label{fig:bestcreatepen_frames}
\end{figure}

\begin{figure}[!ht]
  \centering
    \begin{tabular}{cc}
        \subfloat[]{\includegraphics[width=0.4\linewidth]{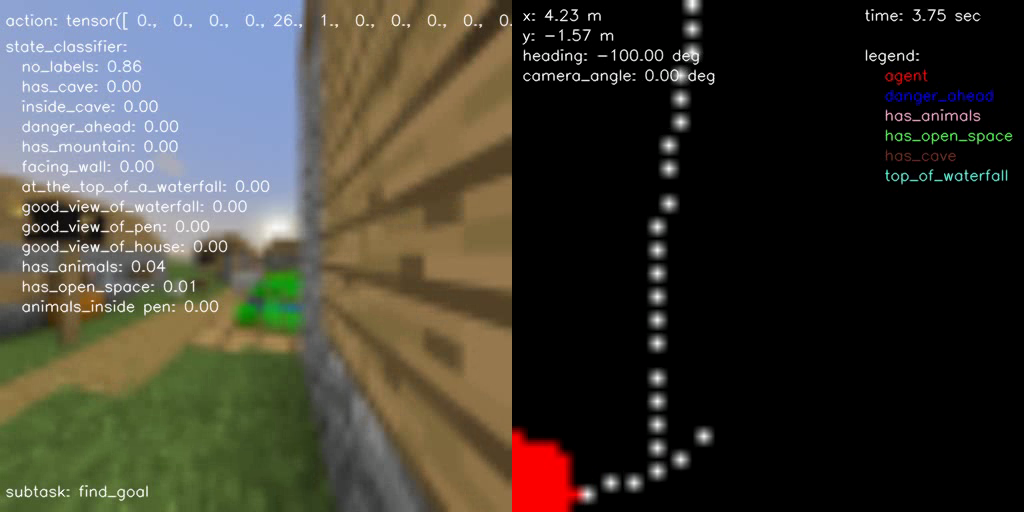}} &
        \subfloat[]{\includegraphics[width=0.4\linewidth]{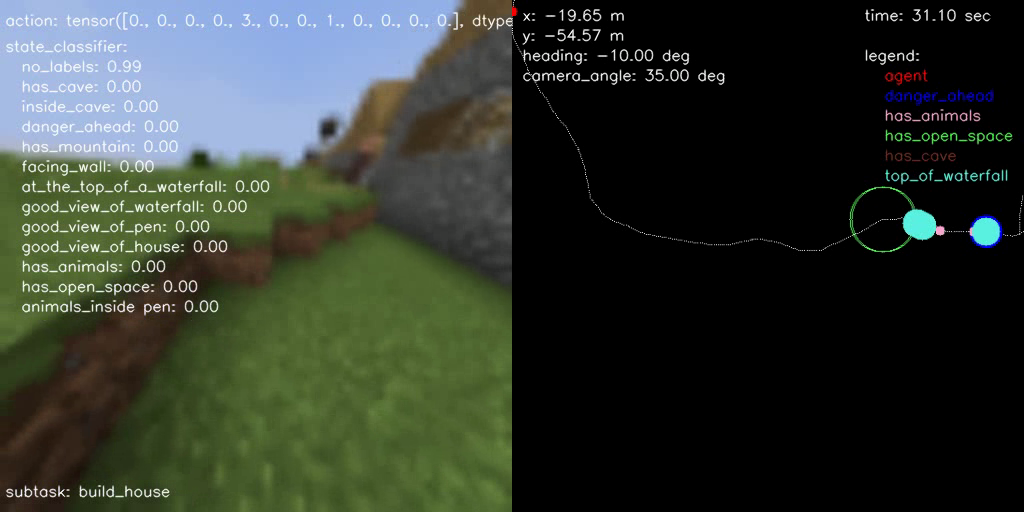}}\\
        \subfloat[]{\includegraphics[width=0.4\linewidth]{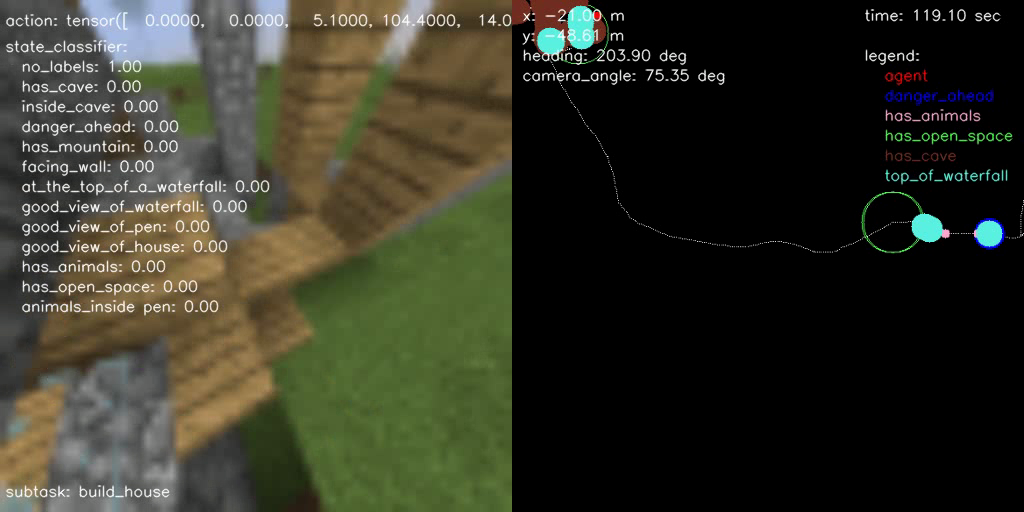}} &
        \subfloat[]{\includegraphics[width=0.4\linewidth]{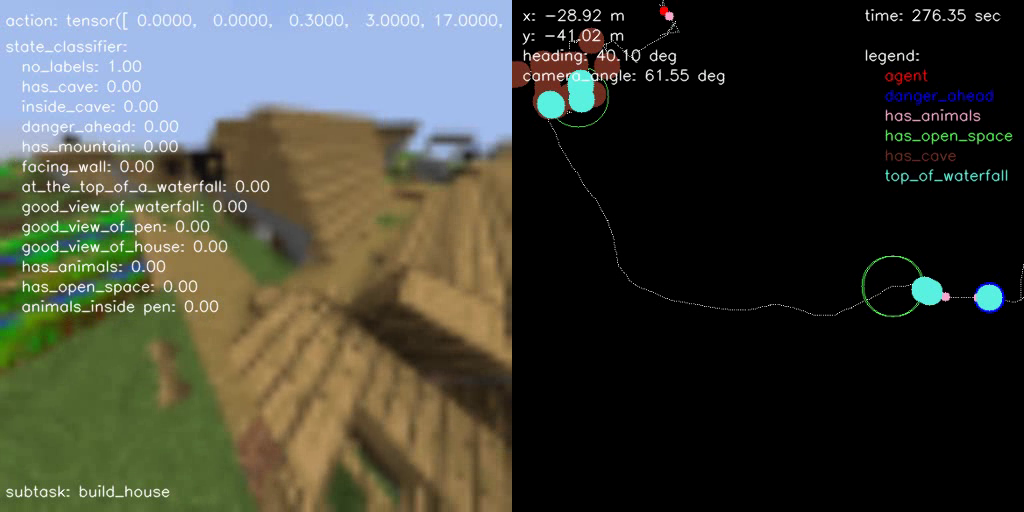}}
    \end{tabular}
  \caption{Sequence of frames of the hybrid agent solving the \textit{BuildVillageHouse} task (complete video available at \url{https://youtu.be/_uKO-ZqBMWQ}).}
  \label{fig:bestbuildhouse_frames}
\end{figure}

In terms of qualitative results, Figures~\ref{fig:bestcave_frames},~\ref{fig:bestwaterfall_frames},~\ref{fig:bestcreatepen_frames}, and~\ref{fig:bestbuildhouse_frames} show a sample episode illustrated by a sequence of frames of the hybrid agent solving the \textit{FindCave}, \textit{MakeWaterfall}, \textit{CreateVillageAnimalPen}, and \textit{BuildVillageHouse} tasks, respectively. Each figure shows the image frames received by the agent (left panel) overlaid with the actions taken (top), output of the state classifier (center), and the subtask currently being followed (bottom). The right panel shows the estimated odometry map overlaid with the location of the relevant states identified by the state classifier.
Links to the videos are provided in the figure captions.

\section{Results and Discussion}

\begin{table}[]
\caption{Summary of the \textit{TrueSkill}\textsuperscript{TM}\cite{herbrich2006trueskill} scores with mean and standard deviation computed from human evaluations separately for each performance metric and agent type averaged out over all tasks. Scores were computed after collecting 268 evaluations from 7 different human evaluators.}
\label{tab:trueskill_summary_average}
\resizebox{\textwidth}{!}{%
\begin{tabular}{@{}cccccc@{}}
\toprule
\multirow{2}{*}{\textbf{Task}} & \multirow{2}{*}{\textbf{\begin{tabular}[c]{@{}c@{}}Performance\ Metric\end{tabular}}} & \multicolumn{4}{c}{\textbf{TrueSkill Rating}} \\ \cmidrule(l){3-6} 
&  & \textbf{Behavior Cloning} & \textbf{Engineered} & \textbf{Hybrid} & \textbf{Human} \\ \midrule

\multirow{3}{*}{\begin{tabular}[c]{@{}c@{}}All Tasks\\ Combined\end{tabular}} & \begin{tabular}[c]{@{}c@{}}Best\\ Performer\end{tabular} & $20.30 \pm 1.81$ & $24.21 \pm 1.46$ & $25.49 \pm 1.40$ & $32.56 \pm 1.85$ \\ \cmidrule(l){2-6} 
& \begin{tabular}[c]{@{}c@{}}Fastest\\ Performer\end{tabular} & $19.42 \pm 1.94$ & $26.92 \pm 1.45$ & $27.59 \pm 1.38$ & $28.36 \pm 1.69$ \\ \cmidrule(l){2-6} 
& \begin{tabular}[c]{@{}c@{}}More Human-like\\ Behavior\end{tabular} & $20.09 \pm 2.04$ & $26.02 \pm 1.56$ & $26.94 \pm 1.57$ & $36.41 \pm 2.12$ \\ \bottomrule
\end{tabular}%
}
\end{table}

Each combination of condition (behavior cloning, engineered, hybrid, human) and performance metric (best performer, fastest performer, most human-like performer) is treated as a separate participant of a one-versus-one competition where skill rating is computed using the \textit{TrueSkill}\textsuperscript{TM}\footnote{Microsoft's \textit{TrueSkill}\textsuperscript{TM} Ranking System: \url{https://www.microsoft.com/en-us/research/project/trueskill-ranking-system}} Bayesian ranking system~\cite{herbrich2006trueskill}. In this Bayesian ranking system, the skill of each participant is characterized by a Gaussian distribution with a mean value $\mu$, representing the average skill of a participant and standard deviation $\sigma$ representing the degree of uncertainty in the participant's skill. There are three outcomes after each comparison: the first agent wins the comparison, the second agent the comparison, or there is a draw (human evaluator selects ''None'' when asked which participant performed better in each metric). Given this outcome, the \textit{TrueSkill}\textsuperscript{TM} ranking system updates the belief distribution of each participant using Bayes' Theorem~\cite{herbrich2006trueskill}, like how scores were computed in the official 2021 NeurIPS MineRL BASALT competition. The open-source \emph{TrueSkill} Python package\footnote{\emph{TrueSkill} Python package: \url{https://github.com/sublee/trueskill} and \url{https://trueskill.org/}.} is used.

The final mean and standard deviation of the \textit{TrueSkill}\textsuperscript{TM} scores computed for each performance metric and agent type are shown in Table~\ref{tab:trueskill_summary_average}. The scores were computed after collecting 268 evaluations from 7 different human evaluators. The main proposed ``Hybrid'' agent, which combines engineered and learned modules, outperforms both pure hand-designed (``Engineered'') and pure learned (``Behavior Cloning'') agents in the ``Best Performer'' category, achieving $5.3 \%$ and $25.6 \%$ higher mean skill rating when compared to the ``Engineered'' and ``Behavior Cloning'' baselines, respectively. However, when compared to the ``Human'' scores, the main proposed agent achieves $21.7 \%$ lower mean skill rating, illustrating that even the best approach is still not able to outperform a human player with respect to best performing the task.

When looking at the ``Fastest Performer'' metric, the ``Hybrid'' agent outperforms both ``Engineered'' and ``Behavior Cloning'' baselines, respectively, scoring only $2.7 \%$ lower than the human players. As expected, in the ``More Human-like Behavior'' performance metric the ``Human'' baseline wins by a large margin, however, the ``Hybrid'' still outperforms all other baselines, including the ``Behavior Cloning'' agent, which is purely learned from human data. The pure learned agent did not make use of the safety-critical engineered subtask, which allowed the agent to escape bodies of water and other obstacles around the environment. Plots showing how the \textit{TrueSkill}\textsuperscript{TM} scores evolved after each match (one-to-one comparison between different agent types) are shown in Section~\ref{appendix:trueskill}.

\begin{table}[]
\caption{Summary of the \textit{TrueSkill}\textsuperscript{TM}\cite{herbrich2006trueskill} scores with mean and standard deviation computed from human evaluations separately for each performance metric, agent type, and task. Scores were computed after collecting 268 evaluations from 7 different human evaluators.}
\label{tab:trueskill_summary}
\resizebox{\textwidth}{!}{%
\begin{tabular}{@{}cccccc@{}}
\toprule
\multirow{2}{*}{\textbf{Task}} & \multirow{2}{*}{\textbf{\begin{tabular}[c]{@{}c@{}}Performance\ Metric\end{tabular}}} & \multicolumn{4}{c}{\textbf{TrueSkill Rating}} \\ \cmidrule(l){3-6} 
&  & \textbf{Behavior Cloning} & \textbf{Engineered} & \textbf{Hybrid} & \textbf{Human} \\ \midrule

\multirow{3}{*}{FindCave} & \begin{tabular}[c]{@{}c@{}}Best\\ Performer\end{tabular} & $24.32 \pm 1.27$ & $24.29 \pm 1.21$ & $25.14 \pm 1.19$ & $32.90 \pm 1.52$ \\ \cmidrule(l){2-6} 
& \begin{tabular}[c]{@{}c@{}}Fastest\\ Performer\end{tabular} & $24.65 \pm 1.27$ & $24.16 \pm 1.21$ & $24.79 \pm 1.19$ & $32.75 \pm 1.54$ \\ \cmidrule(l){2-6} 
& \begin{tabular}[c]{@{}c@{}}More Human-like\\ Behavior\end{tabular} & $21.53 \pm 1.70$ & $26.61 \pm 1.43$ & $28.25 \pm 1.51$ & $38.95 \pm 1.96$ \\ \midrule

\multirow{3}{*}{MakeWaterfall} & \begin{tabular}[c]{@{}c@{}}Best\\ Performer\end{tabular} & $15.16 \pm 2.10$ & $23.16 \pm 1.60$ & $26.53 \pm 1.39$ & $24.39 \pm 1.62$ \\ \cmidrule(l){2-6} 
& \begin{tabular}[c]{@{}c@{}}Fastest\\ Performer\end{tabular} & $14.67 \pm 2.26$ & $28.95 \pm 1.74$ & $28.88 \pm 1.46$ & $18.85 \pm 2.02$ \\ \cmidrule(l){2-6} 
& \begin{tabular}[c]{@{}c@{}}More Human-like\\ Behavior\end{tabular} & $21.27 \pm 1.98$ & $24.51 \pm 1.52$ & $26.91 \pm 1.35$ & $26.48 \pm 1.61$ \\ \midrule

\multirow{3}{*}{\begin{tabular}[c]{@{}c@{}}CreateVillage\\ AnimalPen\end{tabular}} & \begin{tabular}[c]{@{}c@{}}Best\\ Performer\end{tabular} & $21.87 \pm 1.94$ & $23.56 \pm 1.38$ & $26.49 \pm 1.48$ & $33.89 \pm 1.73$ \\ \cmidrule(l){2-6} 
& \begin{tabular}[c]{@{}c@{}}Fastest\\ Performer\end{tabular} & $18.62 \pm 2.27$ & $27.00 \pm 1.32$ & $29.93 \pm 1.50$ & $28.59 \pm 1.53$ \\ \cmidrule(l){2-6} 
& \begin{tabular}[c]{@{}c@{}}More Human-like\\ Behavior\end{tabular} & $21.54 \pm 2.29$ & $25.53 \pm 1.57$ & $27.99 \pm 1.68$ & $40.60 \pm 2.44$ \\ \midrule

\multirow{3}{*}{\begin{tabular}[c]{@{}c@{}}BuildVillage\\ House\end{tabular}} & \begin{tabular}[c]{@{}c@{}}Best\\ Performer\end{tabular} & $19.83 \pm 1.92$ & $25.81 \pm 1.66$ & $23.81 \pm 1.55$ & $39.05 \pm 2.53$ \\ \cmidrule(l){2-6} 
& \begin{tabular}[c]{@{}c@{}}Fastest\\ Performer\end{tabular} & $19.75 \pm 1.97$ & $27.58 \pm 1.54$ & $26.76 \pm 1.35$ & $33.24 \pm 1.67$ \\ \cmidrule(l){2-6} 
& \begin{tabular}[c]{@{}c@{}}More Human-like\\ Behavior\end{tabular} & $16.04 \pm 2.19$ & $27.42 \pm 1.72$ & $24.61 \pm 1.72$ & $39.61 \pm 2.46$ \\ \bottomrule
\end{tabular}%
}
\end{table}

Table~\ref{tab:trueskill_summary} breaks down the results presented in Table~\ref{tab:trueskill_summary_average} for each separate task.
Similar to what was discussed for Table~\ref{tab:trueskill_summary_average}, excluding the ``Human'' baseline, the ``Hybrid'' approach outperforms both ``Behavior Cloning'' and ``Engineered'' baselines in terms of mean skill rating in 8 out of the 12 performance metrics, or in $66.6 \%$ of the comparisons.
Similarly, hybrid intelligence approaches, which include both ``Hybrid'' and ``Engineered'' baselines, outperform the pure learning ``Behavior Cloning'' approach in all 12 performance metrics, not considering the ``Human'' baseline.
The ``Hybrid'' approach only outperforms the ``Human'' baseline in 4 out of the 12 performance metrics, or in $33.3 \%$ of the comparisons.

Particularly for the \textit{MakeWaterfall} task, the proposed hybrid approach outperforms human players for all performance metrics.
The largest margin observed is for the ``Fastest Performer'' metric; the hybrid approach scores $53.2 \%$ higher than the human players.
This large margin comes from human players taking more time to find the best spot to place the waterfall and signal the end of the episode when compared to the engineered subtasks. Plots showing all results for each pairwise comparison are shown in Section~\ref{appendix:barplot}.

When solving the \textit{FindCave} task\footnote{Sample trajectory of hybrid agent solving the \textit{FindCave} task: \url{https://youtu.be/MR8q3Xre_XY}.}, the agent uses the learned navigation policy to search for caves while avoiding water while simultaneously building the map of its environment. Once the agent finds the cave, it throws the snowball to signal the end of the episode. In the \textit{MakeWaterfall} task\footnote{Sample trajectory of hybrid agent solving the \textit{MakeWaterfall} task: \url{https://youtu.be/eXp1urKXIPQ}.}, the hybrid agent uses the learned navigation policy to climb the mountains, detects a suitable location to build the waterfall, builds it, then moves to the picture location using engineered subtasks, and throws the snowball to signal the end of the episode. For the \textit{CreateVillageAnimalPen} task\footnote{Sample trajectory of hybrid agent solving the \textit{CreateVillageAnimalPen} task: \url{https://youtu.be/b8xDMxEZmAE}.}, the agent uses the learned navigation policy and the state classifier to search for an open location to build a pen, builds the pen using an engineered building subtask that repeats the actions taken by the human demonstrators, uses the state classifier and odometry map to go to previously seen animal locations, and then attempts to lure them back to the pen and throws the snowball to signal the end of the episode. Finally, when solving the \textit{BuildVillageHouse} task\footnote{Sample trajectory of hybrid agent solving the \textit{BuildVillageHouse} task: \url{https://youtu.be/_uKO-ZqBMWQ}.}, the hybrid agent spawns nearby a village and uses the learned navigation policy and the state classifier to search for an open location to build a house, builds a house using an engineered building subtask that repeats the actions taken by the human demonstrators, tours the house, and throws the snowball to signal the end of the episode. Each of the described subtasks are shown in Section~\ref{appendix:frames} as a sequence of frames.

\section{Conclusions}

This chapter presents the solution that won first place and was awarded the most human-like agent in the 2021 NeurIPS MineRL BASALT competition, ``Learning from Human Feedback in Minecraft.'' The approach used the available human demonstration data and additional human feedback to train machine learning modules that were combined with engineered ones to solve hierarchical tasks in Minecraft.

The proposed method was compared to both end-to-end machine learning and pure engineered solutions by collecting human evaluations that judged agents in head-to-head matches to answer which agent best solved the task, which agent was the fastest, and which one had the most human-like behavior. These human evaluations were converted to a skill rating score for each question, like how players are ranked in multiplayer online games.

After collecting 268 human evaluations, hybrid intelligence approaches outperformed end-to-end machine learning approaches in all 12 performance metrics computed, even outperforming human players in 4 of them. The results also showed that incorporating machine learning modules for navigation as opposed to engineering navigation policies led to higher scores in 8 out of 12 performance metrics.

Overall, hybrid intelligence approach proves advantageous to solve hierarchical tasks, compared to end-to-end machine learning approaches when the subcomponents of the task are understood by human experts and limited human feedback data is available.

\end{appendices}

\end{document}